\pdfoutput=1
\documentclass[11pt]{article}

\usepackage{CJK}

\usepackage[]{acl}

\usepackage{times}
\usepackage{latexsym}

\usepackage[T1]{fontenc}
\usepackage[utf8]{inputenc}

\usepackage{microtype}

\usepackage{url}

\usepackage[switch]{lineno}
\usepackage{amssymb}
\usepackage{latexsym}
\usepackage{mathtools}
\usepackage{algorithm}
\usepackage[noend]{algpseudocode}
\usepackage{setspace}

\usepackage{graphicx}
\usepackage{amsmath}
\usepackage{amsthm}
\usepackage{amssymb}
\usepackage{booktabs}
\usepackage{graphicx} 
\usepackage{graphics}
\usepackage{amssymb}
\usepackage{amsmath}
\usepackage{pifont}
\usepackage{makecell}
\usepackage{resizegather}
\usepackage{etoolbox}
\usepackage{booktabs,makecell,tabularx} 
\usepackage{enumitem}
\usepackage{mathtools}
\usepackage{cleveref}
\usepackage{multirow}
\usepackage{booktabs,array}
\usepackage{amsmath, bm}
\usepackage{color, colortbl}
\usepackage{xcolor}
\usepackage{booktabs}
\usepackage{algorithm}
\usepackage{array}
\usepackage{subfigure}
\usepackage{multirow}
\usepackage{threeparttable}
\usepackage{tabularx}
\usepackage{booktabs}
\usepackage{colortbl}
\usepackage{xcolor}
\usepackage{stfloats}
\usepackage[noend]{algpseudocode}
\usepackage{mathtools}
\usepackage{makecell}
\usepackage{setspace}
\usepackage{resizegather}
\usepackage{etoolbox}
\usepackage{booktabs,makecell,tabularx} 
\usepackage{enumitem}
\usepackage{mathtools}
\usepackage{cleveref}
\usepackage{multirow}
\usepackage{booktabs,array}
\usepackage{amsmath, bm}
\usepackage{mathrsfs}
\usepackage{wasysym} % 提供\xmark和\cmark符号
\usepackage{soul}
\usepackage[most]{tcolorbox}

\usepackage{minitoc}
\usepackage{titletoc}

\newcolumntype{P}[1]{>{\centering\arraybackslash}p{#1}}

\let\oldnl\nl% Store \nl in \oldnl
\definecolor{Gray}{gray}{0.9}
\definecolor{LightCyan}{rgb}{0.88,1,1}
\newcommand{\nonl}{\renewcommand{\nl}{\let\nl\oldnl}}

\newcolumntype{H}{>{\setbox0=\hbox\bgroup}c<{\egroup}@{}}
\newcommand{\cmark}{\ding{51}}%
\newcommand{\xmark}{\ding{55}}%

\title{MAGA-Bench: Machine-Augment-Generated Text via Alignment Detection Benchmark}

\author{
     \textbf{Anyang Song$^{1,3}$},
    ~\textbf{Ying Cheng$^{2}$},
    ~\textbf{Yiqian Xu}$^{1,3*}$,
    ~\textbf{Rui Feng}$^{1,3}$\thanks{\ \ Corresponding author.} \\
    $^{1}$College of Computer Science and Artificial Intelligence, Fudan University\\
    $^{2}$School of Computer Science and Technology, Tongji University\\
    $^{3}$Shanghai Key Laboratory of Intelligent Information Processing, Fudan University \\
    {\tt aysong24@m.fudan.edu.cn} \\
    {\tt yingcheng@tongji.edu.cn} \\
    {\tt \{xuyiqian,fengrui\}@fudan.edu.cn} \\
}

\begin{document}
%中文
\begin{CJK}{UTF8}{gbsn}

\maketitle

\begin{abstract}
Machine-Generated Text (MGT) is becoming increasingly difficult to distinguish from Human-Written Text (HWT). This trend has exacerbated malicious activities such as fake news and online fraud. The generalization ability of fine-tuned detectors relies heavily on dataset quality, and simply expanding the sources of MGT may become increasingly insufficient. Further augmentation of the generation process is required. Based on HC-Var's theory, enhancing the human-like alignment of MGT not only facilitates robustness testing of existing detectors but also boosts the generalization ability of detectors fine-tuned on such aligned MGT datasets. Therefore, we propose the \textbf{M}achine-\textbf{A}ugment-\textbf{G}enerated Text via \textbf{A}lignment (MAGA) Detection Benchmark. MAGA integrates several alignment methods, ranging from prompt construction to \textbf{G}enerator-\textbf{D}etector \textbf{A}dversarial \textbf{R}einforcement \textbf{L}earning (GDARL) and the reasoning process. In our experiments, the RoBERTa detector fine-tuned on MAGA achieves an average improvement of 4.60\% in generalization AUC. Conversely, the aligned MGTs in MAGA also lead to an average decrease of 8.13\% in the AUC of selected detectors. We hope the MAGA Benchmark will provide valuable insights for future research on the generalization ability of MGT detectors.
\footnote{Our code and datasets are available at \url{https://github.com/anyangsong/MAGA}} 
\end{abstract}

\section{Introduction}

In recent years, Large Language Models (LLMs) have advanced rapidly, with remarkable improvements in their language comprehension capabilities, precision in instruction execution, and ability to generate complex text to fulfill diverse tasks \cite{brown2020language, ouyang2022training}. It has become increasingly difficult for humans to distinguish between Machine-Generated Text (MGT) and Human-Written Text (HWT) \cite{chang2024survey}. This powerful capability, however, poses substantial risks of abuse. \citet{hanley2024machine} noted that in 2022, the proportion of AI-generated news on websites notorious for disseminating false rumors increased by approximately 457\%. Such misinformation is not only used in fake news \cite{zellers2019defending, dugan-etal-2023-roft} but also utilized by numerous malicious actors for various unethical or even illegal activities, including online fraud \cite{weidinger2021ethical, ayoobi2023looming} and academic misconduct \cite{stokel2022ai, kasneci2023chatgpt}. Therefore, to address and mitigate these potential risks in information dissemination, the development of efficient and robust MGT detectors is of paramount importance.

\begin{figure}[!t]
    \centering
    \includegraphics[width=1.0\linewidth]{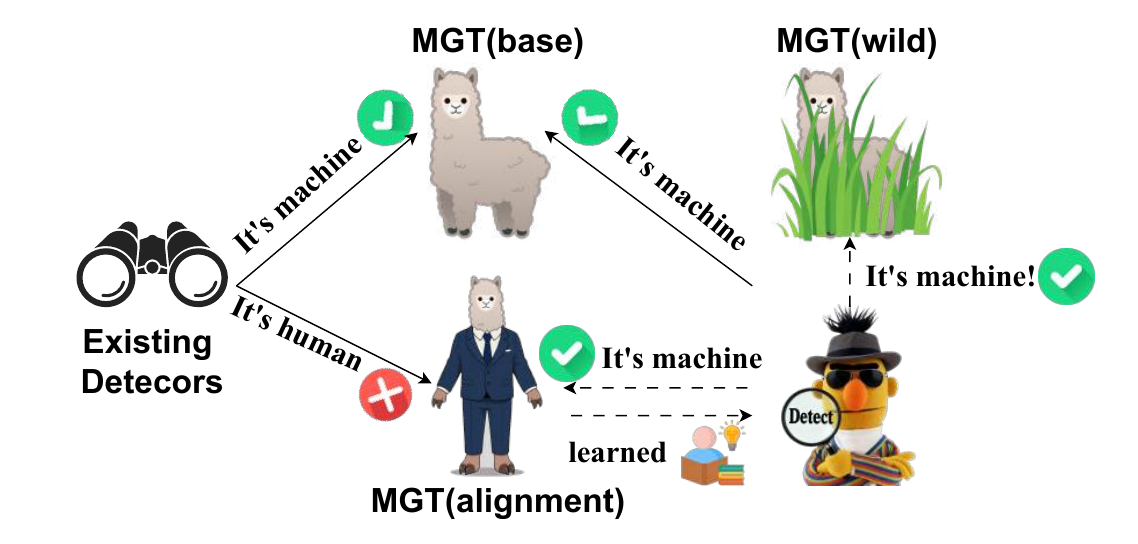}
    \caption{More human-like aligned MGT not only evades detection by existing detectors but also facilitates the fine-tuning of neural-based detectors, thereby enhancing their generalization capability for in-the-wild detection.}
    \label{fig:MAGA Introduction}
\end{figure}

\begin{table*}[!t]
	\resizebox{\linewidth}{!}{
		\centering
		\setlength{\tabcolsep}{6pt} % 调整列之间的间距
		\renewcommand{\arraystretch}{1} % 调整行高，增加表格的空白，使其更易读
		% cyan（青色）、magenta（品红色 / 洋红色）、yellow（黄色）、black（黑色）
		\definecolor{darkgreen}{cmyk}{1, 0, 1, 0}
		\definecolor{darkred}{cmyk}{0, 1, 1, 0.02}
		
		\begin{tabular}{
				l| % 左对齐
				>{\centering\arraybackslash}m{2cm}| % 居中对齐，固定宽度
				>{\centering\arraybackslash}m{2cm}|
				>{\centering\arraybackslash}m{2cm}| 
				>{\centering\arraybackslash}m{2cm}| 
				>{\centering\arraybackslash}m{2cm}|
				>{\centering\arraybackslash}m{2cm}|
				>{\centering\arraybackslash}m{2cm}
			}
			
			\toprule
			
			\textbf{Dataset} & \textbf{Origin Size} & \textbf{Domain Coverage} & \textbf{Model Coverage} & \textbf{Multilingual Coverage} & \textbf{Adversarial Coverage} & \textbf{Sampling Coverage} & \textbf{Alignment Augment} \\ 
			
			\midrule
			
			TuringBench \cite{uchendu2021turingbench}&200k&\textcolor{darkred}{\xmark}&\textcolor{darkgreen}{\cmark}&\textcolor{darkred}{\xmark}&\textcolor{darkred}{\xmark}&\textcolor{darkred}{\xmark}&\textcolor{darkred}{\xmark}\\ 
			HC3 \cite{guo2023close}&26.9k&\textcolor{darkgreen}{\cmark}&\textcolor{darkred}{\xmark}&\textcolor{darkgreen}{\cmark}&\textcolor{darkred}{\xmark}&\textcolor{darkred}{\xmark}&\textcolor{darkred}{\xmark}\\ 
			MGTBench \cite{he2023mgtbench}&2.8k&\textcolor{darkgreen}{\cmark}&\textcolor{darkgreen}{\cmark}&\textcolor{darkred}{\xmark}&\textcolor{darkgreen}{\cmark}&\textcolor{darkred}{\xmark}&\textcolor{darkred}{\xmark}\\ 
			MULTITuDE \cite{macko-etal-2023-multitude}&74.1k&\textcolor{darkred}{\xmark}&\textcolor{darkgreen}{\cmark}&\textcolor{darkgreen}{\cmark}&\textcolor{darkred}{\xmark}&\textcolor{darkred}{\xmark}&\textcolor{darkred}{\xmark}\\ 
			M4 \cite{wang-etal-2024-m4}&122k&\textcolor{darkgreen}{\cmark}&\textcolor{darkgreen}{\cmark}&\textcolor{darkgreen}{\cmark}&\textcolor{darkred}{\xmark}&\textcolor{darkred}{\xmark}&\textcolor{darkred}{\xmark}\\ 
			HC-Var \cite{xu-etal-2024-generalization}&145k&\textcolor{darkgreen}{\cmark}&\textcolor{darkred}{\xmark}&\textcolor{darkred}{\xmark}&\textcolor{darkred}{\xmark}&\textcolor{darkred}{\xmark}&\textcolor{darkred}{\xmark}\\ 
			MAGE \cite{li2024mage}&447k&\textcolor{darkgreen}{\cmark}&\textcolor{darkgreen}{\cmark}&\textcolor{darkred}{\xmark}&\textcolor{darkred}{\xmark}&\textcolor{darkred}{\xmark}&\textcolor{darkred}{\xmark}\\ 
			RAID \cite{dugan2024raid}&570k&\textcolor{darkgreen}{\cmark}&\textcolor{darkgreen}{\cmark}&\textcolor{darkred}{\xmark}&\textcolor{darkgreen}{\cmark}&\textcolor{darkgreen}{\cmark}&\textcolor{darkred}{\xmark}\\
			RealDet \cite{zhu2025reliably}&847k&\textcolor{darkgreen}{\cmark}&\textcolor{darkgreen}{\cmark}&\textcolor{darkgreen}{\cmark}&\textcolor{darkgreen}{\cmark}&\textcolor{darkred}{\xmark}&\textcolor{darkred}{\xmark}\\ 
			
			\midrule
			
			\textbf{MAGA (Ours)}&\textbf{936k}&\textcolor{darkgreen}{\cmark}&\textcolor{darkgreen}{\cmark}&\textcolor{darkgreen}{\cmark}&\textcolor{darkgreen}{\cmark}&\textcolor{darkgreen}{\cmark}&\textcolor{darkgreen}{\cmark}\\
			
			\bottomrule
			
	\end{tabular}}
	\caption{A comparison of publicly available data sources of MGT. The term "Original Size", introduced by RealDet \cite{zhu2025reliably}, refers to the collection of HWTs and direct MGT outputs, excluding rule-based transformations such as those employed in RAID \cite{dugan2024raid}. A more comprehensive comparison is provided in Table \ref{tab:all-datasets}.}
	\label{tab:all-datasets-core}
	\vspace{-4pt}
\end{table*}

\begin{figure*}[!t]
	\centering
	\includegraphics[width=\textwidth]{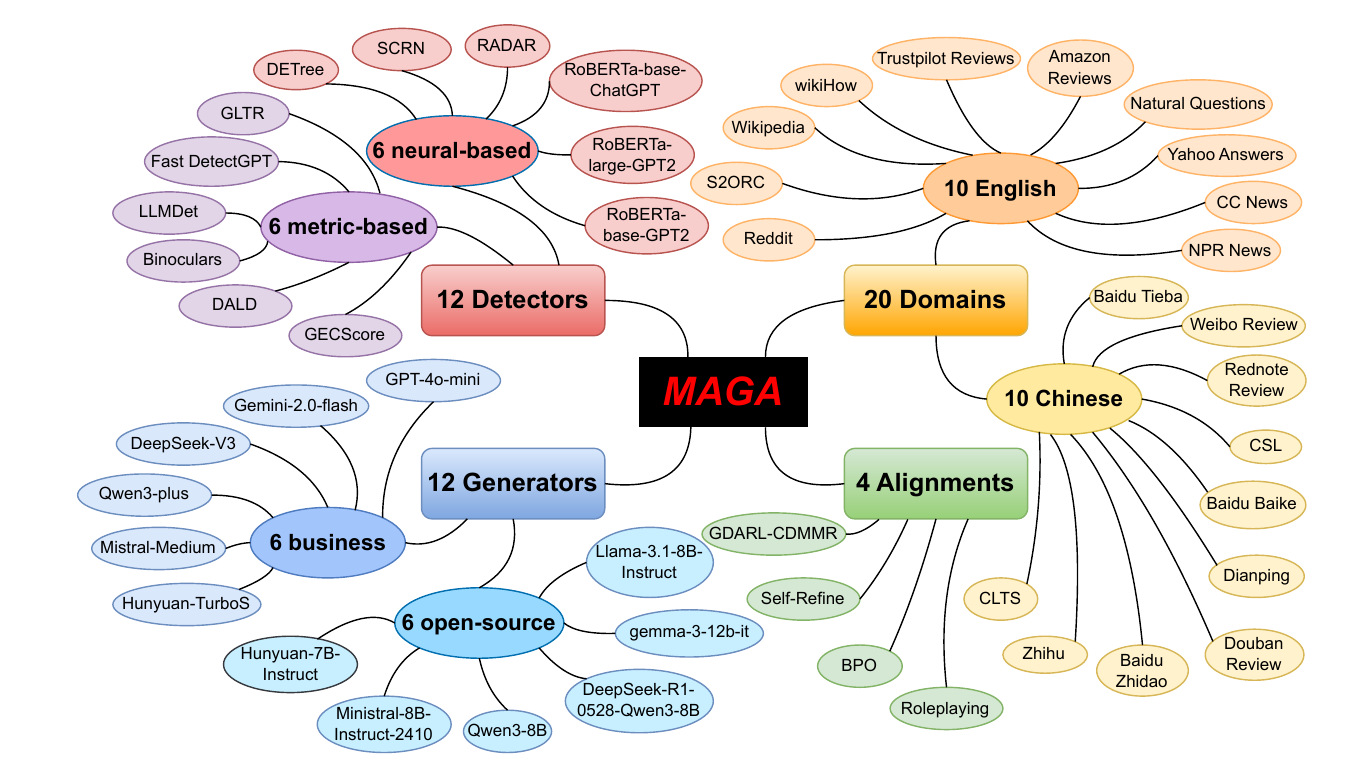}
	\caption{Overview of MAGA-Bench. The construction of the dataset encompasses 20 domains, 12 generators, 2 languages and 4 alignment methods. Selected detectors are evaluated on this dataset. Various decoding strategies are also employed, with details provided in Appendix \ref{sec:decode}.}
	\label{fig:MAGA Overview}
\end{figure*}

To address this need, numerous specialized detection methods and benchmarks have been developed. Prior datasets increased detection difficulty by incorporating diverse distributions, a direction that marked its first pinnacle in M4 \cite{wang-etal-2024-m4}. However, the generation of individual texts in these datasets remained straightforward, relying on a limited set of fixed prompt templates. RAID \cite{dugan2024raid} is the first benchmark to significantly prioritize the robustness of detectors against adversarial attacks. The implemented attacks, such as homoglyph transformations (e.g., randomly transforming 'a'-'z' into '$a$'-'$z$'), successfully compromise detector performance, particularly for metric-based detectors. In line with this, \citet{hans2024spotting} observed that the performance of their proposed Binoculars detector drops to near zero when evaluating highly randomized MGT. Nevertheless, this form of text manipulation disrupts the surface structure of the text, thereby hindering neural-based detectors, exemplified by RoBERTa \cite{liu2019roberta}, from learning and understanding semantics during fine-tuning, even though the semantic meaning remains interpretable to human readers.

HC-Var \cite{xu-etal-2024-generalization} indicates that MGT with a higher degree of human-like alignment exhibits relevant features closer to those of HWT. This proximity assists neural detectors in learning more effective decision boundaries, thereby enhancing generalized detection. Therefore, we augment the generation process while ensuring the diversity of data sources across 20 domains, 12 generators, 2 languages, and various decoding strategies. Our main contributions are: (i) we propose the \textbf{M}achine-\textbf{A}ugment-\textbf{G}enerated Text via \textbf{A}lignment Detection Benchmark (MAGA-Bench), which incorporates 4 human-like alignment methods to enhance MGT generation; (ii) we propose a \textbf{C}ross-\textbf{M}utual \textbf{R}eward (CMR) component, which facilitates the alignment process of large language models in support of the \textbf{G}enerator-\textbf{D}etector \textbf{A}dversarial \textbf{R}einforcement \textbf{L}earning (GDARL) framework; (iii) comprehensive experiments demonstrate that MAGA reduces the average AUC of 12 selected evaluation detectors by 8.13\%, while the RoBERTa detector fine-tuned on MAGA improves the average generalization AUC by 4.60\% across 4 selected external datasets.

\begin{figure*}[!t]
	\centering
	\includegraphics[width=\textwidth]{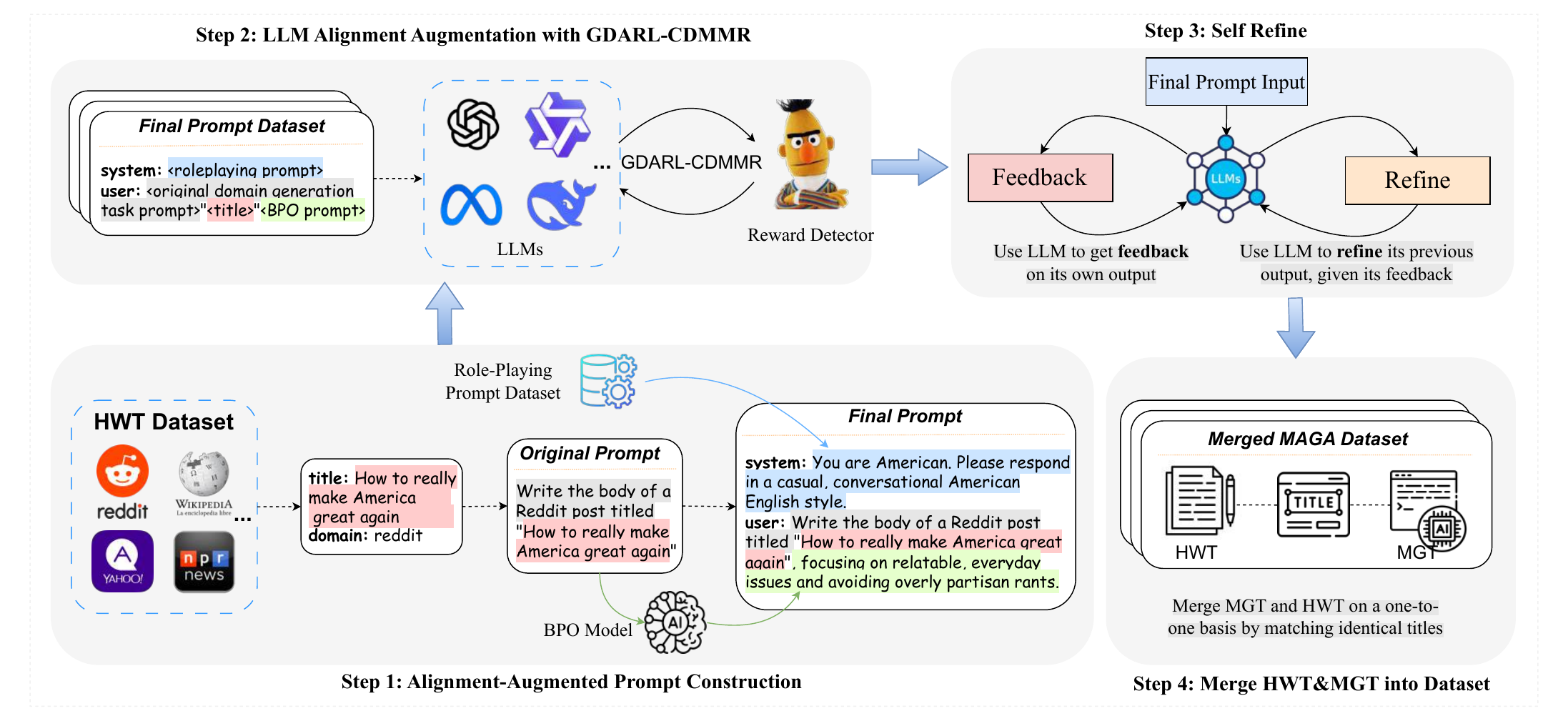}
	\caption{MAGA Pipeline, illustrating the construction process for the MAGA split of the MAGA dataset. MAGA split is the fully enhanced split that integrates all four alignment methods (see Table \ref{tab:data_distribution_overall}, Appendix \ref{sec:statistics maga} for detailed statistics of the splits).}
	\label{fig:MAGA Pipeline}
\end{figure*}

\section{Related Work}
\label{sec:relatedwork}

Previous MGT detection benchmarks (Table \ref{tab:all-datasets-core} for key representatives and Table \ref{tab:all-datasets} in the Appendix for a comprehensive compilation) mainly focused on expanding the diversity of the overall dataset distribution. Since TuringBench \cite{uchendu2021turingbench} extended the multi-generator (or multi-model) dimension, the dimensions used for expansion have become increasingly diverse. RuATD \cite{Shamardina_2022} and RoFT \cite{dugan-etal-2023-roft} were among the earliest works to introduce the multi-domain dimension. HC3 \cite{guo2023close} pioneered the multi-lingual dimension. MGTBench \cite{he2023mgtbench} and CHEAT \cite{yu2023cheat} pioneered the introduction of adversarial considerations, and OpenLLMText \cite{chen2023token} was among the first to explicitly incorporate multiple decoding strategies. This evolutionary trajectory established a significant milestone with M4 \cite{wang-etal-2024-m4}, which fully integrated three core dimensions: multi-domain, multi-generator, and multi-lingual. Subsequently, RAID \cite{dugan2024raid} advanced the field by comprehensively introducing a wide range of adversarial methods for the first time and covering sampling strategies alongside the multi-domain and multi-generator dimensions. RealDet \cite{zhu2025reliably} addressed the absence of multi-lingual support in the original RAID release but did not include sampling coverage.

Other MGT detection benchmarks in Table \ref{tab:all-datasets} concentrate more on specific research scenarios. RuATD, AuText2023 \cite{sarvazyan2023overview}, CCD \cite{wang2023evaluating}, IMDGSP \cite{mosca-etal-2023-distinguishing}, and MultiSocial \cite{macko2025multisocial} are dedicated to MGT detection for Russian, Spanish, source code, academic papers, and social media texts, respectively. LLM-DetectAIve \cite{abassy2024llm}, MixSet \cite{zhang2024llm}, and LAMP \cite{chakrabarty2025can} focus specifically on text paraphrasing attacks, which Beemo \cite{artemova2025beemo} further enhances through expert editing. MULTITuDE \cite{macko-etal-2023-multitude} conducted the first study on cross-linguistic family generalization, and MAGE \cite{li2024mage} was the first to establish multiple out-of-distribution testbeds for generalization research. Despite these targeted efforts, the generalization problem remains one of the most critical unresolved challenges in MGT detection. HC-Var \cite{xu-etal-2024-generalization} found that RoBERTa fine-tuned on more human-like aligned MGTs exhibits stronger generalization in detection tasks. Our work builds upon HC-Var's theory and aims to contribute to the research on the generalization problem in MGT detection.

\begin{figure*}[!t]
	\centering
	\includegraphics[width=\textwidth]{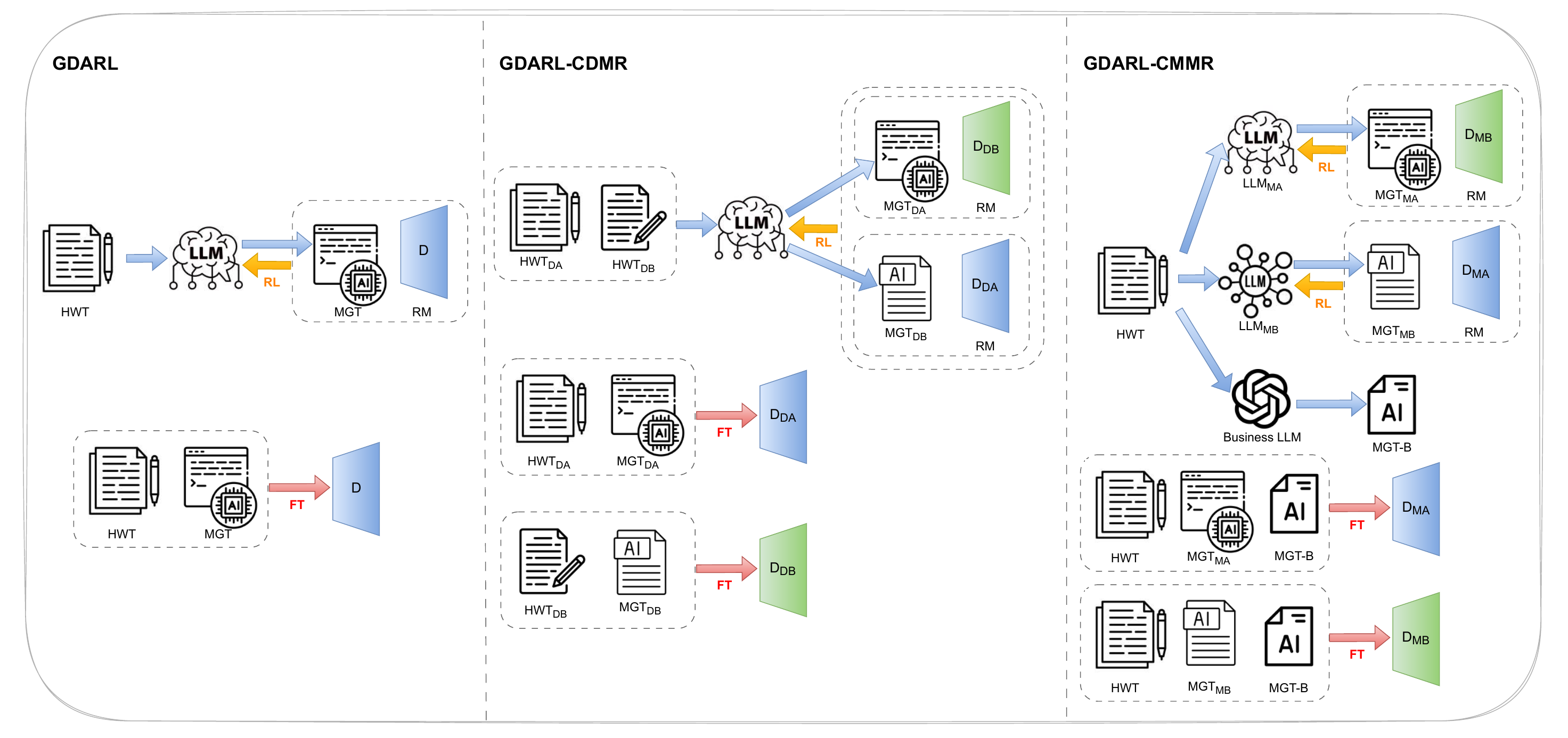}
	\caption{GDARL and its integration with the \textbf{C}ross-\textbf{M}utual \textbf{R}eward (CMR) component. The term GDARL serves as a collective designation for GDARL-like frameworks and also represents their most general form. The illustration depicts the two standard applications of CMR, namely GDARL-CDMR (Cross Domain Mutual Reward) and GDARL-CMMR (Cross Model Mutual Reward).}
	\label{fig:GDARL}
\end{figure*}

\section{The MAGA Dataset}
\label{sec:MAGA}

Figure \ref{fig:MAGA Overview} presents the components of the MAGA dataset. The MAGA dataset consists of two subsets: the English MAGA and MAGA-cn. Each subset comprises six splits, where MGB represents the unenhanced split, MAGA denotes the split integrating four distinct alignment methods, and the remaining four splits individually apply each single alignment method. To construct the MAGA dataset, taking the English subset as an example, we first sample 72k HWTs annotated with titles, which are consistently reused across all splits. For each split, we generate one MGT corresponding to each HWT by embedding the title into a prompt template. Consequently, each split contains 144k texts. Detailed statistics and data examples are provided in Appendix \ref{sec:statistics maga}.

\subsection{Domains}
\label{sec:domains}
Various domains raise detection difficulty \cite{li2024mage, dugan-etal-2023-roft} and improve the fine-tuned RoBERTa's performance \cite{wang2024semeval}. Taking English as an example, our sources cover social conversations (Reddit), summarization (S2ORC), general knowledge (Wikipedia, wikiHow), reviews (Trustpilot Reviews, Amazon Reviews), direct Q\&A (Yahoo Answers, Natural Questions), and news creation (CC News, NPR News), which challenge various capabilities of LLM. Details about all the domains (including the Chinese domains) are provided in Appendix \ref{sec:domain detail}.

\subsection{Generators}
\label{sec:generators}
Following \citet{sarvazyan-etal-2023-supervised}, we first selected the largest (commercial) LLM in each series as possible, then chose models with smaller parameter sizes. We also ensured selected models are as up-to-date as possible to include the latest insights. The models we selected are shown in Figure \ref{fig:MAGA Overview}, with details provided in Appendix \ref{sec:generator detail}.

\subsection{Alignment Methods}
\label{sec:alignment}

\begin{table*}[!t]
	\resizebox{\linewidth}{!}{
		\centering
%		\small
	\begin{tabular}{l|c|c|c|c|c|c|c|c|c}
		\toprule
		& \multicolumn{2}{c|}{ACC} 
		& \multicolumn{2}{c|}{TPR(M.)} 
		& \multicolumn{1}{c|}{TNR(H.)} 
		& \multicolumn{2}{c|}{AUC} 
		& \multicolumn{2}{c}{ACC(@FPR=5\%)} \\
		\midrule
		Alignment? & \xmark & \cmark & \xmark & \cmark & - & \xmark & \cmark & \xmark & \cmark \\
		\midrule
		R-B GPT2   & 58.44 & 51.96\textcolor{red}{(6.48$\downarrow$)} & 33.97 & 21.02\textcolor{red}{(12.95$\downarrow$)} & 82.90 & 56.91 & 45.02\textcolor{red}{(11.89$\downarrow$)} & 59.94 & 54.17\textcolor{red}{(5.77$\downarrow$)} \\
		R-L GPT2   & 56.69 & 51.47\textcolor{red}{(5.22$\downarrow$)}  & 20.34 & 9.91\textcolor{red}{(10.43$\downarrow$)}   & \underline{93.03} & 53.45 & 40.23\textcolor{red}{(13.22$\downarrow$)} & 56.97 & 51.35\textcolor{red}{(5.62$\downarrow$)}  \\
		R-B CGPT   & 57.60 & 47.15\textcolor{red}{(10.45$\downarrow$)} & 33.57 & 12.67\textcolor{red}{(20.90$\downarrow$)}  & 81.62 & 63.97 & 50.31\textcolor{red}{(13.66$\downarrow$)} & 54.73 & 49.01\textcolor{red}{(5.72$\downarrow$)}  \\
		RADAR      & 59.82 & 54.83\textcolor{red}{(4.99$\downarrow$)}  & 29.56 & 19.58\textcolor{red}{(9.98$\downarrow$)}   & 90.08 & 63.43 & 53.49\textcolor{red}{(9.94$\downarrow$)}  & 60.33 & 55.68\textcolor{red}{(4.65$\downarrow$)}  \\
		SCRN       & 74.26 & 73.23\textcolor{red}{(1.03$\downarrow$)}  & \underline{94.23} & \underline{92.17}\textcolor{red}{(2.06$\downarrow$)}   & 54.29 & 84.76 & 81.79\textcolor{red}{(2.97$\downarrow$)}  & 59.97 & 58.93\textcolor{red}{(1.04$\downarrow$)}  \\
		DETree     & 58.05 & 57.94\textcolor{red}{(0.11$\downarrow$)}  & \textbf{99.94} & \textbf{99.72}\textcolor{red}{(0.22$\downarrow$)}   & 16.16 & \textbf{93.92} & \textbf{92.89}\textcolor{red}{(1.03$\downarrow$)}  & \textbf{87.23} & \textbf{85.26}\textcolor{red}{(1.97$\downarrow$)}  \\
		\midrule
		neural.avg& 60.81 & 56.10\textcolor{red}{(4.71$\downarrow$)}  & 51.94 & 42.51\textcolor{red}{(9.42$\downarrow$)}  & 69.68 & 69.41 & 60.62\textcolor{red}{(8.79$\downarrow$)} & 63.20 & 59.07\textcolor{red}{(4.13$\downarrow$)}  \\
		\midrule
		GLTR       & - & -  & - & -  & - & 71.31 & 62.05\textcolor{red}{(9.26$\downarrow$)}  & 70.27 & 62.09\textcolor{red}{(8.18$\downarrow$)}  \\
		F-DetectGPT& 79.97 & \underline{78.41}\textcolor{red}{(1.57$\downarrow$)}  & 76.27 & 73.14\textcolor{red}{(3.13$\downarrow$)}   & 83.67 & 84.01 & \underline{83.38}\textcolor{red}{(0.63$\downarrow$)}  & 79.96 & \underline{78.58}\textcolor{red}{(1.38$\downarrow$)}  \\
		LLMDet     & 50.46 & 43.80\textcolor{red}{(6.66$\downarrow$)} & 28.74 & 15.43\textcolor{red}{(13.31$\downarrow$)}  & 72.17 & 47.89 & 35.58\textcolor{red}{(12.31$\downarrow$)} & 56.89 & 51.02\textcolor{red}{(5.87$\downarrow$)}  \\
		Binoculars & \textbf{84.26} & \textbf{79.03}\textcolor{red}{(5.23$\downarrow$)}  & 83.93 & 73.47\textcolor{red}{(10.46$\downarrow$)}  & 84.59 & \underline{86.76} & 79.68\textcolor{red}{(7.08$\downarrow$)}  & \underline{83.23} & 75.17\textcolor{red}{(8.06$\downarrow$)}  \\
		DALD       & \underline{80.01} & 77.68\textcolor{red}{(2.33$\downarrow$)}  & 78.01 & 73.34\textcolor{red}{(4.67$\downarrow$)}   & 82.01 & 84.49 & 82.97\textcolor{red}{(1.52$\downarrow$)}  & 79.79 & 78.03\textcolor{red}{(1.76$\downarrow$)}  \\
		GECScore   & 56.35 & 50.95\textcolor{red}{(5.40$\downarrow$)}  & 16.42 & 5.63\textcolor{red}{(10.79$\downarrow$)}  & \textbf{96.27} & 65.92 & 51.89\textcolor{red}{(14.03$\downarrow$)}  & 59.57 & 53.06\textcolor{red}{(6.51$\downarrow$)}  \\
		\midrule
		metric.avg & 70.21 & 65.97\textcolor{red}{(4.24$\downarrow$)}  & 56.67 & 48.20\textcolor{red}{(8.47$\downarrow$)}   & 83.74 & 73.40 & 65.93\textcolor{red}{(7.47$\downarrow$)}  & 71.62 & 66.33\textcolor{red}{(5.29$\downarrow$)}  \\
		\midrule
		avg        & 65.08 & 60.59\textcolor{red}{(4.50$\downarrow$)}  & 54.09 & 45.10\textcolor{red}{(8.99$\downarrow$)}  & 76.07 & 71.40 & 63.27\textcolor{red}{(8.13$\downarrow$)}  & 67.41 & 62.70\textcolor{red}{(4.71$\downarrow$)}  \\
		\bottomrule
	\end{tabular}}
	\caption{Attack performance against evaluation detectors on the MAGA split. The "\xmark" symbol in the "Alignment" column denotes the unenhanced MGB split, whereas "\cmark" represents the MAGA split. ACC, TPR, and TNR are evaluated based on the default threshold of each detector. Labels M. and N. are introduced to facilitate understanding. TPR(M.) denotes the ACC exclusively on MGT, and TNR(H.) denotes the ACC exclusively on HWT. The highest values are bolded, and the second highest values are underlined.}
	\label{tab:MAGA Bench}
\end{table*}

A detailed theoretical analysis and introduction of all alignment methods we used, along with their specific execution, are discussed in Appendix \ref{sec:alignment method details}.

\paragraph{Roleplaying} To facilitate roleplaying, we construct a custom prompt set named MAGA-ROLE-80, from which we randomly sample a roleplaying prompt as the system prompt for each original prompt. To the best of our knowledge, among all machine-generated text detection benchmarks listed in Table~\ref{tab:all-datasets}, our work is the only one that leverages system prompts. Our roleplaying prompts incorporate the core philosophy of "Talk Less, Call Right" from RRP~\cite{ruangtanusak2025talklessrightenhancing}, where the first sentence defines the baseline identity and the second sentence specifies the linguistic style.

\paragraph{BPO} \cite{cheng2024black}: BPO is a prompt optimization model. Its optimization aims to ensure that the texts generated by BPO-optimized prompts are better aligned with HWTs. The BPO model always appends additional information to the original prompt. In this work, we directly employ the BPO model released by \citet{cheng2024black}\footnote{\url{https://huggingface.co/zai-org/BPO}}.

\paragraph{Self-Refine} Self Refine\footnote{Although Self Refine supports multi-round iterations, we adopt a single-round execution in our practical implementation, the rationale for which is discussed in Appendix~\ref{sec:selfrefineround}.} denotes an iterative refinement procedure wherein a large language model generates an initial text, reflects upon its own output to provide feedback, and subsequently rewrites the content based on this feedback. The specific prompts deployed for this self-refinement process are provided in Tables \ref{tab:self refine prompts}, \ref{tab:self refine prompts Chinese}, and are discussed in detail in Appendix \ref{sec:prompt detail}.

\paragraph{GDARL-CDMMR} GDARL is the general term for existing GDARL algorithms such as RADAR \cite{hu2023radar} and HUMPA \cite{wang2025humanizing}, and it also represents the vanilla formulation of such approaches. We observe that the vanilla GDARL exhibits minimal change throughout training due to reward overfitting under identical data distributions\footnote{See Appendix \ref{sec:GDARLvs} for further discussion about this as well as why we did not utilize a specific GDARL framework and introduce a CMR component.}. To address this limitation, we introduce a \textbf{Cross-Mutual Reward (CMR) component}. This component incorporates the theoretical insights from HC-Var, which state that cross-distribution generalization primarily relies on the latent essential relative features between HWT and MGT \cite{xu-etal-2024-generalization}.To illustrate CMR, we consider the application of GDARL-CDMMR (Cross Domain\&Model Mutual Reward). Specifically, the text generation models (LLMs) and the target domains of the generated text are divided into two distinct groups, denoted as $GA$ and $GB$. In the adversarial round $t$, the LLMs in group $GA$ first generate MGTs, denoted as $M$. HWTs represented as $H$. The detector $\mathcal{D}$ within group $GA$ is trained with parameters $\phi$ according to the following objective function:

\begin{equation}
	\begin{split}
		\mathcal{J}_\mathcal{D}(\phi^{GA, t}, t) & = - \mathbb{E}_{x_{m}^{GA, t} \sim M^{GA, t}} \left[\log \mathcal{D}_{\phi}^{GA, t}(x_{m}^{GA, t})\right] - \\ & \mathbb{E}_{x_h^{GA} \sim H^{GA}} \left[\log\left(1 - \mathcal{D}_{\phi}^{GA, t}(x_h^{GA})\right)\right]
	\end{split}
\end{equation}

The detector in group $GB$ is trained in an identical manner. Subsequently, the detector from group $GB$ provides rewards for the text generated by the text generation models in group $GA$ to facilitate reinforcement learning. The reward function is formulated as follows:

\begin{equation}
	r^{GA, t}(x, \phi^{GB, t+1}) = 1 - \mathcal{D}_\phi^{GB, t+1}(x)
\end{equation}

The training process for the text generation models in group $GB$ follows the same procedure. Appendix \ref{sec:hyperparameters} provides a detailed discussion regarding the hyperparameter configurations and the final grouping strategy for GDARL-CDMMR.

\begin{table*}[!t]
	\resizebox{\linewidth}{!}{
		\centering
		% \small
		\begin{tabular}{c|c|c|c|c|c|c|c|c|c|c}
			\toprule
			& \multicolumn{2}{c|}{\textbf{S-M4}} 
			& \multicolumn{2}{c|}{\textbf{C-M4GT}}
			& \multicolumn{2}{c|}{\textbf{C-MAGE}}
			& \multicolumn{2}{c|}{\textbf{C-HC3}}
			& \multicolumn{2}{c}{\textbf{AVERAGE}}
			\\
			\toprule
			Metric & AUC & ACC(5\%) & AUC & ACC(5\%) & AUC & ACC(5\%) & AUC & ACC(5\%) & AUC & ACC(5\%) \\
			\midrule
			R-B GPT2        & 75.92     & 68.25       & 78.38      & 63.76       & \underline{76.40} & 55.19     & 98.45       & \underline{94.59}      & 82.29       & 70.45 \\
			R-L GPT2        & 70.44     & 63.98       & 74.02      & 57.04       & \textbf{79.55} & 55.98     & 94.88       & 89.34      & 79.72       & 66.59 \\
			R-B CGPT        & 81.25     & 74.97       & 74.82      & 58.61       & 61.32 & 47.89     & 99.99\textsuperscript{*}       & 99.27\textsuperscript{*}      & 79.35       & 70.19 \\
			RADAR           & 76.25     & 55.46       & 70.89      & 31.13       & 58.43 & 45.08     & 90.16       & 84.74      & 73.93       & 54.10 \\
			SCRN           & 79.37     & 58.13       & \textbf{93.49}      & 33.73       & 66.24 & 48.01     & 93.49       & 88.77      & 83.15       & 57.16 \\
			DETree\textsuperscript{*} & 99.78\textsuperscript{*} & 99.32\textsuperscript{*} & 96.63\textsuperscript{*} & 94.56\textsuperscript{*} & 99.44\textsuperscript{*} & 96.82\textsuperscript{*} & 98.55\textsuperscript{*} & 95.27\textsuperscript{*} & 98.60\textsuperscript{*} & 96.49\textsuperscript{*} \\
			\midrule
			GLTR           & 80.19     & 72.23       & 78.29      & 63.13       & 58.34 & 46.78     & 92.37       & 91.13      & 77.30       & 68.32 \\
			F-DetectGPT     & 88.51     & 85.99       & 86.76      & \underline{79.65}       & 64.67 & 56.67     & \textbf{99.94}       & \textbf{97.19}      & 84.97       & \underline{79.88} \\
			LLMDet          & 70.28     & 50.17       & 74.89      & 53.16       & 50.13 & 47.27     & 90.23       & 81.34      & 71.38       & 57.99 \\
			Binoculars      & 89.73     & 89.09       & \underline{89.14}      & \textbf{84.10}       & 63.79 & \textbf{58.89}     & 99.59       & \textbf{97.19}      & \underline{85.56}       & \textbf{82.32} \\
			DALD      & 88.15     & 85.01       & 85.41      & 78.87       & 64.96 & \underline{56.73}     & \underline{99.76}       & \textbf{97.19}      & 84.57       & 79.45 \\
			GECSCORE      & 89.05     & 68.81       & 65.88      & 50.00       & 53.67 & 40.00     & 90.18       & 86.96      & 74.70       & 61.44 \\
			\midrule
			\textbf{R-B MGB (Ours)}   & \underline{94.63}     & \underline{90.58}       & 82.98      & 73.37       & 60.90 & 46.95     & 93.14       & 91.68      & 82.91       & 75.65 \\
			\textbf{R-B MAGA (Ours)}  & \textbf{98.87} & \textbf{95.41} & 88.40 & 78.99 & 64.63 & 47.67     & 98.14       & 94.06      & \textbf{87.51}       & 79.03 \\
			\textbf{$\Delta$} & \textbf{4.24} & \textbf{4.83} & \textbf{5.42} & \textbf{5.62} &\textbf{3.73} & \textbf{0.72} & \textbf{5.00} & \textbf{2.38} & \textbf{4.60} &
			\textbf{3.39} \\
			\bottomrule
	\end{tabular}}
	\caption{Experiments evaluating generalization capability. Performance of evaluation detectors as baseline alongside our R-B MGB and R-B MAGA on external datasets. R-B CGPT is fine-tuned on HC3 and is marked with an asterisk. DETree utilizes reference vectors constructed from dataset texts including M4, M4GT, and MAGE for detection, and is also marked with an asterisk. The highest values excluding those with asterisks are bolded, and the second highest values are underlined. ACC(5\%) is an abbreviation for ACC(@FPR=5\%).}
	\label{tab:MAGA Generalization}
\end{table*}

\subsection{MAGA Pipeline}

Figure \ref{fig:MAGA Pipeline} illustrates the MAGA pipeline for constructing the MAGA split, which integrates all four methods to elucidate our data construction process. In the first stage, we construct the prompt set utilizing roleplaying and BPO. This prompt set is subsequently applied to the GDARL-CDMMR process in the second stage, followed by self-refining in the third stage. In the final stage, the HWT and MGT are merged on a one-to-one basis to produce the definitive MAGA split.

\begin{table*}[!t]
	\resizebox{\linewidth}{!}{
		% \small
		\centering
		\begin{tabular}{l|c|c|c|c|c|c}
			\toprule
			& \multicolumn{3}{c|}{\textbf{Attack}} 
			& \multicolumn{3}{c}{\textbf{Generalization}}  \\
			\toprule
			Metric & TPR & AUC & ACC(@FPR=5\%) & TPR & AUC & ACC(@FPR=5\%) \\
			\midrule
			MGB & 54.09 & 71.4 & 67.41 & 53.13 & 82.91 & 75.65 \\
			\midrule
			MAGA-extra-GDARL-CDMMR & 49.95\textcolor{red}{(4.14$\downarrow$)} & 67.91\textcolor{red}{(3.49$\downarrow$)} & 65.18\textcolor{red}{(2.23$\downarrow$)} & 60.97\textcolor{green}{(7.84$\uparrow$)} & 85.02\textcolor{green}{(2.11$\uparrow$)} & 77.28\textcolor{green}{(1.63$\uparrow$)} \\
			MAGA-extra-roleplaying & 48.74\textcolor{red}{(5.35$\downarrow$)} & 66.43\textcolor{red}{(4.97$\downarrow$)} & 64.72\textcolor{red}{(2.69$\downarrow$)} & 63.47\textcolor{green}{(10.34$\uparrow$)} & 85.39\textcolor{green}{(2.48$\uparrow$)} & 77.69\textcolor{green}{(2.04$\uparrow$)} \\
			MAGA-extra-BPO & 50.12\textcolor{red}{(3.97$\downarrow$)} & 68.09\textcolor{red}{(3.31$\downarrow$)} & 65.49\textcolor{red}{(1.92$\downarrow$)} & 60.24\textcolor{green}{(7.11$\uparrow$)} & 84.87\textcolor{green}{(1.96$\uparrow$)} & 77.16\textcolor{green}{(1.51$\uparrow$)} \\
			MAGA-extra-self-refine & 48.87\textcolor{red}{(5.22$\downarrow$)} & 66.32\textcolor{red}{(5.08$\downarrow$)} & 64.43\textcolor{red}{(2.98$\downarrow$)} & 64.21\textcolor{green}{(11.08$\uparrow$)} & 85.84\textcolor{green}{(2.93$\uparrow$)} & 77.47\textcolor{green}{(1.82$\uparrow$)} \\
			\midrule
			MAGA & 45.10\textcolor{red}{(8.99$\downarrow$)} & 63.27\textcolor{red}{(8.13$\downarrow$)} & 62.70\textcolor{red}{(4.71$\downarrow$)} & 71.30\textcolor{green}{(18.17$\uparrow$)} & 87.51\textcolor{green}{(4.60$\uparrow$)} & 79.03\textcolor{green}{(3.38$\uparrow$)} \\
			\bottomrule
	\end{tabular}}
	\caption{Experimental results of attack performance and generalization capabilities on the MAGA-extra splits. "Attack" represents the average performance of evaluation detectors under the experimental setup specified in Table \ref{tab:MAGA Bench}. "Generalization" represents the average generalization performance on external datasets of the RoBERTa fine-tuned on the corresponding split, following the identical experimental setup as Table \ref{tab:MAGA Generalization}. This analysis also serves as an ablation study, with more comprehensive details provided in Table \ref{tab:full ablation}, Appendix \ref{sec:more experiment}.}
	\label{tab:ablation}
\end{table*}

\section{Experiments}

\subsection{Experimental Setup}

\paragraph{Evaluation Detectors} Our selection of open-source Evaluation Detectors extends the choices by RAID \cite{dugan2024raid} by incorporating four more recent detectors. The Evaluation Detectors we selected are listed in Figure {fig:MAGA Overview}, where SCRN, DETree, DALD, and GECScore are the four recent detectors. A detailed introduction to these detectors is provided in Appendix \ref{sec:detector detail}.

\paragraph{Evaluation Metrics} Our evaluation spans accuracy (ACC), True Positive Rate (TPR), True Negative Rate (TNR), Area Under the ROC Curve (AUC), and ACC(@FPR=5\%), with False Positive Rate denoted as FPR. Setting the MGT label to 1, TPR and TNR signify the standalone accuracies on MGT and HWT, respectively. ACC, TPR, and TNR are assessed under default probability thresholds (e.g., 0.5 for RoBERTa; see Table \ref{tab:detector_thresholds} for the full list). ACC(@FPR=5\%) is the metric adopted by RAID \cite{dugan2024raid} and denotes the detection accuracy evaluated at a specific threshold, which is determined by fixing the ACC of HWT to 95\%. Given that the default threshold is not always appropriate, we tend to focus more on AUC and ACC(@FPR=5\%).

\begin{figure}[!t]
	%	\vspace{-0.6cm}
	\centering
	\begin{minipage}{0.48\linewidth}
		\centering
		\includegraphics[width=1.0\linewidth]{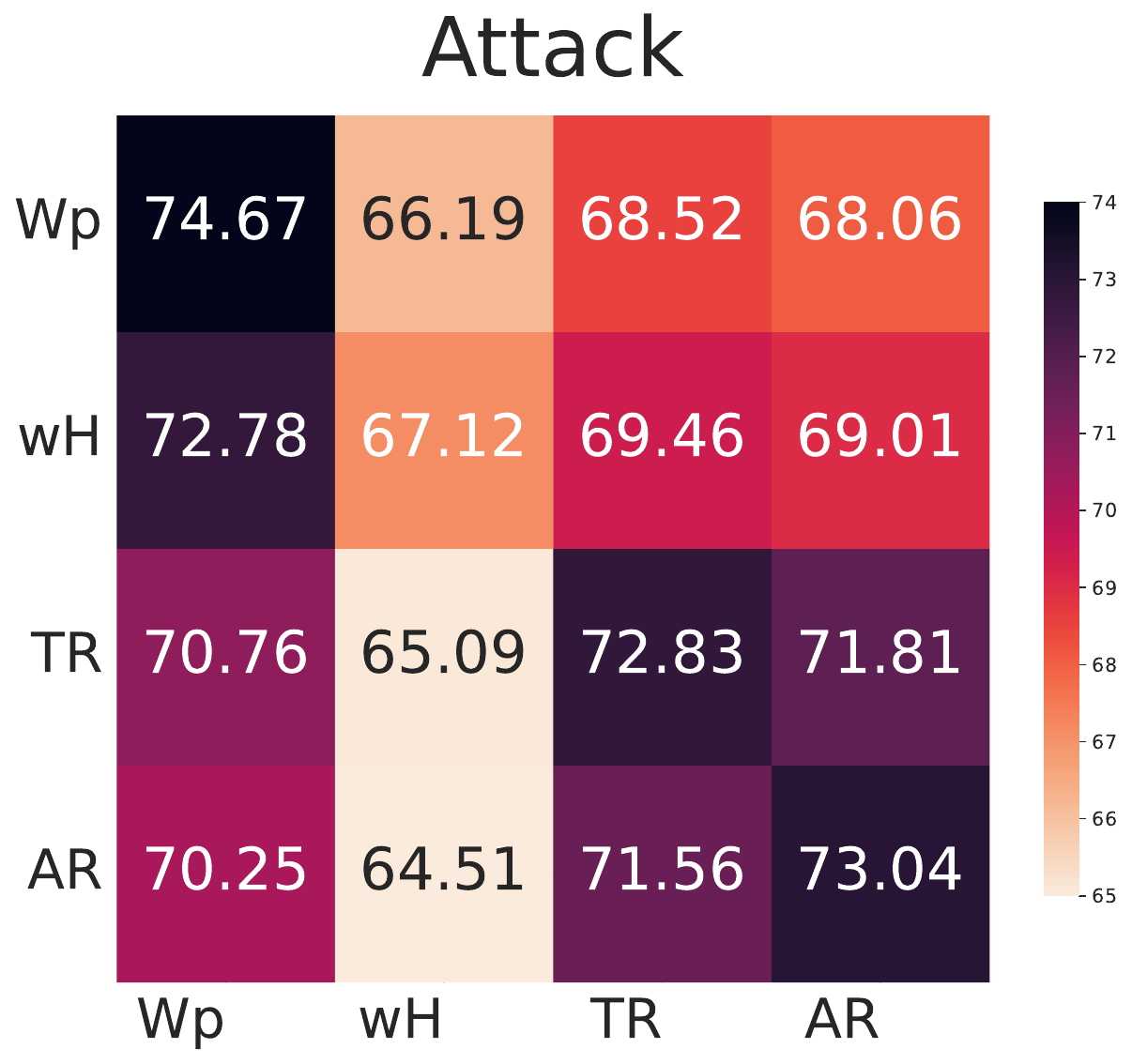}
	\end{minipage}
	\begin{minipage}{0.48\linewidth}
		\centering
		\includegraphics[width=1.0\linewidth]{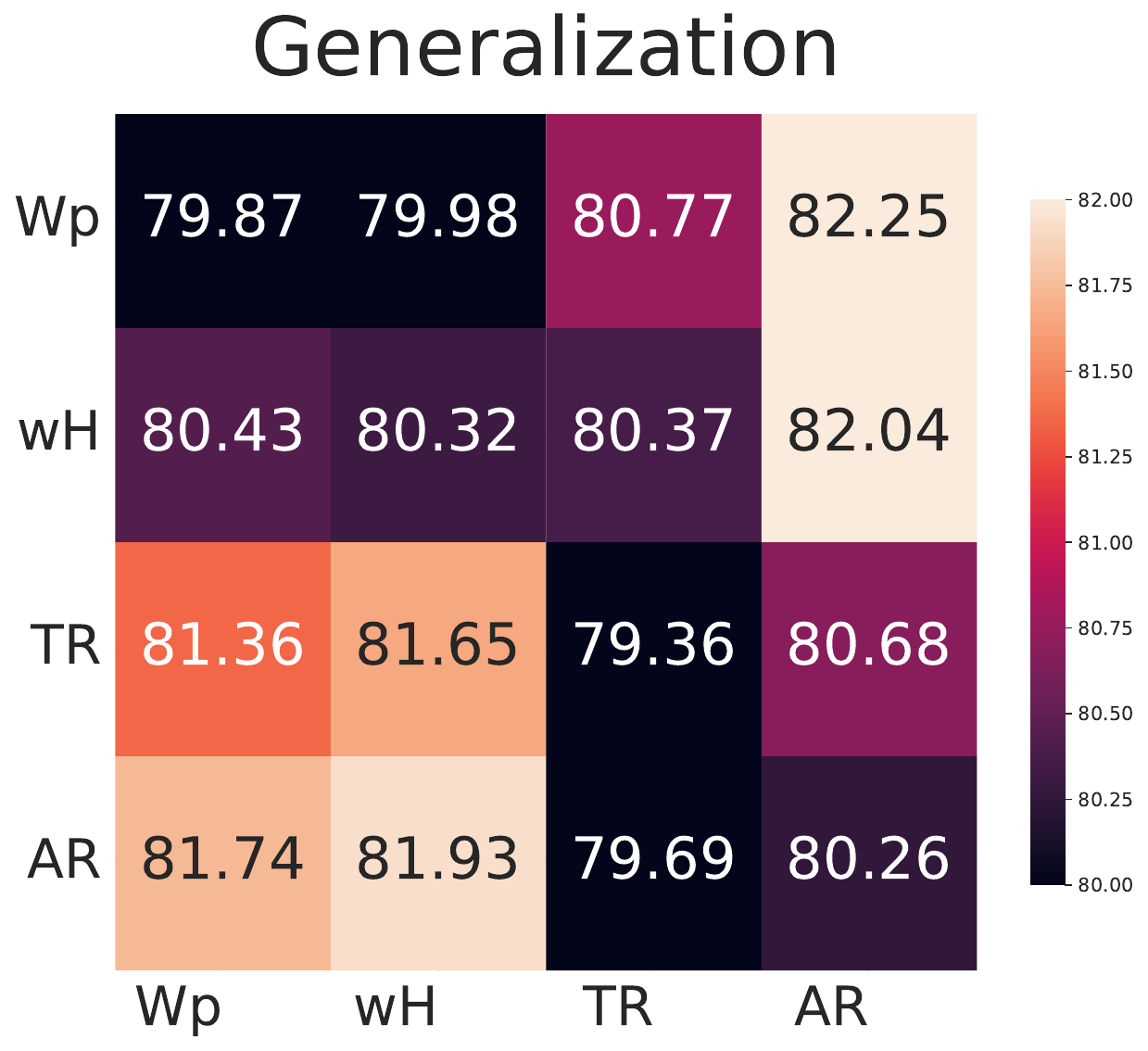}
	\end{minipage}
	\caption{Matrix analysis of GDARL-CDMR for Attack and Generalization (meanings are consistent with Table \ref{tab:ablation}) AUC performance. The horizontal axis denotes the domain of the text used to train the reward detector, while the vertical axis denotes the LLM generation domain. The abbreviations Wp, wH, TR, and AR correspond to Wikipedia, wikiHow, Trustpilot Reviews, and Amazon Reviews, respectively.}
	\label{fig:GDARL-CDMR-matrix-core}
	%	\vspace{-0.5cm}
\end{figure}

\paragraph{External Datasets} We incorporate workshop competition versions\footnote{See Appendix \ref {introduction to external datasets} for detailed introduction to External Datasets.} of four datasets: M4 \cite{wang-etal-2024-m4}, M4GT \cite{wang2024m4gt}, MAGE \cite{li2024mage}, and HC3 \cite{guo2023close}, detailed in Appendix \ref{introduction to external datasets}. For brevity, these are denoted as S-M4 \cite{wang2024semeval}, C-M4GT, C-MAGE, and C-HC3 \cite{wang-etal-2025-genai}, respectively.

\subsection{Main Experiment Results}

\begin{figure}[!t]
	%	\vspace{-0.6cm}
	\centering
	\includegraphics[width=0.8\linewidth]{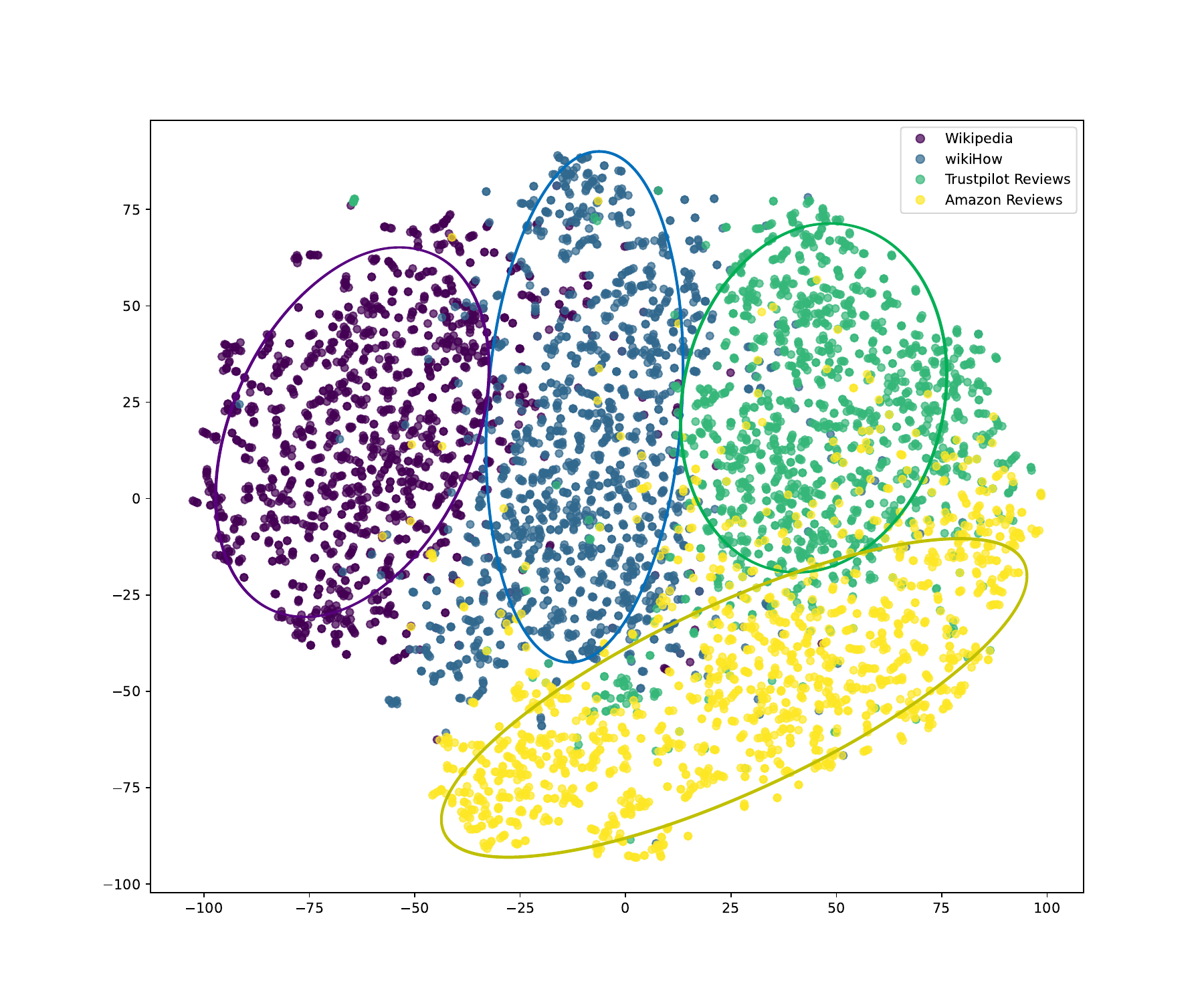}
	\caption{T-SNE visualization of HWTs on the domains specified in Figure~\ref{fig:GDARL-CDMR-matrix-core}.}
	\label{fig:hwt-domain-core}
	%	\vspace{-0.5cm}
\end{figure}

Table \ref{tab:MAGA Bench} presents the empirical results of our Attack experiments. Compared to the MGB split, the Evaluation Detectors on the MAGA split exhibit an average decrease of 8.13\% in AUC and 4.71\% in ACC(@FPR=5\%). The average performance degradation identified in neural-based detectors tightly aligns with that of metric-based detectors. This consistent reduction indicates that our MAGA relatively achieves a comprehensive and stable attack effect across different detection mechanisms rather than showing a biased preference toward a specific category, such as rule-based random text perturbations.

Table \ref{tab:MAGA Generalization} outlines the results of our generalization assessment. On External Datasets, the R-B MAGA delivers an average improvement of 4.60\% in AUC and 3.39\% in ACC(@FPR=5\%) compared to the baseline R-B MGB.

Furthermore, the results on the Chinese MAGA-cn subset are similar\footnote{See Appendix \ref{sec:more experiment} for experiments on Chinese subset and other experiment details.}.

\begin{figure}[!t]
	%	\vspace{-0.6cm}
	\centering
	\includegraphics[width=1.0\linewidth]{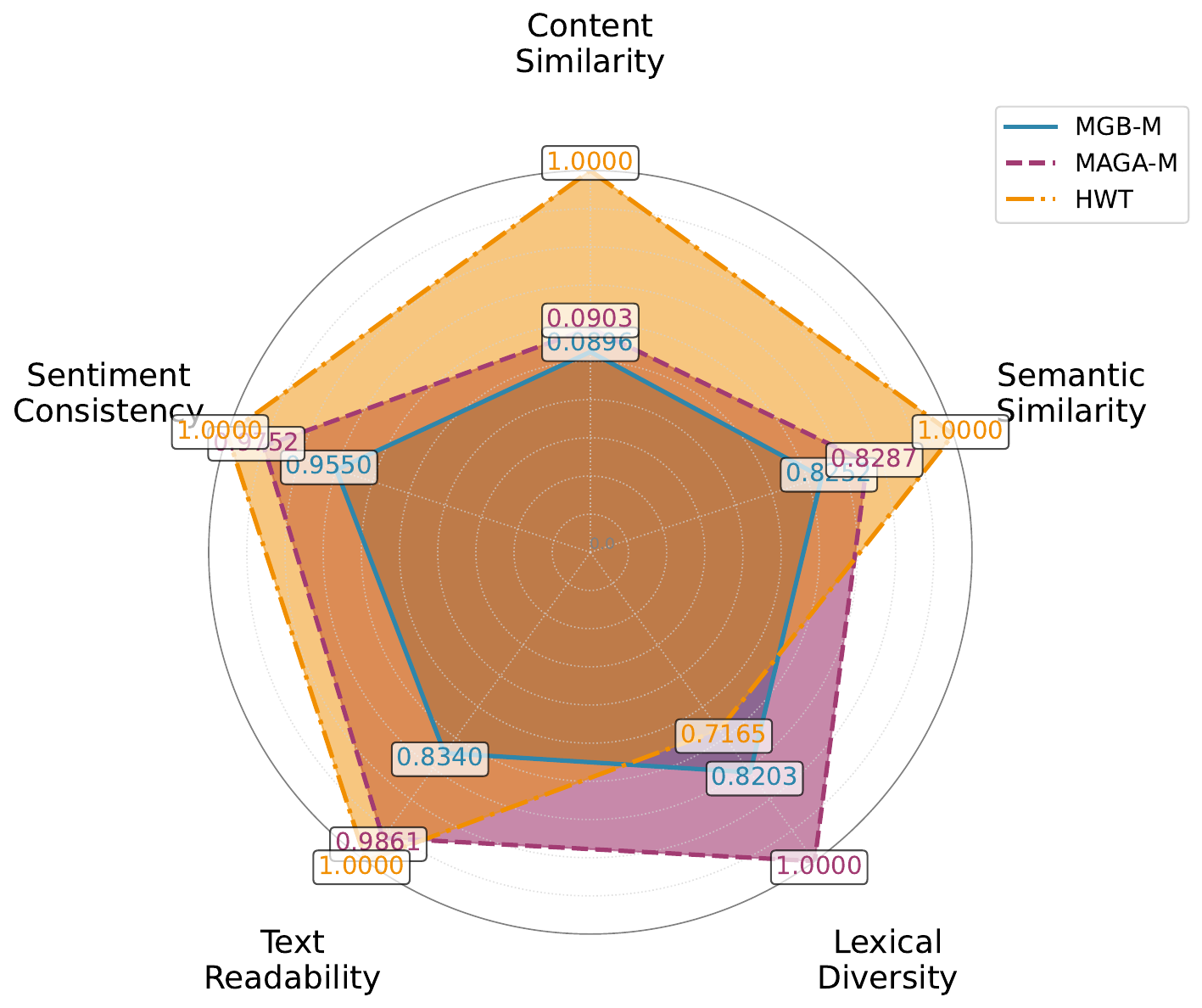}
	\caption{Multidimensional Text Analysis. MGB-M and MAGA-M represent MGTs on the MGB split and MAGA split, respectively. Detailed metrics and specific values ​​for each dimension are presented in Table \ref{tab:multidimensional_analysis} and Appendix \ref{sec:multi analysis detail}.}
	\label{fig:multi-dimension-radar}
	%	\vspace{-0.5cm}
\end{figure}

\subsection{Ablation Study}
\label{sec:ablation}

Table \ref{tab:ablation} presents additional experimental results across various MAGA-extra splits, which also serves as a ablation study to validate the effectiveness of each alignment method. Detailed results are provided in Appendix \ref{sec:more experiment}. Figure \ref{fig:GDARL-CDMR-matrix-core} illustrates the ablation study regarding the grouping configurations of GDARL-CDMR, where the main diagonal represents the performance prior to augmentation. As indicated in Table \ref{tab:auc_w_wo_gdarl}, vanilla GDARL yields nearly no change in performance due to reward overfitting. Figures \ref{fig:GDARL-CDMR-matrix-core} and \ref{fig:hwt-domain-core} demonstrate that larger domain discrepancies lead to more substantial performance gains for GDARL-CDMR\footnote{A more detailed analysis is in Figures \ref{fig:GDARL-CDMR-matrix}, \ref{fig:GDARL-CMMR-matrix}, and \ref{fig:all-tsne}, which are discussed in Appendix \ref{sec:more experiment}.}. These observations demonstrate the significance and efficacy of the CMR component.

\section{Analysis}
\label{sec:analysis}

We consolidate several quantitative analysis metrics from prior literature regarding the discrepancies between MGT and HWT into five dimensions, as illustrated in Figure \ref{fig:multi-dimension-radar}. The value for each dimension represents the mean Normalize Score of all constituent metrics calculated across the entire corpus. Appendix \ref{sec:multi analysis detail} elaborates on the detailed mathematical formulations. 

\paragraph{Content Similarity and Semantic Similarity} HC-Var \cite{xu-etal-2024-generalization} demonstrates that more human-like MGT exhibits higher MAUVE \cite{NEURIPS2021_260c2432} similarity with HWT. Corresponding to MAUVE, we select BERTScore as the analysis metric for semantic similarity. Content similarity is evaluated concurrently via ROUGE, BLEU, and METEOR. The results indicate that MAGA-M achieves slightly higher content and semantic similarity scores with HWT than MGB-M.

\paragraph{Lexical Diversity} M4 \cite{wang-etal-2024-m4} reveals that utilizing multiple generators significantly increases the vocabulary size of unigrams and bigrams, thereby enriching the vocabulary of MGTs and facilitating the training and evaluation of MGT detectors. Furthermore, the unigram and bigram vocabulary size of MGT reported in M4 \cite{wang-etal-2024-m4} already exceeds that of HWT. Extending beyond bigram vocabulary size, we incorporate TTR and Yule's K metric. Our analysis shows that the lexical diversity of MGB-M surpasses that of HWT, and MAGA-M further amplifies this gap. It shows a divergence, although this aligns with the findings in M4 \cite{wang-etal-2024-m4} and may provide benefits.

\paragraph{Text Readability} AA \cite{uchendu-etal-2020-authorship} notes that the Flesch-Kincaid score of MGT is higher than that of HWT. In addition to Flesch-Kincaid, we incorporate SMOG and Dale-Chall into our analysis. The empirical results confirm that the Flesch-Kincaid score of HWT is significantly higher than that of MGB-M. MAGA-M substantially narrows this gap to HWT, demonstrating that its textual expressions are neither overly simplistic nor fragmented. On SMOG and Dale-Chall, MGB-M scores significantly higher than HWT, whereas MAGA-M exhibits a marked decrease, approaching HWT closely. This trend reflects a substantial reduction in obscure and difficult vocabulary.

\paragraph{Sentiment Consistency} HC3 \cite{guo2023close} conducts a comparative sentiment consistency analysis between HWT and MGT across negative, neutral, and positive categories. HC3 concludes that MGT is generally conservative, exhibiting a pronounced deficit in negative sentiment. Our experiments yield consistent findings.

\section{Conclusion}
We introduce MAGA-Bench to address generalization bottlenecks in MGT detection. Evaluations show that aligned MGTs within MAGA degrade existing detectors' AUC by 8.13\% on average. Conversely, fine-tuning on MAGA boosts detector robustness, improving generalization AUC across diverse external datasets by 4.60\%. We hope alignment augmentation serves as a promising path toward robust and generalized MGT detection.

\section*{Limitations}
There remain several limitations to our work: 1. while the combination of Self Refine with other methods proves effective, this integration is still insufficient, suggesting that the introduction of agentic reinforcement learning may be necessary. 2. MAGA-M exhibits a further and significant increase in lexical diversity for reasons that currently remain unclear, although this trend aligns to some extent with the findings reported in M4~\cite{wang-etal-2024-m4}. 3. a notable limitation in our experiments is that R-B MAGA yields a marginal improvement in ACC(@FPR=5\%) on C-MAGE. Clarifying the underlying cause of this limited gain could potentially advance research on the generalization of machine-generated text detection. 4. regarding multilingual evaluation, our current study only incorporates English and Chinese, leaving the efficacy of our method on other languages undetermined. 5. our focus primarily centers on vanilla MGT, without accounting for specialized text forms such as continued writing or paraphrased text.

\bibliography{custom}

\clearpage
\appendix

\startcontents[appendices]
\section*{Contents of Appendix}
\printcontents[appendices]{}{1}{\normalsize}

\section{Details and Statistics of MAGA Dataset}
\label{sec:statistics maga}

In the main paper, we primarily focus on the MAGA split, which represents the fully enhanced MAGA split of the English subset. This split is constructed via the MAGA pipeline integrated with all four alignment methods we adopted.

Our generalized MAGA dataset comprises two language-specific subsets, namely the English subset MAGA and the Chinese subset MAGA-cn. Specifically, the English MAGA subset contains six distinct splits: MGB, MAGA, MAGA-extra-GDARL-CDMMR, MAGA-extra-BPO, MAGA-extra-roleplaying, and MAGA-extra-self-refine. The MAGA-cn subset follows an identical structure with these six splits. Each split consists of a training subsplit and a validation subsplit. MGB serves as the baseline split constructed without any alignment methods, while each extra split is constructed using a single corresponding alignment technique.

Table \ref{tab:data_distribution_overall} presents the overall distribution of the generalized MAGA dataset. Each split contains 72,000 machine-generated texts and 72,000 human-written texts (60,000 training + 12,000 validation). It should be noted that the 72,000 human texts are reused across all splits and remain identical. The machine-generated texts vary across splits with one exception: in the MAGA-extra-GDARL-CDMMR split, since GDARL-CDMMR only applies to non-commercial large language models and does not integrate other alignment methods, the 36,000 machine texts generated by commercial models are sourced from MGB.

Our MAGA dataset maintains a high degree of balance. Table \ref{tab:data split distribution on train set} and Table \ref{tab:data split distribution on val set} detail the number of samples generated by each model within each domain. For instance, in a single training subsplit, each model generates 500 samples per domain.

Table \ref{tab:data schema} provides the detailed annotation schema for our MAGA dataset to ensure comprehensive documentation. Figure \ref{fig:sample-human} presents a HWT sample from MGB\_train.jsonl, which also appears across all data splits. Figures \ref{fig:sample-mgb}, \ref{fig:sample-bpo}, \ref{fig:sample-gdarl}, \ref{fig:sample-roleplaying}, \ref{fig:sample-self-refine}, \ref{fig:sample-maga} illustrate MGT examples from each respective split, all of which correspond directly to the same HWT shown in Figure \ref{fig:sample-human}. These examples facilitate a more intuitive understanding of the meaning of our annotations.

\begin{table*}[!t]
	\resizebox{\linewidth}{!}{
		\centering
		\setlength{\tabcolsep}{6pt} % 调整列之间的间距
		\renewcommand{\arraystretch}{1} % 调整行高，增加表格的空白，使其更易读
		% cyan（青色）、magenta（品红色 / 洋红色）、yellow（黄色）、black（黑色）
		\definecolor{darkgreen}{cmyk}{1, 0, 1, 0}
		\definecolor{darkred}{cmyk}{0, 1, 1, 0.02}
		
		\begin{tabular}{
				l| % 左对齐
				>{\centering\arraybackslash}m{2cm}| % 居中对齐，固定宽度
				>{\centering\arraybackslash}m{2cm}|
				>{\centering\arraybackslash}m{2cm}| 
				>{\centering\arraybackslash}m{2cm}| 
				>{\centering\arraybackslash}m{2cm}|
				>{\centering\arraybackslash}m{2cm}|
				>{\centering\arraybackslash}m{2cm}
			}
			
			\toprule
			
			\textbf{Dataset} & \textbf{Origin Size} & \textbf{Domain Coverage} & \textbf{Model Coverage} & \textbf{Multilingual Coverage} & \textbf{Adversarial Coverage} & \textbf{Sampling Coverage} & \textbf{Alignment Augment} \\ 
			
			\midrule
			
			TuringBench \cite{uchendu2021turingbench}&200k&\textcolor{darkred}{\xmark}&\textcolor{darkgreen}{\cmark}&\textcolor{darkred}{\xmark}&\textcolor{darkred}{\xmark}&\textcolor{darkred}{\xmark}&\textcolor{darkred}{\xmark}\\ 
			RuATD \cite{Shamardina_2022}&215k&\textcolor{darkgreen}{\cmark}&\textcolor{darkgreen}{\cmark}&\textcolor{darkred}{\xmark}&\textcolor{darkred}{\xmark}&\textcolor{darkred}{\xmark}&\textcolor{darkred}{\xmark}\\
			RoFT \cite{dugan-etal-2023-roft}&21k&\textcolor{darkgreen}{\cmark}&\textcolor{darkgreen}{\cmark}&\textcolor{darkred}{\xmark}&\textcolor{darkred}{\xmark}&\textcolor{darkred}{\xmark}&\textcolor{darkred}{\xmark}\\ 
			HC3 \cite{guo2023close}&26.9k&\textcolor{darkgreen}{\cmark}&\textcolor{darkred}{\xmark}&\textcolor{darkgreen}{\cmark}&\textcolor{darkred}{\xmark}&\textcolor{darkred}{\xmark}&\textcolor{darkred}{\xmark}\\ 
			MGTBench \cite{he2023mgtbench}&2.8k&\textcolor{darkgreen}{\cmark}&\textcolor{darkgreen}{\cmark}&\textcolor{darkred}{\xmark}&\textcolor{darkgreen}{\cmark}&\textcolor{darkred}{\xmark}&\textcolor{darkred}{\xmark}\\ 
			CHEAT \cite{yu2023cheat}&50k&\textcolor{darkred}{\xmark}&\textcolor{darkred}{\xmark}&\textcolor{darkred}{\xmark}&\textcolor{darkgreen}{\cmark}&\textcolor{darkred}{\xmark}&\textcolor{darkred}{\xmark}\\ 
			MULTITuDE \cite{macko-etal-2023-multitude}&74.1k&\textcolor{darkred}{\xmark}&\textcolor{darkgreen}{\cmark}&\textcolor{darkgreen}{\cmark}&\textcolor{darkred}{\xmark}&\textcolor{darkred}{\xmark}&\textcolor{darkred}{\xmark}\\ 
			OpenLLMText \cite{chen2023token}&340k&\textcolor{darkgreen}{\cmark}&\textcolor{darkgreen}{\cmark}&\textcolor{darkred}{\xmark}&\textcolor{darkred}{\xmark}&\textcolor{darkgreen}{\cmark}&\textcolor{darkred}{\xmark}\\ 
			AuText2023 \cite{sarvazyan2023overview}&160k&\textcolor{darkgreen}{\cmark}&\textcolor{darkred}{\xmark}&\textcolor{darkgreen}{\cmark}&\textcolor{darkred}{\xmark}&\textcolor{darkred}{\xmark}&\textcolor{darkred}{\xmark}\\ 
			CCD \cite{wang2023evaluating}&467k&\textcolor{darkred}{\xmark}&\textcolor{darkred}{\xmark}&\textcolor{darkgreen}{\cmark}&\textcolor{darkgreen}{\cmark}&\textcolor{darkred}{\xmark}&\textcolor{darkred}{\xmark}\\ 
			IMDGSP \cite{mosca-etal-2023-distinguishing}&29k&\textcolor{darkred}{\xmark}&\textcolor{darkgreen}{\cmark}&\textcolor{darkred}{\xmark}&\textcolor{darkred}{\xmark}&\textcolor{darkred}{\xmark}&\textcolor{darkred}{\xmark}\\ 
			M4 \cite{wang-etal-2024-m4}&122k&\textcolor{darkgreen}{\cmark}&\textcolor{darkgreen}{\cmark}&\textcolor{darkgreen}{\cmark}&\textcolor{darkred}{\xmark}&\textcolor{darkred}{\xmark}&\textcolor{darkred}{\xmark}\\ 
			HC-Var \cite{xu-etal-2024-generalization}&145k&\textcolor{darkgreen}{\cmark}&\textcolor{darkred}{\xmark}&\textcolor{darkred}{\xmark}&\textcolor{darkred}{\xmark}&\textcolor{darkred}{\xmark}&\textcolor{darkred}{\xmark}\\ 
			BUST \cite{cornelius2024bust}&25k&\textcolor{darkgreen}{\cmark}&\textcolor{darkgreen}{\cmark}&\textcolor{darkred}{\xmark}&\textcolor{darkred}{\xmark}&\textcolor{darkgreen}{\cmark}&\textcolor{darkred}{\xmark}\\ 
			M4GT-Bench \cite{wang2024m4gt}&217k&\textcolor{darkgreen}{\cmark}&\textcolor{darkgreen}{\cmark}&\textcolor{darkgreen}{\cmark}&\textcolor{darkred}{\xmark}&\textcolor{darkred}{\xmark}&\textcolor{darkred}{\xmark}\\ 
			LLM-DetectAIve \cite{abassy2024llm}&303k&\textcolor{darkgreen}{\cmark}&\textcolor{darkgreen}{\cmark}&\textcolor{darkred}{\xmark}&\textcolor{darkgreen}{\cmark}&\textcolor{darkred}{\xmark}&\textcolor{darkred}{\xmark}\\ 
			MixSet \cite{zhang2024llm}&3.6k&\textcolor{darkgreen}{\cmark}&\textcolor{darkgreen}{\cmark}&\textcolor{darkred}{\xmark}&\textcolor{darkgreen}{\cmark}&\textcolor{darkred}{\xmark}&\textcolor{darkred}{\xmark}\\ 
			HC3 Plus \cite{su2024hc3}&210k&\textcolor{darkgreen}{\cmark}&\textcolor{darkred}{\xmark}&\textcolor{darkgreen}{\cmark}&\textcolor{darkred}{\xmark}&\textcolor{darkred}{\xmark}&\textcolor{darkred}{\xmark}\\ 
			MAGE \cite{li2024mage}&447k&\textcolor{darkgreen}{\cmark}&\textcolor{darkgreen}{\cmark}&\textcolor{darkred}{\xmark}&\textcolor{darkred}{\xmark}&\textcolor{darkred}{\xmark}&\textcolor{darkred}{\xmark}\\ 
			RAID \cite{dugan2024raid}&570k&\textcolor{darkgreen}{\cmark}&\textcolor{darkgreen}{\cmark}&\textcolor{darkred}{\xmark}&\textcolor{darkgreen}{\cmark}&\textcolor{darkgreen}{\cmark}&\textcolor{darkred}{\xmark}\\
			LAMP \cite{chakrabarty2025can}&1k&\textcolor{darkgreen}{\cmark}&\textcolor{darkgreen}{\cmark}&\textcolor{darkred}{\xmark}&\textcolor{darkgreen}{\cmark}&\textcolor{darkred}{\xmark}&\textcolor{darkred}{\xmark}\\ 
			Beemo \cite{artemova2025beemo}&19.6k&\textcolor{darkgreen}{\cmark}&\textcolor{darkgreen}{\cmark}&\textcolor{darkred}{\xmark}&\textcolor{darkgreen}{\cmark}&\textcolor{darkred}{\xmark}&\textcolor{darkred}{\xmark}\\ 
			MultiSocial \cite{macko2025multisocial}&472k&\textcolor{darkgreen}{\cmark}&\textcolor{darkgreen}{\cmark}&\textcolor{darkgreen}{\cmark}&\textcolor{darkred}{\xmark}&\textcolor{darkred}{\xmark}&\textcolor{darkred}{\xmark}\\ 
			RealDet \cite{zhu2025reliably}&847k&\textcolor{darkgreen}{\cmark}&\textcolor{darkgreen}{\cmark}&\textcolor{darkgreen}{\cmark}&\textcolor{darkgreen}{\cmark}&\textcolor{darkred}{\xmark}&\textcolor{darkred}{\xmark}\\ 
			
			\midrule
			
			\textbf{MAGA (Ours)}&\textbf{936k}&\textcolor{darkgreen}{\cmark}&\textcolor{darkgreen}{\cmark}&\textcolor{darkgreen}{\cmark}&\textcolor{darkgreen}{\cmark}&\textcolor{darkgreen}{\cmark}&\textcolor{darkgreen}{\cmark}\\
			
			\bottomrule
			
	\end{tabular}}
	\caption{A more detailed comparison of the publicly available sources of machine-generated text.}
	\label{tab:all-datasets}
	\vspace{-4pt}
\end{table*}

\begin{table*}[!t]
	\resizebox{\linewidth}{!}{
		\centering
		\begin{tabular}{c|c|c|c|c|c|c|c|c|c|c}
			\toprule
			\multirow{2}{*}{Subset (Language)} & \multirow{2}{*}{Split} & \multicolumn{3}{c|}{Training Subsplit} & \multicolumn{3}{c|}{Validation Subsplit} & \multicolumn{3}{c}{\textbf{Total}} \\
			\cline{3-11}
			& & Machine & Human & Total & Machine & Human & Total & Machine & Human & Total \\
			\toprule
			\multirow{7}{*}{\makecell{MAGA (English)}} 
			& MGB & 60k & 60k & 120k & 12k & 12k & 24k & 72k & 72k & 144k \\
			& MAGA & 60k & 60k & 120k & 12k & 12k & 24k & 72k & 72k & 144k \\
			& MAGA-extra-GDARL-CDMMR & 60k & 60k & 120k & 12k & 12k & 24k & 72k & 72k & 144k \\
			& MAGA-extra-BPO & 60k & 60k & 120k & 12k & 12k & 24k & 72k & 72k & 144k \\
			& MAGA-extra-roleplaying & 60k & 60k & 120k & 12k & 12k & 24k & 72k & 72k & 144k \\
			& MAGA-extra-self-refine & 60k & 60k & 120k & 12k & 12k & 24k & 72k & 72k & 144k \\
			\cmidrule(lr){2-11}
			& \textbf{Grand Total} & 330k & 60k & 390k & 66k & 12k & 78k & 396k & 72k & 468k \\
			\midrule
			\multirow{7}{*}{\makecell{MAGA-cn (Chinese)}} 
			& MGB-cn & 60k & 60k & 120k & 12k & 12k & 24k & 72k & 72k & 144k \\
			& MAGA-cn & 60k & 60k & 120k & 12k & 12k & 24k & 72k & 72k & 144k \\
			& MAGA-cn-extra-GDARL-CDMMR & 60k & 60k & 120k & 12k & 12k & 24k & 72k & 72k & 144k \\
			& MAGA-cn-extra-BPO & 60k & 60k & 120k & 12k & 12k & 24k & 72k & 72k & 144k \\
			& MAGA-cn-extra-roleplaying & 60k & 60k & 120k & 12k & 12k & 24k & 72k & 72k & 144k \\
			& MAGA-cn-extra-self-refine & 60k & 60k & 120k & 12k & 12k & 24k & 72k & 72k & 144k \\
			\cmidrule(lr){2-11}
			& \textbf{Grand Total} & 330k & 60k & 390k & 66k & 12k & 78k & 396k & 72k & 468k \\
			\midrule
			\multicolumn{2}{c|}{\textbf{Grand Total}} & 660k & 120k & 780k & 132k & 24k & 156k & 792k & 144k & \textbf{936k} \\
			\bottomrule
		\end{tabular}
	}
	\caption{Overall Instance Distribution of MAGA Dataset by Language and Split (MGB, MAGA, MAGA-extra Alignment Methods). The table summarizes sample counts across training and validation sets, including machine-generated, human-annotated and aggregated totals, with notes that human samples are shared across all splits and partial machine data of MAGA-extra-GDARL-CDMMR is derived from MGB.}
	\label{tab:data_distribution_overall}
\end{table*}

\begin{table*}[!t]
	\resizebox{\linewidth}{!}{
		\centering
		\begin{tabular}{c|c|c|c|c|c|c|c|c|c|c|c|c|c|c}
			\toprule
			Domain & \makecell{GPT-4o-mini} & \makecell{Gemini-2.0-flash} & \makecell{DeepSeek-V3} & \makecell{Qwen3-plus} & \makecell{Mistral-Medium} & \makecell{Hunyuan-TurboS} & \makecell{Llama-3.1-8B\\Instruct} & \makecell{gemma-3-12b-it} & \makecell{DeepSeek-R1-0528\\Qwen3-8B} & \makecell{Qwen3-8B} & \makecell{Ministral-8B\\Instruct-2410} & \makecell{Hunyuan-7B\\Instruct} & \makecell{\textbf{Machine.Total}} & \makecell{\textbf{Human}} \\
			\toprule
			Reddit & 500 & 500 & 500 & 500 & 500 & 500 & 500 & 500 & 500 & 500 & 500 & 500 & 6000 & 6000 \\
			S2ORC & 500 & 500 & 500 & 500 & 500 & 500 & 500 & 500 & 500 & 500 & 500 & 500 & 6000 & 6000 \\
			Wikipedia & 500 & 500 & 500 & 500 & 500 & 500 & 500 & 500 & 500 & 500 & 500 & 500 & 6000 & 6000 \\
			wikiHow & 500 & 500 & 500 & 500 & 500 & 500 & 500 & 500 & 500 & 500 & 500 & 500 & 6000 & 6000 \\
			Trustpilot Reviews & 500 & 500 & 500 & 500 & 500 & 500 & 500 & 500 & 500 & 500 & 500 & 500 & 6000 & 6000 \\
			Amazon Reviews & 500 & 500 & 500 & 500 & 500 & 500 & 500 & 500 & 500 & 500 & 500 & 500 & 6000 & 6000 \\
			Yahoo Answers & 500 & 500 & 500 & 500 & 500 & 500 & 500 & 500 & 500 & 500 & 500 & 500 & 6000 & 6000 \\
			Natural Questions & 500 & 500 & 500 & 500 & 500 & 500 & 500 & 500 & 500 & 500 & 500 & 500 & 6000 & 6000 \\
			CC News & 500 & 500 & 500 & 500 & 500 & 500 & 500 & 500 & 500 & 500 & 500 & 500 & 6000 & 6000 \\
			NPR News & 500 & 500 & 500 & 500 & 500 & 500 & 500 & 500 & 500 & 500 & 500 & 500 & 6000 & 6000 \\
			\midrule
			\textbf{EN.Total} & 5000 & 5000 & 5000 & 5000 & 5000 & 5000 & 5000 & 5000 & 5000 & 5000 & 5000 & 5000 & 60000 & 60000 \\
			\midrule
			Baidu Tieba (百度贴吧) & 500 & 500 & 500 & 500 & 500 & 500 & 500 & 500 & 500 & 500 & 500 & 500 & 6000 & 6000 \\
			Weibo Review (微博评论) & 500 & 500 & 500 & 500 & 500 & 500 & 500 & 500 & 500 & 500 & 500 & 500 & 6000 & 6000 \\
			Rednote Review (小红书评论) & 500 & 500 & 500 & 500 & 500 & 500 & 500 & 500 & 500 & 500 & 500 & 500 & 6000 & 6000 \\
			CSL (中文核心期刊摘要) & 500 & 500 & 500 & 500 & 500 & 500 & 500 & 500 & 500 & 500 & 500 & 500 & 6000 & 6000 \\
			Baidu Baike (百度百科) & 500 & 500 & 500 & 500 & 500 & 500 & 500 & 500 & 500 & 500 & 500 & 500 & 6000 & 6000 \\
			Dianping (大众点评) & 500 & 500 & 500 & 500 & 500 & 500 & 500 & 500 & 500 & 500 & 500 & 500 & 6000 & 6000 \\
			Douban Review (豆瓣评论) & 500 & 500 & 500 & 500 & 500 & 500 & 500 & 500 & 500 & 500 & 500 & 500 & 6000 & 6000 \\
			Baidu Zhidao (百度知道) & 500 & 500 & 500 & 500 & 500 & 500 & 500 & 500 & 500 & 500 & 500 & 500 & 6000 & 6000 \\
			Zhihu (知乎) & 500 & 500 & 500 & 500 & 500 & 500 & 500 & 500 & 500 & 500 & 500 & 500 & 6000 & 6000 \\
			CLTS (澎湃新闻) & 500 & 500 & 500 & 500 & 500 & 500 & 500 & 500 & 500 & 500 & 500 & 500 & 6000 & 6000 \\
			\midrule
			\textbf{CN.Total} & 5000 & 5000 & 5000 & 5000 & 5000 & 5000 & 5000 & 5000 & 5000 & 5000 & 5000 & 5000 & 60000 & 60000 \\
			\bottomrule
		\end{tabular}
	}
	\caption{Training Set Instance Distribution of MAGA Dataset Splits (MGB, MAGA and every split of MAGA-extra). The table summarizes sample counts across English and Chinese domains, including per-model samples, machine sample totals and human samples, with EN.Total and CN.Total for aggregated domain counts.}
	\label{tab:data split distribution on train set}
\end{table*}

\begin{table*}[!t]
	\resizebox{\linewidth}{!}{
		\centering
		\begin{tabular}{c|c|c|c|c|c|c|c|c|c|c|c|c|c|c}
			\toprule
			Domain & \makecell{GPT-4o-mini} & \makecell{Gemini-2.0-flash} & \makecell{DeepSeek-V3} & \makecell{Qwen3-plus} & \makecell{Mistral-Medium} & \makecell{Hunyuan-TurboS} & \makecell{Llama-3.1-8B\\Instruct} & \makecell{gemma-3-12b-it} & \makecell{DeepSeek-R1-0528\\Qwen3-8B} & \makecell{Qwen3-8B} & \makecell{Ministral-8B\\Instruct-2410} & \makecell{Hunyuan-7B\\Instruct} & \makecell{\textbf{Machine.Total}} & \makecell{\textbf{Human}} \\
			\toprule
			Reddit & 100 & 100 & 100 & 100 & 100 & 100 & 100 & 100 & 100 & 100 & 100 & 100 & 1200 & 1200 \\
			S2ORC & 100 & 100 & 100 & 100 & 100 & 100 & 100 & 100 & 100 & 100 & 100 & 100 & 1200 & 1200 \\
			Wikipedia & 100 & 100 & 100 & 100 & 100 & 100 & 100 & 100 & 100 & 100 & 100 & 100 & 1200 & 1200 \\
			wikiHow & 100 & 100 & 100 & 100 & 100 & 100 & 100 & 100 & 100 & 100 & 100 & 100 & 1200 & 1200 \\
			Trustpilot Reviews & 100 & 100 & 100 & 100 & 100 & 100 & 100 & 100 & 100 & 100 & 100 & 100 & 1200 & 1200 \\
			Amazon Reviews & 100 & 100 & 100 & 100 & 100 & 100 & 100 & 100 & 100 & 100 & 100 & 100 & 1200 & 1200 \\
			Yahoo Answers & 100 & 100 & 100 & 100 & 100 & 100 & 100 & 100 & 100 & 100 & 100 & 100 & 1200 & 1200 \\
			Natural Questions & 100 & 100 & 100 & 100 & 100 & 100 & 100 & 100 & 100 & 100 & 100 & 100 & 1200 & 1200 \\
			CC News & 100 & 100 & 100 & 100 & 100 & 100 & 100 & 100 & 100 & 100 & 100 & 100 & 1200 & 1200 \\
			NPR News & 100 & 100 & 100 & 100 & 100 & 100 & 100 & 100 & 100 & 100 & 100 & 100 & 1200 & 1200 \\
			\midrule
			\textbf{EN.Total} & 1000 & 1000 & 1000 & 1000 & 1000 & 1000 & 1000 & 1000 & 1000 & 1000 & 1000 & 1000 & 12000 & 12000 \\
			\midrule
			Baidu Tieba (百度贴吧) & 100 & 100 & 100 & 100 & 100 & 100 & 100 & 100 & 100 & 100 & 100 & 100 & 1200 & 1200 \\
			Weibo Review（微博评论） & 100 & 100 & 100 & 100 & 100 & 100 & 100 & 100 & 100 & 100 & 100 & 100 & 1200 & 1200 \\
			Rednote Review（小红书评论） & 100 & 100 & 100 & 100 & 100 & 100 & 100 & 100 & 100 & 100 & 100 & 100 & 1200 & 1200 \\
			CSL（中文核心期刊摘要） & 100 & 100 & 100 & 100 & 100 & 100 & 100 & 100 & 100 & 100 & 100 & 100 & 1200 & 1200 \\
			Baidu Baike (百度百科) & 100 & 100 & 100 & 100 & 100 & 100 & 100 & 100 & 100 & 100 & 100 & 100 & 1200 & 1200 \\
			Dianping（大众点评） & 100 & 100 & 100 & 100 & 100 & 100 & 100 & 100 & 100 & 100 & 100 & 100 & 1200 & 1200 \\
			Douban Review (豆瓣评论) & 100 & 100 & 100 & 100 & 100 & 100 & 100 & 100 & 100 & 100 & 100 & 100 & 1200 & 1200 \\
			Baidu Zhidao (百度知道) & 100 & 100 & 100 & 100 & 100 & 100 & 100 & 100 & 100 & 100 & 100 & 100 & 1200 & 1200 \\
			Zhihu（知乎） & 100 & 100 & 100 & 100 & 100 & 100 & 100 & 100 & 100 & 100 & 100 & 100 & 1200 & 1200 \\
			CLTS（澎湃新闻） & 100 & 100 & 100 & 100 & 100 & 100 & 100 & 100 & 100 & 100 & 100 & 100 & 1200 & 1200 \\
			\midrule
			\textbf{CN.Total} & 1000 & 1000 & 1000 & 1000 & 1000 & 1000 & 1000 & 1000 & 1000 & 1000 & 1000 & 1000 & 12000 & 12000 \\
			\bottomrule
		\end{tabular}
	}
	\caption{Validation Set Instance Distribution of MAGA Dataset Splits (MGB, MAGA and every split of MAGA-extra). The table summarizes sample counts across English and Chinese domains, including per-model samples, machine sample totals and human samples, with EN.Total and CN.Total for aggregated domain counts.}
	\label{tab:data split distribution on val set}
\end{table*}

\begin{table*}[!t]
	\centering
	\tiny
	\resizebox{\linewidth}{!}{
		\begin{tabular}{l|l|l}
			\toprule
			\textbf{Field} & \textbf{Type} & \textbf{Description} \\
			\midrule
			\textbf{id} & \texttt{str} & A UUID4 that uniquely identifies the content of the text. \\
			\textbf{title} & \texttt{str} & Title from the annotation of the raw HWT data, used in our prompt. \\
			\textbf{text} & \texttt{str} & The content of the text. \\
			\textbf{domain} & \texttt{str} & The genre or platform of the text (e.g., Reddit). \\
			\textbf{human\_source\_id} & \texttt{str} & A UUID4 referencing the human-written source text. \\
			\textbf{prompt\_id} & \texttt{str} & A UUID4 uniquely identifying the prompt (system+user). \\
			\textbf{system\_prompt} & \texttt{str} & The system prompt used for generation. \\
			\textbf{user\_prompt} & \texttt{str} & The user prompt used for generation. \\
			\textbf{model} & \texttt{str} & The LLM that generated the text; labeled as "human" for HWT. \\
			\textbf{label} & \texttt{int} & Binary indicator: 0 for HWT and 1 for MGT. \\
			\textbf{temperature} & \texttt{float} & The decoding parameter for generation. \\
			\textbf{top\_p} & \texttt{float} & The decoding parameter for generation. \\
			\textbf{top\_k} & \texttt{int} & The decoding parameter for generation. \\
			\textbf{repetition\_penalty} & \texttt{float} & The decoding parameter for generation. \\
			\bottomrule
		\end{tabular}
	}
	\caption{Annotation Data Schema of the MAGA Dataset}
	\label{tab:data schema}
\end{table*}

\begin{table}[!t]
	\centering 
	\small
	\resizebox{\linewidth}{!}{
		\begin{tabular}{c|c} 
			\toprule
			\textbf{Model} & \textbf{Identifier}\\
			\midrule
			\textbf{GPT-4o-mini} & \texttt{gpt-4o-mini-2024-07-18}\\
			\textbf{Gemini-2.0-flash} & \texttt{models/gemini-2.0-flash}\\
			\textbf{DeepSeek-V3} & \texttt{deepseek-v3}\\
			\textbf{Qwen3-plus} & \texttt{qwen-plus-2025-04-28}\\
			\textbf{Mistral-Medium} & \texttt{mistral-medium}\\
			\textbf{Hunyuan-TurboS} & \texttt{hunyuan-turbos-20250313}\\
			\textit{---} & \texttt{Llama-3.1-8B-Instruct}\\
			\textit{---} & \texttt{gemma-3-12b-it}\\
			\textit{---} & \texttt{DeepSeek-R1-0528-Qwen3-8B}\\
			\textit{---} & \texttt{Qwen3-8B}\\
			\textit{---} & \texttt{Ministral-8B-Instruct-2410}\\
			\textit{---} & \texttt{Hunyuan-7B-Instruct}\\
			\bottomrule
		\end{tabular}
	}
	\caption{The generative models used in our project}
	\label{tab:generative_models}
\end{table}

Table \ref{tab:generative_models} lists the abbreviations for the models used in this study. The model names appearing in the text and dataset annotations correspond to the Model column in Table \ref{tab:generative_models}, while the Identifier column specifies the actual model names used during implementation, such as API calls. We employ abbreviations for all commercial models, whereas for non-commercial models, the actual identifiers remain consistent with the annotations.

\section{Domain Details}
\label{sec:domain detail}

We prioritize diversity as a major consideration for domain selection. As discussed in \S\ref{sec:domains}, we select domains for the English subset to challenge various model capabilities, and we apply a similar principle to the Chinese subset. Specifically, we include ten distinct Chinese domains: social conversations (Baidu Tieba, Weibo Review, Rednote Review), summarization (CSL), general knowledge (Baidu Baike), reviews (Dianping, Douban Review), direct Q\&A (Baidu Zhidao, Zhihu), and news creation (CLTS).

To enhance coverage, we prioritize selecting two distinct data sources with certain differences for each capability where possible. For the English datasets, Wikipedia provides rigorous, fact-oriented content, whereas wikiHow offers more informal, instructional text. Regarding reviews, Amazon focuses on specific product transactions and feedback, while Trustpilot Reviews maintains a broader scope by including evaluations of services, brands, and physical stores. In the context of question answering, Natural Questions emphasizes scientific accuracy, while Yahoo Answers exhibits a more community-driven style. Similarly, CC News consists of general textual news, whereas NPR News adopts a specialized broadcast style. We note that the selection process for Chinese domains does not strictly adhere to this dual-source strategy due to the relative scarcity of open-source human-written Chinese text datasets.

All our source HWT datasets across selected domains possess a common attribute representing the ground-truth title (even if the specific field naming varies, such as the \texttt{Movie\_Name} field in the Douban Review domain). This enables the integration of titles into prompt templates to generate the corresponding MGT.

The source HWTs we collected are generally clean, yet minimal further inspection and cleaning are required. We provide a detailed description of our data cleaning procedures in Appendix \ref{sec:data clean detail}.

Furthermore, as discussed in Appendix \ref{sec:statistics maga}, taking the MAGA English subset as an example, this subset is constructed by initially sampling 7,200 human-written texts from the original human text dataset of each domain, resulting in a total of 72,000 human texts across all domains. All subsequent experiments and the construction of every splits rely exclusively on these 72,000 human texts. This sampling procedure exerts no influence on the experimental results, and no further sampling for data filtering occurs in the following experimental stages.

The following provides a detailed introduction to all our selected domains:

\paragraph{Reddit} \cite{sentence_transformers_reddit_title_body} This dataset consists of posts sourced from Reddit. We chose this dataset due to its highly informal, conversational style and diverse range of topics, which we hypothesize will pose challenges for detecting machine-generated content due to the lack of strict structural constraints. For this domain, we use the post title as the title and the specific content of each post as the target text.

\paragraph{S2ORC} \cite{lo-wang-2020-s2orc} This Semantic Scholar Open Research Corpus (S2ORC) dataset consists of scholarly paper metadata and full-text content from a wide range of academic disciplines. We include this domain because it tests models' ability to generate and detect text with formal academic tone, precise terminology, and logical argumentation, which relies heavily on domain-specific knowledge. For this domain, we use the paper title as the title and the abstract as the target text.

\paragraph{Wikipedia} \cite{aaditya_bhat_2023} This dataset contains introductory sections of Wikipedia articles covering diverse topics such as history, science, and culture. This domain is challenging as it requires accurate recall of factual information and neutral, encyclopedic writing style, making it a strong testbed for evaluating the factuality of machine-generated text. For this domain, we use the title of a Wikipedia page as the title and the article content as the target text.

\paragraph{wikiHow} \cite{Koupaee2018WikiHowAL} This dataset contains the introductory text of wikiHow articles, spanning from daily life skills to complex technical operations. We selected this domain because it demands practical common sense and an inherent procedural logic, as generating the content requires an understanding of the necessary steps and underlying structure of the task. For this domain, we use the title of a wikiHow page as the title and the article content as the target text.

\paragraph{Trustpilot Reviews} \cite{Kerassy_trustpilot_reviews_123k} This dataset contains user-written reviews of businesses, products, and services. The reviews exhibit a wide range of sentiment polarity and subjective opinions, and we hypothesize that machine-generated reviews may lack the nuanced personal experiences that characterize authentic user feedback. For this domain, we use the title of a review as the title and the review content as the target text.

\paragraph{Amazon Reviews} \cite{Ni2019JustifyingRU} This dataset consists of customer reviews for products sold on Amazon, spanning multiple categories such as electronics, apparel, and household goods. The reviews combine subjective opinions, specific product feature evaluations, and personal usage scenarios, testing models' ability to generate realistic, detailed consumer feedback. For this domain, we use the title of a review as the title and the review content as the target text.

\paragraph{Yahoo Answers} \cite{sentence_transformers_yahoo_answers} This dataset contains user-submitted questions and their corresponding answers on a broad array of topics. We include this domain due to its mix of casual conversational tone and informative content; machine-generated answers may struggle to match the diverse, sometimes idiosyncratic problem-solving approaches of human respondents. For this domain, we use the question as the title and the answer as the target text.

\paragraph{Natural Questions} \cite{sentence_transformers_natural_questions} This dataset comprises real user questions posed to Google Search and their corresponding answers extracted from Wikipedia. This domain is designed to test models' ability to generate accurate, concise factual responses to information-seeking queries, with a focus on alignment between questions and answer relevance. For this domain, we use the question as the title and the answer as the target text.

\paragraph{CC News} \cite{Hamborg2017} This dataset contains news articles sourced from Common Crawl, covering global events across politics, economy, technology, and more. The articles are collected from a vast number of online news outlets, providing a diverse sample of journalistic writing styles and testing models' capacity to generate factually consistent news content. For this domain, we use the title of the news article as the title and the article content as the target text.

\paragraph{NPR News} \cite{sentence_transformers_npr} This dataset consists of news reports and feature articles from National Public Radio, known for its in-depth, balanced coverage of domestic and international issues. We chose this domain because it requires formal journalistic tone, rigorous fact-checking, and narrative coherence, highlighting the challenge of machine-generated text in mimicking professional news writing. For this domain, we use the title of the news article as the title and the article content as the target text.

\paragraph{Baidu Tieba} \cite{HongzheBi_2023} This dataset contains posts and comments from Baidu Tieba, a Chinese-language online forum with topic-specific communities. The content features colloquial Chinese expressions, slang, and community-specific jargon, making it challenging for models to replicate the authentic conversational style of Chinese internet users. For this domain, we use the title of a post as the title and a reply to that post as the target text.

\paragraph{Weibo Review} \cite{LZYFirecn_2025} This dataset comprises user comments from Weibo, a major Chinese microblogging platform. The reviews are typically concise, sentiment-laden, and often tied to trending topics, testing models' ability to generate short-form, context-aware Chinese text that captures the nuances of online public opinion.  For this domain, we use the title of a Weibo article as the title and a review of that article as the target text.

\paragraph{Rednote Review} \cite{CHASM} This dataset consists of user-written reviews and lifestyle sharing content from Xiaohongshu (Rednote), a leading Chinese social media platform with a predominantly female user base. Centered on authentic personal experiences, the platform emphasizes highly subjective, first-person narratives and visual cues rooted in daily life. We hypothesize that machine-generated content may struggle to replicate the nuanced, relatable details and the emotional resonance characteristic of these organic community posts. For this domain, we use the title of a Rednote post as the title and a review of the post as the target text.

\paragraph{CSL} \cite{li-etal-2022-csl} This Chinese Scientific Literature (CSL) dataset contains abstracts from Chinese Core Journals, covering academic research in fields such as humanities, social sciences, natural sciences, and engineering. We include this domain as it requires mastery of formal Chinese academic writing conventions, precise disciplinary terminology, and logical research framework presentation. For this domain, we use the paper title as the title and the abstracts as the target text.

\paragraph{Baidu Baike} \cite{lars1234_2025} This dataset contains entry content from Baidu Baike, a Chinese-language online encyclopedia. Similar to Wikipedia, this domain demands neutral, factual writing and accurate representation of Chinese-specific cultural, historical, and scientific knowledge, posing challenges for models' cross-lingual factual recall ability. For this domain, we use the title of a Baidu Baike page as the title and the article content as the target text.

\paragraph{Dianping} \cite{twn39_2025} This dataset comprises user reviews of restaurants, shops, and local services from Dianping, a Chinese local life service platform. The reviews focus on detailed experience descriptions, such as food taste, service quality, and store environment, testing models' ability to generate realistic, location-specific Chinese consumer feedback. For this domain, the title we use consists of the products and services user purchased, while the target text is the review provided by the user for this purchase.

\paragraph{Douban Review} \cite{dirtycomputer_2023} This dataset contains user reviews of movies from Douban, a Chinese cultural community platform. The reviews are characterized by subjective, insightful critical analysis and emotional expression, making it difficult for machine-generated text to replicate the depth of human cultural perception and personal taste. For this domain,  we use the movie name as the title and the user review as the target text.

\paragraph{Baidu Zhidao} \cite{HongzheBi_2023} This dataset consists of Q\&A pairs from Baidu Zhidao. Characterized by rigorous content moderation and professional verification, the answers provide a source of high-quality, standardized Chinese text. This helps evaluated models distinguish between machine-generated content and highly structured, verified human expertise. For this domain, we use the question as the title and the answer as the target text.

\paragraph{Zhihu} \cite{Liu2020CLTSAN} This dataset contains content from Zhihu, a platform known for its casual and spontaneous user interactions. Unlike professional databases, these answers are often subjective and conversational, reflecting the unconstrained writing styles of general users. Such data is crucial for testing whether detection models can identify the subtle gap between AI outputs and fluid, non-expert human expression. For this domain, we use the question as the title and the answer as the target text.

\paragraph{CLTS} \cite{Liu2020CLTSAN} This Chinese Long Text Summarization (CLTS) dataset comprises news articles from "Pengpai Xinwen" (thepaper.cn), a Chinese mainstream digital news outlet, covering politics, society, culture, and technology. The articles adhere to rigorous journalistic standards and focus on in-depth investigative reporting, testing models' ability to generate formal, factually accurate Chinese news content. For this domain, we use the title of the news article as the title and the article content as the target text.

\section{Generator Details}
\label{sec:generator detail}

In Table \ref{tab:generative_models} we list the exact generative models used in our project along with their unique identifiers. All open-source models were run using the vLLM library \footnote{\url{https://docs.vllm.ai/en/stable/}} and all closed-source models were run using the proprietary APIs from OpenAI \footnote{\url{https://platform.openai.com/docs/introduction}}. The following is a detailed list of the generative models used in the project.

\paragraph{GPT-4o-mini} \cite{Hurst2024GPT4oSC} is a lightweight variant of GPT-4o developed by OpenAI. We utilize the \texttt{gpt-4o-mini-2024-07-18} checkpoint released on July 18, 2024. This model significantly outperforms GPT-3.5 Turbo in reasoning and coding tasks while maintaining high efficiency and low latency, making it the industry standard for cost-effective intelligence.

\paragraph{Gemini-2.0-flash} \cite{google_deepmind_2024_gemini} is Google DeepMind's multimodal model released on March 12, 2025. It is designed for high-speed inference across text, image, and audio. It offers state-of-the-art multimodal reasoning with near-instantaneous response times. We queried the model for the MAGA English subset from September 6 to 7, 2025, and for the MAGA-cn Chinese subset from October 13 to 14, 2025.

\paragraph{DeepSeek-V3} \cite{DeepSeekAI2024DeepSeekV3TR} is a strong decoder-only model developed by DeepSeek AI, released on December 16, 2024. It employs a Mixture-of-Experts (MoE) architecture, excelling in math, coding, and logical reasoning to a level comparable with top-tier proprietary models. We queried the model for the MAGA English subset from September 6 to 7, 2025, and for the MAGA-cn Chinese subset from October 13 to 14, 2025.

\paragraph{Qwen3-plus} \cite{Yang2025Qwen3TR} is a large-scale model developed by Alibaba Cloud. We utilize the \texttt{qwen-plus-2025-04-28} checkpoint released on April 28, 2025. As a flagship model in the Qwen3 series, it demonstrates superior performance in complex instruction following and multi-turn dialogue compared to its predecessors.

\paragraph{Mistral-Medium} \cite{mistral_ai_2025_medium3} is developed by Mistral AI and was released on May 7, 2025. It strikes an optimal balance between the efficiency of smaller models and the reasoning power of the Large series, particularly excelling in multilingual understanding. We queried the model for the MAGA English subset from September 6 to 7, 2025, and for the MAGA-cn Chinese subset from October 13 to 14, 2025.

\paragraph{Hunyuan-TurboS} \cite{tencenthunyuanteam2025hunyuanturbosadvancinglargelanguage} is Tencent's lightweight variant optimized for low-latency streaming. We use the \texttt{hunyuan-turbos-20250313} checkpoint released on March 13, 2025. This model is specifically fine-tuned for high-concurrency scenarios and daily conversational interactions.

\paragraph{Llama-3.1-8B-Instruct} \cite{grattafiori2024llama3herdmodels} is an 8B-parameter instruction-tuned model developed by Meta, released in July 2024. It features an expanded 128K context window and significantly improved tool-use capabilities, setting a benchmark for open-source models in its weight class.

\paragraph{gemma-3-12b-it} \cite{gemmateam2025gemma3technicalreport} is an open-weight instruction-tuned model developed by Google, released in March 2025. Building on the Gemma 3 architecture, it introduces native multimodal support and exhibits robust performance in logical deduction and mathematical problem-solving.

\paragraph{DeepSeek-R1-0528-Qwen3-8B} \cite{DeepSeekAI2025DeepSeekR1IR} is developed by DeepSeek AI and released in May 2025. It integrates DeepSeek's R1 reinforcement learning methodology into the Qwen3-8B base, specifically designed to enhance "Chain-of-Thought" (CoT) reasoning for complex tasks.

\paragraph{Qwen3-8B} \cite{Yang2025Qwen3TR} is Alibaba Cloud's 8B-parameter open-source model released in April 2025. It serves as a highly versatile base model with competitive performance across general knowledge and coding benchmarks for small-scale deployment.

\paragraph{Ministral-8B-Instruct-2410} \cite{mistral_ai_team_2024_ministral} is a specialized "edge" model developed by Mistral AI, released in October 2024. It is optimized for local inference on resource-constrained devices while maintaining high instruction-following accuracy.

\paragraph{Hunyuan-7B-Instruct} \cite{tencent_2025_hunyuan} is a 7B-parameter model developed by Tencent, released in July 2025. It is highly optimized for Chinese linguistic nuances and cultural context, making it particularly effective for Chinese content generation tasks.

\section{Alignment Method Details}
\label{sec:alignment method details}
\subsection{Roleplaying}
\label{sec:roleplaying detail}

To implement coarse-grained roleplaying within the MAGA-extra-roleplaying and MAGA splits, we first constructed the MAGA-ROLE-80 prompt set. As discussed in Section \ref{sec:MAGA} and Appendix \ref{sec:statistics maga}, a one-to-one basis exists between MGT and HWT. Taking the MAGA English Subset as an example, we utilize the title fields from 72k HWT data to create 72k original prompts, which serve as the basis for the MGTs in the MGB split. Then each original prompt is randomly assigned one of the 80 roleplaying prompts. Once this assignment is complete, the prompt set remains fixed for all subsequent uses to eliminate sampling randomness and we believe it is helpful for GDARL's learning stability. Details about all prompts including rolepaying prompts are provided in Appendix \ref{sec:prompt detail}.

All previous MGT detection benchmarks listed in Table \ref{tab:all-datasets} do not explicitly report the use of system prompt. Based on our reading and analysis and to the best of our knowledge, some additional instructions found in these  previous works appear to have been included within the user prompt, rather than being utilized as a system prompt. In our work, roleplaying prompts serve as system prompts to provide high-level instructions and persona settings.

In the design of LLM conversational API, the structured input method that distinguishes the three core "Messages" roles—System, User, and Assistant—is known as Chat Markup Language (ChatML)\footnote{\url{https://github.com/openai/openai-python/blob/120d225b91a8453e15240a49fb1c6794d8119326/chatml.md}}. ChatML addresses the difficulty models face in distinguishing between instructions and conversational content, a framework originally introduced with GPT-3.5\footnote{\url{https://openai.com/index/introducing-chatgpt-and-whisper-apis/}}. System prompts are now frequently used to define a model's behavioral guidelines, identity and background, output constraints, or specific knowledge scope, effectively allowing the model to maintain a specific tone.

Regarding its mode of input operation, taking the example in Figure \ref{fig:MAGA Pipeline} as an illustration, when only a user prompt is present, the actual input fed to the LLM after being formatted into the ChatML template is: '<|im\_start|>user\textbackslash nWrite the body of a Reddit post titled "How to really make America great again".<|im\_end|>\textbackslash n'. When a system prompt is added, the actual input fed to the LLM after formatting into the ChatML template becomes: '<|im\_start|>system\textbackslash nYou are American. Please respond in a casual, conversational American English style.<|im\_end|>\textbackslash n<|im\_start|>user\textbackslash nWrite the body of a Reddit post titled "How to really make America great again".<|im\_end|>\textbackslash n'.

Why do we adopt coarse-grained role-playing? What are the characteristics of MAGA-ROLE-80, and why is it effective? The following is a detailed discussion and analysis, grounded in relevant prior work and presented according to a logical theoretical progression:

\paragraph{Coarse-grained and Fine-grained Role-playing} \citet{chen2024oscars} provide a systematic survey of the role-playing landscape in LLMs, categorizing the granularity of role-playing into coarse-grained and fine-grained approaches. Coarse-grained role-playing is typically persona-based, focusing on basic demographic information such as age, gender, location, and brief personality traits, or in even coarser-grained instances, it may be limited solely to identity traits. In contrast, fine-grained role-playing is character-based and involves comprehensive attributes including lineage, talents, belongings, and experiences, as well as complex interpersonal relations, environmental scenes, and temporal personas that evolve over time. The evaluation of role-playing generally encompasses four dimensions: conversational ability, role-persona consistency, role-behavior consistency, and role-playing attractiveness. According to \citet{chen2024oscars}, the primary objective of fine-grained role-playing is to achieve a deep restoration of specific characters while maintaining consistency across these four dimensions. However, relying solely on prompt engineering for fine-grained tasks often proves problematic. Extensive prompts tend to saturate the context window, which leads to a decline in generation quality and coherence. Furthermore, excessively detailed personas can interfere with the native conversational capabilities of the model, resulting in rigid, unnatural, or logically fractured responses. Fine-grained role-playing typically requires dimension expanded high-quality data collection, and fine-tuning to improve dialogue quality. Without fine-tuning, LLMs often exhibit a decrease in fluency and human-like performance when performing fine-grained tasks. For improvements in basic style and human-likeness, coarse-grained role-playing is generally sufficient and yields satisfactory results through optimized prompt engineering. Experiments in RoleLLM \cite{wang-etal-2024-rolellm} report similar findings, demonstrating that coarse-grained role-playing achieves excellent performance using only prompt engineering.

\paragraph{RRP} (Rule-based Role Prompting, \citet{ruangtanusak2025talklessrightenhancing} RRP focuses on enhancing coarse-grained role-playing within the Commonsense Persona-grounded Dialogue Challenge (CPDC) 2025. The core philosophy of RRP is "Talk Less, Call Right". Beyond the specific tool-calling challenges addressed in their work, their experimental results and analysis indicate that optimization in prompt design alone significantly improves the coarse-grained role-playing performance of LLMs. Specifically, prompts that are brief, clearly structured, explicit in intention, and possess a certain rule-based nature facilitate better role-playing outcomes.

\paragraph{ROLE-88} \cite{ettinger-2020-bert, chow2016bag} ROLE-88 is originally designed for verb prediction tasks involving argument role reversals in sentences. For example, in the sentences "the restaurant owner forgot which customer the waitress had \_\_\_" and "the restaurant owner forgot which waitress the customer had \_\_\_"---where the role nouns *customer* and *waitress* have been swapped---predict the verb. Setting aside the specific tasks associated with ROLE-88, let us focus solely on the roles themselves (excluding the very few non-human entities, such as birds or dragons). Generally speaking, ROLE-88 primarily encompasses a curated selection of the most typical, high-frequency professional roles comparatively speaking encountered in everyday social life roles that, to a significant extent, serve as a representative sample of daily social interactions. Moreover, these roles are sufficiently easy to comprehend and well-suited for role-playing.

\paragraph{MAGA-ROLE-80} Consequently, MAGA adopts a coarse-grained role-playing approach. As suggested by the findings in \citet{chen2024oscars}'s survey, RoleLLM \cite{wang-etal-2024-rolellm}, and RRP \cite{ruangtanusak2025talklessrightenhancing}, coarse-grained role-playing is the appropriate choice for our objective of human-like alignment in our task, offering enough satisfactory performance via prompt engineering.

To ensure social coverage and diversity, we select the majority of roles from ROLE-88 \cite{ettinger-2020-bert, chow2016bag}. Since ROLE-88 is not directly applicable as a set of conversational prompts, we construct the MAGA-ROLE-80 prompt set based on these roles. Our prompt design incorporates the core ideas of RRP's "Talk Less, Call Right" principle \cite{ruangtanusak2025talklessrightenhancing}, ensuring that the prompts are concise, structured, and easy to inhabit. Table \ref{tab:role playing prompts} provides ten examples of these prompts. These prompts possess a simple, uniform rule-based nature, such as "You are American. Please respond in a casual, conversational American English style.", or "You are a teacher. Respond in a patient, firm, and caring tone.". Each prompt basically consists of two sentences: the first defines the basic identity, and the second specifies the speaking style, which aligns with the requirements of our task.

The MAGA-ROLE-80 set contains 80 prompts each for English and Chinese. While primarily professional roles, we include a few well-known figures like Sherlock Holmes to enhance diversity. The English and Chinese versions are largely translations of each other, with minor cultural adaptations, such as mapping an American role in English to a Beijinger role in Chinese. All prompts are manually authored.

Similar to ROLE-88, a set of 80 roles is considered sufficient for broad representation. As detailed in Appendix \ref{sec:statistics maga}, each LLM generates 7200 samples per split (6000 for training and 1200 for validation). The square root of this total is approximately 80, which ensures a balance between the number of roles and the amount of text generated per role. We believe this balance is conducive to the learning process of GDARL.

Additionally, it should also be noted that, our MAGA-ROLE-80 prompt set actually contains 79 specific roles plus one null string, representing a default or role-less role. We believe this inclusion benefits the overall quality of the dataset and the stability of GDARL training.

\subsection{BPO}
\label{sec:bpo detail}
The Black-box Prompt Optimization (BPO) \cite{cheng2024black} method is an automated alignment model that optimizes user instructions to bridge the understanding gap between humans and large language models. The underlying principle involves constructing a preference dataset where an initial prompt is paired with both preferred and rejected responses. By utilizing a powerful teacher model to reverse engineer optimized instructions that elicit the preferred responses, the authors train a sequence to sequence model to map original user inputs to refined versions. A defining characteristic of the BPO model is its operational tendency to preserve the core intent of the user while appending a descriptive suffix that clarifies the task requirements. For example, a simple instruction such as "Tell me about Harry Potter" is transformed into an expanded version such as "Tell me about Harry Potter, providing a comprehensive summary of the book series including its main characters, central plot, and major themes."

In this work, we directly employ the BPO model released by \citet{cheng2024black}\footnote{\url{https://huggingface.co/zai-org/BPO}}. To ensure the reliability and consistency of the optimized instructions, we specifically utilize the stable mode of the model. The execution process involves wrapping the user prompt into a specialized template: "[INST] You are an expert prompt engineer. Please help me improve this prompt to get a more helpful and harmless response: {user\_prompt} [/INST]". During the inference stage, the model generates a suffix to optimize the original prompt using a temperature of 0.6, a top\_p value of 0.9, and a max\_new\_tokens of 1024. This configuration prioritizes stable and grounded refinements over the more stochastic outputs of the aggressive mode, thereby providing a robust foundation for the subsequent generation tasks of the target model.

To the best of our knowledge, all previous machine-generated text (MGT) detection benchmarks listed in Table \ref{tab:all-datasets} exclusively utilize static, human-designed prompts. The primary advantage of BPO lies in its capacity to ensure that each individual prompt is unique, substantially enhancing prompt diversity.

It is essential to note that when incorporating a title into the original prompt, human designers cannot anticipate the specific content of each title in advance, or more precisely, it is simply impossible for human to tailor the prompt individually for every conceivable title. BPO addresses this limitation by providing specific optimization suggestions for every unique title. Additionally, it should also be noted that, role-playing do not encounter such unpredictable elements as titles and possess inherent specificity, thereby reducing and therefore do not necessitate such highly automated prompt optimization.

Additionally, the BPO model \citet{cheng2024black} released only supports English inputs and outputs, lacking native support for Chinese. To address this limitation, we employ the Hunyuan-MT-7B \cite{zheng2025hunyuanmttechnicalreport} translation model. Our selection of Hunyuan-MT-7B is motivated by two primary factors. First, it demonstrates exceptional translation quality, as \citet{zheng2025hunyuanmttechnicalreport} report that its performance is comparable to powerful commercial models such as GPT-4.1 and DeepSeek-V3-0324. Second, as a specialized translation model, it directly produces translation outputs without generating redundant content, and eliminates the need for complex prompt engineering or post-processing controls. Specifically, we first translate the original Chinese prompts into English using Hunyuan-MT-7B and then process them through the BPO model for optimization. The resulting English suffixes generated by BPO are translated back into Chinese via Hunyuan-MT-7B. Finally, these optimized suffixes are appended to the original prompts to construct the BPO-optimized Chinese prompts.

\subsection{Self Refine}
\label{sec:selfrefine detail}
The Self Refine \cite{madaan2023self} framework enables the autonomous enhancement of large language model outputs through a recursive process involving critique and revision. The procedure begins with the generation of an initial response $y_{0}$ for a given input $x$. At each iteration $t$, the model performs self-evaluation via in-context learning (ICL) to produce multi-aspect feedback that identifies necessary corrections. Based on this feedback and the preceding output $y_{t}$, the model constructs a refined response $y_{t+1}$. This sequence continues iteratively, where the model employs in-context learning to determine whether the current output satisfies the predefined objectives. The loop terminates when the self-judgment indicates that the goal is reached or when the process hits a maximum number of allowed iterations. Since the framework utilizes a single model to function as the generator, feedback provider, and refiner, it avoids the need for supervised fine-tuning or reinforcement learning.

In our implementation for machine-generated text detection, we adhere to the original algorithmic structure while introducing a specific modification. While large language models are reliable as general evaluators for self-critique, their performance in the explicit detection of machine-generated patterns remains a challenging task that can exhibit inconsistency across multiple rounds. To ensure the stability of the detection results, we restrict the process to a single refinement round instead of multiple iterations. This approach leverages the structural benefits of feedback-informed refinement while mitigating the potential for error accumulation or instability in multi-round detection. We provide a more comprehensive discussion and experimental analysis regarding the choice of iteration rounds in Appendix \ref{sec:selfrefineround}. The Self-Refine prompts for English and Chinese are shown in Table \ref{tab:self refine prompts} and Table \ref{tab:self refine prompts Chinese}, respectively, which are described in Appendix \ref{sec:prompt detail}.

\subsection{GDARL}
\label{sec:GDARL detail}
It is necessary to emphasize that Generator-Detector Adversarial Reinforcement Learning (GDARL) is the general term for existing GDARL algorithms such as RADAR \cite{hu2023radar} and HUMPA \cite{wang2025humanizing} in the field of MGT detection, and it represents the vanilla formulation of such approaches rather than a newly proposed innovative method.

However, We observe that the vanilla GDARL suffers from reward overfitting, which leads to negligible updates of the trained LLM. Based on the HC-Var theory, we further propose the Cross-Mutual Reward (CMR) component and present its three basic application, namely GDARL-CDMR, GDARL-CMMR and GDARL-CDMMR.

The CMR component operates as an external process attached to the ARL paradigm and does not modify the internal algorithmic framework of GDARL. Therefore, it does not conflict with algorithms such as RADAR or HUMPA and is able to be used in conjunction with them.

This section formally defines the algorithmic procedures of GDARL, GDARL-CDMR, GDARL-CMMR and GDARL-CDMMR. Detailed discussion about the CMR component and existing ARL frameworks is provided in Appendix \ref{sec:GDARLvs}.

First, we predefine the following notations and variables:

$Q$: The set of input prompts.

$t$: The current round of adversarial learning, where $t \geq 0$ is a non-negative integer and $t=0$ denotes the initial model state with pre-trained weights.

$\mathcal{M}_{\theta}^{t}$: The large language model (LLM) to be fine-tuned at round $t$ with parameters $\theta^{t}$; its generation distribution is $\pi_{\theta}^{t}(o|q) = \mathcal{M}_{\theta}^{t}(o|q)$, where $q$ is the input prompt and $o$ is the generated text.

$\mathcal{D}_{\phi}^{t}$: The RoBERTa reward detector at round $t$ with parameters $\phi^{t}$ that serves as GRPO's reward model (RM) outputs the probability $\mathcal{D}_{\phi}^{t}(x) \in [0,1]$ that text $x$ is machine-generated.

$H$: The set of human-written text (HWT), labeled as 0.

$M^{t}$: The set of machine-generated text (MGT) at round $t$, labeled as 1. It is composed of one single corresponding text generated by \(\mathcal{M}_{\theta}^{t}\) for each individual prompt in Q on a one-to-one basis.

$\eta$: The learning rate for updates of the LLM parameters $\theta$.

$\alpha$: The learning rate for updates of the detector parameters $\phi$.

\paragraph{GDARL} First, feed each prompt in $Q$ into $\mathcal{M}_{\theta}^{t}$; the model generates one MGT for each prompt on a one-to-one basis to form the MGT set $\mathcal{M}^{t}$. The set $\mathcal{M}^{t}$ is then used to train the detector $\mathcal{D}^{t}$ and yield the updated detector $\mathcal{D}^{t+1}$:

\begin{equation}
	\begin{split}
		\mathcal{J}_\mathcal{D}(\phi^{t}, t) & =  - \mathbb{E}_{x_{m}^{t} \sim M^{t}} \left[\log \mathcal{D}_{\phi}^{t}(x_{m}^{t})\right] - \\ & \mathbb{E}_{x_h \sim H} \left[\log\left(1 - \mathcal{D}_{\phi}^{t}(x_h)\right)\right]
	\end{split}
\end{equation}

\begin{equation}\phi^{t+1} = \phi^{t} - \alpha \nabla_\phi \mathcal{J}_\mathcal{D}(\phi^{t}, t)\end{equation}

The trained detector $\mathcal{D}_{\phi}^{t+1}$ subsequently functions as the reward model to assign human-like scores. We employ Group Relative Policy Optimization (GRPO) to train the LLM, thereby achieving adversarial RL (ARL). Specifically, for each question $q$, GRPO samples a group of outputs $O=\{o_{1}, o_{2}, \dots, o_{G}\}$ from the old policy $\pi_{\theta_{old}}^{t}$ and optimizes the policy model $\pi_{\theta}^{t}$ by maximizing the following objective:

\begin{equation}
	\begin{split}
		\mathcal{J}_{GRPO}(\theta^{t}, t) & = \mathbb{E} \left[ q \sim P(Q), \{o_i\}_{i=1}^G \sim \pi_{\theta_{old}}^{t}(O|q) \right] \frac{1}{G} \sum_{i=1}^G \Biggl( \\
		& \min \left( \frac{\pi_\theta^{t}(o_i|q)}{\pi_{\theta_{old}}^{t}(o_i|q)} A_{i}^{t}, \text{clip} \left( \frac{\pi_\theta^{t}(o_i|q)}{\pi_{\theta_{old}}^{t}(o_i|q)}, 1-\varepsilon, 1+\varepsilon \right) A_{i}^{t} \right) \\
		& - \beta \mathbb{D}_{KL} (\pi_{\theta}^{t} || \pi_{ref}^{t}) \Biggr),
	\end{split}
\end{equation}

\begin{equation}\mathbb{D}_{KL} (\pi_\theta^{t} || \pi_{ref}^{t}) = \frac{\pi_{ref}^{t}(o_i|q)}{\pi_\theta^{t}(o_i|q)} - \log \frac{\pi_{ref}^{t}(o_i|q)}{\pi_\theta^{t}(o_i|q)} - 1,\end{equation}

where $\varepsilon$, $\beta$, and $G$ denote the PPO clipping threshold, the KL regularization weight, and the number of responses generated per question for group-wise averaging, respectively. $A_{i}^{t}$ is the advantage, computed using a group of rewards $\{r_1^{t}, r_2^{t}, \dots, r_G^{t}\}$ corresponding to the outputs within each group:

\begin{equation}A_i^{t} = \frac{r_i^{t} - \text{mean}(\{r_1^{t}, r_2^{t}, \dots, r_G^{t}\})}{\text{std}(\{r_1^{t}, r_2^{t}, \dots, r_G^{t}\})},\end{equation}

\begin{equation}r^{t}(x, \phi^{t+1}) = 1 - \mathcal{D}_\phi^{t+1}(x).\end{equation}

Then we can obtain the LLM $\mathcal{M}_{\theta}^{t+1}$:

\begin{equation}\theta^{t+1} = \theta^{t} + \eta \nabla_\theta \mathcal{J}_{GRPO}(\theta^{t}, t)\end{equation}

\paragraph{GDARL-CDMR} GDARL-CDMR (Cross Domain Mutual Reward) divides all texts into two groups: Domains $DA$ and Domains $DB$. Each group has its own separately trained reward detectors $\mathcal{D}_{\phi}^{DA, t}$ and $\mathcal{D}_{\phi}^{DB, t}$. During the training of the larger model, Cross-Mutual Rewards are applied: $\mathcal{D}_{\phi}^{DB, t}$ scores text in $DA$, and $\mathcal{D}_{\phi}^{DA, t}$ scores text in $DB$. Specifically:

We divide $H$ into $H^{DA}$ and $H^{DB}$, with corresponding $Q$ into $Q^{DA}$ and $Q^{DB}$. The MGT set generated using $\mathcal{M}_{\theta}^{t}$ is also divided into $M^{DA, t}$ and $M^{DB, t}$, respectively. First, the detectors are trained as follows:

\begin{equation}
	\begin{split}
	\mathcal{J}_\mathcal{D}(\phi^{DA, t}, t) & =  - \mathbb{E}_{x_{m}^{DA, t} \sim M^{DA, t}} \left[\log \mathcal{D}_{\phi}^{DA, t}(x_{m}^{DA, t})\right] - \\ & \mathbb{E}_{x_h^{DA} \sim H^{DA}} \left[\log\left(1 - \mathcal{D}_{\phi}^{DA, t}(x_h^{DA})\right)\right]
	\end{split}
\end{equation}

\begin{equation}
	\phi^{DA, t+1} = \phi^{DA, t} - \alpha \nabla_\phi \mathcal{J}_\mathcal{D}(\phi^{DA, t}, t)
\end{equation}

\begin{equation}
	\begin{split}
	\mathcal{J}_\mathcal{D}(\phi^{DB, t}, t) & =  - \mathbb{E}_{x_{m}^{DB, t} \sim M^{DB, t}} \left[\log \mathcal{D}_{\phi}^{DB, t}(x_{m}^{DB, t})\right] - \\ & \mathbb{E}_{x_h^{DB} \sim H^{DB}} \left[\log\left(1 - \mathcal{D}_{\phi}^{DB, t}(x_h^{DB})\right)\right]
	\end{split}
\end{equation}

\begin{equation}
	\phi^{DB, t+1} = \phi^{DB, t} - \alpha \nabla_\phi \mathcal{J}_\mathcal{D}(\phi^{DB, t}, t)
\end{equation}

Then the trained detector $\mathcal{D}_{\phi}^{DA, t+1}, \mathcal{D}_{\phi}^{DB, t+1}$ is used as the reward model cross-scoring for the text $x^{DB}, x^{DA}$ on Domains $DB$ and Domains $DA$. LLM is trained using GRPO, with the following reward:

\begin{equation}
	r^{t} =
	\begin{cases}
		1 - \mathcal{D}_{\phi}^{DB, t+1}\big(x^{DA}\big), & x^{DA} \sim \mathcal{M}_{\theta}^{t}(o|q^{DA}),\ q^{DA}\in Q^{DA} \\[6pt]
		1 - \mathcal{D}_{\phi}^{DA, t+1}\big(x^{DB}\big), & x^{DB} \sim \mathcal{M}_{\theta}^{t}(o|q^{DB}),\ q^{DB}\in Q^{DB}
	\end{cases}
\end{equation}

Then we can obtain the LLM $\mathcal{M}_{\theta}^{t+1}$:

\begin{equation}
	\theta^{t+1} = \theta^{t} + \eta \nabla_\theta \mathcal{J}_{GRPO}(\theta^{t}, t)
\end{equation}

\paragraph{GDARL-CMMR} In our task, in practice, we adopt multiple LLMs and split the entire generation work, with each LLM responsible for a separate part. GDARL-CMMR (Cross Model Mutual Reward) directly leverages this pipeline.

We divide all texts into two groups corresponding to LLM Models $MA$ and LLM Models $MB$. A detector $\mathcal{D}_{\phi}^{MA, t}$ and a detector $\mathcal{D}_{\phi}^{MB, t}$ are trained independently for each group respectively. During the training of LLMs, cross-model mutual reward is implemented: $\mathcal{D}_{\phi}^{MB, t}$ scores the texts generated by $MA$, and $\mathcal{D}_{\phi}^{MA, t}$ scores the texts generated by $MB$. It is worth noting that the pipeline also involves texts generated by commercial LLMs, which is simply denoted as MGT-B in Figure \ref{fig:GDARL}. This part is also grouped in our work, as described in Appendix \ref{sec:gdarl-cdmmr group}. For ease of understanding and to avoid overly cumbersome formula expressions, we omit the description of this component in the formulas herein. Specifically:

We partition $H$ into $H^{MA}$ and $H^{MB}$, with corresponding prompts $Q^{MA}$ and $Q^{MB}$. These prompts are then assigned to $\mathcal{M}_{\theta}^{MA, t}$ and $\mathcal{M}_{\theta}^{MB, t}$, respectively, for generation; the resulting sets of generated MGTs are $M^{MA, t}$ and $M^{MB, t}$. First, we train the detectors:

\begin{equation}
	\begin{split}
		\mathcal{J}_\mathcal{D}(\phi^{MA, t}, t) & =  - \mathbb{E}_{x_{m}^{MA, t} \sim M^{MA, t}} \left[\log \mathcal{D}_{\phi}^{MA, t}(x_{m}^{MA, t})\right] - \\ & \mathbb{E}_{x_h^{MA} \sim H^{MA}} \left[\log\left(1 - \mathcal{D}_{\phi}^{MA, t}(x_h^{MA})\right)\right]
	\end{split}
\end{equation}

\begin{equation}
	\phi^{MA, t+1} = \phi^{MA, t} - \alpha \nabla_\phi \mathcal{J}_\mathcal{D}(\phi^{MA, t}, t)
\end{equation}

\begin{equation}
	\begin{split}
		\mathcal{J}_\mathcal{D}(\phi^{MB, t}, t) & =  - \mathbb{E}_{x_{m}^{MB, t} \sim M^{MB, t}} \left[\log \mathcal{D}_{\phi}^{MB, t}(x_{m}^{MB, t})\right] - \\ & \mathbb{E}_{x_h^{MB} \sim H^{MB}} \left[\log\left(1 - \mathcal{D}_{\phi}^{MB, t}(x_h^{MB})\right)\right]
	\end{split}
\end{equation}

\begin{equation}
	\phi^{MB, t+1} = \phi^{MB, t} - \alpha \nabla_\phi \mathcal{J}_\mathcal{D}(\phi^{MB, t}, t)
\end{equation}

Subsequently, the trained detectors $\mathcal{D}_{\phi}^{MB, t+1}$ and $\mathcal{D}_{\phi}^{MA, t+1}$ serve as reward models to cross-score Models $\mathcal{M}_{\theta}^{MA, t}$ with parameters $\theta^{MA, t}$ and Models $\mathcal{M}_{\theta}^{MB, t}$ with parameters $\theta^{MB, t}$, respectively. The large models are then trained via GRPO, with the rewards defined as follows:

\begin{equation}
	r^{MA, t}(x, \phi^{MB, t+1}) = 1 - \mathcal{D}_\phi^{MB, t+1}(x)
\end{equation}

\begin{equation}
	r^{MB, t}(x, \phi^{MA, t+1}) = 1 - \mathcal{D}_\phi^{MA, t+1}(x)
\end{equation}

Here, $r^{MA, t}$ is utilized for the GRPO training of Models $\mathcal{M}_{\theta}^{MA, t}$, while $r^{MB, t}$ is utilized for the GRPO training of Models $\mathcal{M}_{\theta}^{MB, t}$, yielding the following updates:

\begin{equation}
	\theta^{MA, t+1} = \theta^{MA, t} + \eta \nabla_\theta \mathcal{J}_{GRPO}(\theta^{MA, t}, t)
\end{equation}

\begin{equation}
	\theta^{MB, t+1} = \theta^{MB, t} + \eta \nabla_\theta \mathcal{J}_{GRPO}(\theta^{MB, t}, t)
\end{equation}

\paragraph{GDARL-CDMMR} The final form of GDARL-CDMMR (Cross Domain \& Model Mutual Reward) is constructed by integrating GDARL-CDMR and GDARL-CMMR. More generally, we can simply designate the two groups as $GA$ and $GB$. In group $GA$, the LLMs responsible for generation are $MA$, and the domains of the generated texts are $DA$; in group $GB$, the LLMs responsible for generation are $MB$, and the domains of the generated texts are $DB$. The specific grouping scheme utilized by MAGA is detailed in Appendix \label{sec:gdarl cdmmr group}.

We partition $H$ into $H^{GA}$ and $H^{GB}$, with the corresponding prompts being $Q^{GA}$ and $Q^{GB}$. These prompts are then assigned to $\mathcal{M}_{\theta}^{GA, t}$ and $\mathcal{M}_{\theta}^{GB, t}$, respectively, for text generation; the corresponding sets of MGTs produced are $M^{GA, t}$ and $M^{GB, t}$. We first train the detectors:

\begin{equation}
	\begin{split}
		\mathcal{J}_\mathcal{D}(\phi^{GA, t}, t) & =  - \mathbb{E}_{x_{m}^{GA, t} \sim M^{GA, t}} \left[\log \mathcal{D}_{\phi}^{GA, t}(x_{m}^{GA, t})\right] - \\ & \mathbb{E}_{x_h^{GA} \sim H^{GA}} \left[\log\left(1 - \mathcal{D}_{\phi}^{GA, t}(x_h^{GA})\right)\right]
	\end{split}
\end{equation}

\begin{equation}
	\phi^{GA, t+1} = \phi^{GA, t} - \alpha \nabla_\phi \mathcal{J}_\mathcal{D}(\phi^{GA, t}, t)
\end{equation}

\begin{equation}
	\begin{split}
		\mathcal{J}_\mathcal{D}(\phi^{GB, t}, t) & =  - \mathbb{E}_{x_{m}^{GB, t} \sim M^{GB, t}} \left[\log \mathcal{D}_{\phi}^{GB, t}(x_{m}^{GB, t})\right] - \\ & \mathbb{E}_{x_h^{GB} \sim H^{GB}} \left[\log\left(1 - \mathcal{D}_{\phi}^{GB, t}(x_h^{GB})\right)\right]
	\end{split}
\end{equation}

\begin{equation}
	\phi^{GB, t+1} = \phi^{GB, t} - \alpha \nabla_\phi \mathcal{J}_\mathcal{D}(\phi^{GB, t}, t)
\end{equation}

Subsequently, the trained detectors $\mathcal{D}_{\phi}^{GB, t+1}$ and $\mathcal{D}_{\phi}^{GA, t+1}$ serve as reward models to cross-score Models $\mathcal{M}_{\theta}^{GA, t}$ with parameters $\theta^{GA, t}$ and Models $\mathcal{M}_{\theta}^{GB, t}$ with parameters $\theta^{GB, t}$, respectively. The large models are then trained via GRPO, with the rewards defined as follows:

\begin{equation}
	r^{GA, t}(x, \phi^{GB, t+1}) = 1 - \mathcal{D}_\phi^{GB, t+1}(x)
\end{equation}

\begin{equation}
	r^{GB, t}(x, \phi^{GA, t+1}) = 1 - \mathcal{D}_\phi^{GA, t+1}(x)
\end{equation}

Here, $r^{GA, t}$ is utilized for the GRPO training of Models $\mathcal{M}_{\theta}^{GA, t}$, while $r^{GB, t}$ is utilized for the GRPO training of Models $\mathcal{M}_{\theta}^{GB, t}$, yielding the following updates:

\begin{equation}
	\theta^{GA, t+1} = \theta^{GA, t} + \eta \nabla_\theta \mathcal{J}_{GRPO}(\theta^{GA, t}, t)
\end{equation}

\begin{equation}
	\theta^{GB, t+1} = \theta^{GB, t} + \eta \nabla_\theta \mathcal{J}_{GRPO}(\theta^{GB, t}, t)
\end{equation}

\section{Decoding Strategies}

\begin{table*}[!t]
	\tiny
	\resizebox{\linewidth}{!}{
		\centering
		\begin{tabular}{l|c|c|c|c}
			\toprule
			\textbf{Model} & \textbf{Temperature} & \textbf{Top-p} & \textbf{Top-k} & \textbf{Repetition Penalty} \\
			\toprule
			GPT-4o-mini & 0.6 & 1 & -1 & 1 \\
			Gemini-2.0-flash & 1 & 1 & -1 & 1 \\
			DeepSeek-V3 & 1 & 1 & -1 & 1 \\
			Qwen3-plus{*}  & 0.7 & 0.8 & -1 & 1{*} \\
			Mistral-Medium & 1 & 1 & -1 & 1 \\
			Hunyuan-TurboS & 1 & 1 & -1 & 1 \\
			Llama-3.1-8B-Instruct & 0.6 & 0.9 & -1 & 1 \\
			gemma-3-12b-it & 1 & 0.95 & 64 & 1 \\
			DeepSeek-R1-0528-Qwen3-8B & 0.6 & 0.95 & -1 & 1 \\
			Qwen3-8B & 0.6 & 0.95 & 20 & 1 \\
			Ministral-8B-Instruct-2410 & 1 & 1 & -1 & 1 \\
			Hunyuan-7B-Instruct & 0.7 & 0.8 & 20 & 1.05 \\
			\bottomrule
		\end{tabular}
	}
	\caption{Decoding strategies used for each generative model. 
		Qwen3-plus{*} additionally uses a presence penalty of 1.5, which is not listed in the table. 
		A value of -1 indicates that the corresponding parameter is not used (i.e., disabled) for that model.}
	\label{tab:decoding_strategies}
\end{table*}

\label{sec:decode}
Decoding strategies play a crucial role in controlling the behavior of large language models (LLMs) during text generation, by modulating the probability distribution over the vocabulary to balance diversity and coherence. 

Our goal is to generate more aligned machine texts. We employed various decoding strategies, as detailed in Table \ref{tab:decoding_strategies}. For open-source models, the official recommended decoding parameters are typically provided in the \texttt{generation\_config.json} file to ensure optimal or near-optimal generation quality. For commercial models, such recommendations are usually disclosed in their Technical Reports; for instance, the recommended parameters for Qwen3-Plus are specified in the Qwen3 Technical Report \cite{Yang2025Qwen3TR}. In cases where specific recommended parameters are unavailable, we follow empirical sub-optimal or near-sub-optimal practices by adopting the default configurations of vLLM (e.g., setting top\_k to -1 and keeping other parameters at 1), which generally yield diverse and stable results.

\subsection{Temperature}
\label{subsec:temperature}
Temperature $T > 0$ is a scaling factor that adjusts the sharpness of the token probability distribution. Given raw logits $x_i \in \mathbb{R}^d$ for token $i$, the probability $p_i$ is computed as:
$$p_i = \frac{\text{exp}(x_i / T)}{\sum_j\text{exp}(x_j / T)}$$
A higher $T$ flattens the distribution to promote diverse generation, while a lower $T$ sharpens the distribution, favoring high-probability tokens for more deterministic outputs.

\subsection{Top-k and Top-p Sampling}
\label{subsec:topk_topp}
Top-$k$ and top-$p$ (nucleus sampling) are two constrained sampling strategies that restrict the candidate token set to reduce irrelevant generation.

For top-$k$ sampling, the model only samples from the $k$ most probable tokens (sorted by $\text{exp}(x_i/T)$). Let $S_k$ denote this set of top-$k$ tokens. The adjusted probability is:
$$p_i = \begin{cases} 
\frac{\text{exp}(x_i / T)}{\sum_{j \in S_k}\text{exp}(x_j / T)} & \text{if } i \in S_k \\
0 & \text{otherwise}
\end{cases}$$
A larger $k$ increases diversity, while $k=1$ reduces to greedy decoding.

For top-$p$ sampling, the candidate set $S_p$ is defined as the smallest subset of tokens whose cumulative probability exceeds a threshold $p \in (0,1]$. The probability distribution is:
$$p_i = \begin{cases} 
\frac{\text{exp}(x_i / T)}{\sum_{j \in S_p}\text{exp}(x_j / T)} & \text{if } i \in S_p \\
0 & \text{otherwise}
\end{cases}$$
This strategy adapts the size of the candidate set dynamically based on probability distribution.

\subsection{Penalty}
\label{subsec:penalty}
Penalty mechanisms are designed to discourage repetitive token generation by modifying the logit-based probability distribution, with three common variants: repetition penalty, presence penalty, and frequency penalty. Let $g$ represent the list of previously generated tokens.

Repetition penalty \citep{keskar2019ctrl} applies a multiplicative adjustment to the logits of tokens in $g$. The modified probability is:
$$p_i = \frac{\text{exp}(x_i / (T \cdot I(i\in g)))}{\sum_j\text{exp}(x_j / (T \cdot I(j\in g)))}$$
where $I(c) = \theta$ if condition $c$ is true (token $i$ is in $g$) and $I(c)=1$ otherwise, with $\theta > 1$ as the penalty coefficient.

Presence penalty (implemented by OpenAI) uses an additive penalty to reduce the likelihood of any token that has appeared in $g$, regardless of occurrence frequency:
$$p_i = \frac{\text{exp}((x_i / T) - \alpha \cdot I(i\in g))}{\sum_j\text{exp}((x_j / T) - \alpha \cdot I(j\in g))}$$
where $\alpha > 0$ is the penalty weight, and $I(i\in g)$ is an indicator function that equals 1 if token $i$ is in $g$, and 0 otherwise.

Frequency penalty extends presence penalty by penalizing tokens in proportion to their occurrence frequency in $g$. Let $f_i$ denote the frequency of token $i$ in $g$. The adjusted probability is:
$$p_i = \frac{\text{exp}((x_i / T) - \beta \cdot f_i \cdot I(i\in g))}{\sum_j\text{exp}((x_j / T) - \beta \cdot f_j \cdot I(j\in g))}$$
where $\beta > 0$ is the penalty weight, so tokens with higher frequency receive stronger penalties.

\section{Evaluation Detector Details}
\label{sec:detector detail}

\begin{table}[!t]
	\centering 
	\tiny
	\resizebox{\linewidth}{!}{
		\begin{tabular}{c|c} 
			\toprule
			\textbf{Detector} & \textbf{Default Threshold}\\
			\midrule
			R-B GPT2 & 0.5\\
			R-L GPT2 & 0.5\\
			R-B CGPT & 0.5\\
			RADAR & 0.5\\
			SCRN & 0.5\\
			DETree & 0.03\\
			Fast DetectGPT & 0.5\\
			LLMDet & 0.5\\
			Binoculars & 0.901531074927684\\
			DALD & 0.5\\
			GECSCORE & 0.9243697428995128\\
			R-B MPU & 0.5\\
			\bottomrule
		\end{tabular}
	}
	\caption{Default thresholds of evaluation detectors. It should be noted that Binoculars Score is a human-like score, where texts exceeding the threshold are classified as HWT. In contrast, others are machine-like scores (or specifically, RoBERTa-based detector consists of two-dimensional probability scores, the dimension corresponding to the machine-like likelihood can be selected).}
	\label{tab:detector_thresholds}
\end{table}

In this section we provide a detailed description of all evaluation detectors used in the evaluations of the MAGA dataset. Table \ref{tab:detector_thresholds} presents the default thresholds for these detectors, which are established by their respective original authors.

\paragraph{R-B GPT2 \& R-L GPT2} \cite{solaiman-etal-2019} These two detectors\footnote{We download the model hosted by OpenAI from the following links \url{https://openaipublic.azureedge.net/gpt-2/detector-models/v1/detector-base.pt} and \url{https://openaipublic.azureedge.net/gpt-2/detector-models/v1/detector-large.pt}} are RoBERTa-base and RoBERTa-large models respectively, fine-tuned on GPT2 open-domain outputs, which were generated via greedy decoding, top-k=50 sampling, and full random sampling. As a long-standing baseline, both base and large variants are used for comparison.

\paragraph{R-B CGPT} \cite{guo2023close} This detector is a RoBERTa-base model fine-tuned on the HC3 dataset (roughly 27,000 human-ChatGPT answer pairs across multiple domains).

\paragraph{RADAR} \cite{hu2023radar} This detector is a RoBERTa-based model trained via adversarial reinforcement learning between a RoBERTa classifier and a Vicuna-7B paraphraser. This framework enables the detector to better distinguish between human text, machine-generated text, and paraphrased machine outputs.

\paragraph{GLTR} \cite{gehrmann-etal-2019-gltr} This detector was originally an auxiliary interface for human detection, now a standard robustness baseline. It evaluates token likelihood via an LM, bins tokens by likelihood, and uses these as features. We adopt its default settings (rank=10 cutoff, GPT2 small) for detection.

\paragraph{FastDetectGPT} \cite{bao2024fastdetectgpt} This detector is an optimized DetectGPT \cite{pmlr-v202-mitchell23a} variant that achieves 340x faster inference without any reduction in accuracy. We use their default settings: GPT-Neo-2.7B as the scoring model and GPT-J-7B as the reference model. We felt that this was a reasonable choice, as neither model was used for generating in our dataset.

\paragraph{Binoculars} \cite{hans2024spotting} This detector uses the ratio of perplexity to cross-entropy between two similar language models as the detection metric. Adopts official default Falcon-7B and Falcon-7B-Instruct models; neither was used for dataset generation, ensuring fairness.

\paragraph{LLMDet} \cite{wu-etal-2023-llmdet} This detector computes proxy-perplexity (n-gram sampling-based approximation) of input text via 10 small language models, using these as detection features. None of the language models were used for dataset generation, ensuring fair comparison.

\paragraph{SCRN} \cite{Huang2024AreAT} This detector is a robust model with a reconstruction network for denoising and siamese calibration. It resists adversarial perturbations across in-domain, cross-domain, and mixed-source scenarios.

\paragraph{DETree} \cite{He2025DETreeDH} This detector is a representation learning-based model that models text category relationships via a Hierarchical Affinity Tree and contrastive loss. Trained on RealBench, it excels at hybrid text detection and out-of-distribution generalization. RealBench aggregates several existing datasets, including MAGE \cite{li2024mage}, M4 \cite{wang-etal-2024-m4}, M4GT-Bench \cite{wang2024m4gt}, TuringBench \cite{uchendu2021turingbench}, OUTFOX \cite{Koike_Kaneko_Okazaki_2024}, and RAID \cite{dugan2024raid}, and undergoes data augmentation by simulating various realistic collaborative writing paradigms such as rewriting, expansion, polishing, and translation. DETree represents the text in RealBench as vectors and stores them in a Faiss vector database. During the specific detection phase, DETree transforms the input text into a vector representation and retrieves $k$ nearest neighbors via the k-nearest neighbors algorithm, which subsequently yields the probability of the text being human-written or AI-generated through a probability-weighted voting mechanism. This underlying intuition is conceptually similar to Retrieval-Augmented Generation (RAG, \citet{NEURIPS2020_6b493230}. We set the parameter $k$ to their default value of 10, and the reference vectors constructed from RealBench are sourced from DETree's official Hugging Face repository \footnote{\url{https://huggingface.co/datasets/heyongxin233/RealBench}}.

\paragraph{DALD} \cite{Zeng2024ImprovingLD} This distribution-aligned detector mitigates the distribution mismatch in black-box scenarios by fine-tuning a surrogate model with limited target samples. It serves as a flexible enhancement to boost zero-shot detectors against model iterations. In this work, we adopt its default and preferred configuration, which operates on top of Fast-DetectGPT \cite{bao2024fastdetectgpt} for conditional probability curvature estimation.

\paragraph{GECSCORE} \cite{Wu2024WhoWT} This detector is a zero-shot model using grammar error correction similarity. It distinguishes texts by comparing input with corrected versions, boasting strong generalization and paraphrase attack resistance.

\paragraph{RoBERTa-MPU-zhv3} \cite{tian2024multiscale} This detector is a multiscale model with length-sensitive PU loss and text multiscaling. It improves short-text detection while maintaining long-text performance. We select its Chinese version as an additional evaluation detector, for MAGA-cn evaluation.

It should be noted that the base language models employed by nearly all detectors are primarily competent only in English. In some cases, these base language models lack any Chinese language capability whatsoever; in others, their Chinese recognition ability is weak like low-resource languages, with no practical proficiency in Chinese. Among all our selected evaluation detectors, R-B CGPT offers a standalone Chinese version, while GECScore can be directly applied to Chinese detection tasks simply by translating the prompt text into Chinese. Notably, Binoculars is capable of Chinese recognition, and despite its base language models Falcon-7B and Falcon-7B-Instruct lack functional Chinese language proficiency, we found that its detection performance is surprisingly satisfactory. However, we did not individually verify whether other metric-based detectors still retain a certain level of capability in detecting Chinese text. As a supplementary addition for training-based detectors, we have additionally incorporated RoBERTa-MPU-zhv3 \cite{tian2024multiscale}. The four aforementioned models are utilized by us for Chinese detection tasks. Then the number of neural-based and metric-based detectors used in our MAGA-cn evaluation is consistent.

\section{Prompt Details}
\label{sec:prompt detail}
\paragraph{Original Prompt and BPO-Optimized Prompt} Table \ref{tab:generation prompts} and Table \ref{tab:generation prompts-cn} present the English and Chinese versions of the orginal prompts respectively. These original prompts serve as the user prompts to construct the MGB, MAGA-extra-GDARL-CDMMR, and MAGA-extra-roleplaying splits (roleplaying has additional system prompts). Also, these original prompts are utilized to generate the initial MGT for the MAGA-extra-self-refine split before the feedback and refinement stage. BPO automatically optimizes these original prompts and the MAGA-extra-BPO split adopts the BPO-optimized prompts as user prompts (Figures \ref{fig:sample-mgb} and \ref{fig:sample-bpo} serve as a concrete example.). Similarly, the MAGA split (integrates all our alignment methods) also utilizes BPO-optimized prompts as user prompts for generating the initial text prior to get feedback and refinement.

\paragraph{MAGA Additional Instruction} The MAGA Additional Instruction, as detailed in Table \ref{tab:generation prompts} and Table \ref{tab:generation prompts-cn}, functions as a suffix to the original prompt. This design is inspired by RAID \cite{dugan2024raid}; for instance, the additional instruction for the Reddit domain, "Do not repeat the title.", is also utilized within RAID's Reddit prompts. We observe that LLMs tend to include introductory remarks during text generation, often incorporating the title in unpredictable ways. Such patterns are absent in human-written texts and are nearly impossible to remove by rule-based cleaning. Despite repeated attempts to provide explicit instructions forbidding prologues or title repetitions, models often fail to comply. We find that as prompt complexity and length increase, model adherence tends to deteriorate. Conversely, concise instructions such as Do not repeat the title'' suggested by RAID \cite{dugan2024raid} result in a lower probability of the model outputting either the title or unnecessary introductory phrases. Furthermore, as illustrated in Figures \ref{fig:sample-mgb} and \ref{fig:sample-bpo}, the prompts optimized by BPO also incorporate the MAGA Additional Instruction. To maintain clarity regarding the core BPO mechanism, this detail is omitted in Figure \ref{fig:MAGA Pipeline}. Specifically, the implementation involves a two-step process. The Original User Prompt in Table \ref{tab:generation prompts} and Table \ref{tab:generation prompts-cn}, which is used directly in the MGB split, already contains the MAGA Additional Instruction. For the BPO optimization, we first remove this instruction to facilitate the optimization process. Once BPO generates the refined prompt, the MAGA Additional Instruction is re-appended as a suffix.

\paragraph{Self-Refine Prompt} Table \ref{tab:self refine prompts} and Table \ref{tab:self refine prompts Chinese} illustrate the Self-Refine prompts for English and Chinese contexts. For the initial raw text generated before refinement, a Feedback Prompt is first introduced to produce specific feedback. This feedback is subsequently embedded into the Refine Prompt to generate the final self-refined text. The self-refinement prompts are applied within both the MAGA-extra-self-refine and MAGA splits. To facilitate a clearer understanding of the operational process, specifically we illustrate the procedure using the ChatML template. The large language model first receives the input sequence '<|im\_start|>user\textbackslash n\{user\_prompt\}<|im\_end|>' to generate the initial text. Subsequently, we employ a multi-turn dialogue approach where the input '<|im\_start|>user\textbackslash n\{user\_prompt\}<|im\_end|>\textbackslash n
<|im\_start|>assistant\textbackslash n\{initial\_text\}<|im\_end|>\textbackslash n
<|im\_start|>user\textbackslash n\{feedback\_prompt\}<|im\_end|>' yields the feedback response. The process then reverts to the second dialogue turn, where the obtained feedback is embedded into the Refine Prompt to produce the polished output. Specifically, the model processes the sequence '<|im\_start|>user\textbackslash n\{user\_prompt\}<|im\_end|>\textbackslash n
<|im\_start|>assistant\textbackslash n\{initial\_text\}<|im\_end|>\textbackslash n
<|im\_start|>user\textbackslash n\{refine\_prompt\}<|im\_end|>' to generate the final refined text. The two-turn dialogues can be compressed into a single turn using template. Two-turn may be clearer to LLM. Actually, this choice is not important, merely a minor implementation detail worth mentioning. Figures \ref{fig:sample-self-refine}, \ref{fig:sample-self-refine-detail}, \ref{fig:sample-maga}, \ref{fig:sample-maga-self-refine-detail} provide specific examples of the outputs.

\paragraph{Roleplaying Prompt} Table \ref{tab:role playing prompts} displays samples of our role-playing prompts, which function as system prompts for the MAGA-extra-roleplaying and MAGA splits. We provide a detailed introduction to our role-playing prompts' characteristics and usage methods in Appendix \ref{sec:roleplaying detail}.

It should be noted that, as we have repeatedly mentioned across various sections, such as Appendices \ref{sec:roleplaying detail} and \ref{sec:GDARL detail}, that our prompt set is pre-constructed; in this section, we provide a systematic explanation of this aspect. Taking the MAGA English Subset as an example, all splits contain the same 72k Human-Written Text (HWT) samples (see Table \ref{tab:data split distribution on train set}), each associated with a title annotation. We consistently utilize these titles to construct the prompt sets. Specifically, embedding the 72k titles into the original prompt template yields the original prompt set for the MGB split. Each original prompt generates one MGT, and the combination of these 72k MGTs with their corresponding 72k HWTs constitutes the MGB split. Similarly, for the MAGA-extra-roleplaying split, each of the 72k original prompts is randomly assigned a roleplaying system prompt according to the procedure in Appendix \ref{sec:roleplaying detail}. In the MAGA-extra-BPO split, the 72k original prompts are automatically optimized by BPO model and the 72k optimized prompts are served as the generation prompts. For the MAGA-extra-GDARL-CDMMR split, we fix the set of 72k prompts to generate MGT for training the RoBERTa detector and to serve as the prompt set for reinforcement learning as detailed in Appendix \ref{sec:GDARL detail}. The MAGA split represents a synthesis of these methodologies, where all 72k prompts undergo BPO optimization and include roleplaying system prompts before executing the GDARL-CDMMR algorithm to train the model, followed by self-refinement to enhance the inference process. Consequently, we emphasize that our prompt sets are pre-constructed and remain invariant, maintaining a strict one-to-one correspondence where each HWT sample relates to exactly one prompt and one generated MGT.

\section{Data Clean Details}
\label{sec:data clean detail}
It is essential to clarify that we perform identical data cleaning procedures across all texts in the entire MAGA dataset. Consequently, the cleaning process is orthogonal to our comparative experiments and does not influence the core conclusions. For instance, the primary experiments in the main paper compare the attack and generalization performance between the MGB and MAGA splits before and after human-like alignment. Since both splits undergo the same cleaning pipeline, the results remain consistent. Our cleaning efforts are solely intended to ensure the overall quality of the dataset.

The source HWTs we collected are generally clean, as they were pre-processed by their respective publishers, particularly for the English data. Thus, only minimal further inspection and cleaning are required. For machine-generated texts (MGTs), inspired by M4 \cite{wang-etal-2024-m4}, we observe that simple artifacts such as multiple newlines and bullet points occasionally persist. These patterns may allow detectors to exploit superficial shortcuts rather than learning robust features for generalization.

We checked the text quality by randomly sampling 10 to 20 cases for each domain and generator. Our key observations and subsequent processing steps are detailed below.

\subsection{HWT Clean}

Based on the sample inspection findings, the major domains under data cleaning and their corresponding cleaning details are presented below.

\paragraph{S2ORC} We unify the format by removing leading words such as "Abstract" and "abstract" in the string prefixes, along with any redundant whitespace.

\paragraph{Wikipedia} Although the original publishers removed references (e.g., [1], [2]), URLs, multiple newlines, and bullet points, we identify some instances where article titles appear in the string prefixes. We remove these titles to ensure consistency.

\paragraph{wikiHow} It should be noted that the original publishers retained only the first paragraph, the introduction of a wikiHow article, and this affects how we clean MGTs on wikiHow. In addition to processing the HWTs, the original publisher removed the "How to" prefix present in every wikiHow title from the title field. We restore the "How to" prefix to the title field. We believe this prefix potentially assists large language models in contextual understanding.

\paragraph{Baidu Tieba and Baidu Zhidao} These domains are part of the DuReader2.0 \cite{HongzheBi_2023} dataset. We separate the data for these two sources. For Baidu Tieba, where multiple user replies are provided for each post, we retain only the first reply because subsequent entries may often consist of nested replies.

\paragraph{Weibo Review} We remove unnecessary usernames and Chinese colons from the prefixes of each review string.

\paragraph{CLTS} The original data contains numerous web-based whitespace characters, such as '\textbackslash t', '\textbackslash v', '\textbackslash f', '\textbackslash xa0', '\textbackslash u2000' to '\textbackslash u200b', "\&nbsp;", "\&ensp;", "\&thinsp;". We remove these characters to the greatest extent possible. Furthermore, we identify and remove specific journalist remarks at the end of some news articles, such as "（澎湃新闻记者 xxx 整理）", as these serve as explicit hints of HWT.

\subsection{MGT Clean}

Based on the sample inspection findings, the major domains under data cleaning and their corresponding cleaning details are presented below.

\paragraph{S2ORC} We observe that machines frequently generate various title-like prefixes such as "**Abstract**". We remove "Abstract", "abstract", all special characters such as '*' and '\#', and redundant newlines from the prefixes. Notably, machine-generated S2ORC abstracts frequently begin with the phrase "This paper". Based on sampling survey, we found that---while not numerous---a distinct subset of scientific journal abstracts in human written papers indeed begin with the phrase "This paper." Inspired by the additional instructions in RAID \cite{dugan2024raid} (e.g., they appended the additional instruction "Do not repeat the title." to the prompt for the Reddit domain), we append "It is preferable not to start with \textbackslash "This paper\textbackslash "" to our prompt for S2ORC (specifically detailed in Appendix \ref{sec:prompt detail}.).

\paragraph{Wikipedia} To align with HWTs, we remove redundant titles from the string prefixes and eliminate markdown syntax, including bullet points (-), bold markers (**), and header symbols (\#, \#\#, \#\#\#).

\paragraph{wikiHow} We remove redundant titles and whitespace from the string prefixes and retain only the first paragraph to align with HWTs.

\paragraph{CSL} We remove all occurrences of "《\{title\}》" and "摘要：" from the string prefixes, along with redundant whitespace. Similar to the S2ORC English abstracts, we add the additional instruction to the CSL prompt: "最好不要以“本文”开头。".

\paragraph{Baidu Baike} We remove titles appearing in various formats from the string prefixes (including "标题：\{title\}", "**\{title\}**", "【\{title\}】", "\# \{title\}") and markdown symbols, including bullet points (-), bold markers (**), and header symbols (\#, \#\#, \#\#\#). We also eliminate leading numbers in paragraphs (e.g., 1., 2.) and consolidate multiple consecutive newlines into a single newline.

\paragraph{Zhihu} To align with HWTs, we replace all newlines with spaces. Specifically, multiple consecutive newlines are collapsed into a single space.

\paragraph{CLTS} To align with HWTs, we remove all newlines from the MGTs.

\section{Introduction to External Datasets for Generalization Ability Evaluation}
\label{introduction to external datasets}
We select the optimized versions of four datasets, M4 \cite{wang-etal-2024-m4}, M4GT \cite{wang2024m4gt}, MAGE \cite{li2024mage}, and HC3 \cite{guo2023close}, as external datasets to evaluate the generalization ability of detectors. These datasets are four representatives of the most diverse and challenging datasets among all benchmarks listed in Table \ref{tab:all-datasets}.

The optimized versions specifically refer to the adjusted and refined releases provided for academic competition workshops focusing on MGT detection. For M4 \cite{wang-etal-2024-m4}, we adopt the test set they released for SemEval-2024 Task 8 \cite{wang2024semeval}. We denoted it as S-M4. Detailed statistics regarding its size and domain distribution are available in the corresponding overview paper \cite{wang2024semeval} and the official repository\footnote{\url{https://github.com/mbzuai-nlp/SemEval2024-task8}}. Regarding M4GT \cite{wang2024m4gt}, MAGE \cite{li2024mage}, and HC3 \cite{guo2023close}, we utilize the validation sets from the COLING 2025 Workshop on MGT Detection Task 1 \cite{wang-etal-2025-genai}, which we denote as C-M4GT, C-MAGE, and C-HC3, respectively. Detailed statistics regarding its size and domain distribution of these datasets are available in the corresponding overview paper \cite{wang-etal-2025-genai} and the official repository\footnote{\url{https://github.com/mbzuai-nlp/COLING-2025-Workshop-on-MGT-Detection-Task1}}.

For English, we use the original English versions of these datasets. For Chinese, we extract all Chinese data from their original multilingual releases based on provided annotations. Notably, C-MAGE contains no Chinese data, so it is omitted from the experiments on our Chinese version.

\begin{table*}[!t]
	\resizebox{\linewidth}{!}{
		\centering
		%    	\small
		\begin{tabular}{l|c|c|c|c|c|c|c|c|c}
			\toprule
			& \multicolumn{2}{c|}{ACC} 
			& \multicolumn{2}{c|}{TPR} 
			& \multicolumn{1}{c|}{TNR} 
			& \multicolumn{2}{c|}{AUC} 
			& \multicolumn{2}{c}{ACC(@FPR=5\%)} \\
			\midrule
			Alignment? & \xmark & \cmark & \xmark & \cmark & - & \xmark & \cmark & \xmark & \cmark \\
			\midrule
			R-B CGPT CN & 52.19 & 51.81\textcolor{red}{(0.38$\downarrow$)} & \textbf{99.54} & \textbf{98.79}\textcolor{red}{(0.75$\downarrow$)} & 4.83 & 73.01 & 70.88\textcolor{red}{(2.13$\downarrow$)} & 53.64 & 53.06\textcolor{red}{(0.58$\downarrow$)} \\
			R-B MPU zhv3 & 53.38 & 52.86\textcolor{red}{(0.52$\downarrow$)} & 7.06 & 6.03\textcolor{red}{(1.03$\downarrow$)} & \textbf{99.69} & 79.89 & 76.78\textcolor{red}{(3.11$\downarrow$)} & 57.13 & 56.24\textcolor{red}{(0.89$\downarrow$)} \\
			\midrule
			neural.avg & 52.78 & 52.34\textcolor{red}{(0.45$\downarrow$)} & 53.30 & 52.41\textcolor{red}{(0.89$\downarrow$)} & 52.26 & 76.45 & 73.83\textcolor{red}{(2.62$\downarrow$)} & 55.39 & 54.65\textcolor{red}{(0.73$\downarrow$)} \\
			\midrule
			Binoculars & \textbf{71.84} & \textbf{67.54}\textcolor{red}{(4.30$\downarrow$)} & 48.67 & 40.08\textcolor{red}{(8.59$\downarrow$)} & 95.00 & \textbf{83.38} & \textbf{81.69}\textcolor{red}{(1.69$\downarrow$)} & \textbf{71.21} & \textbf{66.71}\textcolor{red}{(4.50$\downarrow$)} \\
			GECScore & 49.59 & 49.59\textcolor{red}{(0.00$\downarrow$)} & 0.01 & 0.01\textcolor{red}{(0.00$\downarrow$)} & 99.17 & 51.33 & 48.94\textcolor{red}{(2.39$\downarrow$)} & 52.08 & 51.25\textcolor{red}{(0.83$\downarrow$)} \\
			\midrule
			metric.avg & 60.71 & 58.57\textcolor{red}{(2.15$\downarrow$)} & 24.34 & 20.05\textcolor{red}{(4.30$\downarrow$)} & 97.09 & 67.36 & 65.32\textcolor{red}{(2.04$\downarrow$)} & 61.65 & 58.98\textcolor{red}{(2.67$\downarrow$)} \\
			\midrule
			avg & 56.75 & 55.45\textcolor{red}{(1.30$\downarrow$)} & 38.82 & 36.23\textcolor{red}{(2.59$\downarrow$)} & 74.67 & 71.90 & 69.57\textcolor{red}{(2.33$\downarrow$)} & 58.52 & 56.82\textcolor{red}{(1.70$\downarrow$)} \\
			\bottomrule
	\end{tabular}}
	\caption{Attack Experiment on MAGA-cn. An "\xmark" in the "Alignment" column indicates the MGB-cn split, while the opposite indicates the MAGA-cn split.}
	\label{tab:MAGA-cn Bench}
\end{table*}

\begin{table*}[!t]
	% \small
	\resizebox{\linewidth}{!}{
		\centering
		\begin{tabular}{c|c|c|c|c|c|c|c|c}
			\toprule
			& \multicolumn{2}{c|}{\textbf{S-M4-CN}} 
			& \multicolumn{2}{c|}{\textbf{C-M4GT-CN}}
			& \multicolumn{2}{c|}{\textbf{C-HC3-CN}}
			& \multicolumn{2}{c}{\textbf{AVERAGE}}
			\\
			\toprule
			Metric & AUC & ACC(5\%) & AUC & ACC(5\%) & AUC & ACC(5\%) & AUC & ACC(5\%) \\
			\midrule
			R-B CGPT CN     & \underline{99.24}     & \underline{95.19}       & \underline{98.95}      & \underline{94.29}       & 99.96\textsuperscript{*}       & 97.15\textsuperscript{*}      & \underline{99.38}       & \underline{95.54} \\
			R-B MPU zhv3    & 95.94     & 89.10       & 95.40      & 87.09       & 95.09       & 88.65      & 95.48       & 88.28 \\
			Binoculars      & 97.87     & 94.18       & 97.73      & 94.21       & 97.02       & 92.21      & 97.54       & 93.53 \\
			GECScore        & 45.34     & 51.26       & 54.69      & 47.06       & 47.94       & 56.64      & 49.32       & 51.65 \\
			\midrule
			\textbf{R-B MGB CN (Ours)}   & 98.68     & 93.77       & 98.04      & 91.42       & \underline{99.04}       & \underline{94.90}      & 98.59       & 93.36 \\
			\textbf{R-B MAGA CN (Ours)}  & \textbf{99.41} & \textbf{96.33} & \textbf{99.42} & \textbf{96.23} & \textbf{99.67}       & \textbf{96.64}      & \textbf{99.50}       & \textbf{96.40} \\
			\textbf{$\Delta$} & \textbf{0.73} & \textbf{2.56} & \textbf{1.38} & \textbf{4.81} & \textbf{0.63} & \textbf{1.74} & \textbf{0.91} & \textbf{3.04} \\
			\bottomrule
	\end{tabular}}
	\caption{Three Chinese datasets including S-M4-CN, C-M4GT-CN, and C-HC3-CN. ACC(5\%) here is a short for ACC(@FPR=5\%). Bold indicates the largest value, underline indicates the second largest value. R-B CGPT CN is trained on c-HC3-CN}
	\label{tab:MAGA Generalization CN}
\end{table*}

\begin{table*}[!t]
	\resizebox{\linewidth}{!}{
		% \small
		\centering
		\begin{tabular}{l|c|c|c|c|c|c}
			\toprule
			& \multicolumn{3}{c|}{\textbf{Attack}} 
			& \multicolumn{3}{c}{\textbf{Generalization}}  \\
			\toprule
			Metric & TPR & AUC & ACC(@FPR=5\%) & TPR & AUC & ACC(@FPR=5\%) \\
			\midrule
			MGB\textsuperscript{*} & 54.09 & 71.41 & 67.38 & 50.34 & 80.88 & 73.32 \\
			\midrule
			MGB /w GDARL-CDMMR\textsuperscript{*} & 50.92\textcolor{red}{(3.17$\downarrow$)} & 68.83\textcolor{red}{(2.58$\downarrow$)} & 66.25\textcolor{red}{(1.13$\downarrow$)} & 57.24\textcolor{green}{(6.90$\uparrow$)} & 82.05\textcolor{green}{(1.17$\uparrow$)} & 74.04\textcolor{green}{(0.72$\uparrow$)} \\
			MGB /w roleplaying\textsuperscript{*} & 48.76\textcolor{red}{(5.33$\downarrow$)} & 66.43\textcolor{red}{(4.98$\downarrow$)} & 64.71\textcolor{red}{(2.67$\downarrow$)} & 60.68\textcolor{green}{(10.34$\uparrow$)} & 83.37\textcolor{green}{(2.49$\uparrow$)} & 75.36\textcolor{green}{(2.04$\uparrow$)} \\
			MGB /w BPO\textsuperscript{*} & 50.12\textcolor{red}{(3.97$\downarrow$)} & 68.07\textcolor{red}{(3.34$\downarrow$)} & 65.48\textcolor{red}{(1.90$\downarrow$)} & 57.49\textcolor{green}{(7.15$\uparrow$)} & 82.84\textcolor{green}{(1.96$\uparrow$)} & 74.82\textcolor{green}{(1.50$\uparrow$)} \\
			MGB /w self-refine\textsuperscript{*} & 48.88\textcolor{red}{(5.21$\downarrow$)} & 66.33\textcolor{red}{(5.08$\downarrow$)} & 64.43\textcolor{red}{(2.95$\downarrow$)} & 61.41\textcolor{green}{(11.07$\uparrow$)} & 83.83\textcolor{green}{(2.95$\uparrow$)} & 75.11\textcolor{green}{(1.79$\uparrow$)} \\
			\midrule
			MGB /w BPO /w GDARL-CDMMR & 48.89\textcolor{red}{(5.20$\downarrow$)} & 67.05\textcolor{red}{(4.36$\downarrow$)} & 64.93\textcolor{red}{(2.45$\downarrow$)} & 60.39\textcolor{green}{(10.05$\uparrow$)} & 83.33\textcolor{green}{(2.45$\uparrow$)} & 74.96\textcolor{green}{(1.64$\uparrow$)} \\
			MGB /w BPO /w self-refine & 47.45\textcolor{red}{(6.64$\downarrow$)} & 65.40\textcolor{red}{(6.01$\downarrow$)} & 64.02\textcolor{red}{(3.36$\downarrow$)} & 63.92\textcolor{green}{(13.58$\uparrow$)} & 84.25\textcolor{green}{(3.37$\uparrow$)} & 75.50\textcolor{green}{(2.18$\uparrow$)} \\
			MGB /w BPO /w roleplaying & 47.48\textcolor{red}{(6.61$\downarrow$)} & 65.43\textcolor{red}{(5.98$\downarrow$)} & 64.21\textcolor{red}{(3.17$\downarrow$)} & 64.07\textcolor{green}{(13.73$\uparrow$)} & 83.98\textcolor{green}{(3.10$\uparrow$)} & 75.42\textcolor{green}{(2.10$\uparrow$)} \\
			MGB /w roleplaying /w GDARL-CDMMR & 47.75\textcolor{red}{(6.34$\downarrow$)} & 65.11\textcolor{red}{(6.30$\downarrow$)} & 64.32\textcolor{red}{(3.06$\downarrow$)} & 64.79\textcolor{green}{(14.45$\uparrow$)} & 83.84\textcolor{green}{(2.96$\uparrow$)} & 75.34\textcolor{green}{(2.02$\uparrow$)} \\
			MGB /w roleplaying /w self-refine & 47.36\textcolor{red}{(6.73$\downarrow$)} & 65.51\textcolor{red}{(5.90$\downarrow$)} & 63.93\textcolor{red}{(3.45$\downarrow$)} & 64.10\textcolor{green}{(13.76$\uparrow$)} & 84.13\textcolor{green}{(3.25$\uparrow$)} & 75.46\textcolor{green}{(2.14$\uparrow$)} \\
			MGB /w GDARL-CDMMR /w self-refine & 47.85\textcolor{red}{(6.24$\downarrow$)} & 65.38\textcolor{red}{(6.03$\downarrow$)} & 64.00\textcolor{red}{(3.38$\downarrow$)} & 63.96\textcolor{green}{(13.62$\uparrow$)} & 84.26\textcolor{green}{(3.38$\uparrow$)} & 75.42\textcolor{green}{(2.10$\uparrow$)} \\
			\midrule
			MAGA w/o self-refine & 46.10\textcolor{red}{(7.99$\downarrow$)} & 64.22\textcolor{red}{(7.19$\downarrow$)} & 63.56\textcolor{red}{(3.82$\downarrow$)} & 67.17\textcolor{green}{(16.83$\uparrow$)} & 84.72\textcolor{green}{(3.84$\uparrow$)} & 75.92\textcolor{green}{(2.60$\uparrow$)} \\
			MAGA w/o BPO & 46.69\textcolor{red}{(7.40$\downarrow$)} & 64.56\textcolor{red}{(6.85$\downarrow$)} & 63.62\textcolor{red}{(3.76$\downarrow$)} & 66.34\textcolor{green}{(16.00$\uparrow$)} & 84.50\textcolor{green}{(3.62$\uparrow$)} & 75.88\textcolor{green}{(2.56$\uparrow$)} \\
			MAGA w/o roleplaying & 46.92\textcolor{red}{(7.17$\downarrow$)} & 64.79\textcolor{red}{(6.62$\downarrow$)} & 63.79\textcolor{red}{(3.59$\downarrow$)} & 65.45\textcolor{green}{(15.11$\uparrow$)} & 84.33\textcolor{green}{(3.45$\uparrow$)} & 75.80\textcolor{green}{(2.48$\uparrow$)} \\
			MAGA w/o GDARL-CDMMR & 46.56\textcolor{red}{(7.53$\downarrow$)} & 64.62\textcolor{red}{(6.79$\downarrow$)} & 63.76\textcolor{red}{(3.62$\downarrow$)} & 65.86\textcolor{green}{(15.52$\uparrow$)} & 84.59\textcolor{green}{(3.71$\uparrow$)} & 75.85\textcolor{green}{(2.53$\uparrow$)} \\
			\midrule
			MAGA\textsuperscript{*} & 45.59\textcolor{red}{(8.50$\downarrow$)} & 63.71\textcolor{red}{(7.70$\downarrow$)} & 63.31\textcolor{red}{(4.07$\downarrow$)} & 68.03\textcolor{green}{(17.69$\uparrow$)} & 85.03\textcolor{green}{(4.15$\uparrow$)} & 76.13\textcolor{green}{(2.81$\uparrow$)} \\
			\bottomrule
	\end{tabular}}
	\caption{A more detailed ablation study on 10\% data.}
	\label{tab:full ablation}
\end{table*}

\section{More Experiment Details}
\label{sec:more experiment}
All our MGTs are generated in a single run, and all evaluation experiments process the entire split continuously within a single execution. We supplement additional experiment details as follows:

\paragraph{Experiments on the MAGA-cn Subset} Table~\ref{tab:MAGA-cn Bench} presents the benchmarking results on the MAGA-cn split, which corresponds to Table~\ref{tab:MAGA Bench} for the English split. Table~\ref{tab:MAGA Generalization CN} displays the generalization experiments for the final trained RoBERTa detector evaluated on MAGA-cn, serving as the Chinese counterpart to Table~\ref{tab:MAGA Generalization}. The evaluation metrics and methodologies remain strictly identical to those used in the English version. At the end of Appendix~\ref{sec:detector detail}, we justify the selection of the four specific evaluation detectors, namely R-B CGPT CN, R-B MPU zhv3, Binoculars, and GECScore, because the vast majority of our selected detectors lack support for Chinese text detection. For the generalization experiments, we note at the end of Appendix~\ref{introduction to external datasets} that C-MAGE is omitted due to the absence of Chinese data.

As shown in Table~\ref{tab:MAGA-cn Bench} and Table~\ref{tab:MAGA Generalization CN}, the experimental results on the MAGA-cn split remain consistent with those observed on the MAGA split. On the MAGA-cn split, the average AUC of evaluation detectors decreases by 2.33\% compared to the MGB-cn split, where neural-based detectors and metric-based detectors drop by 2.62\% and 2.04\%, respectively. Similarly, the average ACC(@FPR=5\%) of evaluation detectors decreases by 1.70\%, with neural-based detectors dropping by 0.73\% and metric-based detectors decreasing by 2.67\%. Similar to the trends on the MAGA split, the performance decrease discrepancy between neural-based detectors and metric-based detectors on the MAGA-cn split is marginal. In generalization experiments, R-B MAGA CN improves the average AUC on external datasets by 0.91\% and the average ACC(@FPR=5\%) by 3.04\% compared to R-B MGB CN.

The relatively modest performance degradation in evaluation detectors is potentially due to the imbalanced detection capabilities of existing Chinese detectors. For instance, R-B CGPT CN exhibits a TPR close to 100\% and a TNR close to 0\%, whereas R-B MPU zhv3 and GECScore show a TNR close to 100\% and a TPR close to 0\%. This scarcity of available Chinese detectors might lead to an insufficient evaluation of the MAGA-cn split, which constitutes one of our limitations. Regarding generalization experiments, R-B MGB CN already achieves an average AUC of 98.59\%, which restricts the margin for potential improvement. Consequently, the AUC enhancement achieved by R-B MAGA CN may be more substantial than the absolute numerical increase suggests. But anyway, a joint analysis of Table~\ref{tab:MAGA-cn Bench} and Table~\ref{tab:MAGA Generalization CN} may reveal that both MGB-cn and MAGA-cn pose significantly greater detection challenges compared to existing Chinese datasets.

\paragraph{Experimental Results on MAGA-extra Splits} The ablation studies presented in Table~\ref{tab:ablation} provide a consolidated summary of the average experimental results across all individual MAGA-extra splits. Tables~\ref{tab:MAGA-extra-GDARL-CDMMR Bench}, \ref{tab:MAGA-extra-roleplaying Bench}, \ref{tab:MAGA-extra-BPO Bench}, \ref{tab:MAGA-extra-self-refine Bench} report the performance of evaluation detectors across all four extra splits under the English MAGA subset, which include MAGA-extra-CDMMR, MAGA-extra-roleplaying, MAGA-extra-BPO, and MAGA-extra-self-refine. Analogously, Tables~\ref{tab:MAGA-cn-extra-GDARL-CDMMR Bench}, \ref{tab:MAGA-cn-extra-roleplaying Bench}, \ref{tab:MAGA-cn-extra-BPO Bench}, \ref{tab:MAGA-cn-extra-self-refine Bench} detail the detection results on the four corresponding Chinese extra splits. Minor performance fluctuations occur occasionally in these experiments, where specific evaluation detectors show either a marginal decrease or a slight increase in detection efficacy on certain extra splits.

\paragraph{Detailed Experimental Results on External Datasets} Table~\ref{tab:MAGA Generalization} and Table~\ref{tab:MAGA Generalization CN} provide the comprehensive generalizations for the English and Chinese subsets, respectively, focusing primarily on the critical metrics of AUC and ACC(@FPR=5\%). In the Tables described in the below in this paragraph, we provide additional documentation for the complete evaluation results, which encompass all four metrics of ACC, TPR, AUC, and ACC(@FPR=5\%). Specifically, Tables~\ref{tab:S-M4 Performance}, \ref{tab:C-M4GT Performance}, \ref{tab:C-MAGE Performance}, \ref{tab:C-HC3 Performance} compile the granular outcomes on S-M4, C-M4GT, C-MAGE, and C-HC3. Furthermore, Tables~\ref{tab:S-M4-CN Performance}, \ref{tab:C-M4GT-CN Performance}, \ref{tab:C-HC3-CN Performance} display the detailed breakdown on S-M4-CN, C-M4GT-CN, and C-HC3-CN, respectively.

\begin{figure}[!t]
	\centering
	\includegraphics[width=1.0\linewidth]{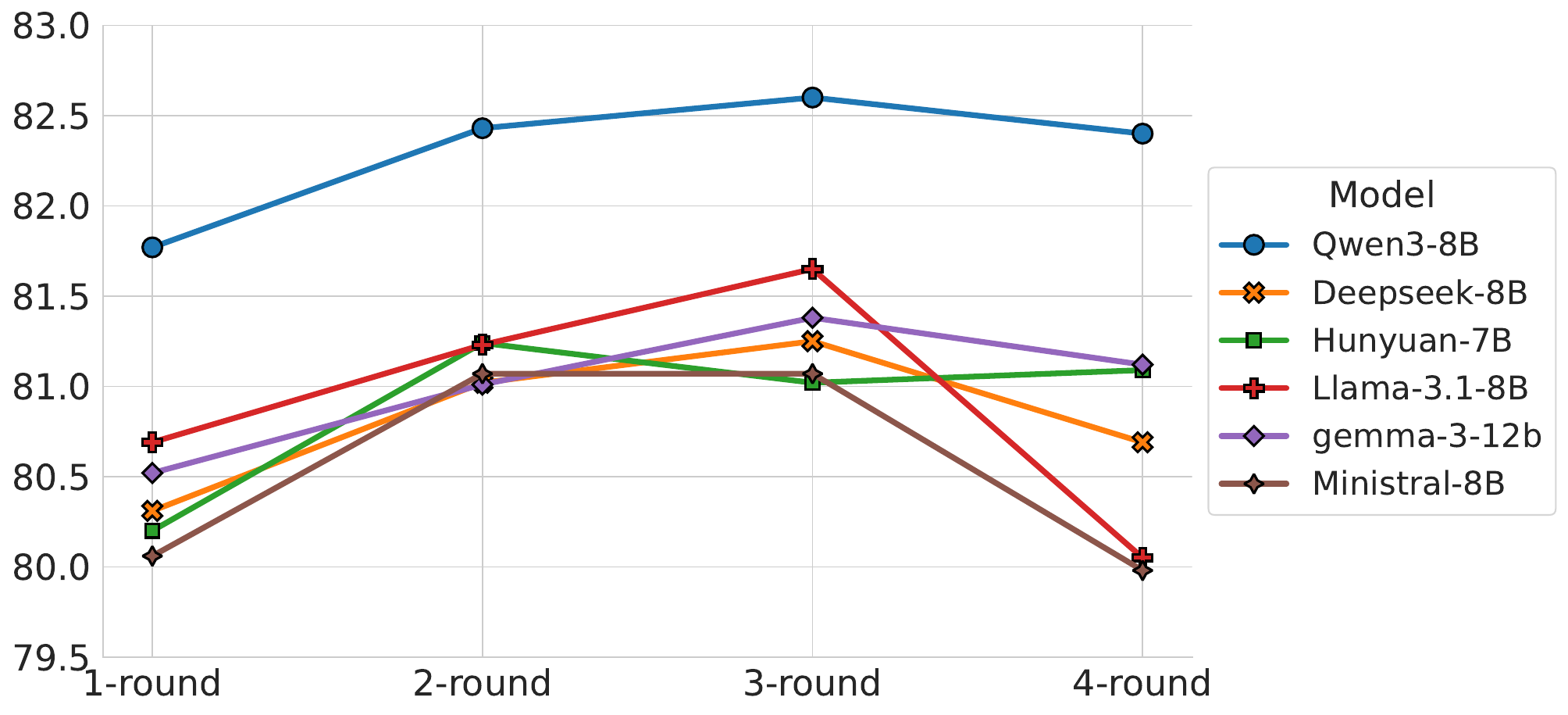}
	\caption{GDARL-CMMR generalization AUC by GDARL round. The CM group configuration use the most significant increase in Figure \ref{fig:GDARL-CMMR-matrix}.}
	\label{fig:GDARL-round-generalization}
\end{figure}

\paragraph{Ablation Study Details} Table~\ref{tab:full ablation} provides a more detailed expansion of the experiments in Table~\ref{tab:ablation}. This analysis is based on 10\% of the data, that is, based on 10\% of the prompts and 10\% of the HWTs, corresponding to the generation of 10\% of the MGTs. The following are several key observations from Table~\ref{tab:full ablation}.

First, removing any of the four alignment methods individually from MAGA leads to a performance decline compared to the full framework. This ablation study demonstrates that each of the four alignment methods contributes positively to MAGA.

Second, the performance improvement becomes increasingly marginal as the number of integrated methods increases. Overall, this trend indicates that substantial mutual dilution exists among these alignment methods.

Third, removing either self-refine or GDARL-CDMMR from MAGA leads to a performance decline, which demonstrates that integrating both self-refine and GDARL-CDMMR is beneficial and that they can mutually enhance performance beyond each other.

Fourth, the performance drop upon removing self-refine from MAGA is extremely marginal, despite being present. This suggests that the impact of self-refine is largely diluted, potentially because self-refine does not participate in the RL process, a limitation that agentic-RL may need to be introduced.

Fifth, taking the AUC under attack as an example, MGB /w roleplaying /w GDARL-CDMMR induces an additional 1.32\% decrease in evaluation detector performance compared to MGB /w roleplaying, while MGB /w BPO /w GDARL-CDMMR leads to an extra 1.02\% drop compared to MGB /w BPO. Meanwhile, MAGA w/o self-refine, which is equivalent to MGB /w roleplaying /w BPO /w GDARL-CDMMR, causes an additional 1.21\% reduction compared to MGB /w roleplaying /w BPO. The marginal degradation brought by GDARL-CDMMR when introduced after combining BPO and roleplaying is largely comparable to when it is introduced after selecting only one of them. Despite the increased number of integrated methods, GDARL-CDMMR exhibits no further dilution, indicating to some extent that the combination of BPO and roleplaying is helpful to promote the efficacy of GDARL-CDMMR.

\begin{figure*}[!t]
%    \vspace{-2pt}
    \centering
    \begin{minipage}{\textwidth}
        \centering  % 确保minipage内所有元素居中
        Attack  % 独立的标注块
        \par  % 强制换行，将标注和图片拆分为两个垂直块
        \vspace{6pt}  % 仅控制标注和图片的垂直间距（可改15/20pt）
        \includegraphics[width=\textwidth]{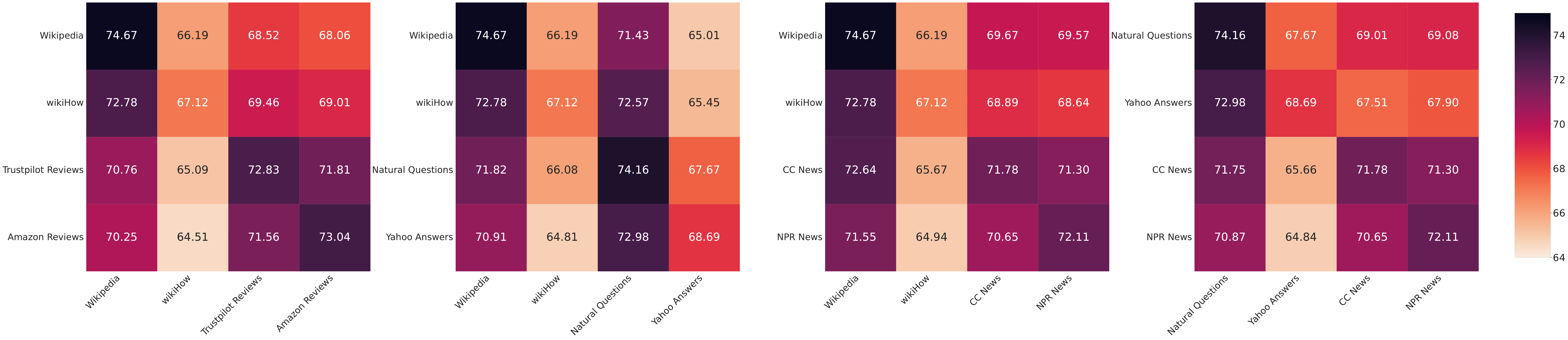}  % 图片占满宽度，不缩小
    \end{minipage}
    \\[1pt]  % 两个子图之间的间距（独立控制，不受内部影响）
    
    \begin{minipage}{\textwidth}
        \centering
        Generalization
        \par  % 强制换行，拆分标注和图片
        \vspace{6pt}  % 标注-图片间距，和上一个子图保持一致
        \includegraphics[width=\textwidth]{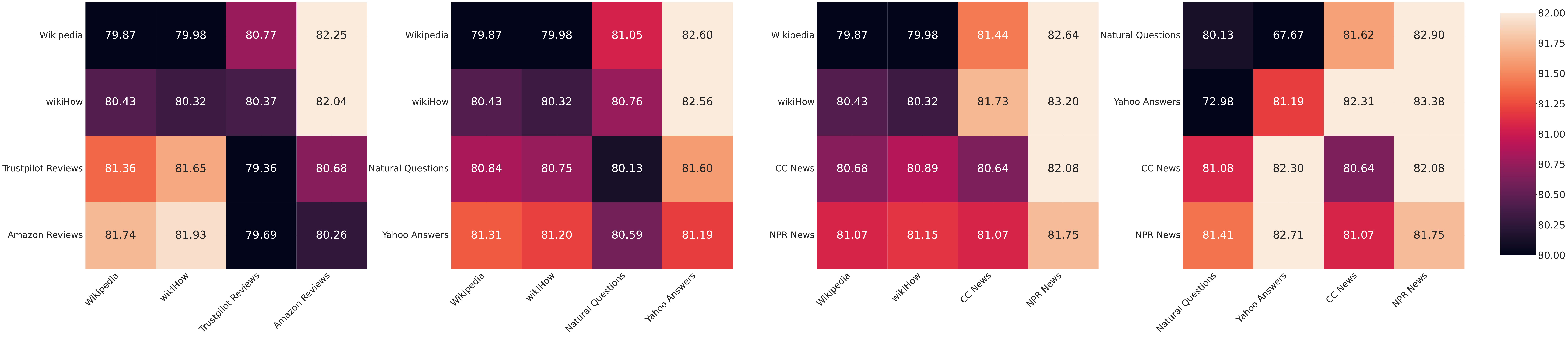}
    \end{minipage}
    
    % 整体caption（间距不受子图内部调整影响）
    \caption{GDARL-CDMR-matrix results. Top: Attack performance; Bottom: Generalization performance.}
    \label{fig:GDARL-CDMR-matrix}
%    \vspace{-2pt}
\end{figure*}

\begin{figure*}[!t]
%	\vspace{-2pt}
    \centering
    \begin{minipage}{0.44\linewidth}
        \centering
        \includegraphics[width=1.0\linewidth]{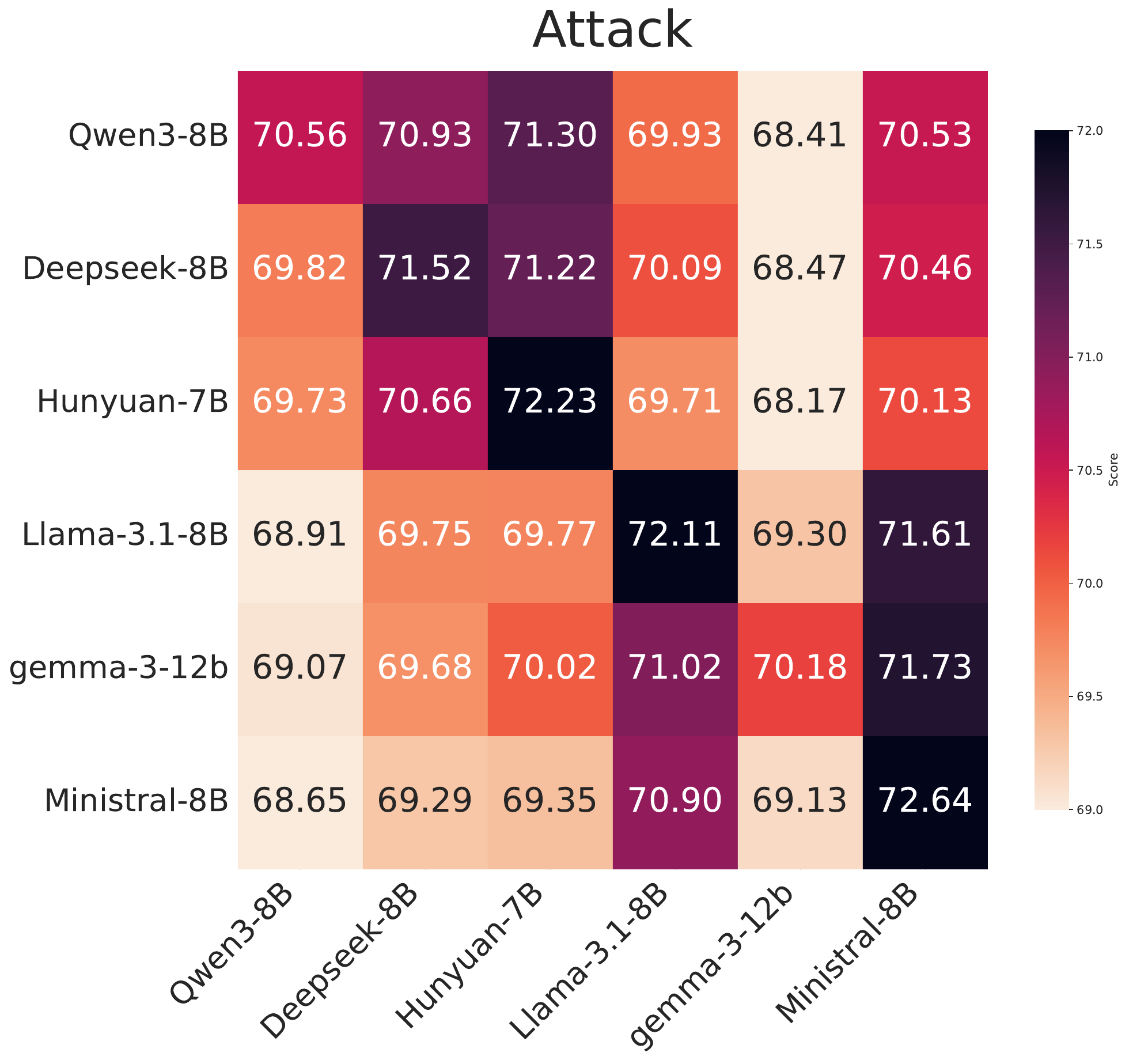}
    \end{minipage}
    \begin{minipage}{0.44\linewidth}
        \centering
        \includegraphics[width=1.0\linewidth]{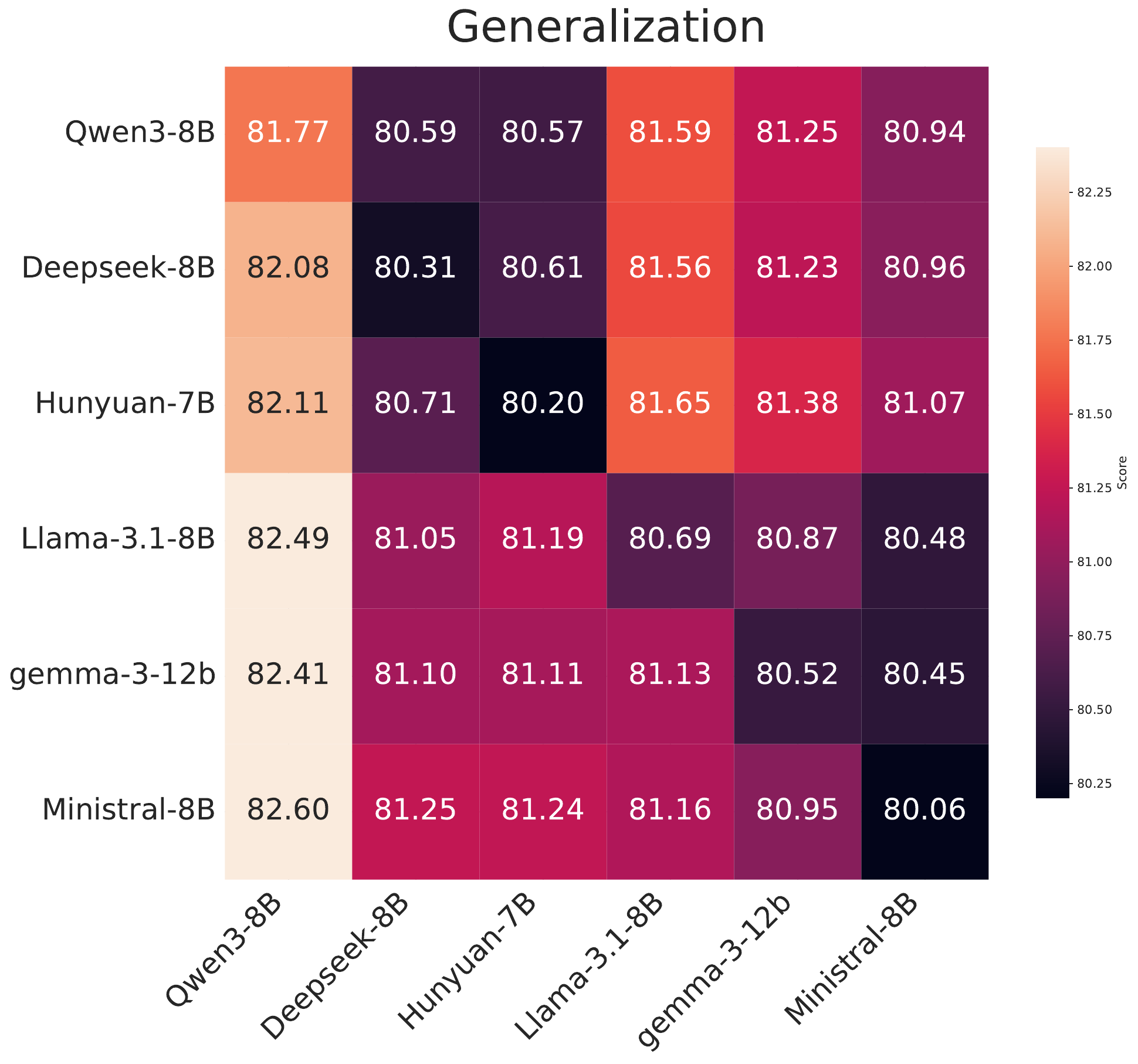}
    \end{minipage}
    \caption{GDARL-CMMR matrix analysis for attack and generalization}
    \label{fig:GDARL-CMMR-matrix}
%    \vspace{-2pt}
\end{figure*}

\begin{figure*}[!t]
    \centering  % 整体居中
    
    % 四个minipage在代码层面保持同一行（关键：用%注释换行，避免LaTeX解析为分行）
    \begin{minipage}{0.24\textwidth}
        \centering
        \scriptsize HWT-Domain
        \par
        \vspace{-1pt}
        \includegraphics[width=\textwidth]{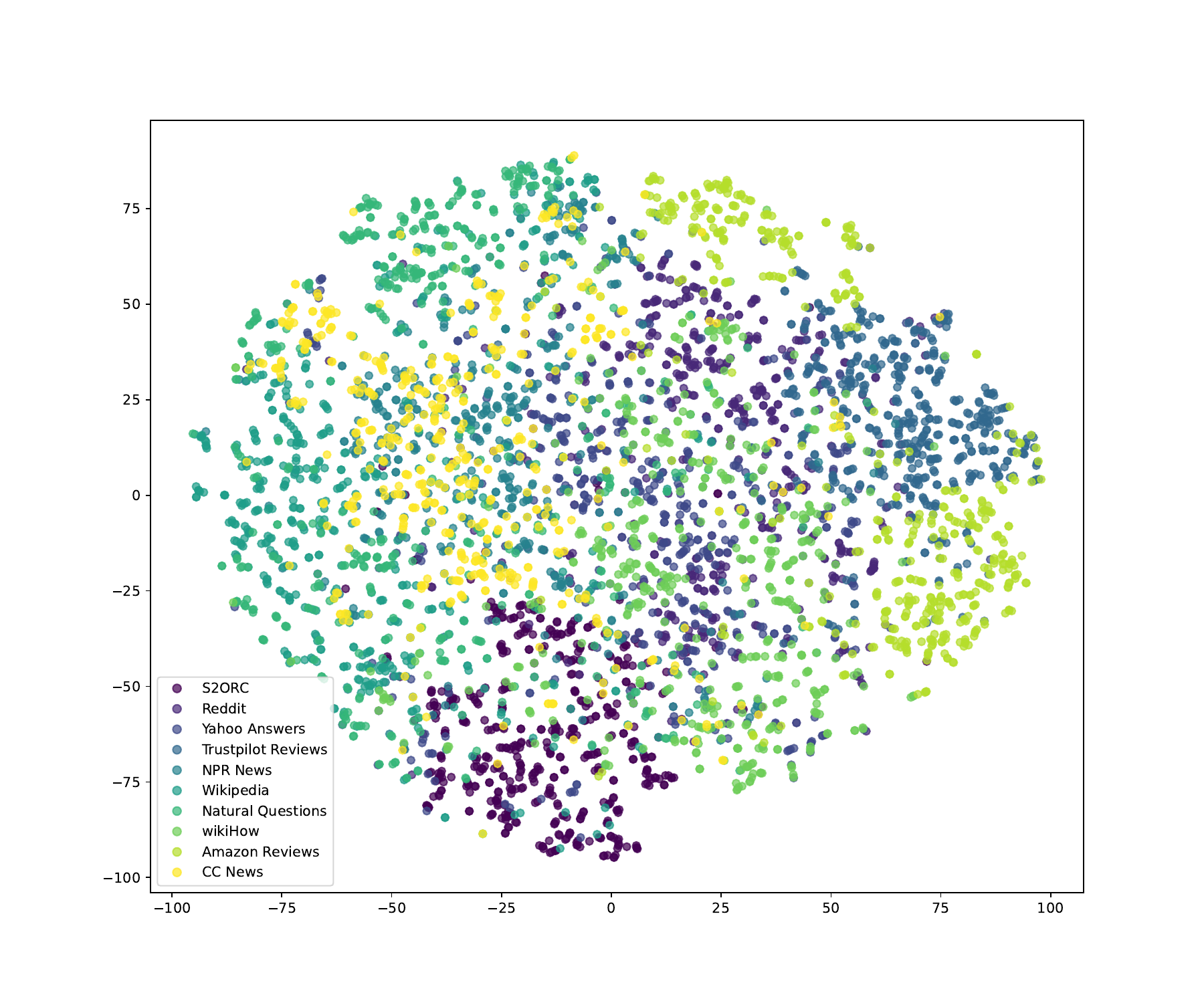}
    \end{minipage}% 关键：注释掉换行，避免LaTeX识别为分行符
    \hfill% 水平填充间距
    \begin{minipage}{0.24\textwidth}
        \centering
        \scriptsize MGB-M-Domain
        \par
        \vspace{-1pt}
        \includegraphics[width=\textwidth]{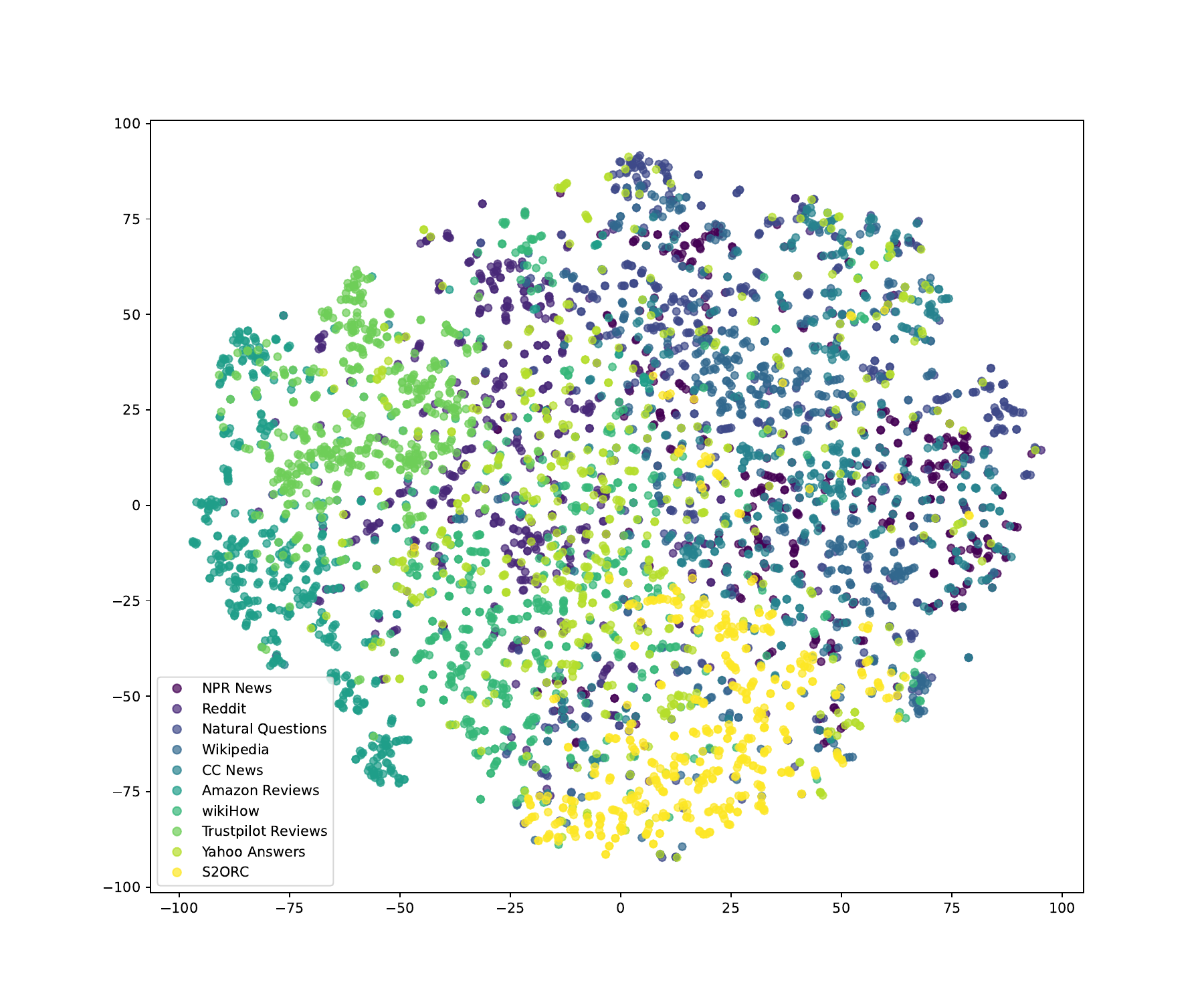}
    \end{minipage}% 关键：注释换行
    \hfill% 水平填充间距
    \begin{minipage}{0.24\textwidth}
        \centering
        \scriptsize MAGA-M-Domain
        \par
        \vspace{-1pt}
        \includegraphics[width=\textwidth]{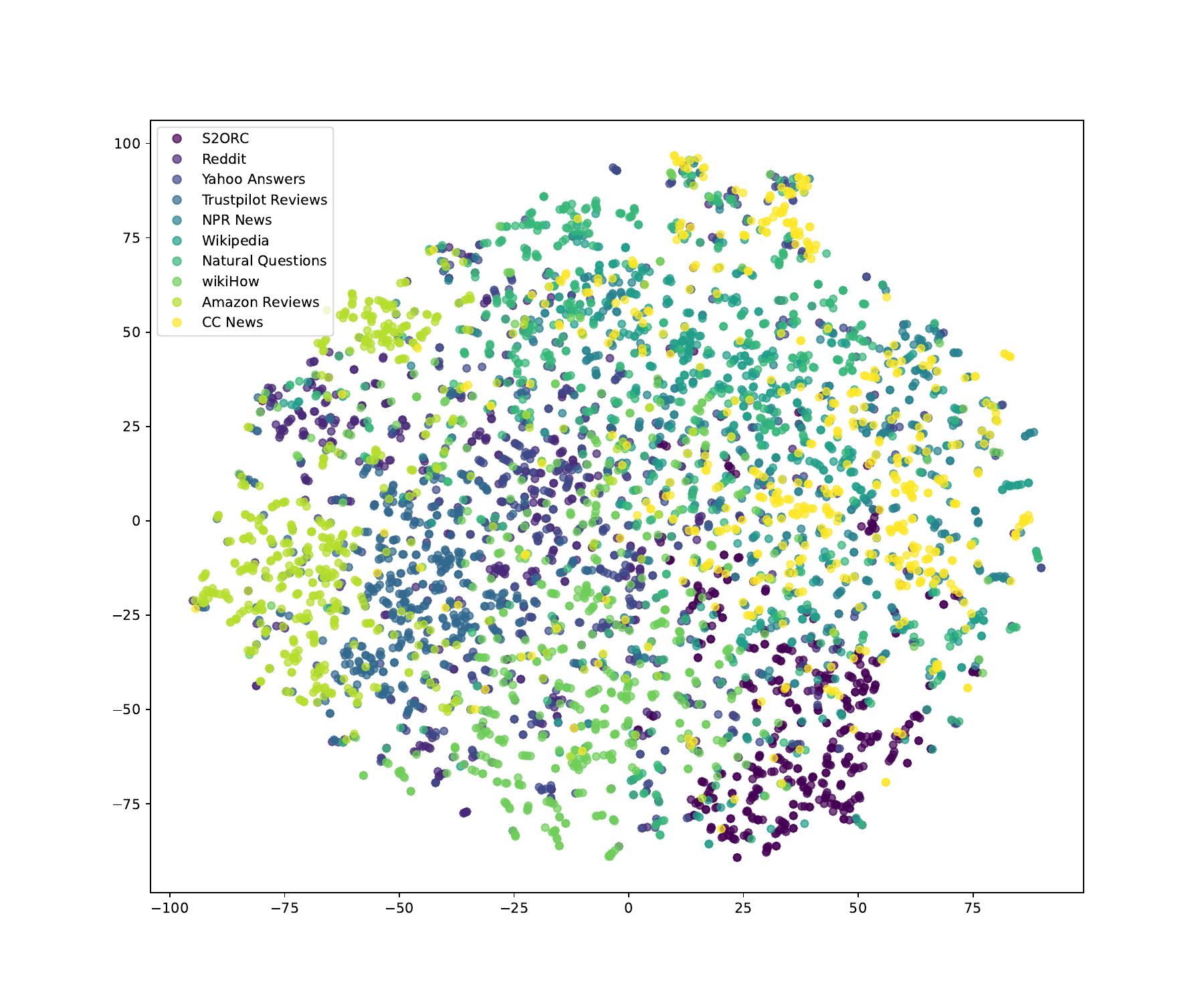}
    \end{minipage}% 关键：注释换行
    \hfill% 水平填充间距
    \begin{minipage}{0.24\textwidth}
        \centering
        \scriptsize MGB-M-Model
        \par
        \vspace{-1pt}
        \includegraphics[width=\textwidth]{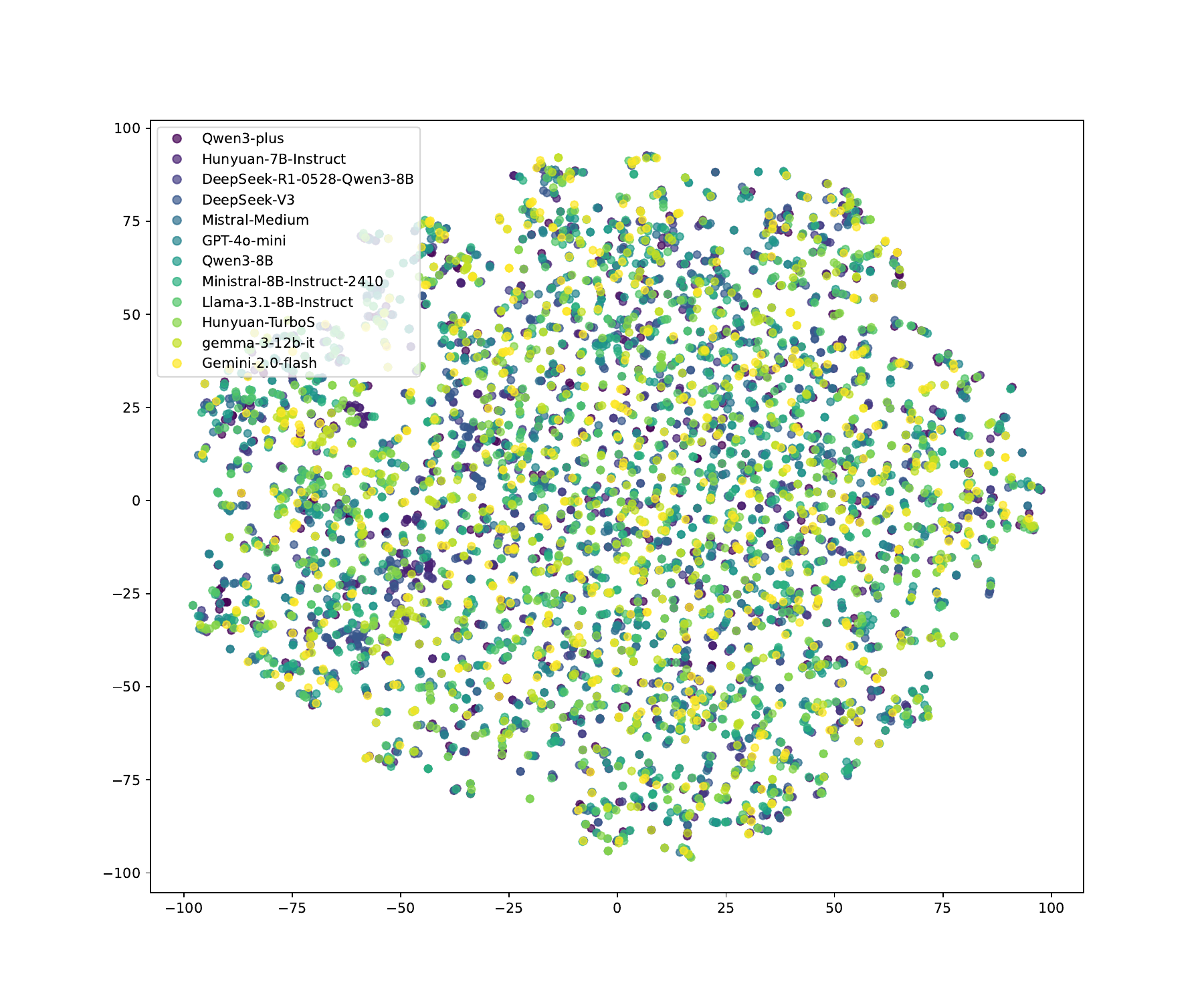}
    \end{minipage}

    % 整体caption
    \caption{T-SNE visualization results of different domains and models. From left to right: HWT domain, MGB-M domain, MAGA-M domain, MGB-M Model.}
    \label{fig:all-tsne}
\end{figure*}

Figure~\ref{fig:GDARL-round-generalization} presents the ablation study on the number of adversarial training rounds in GDARL-CMMR. The overall performance generally converges after 3 rounds, whereas a 4-round configuration leads to apparent overfitting, resulting in a consistent performance decline. Figure~\ref{fig:GDARL-CDMR-matrix} and Figure~\ref{fig:GDARL-CMMR-matrix} explore the grouping mechanisms across domains and models for GDARL-CDMR and GDARL-CMMR, respectively. Specifically, Figure~\ref{fig:GDARL-CMMR-matrix} provides a more comprehensive exploration that complements Figure~\ref{fig:GDARL-CDMR-matrix-core}. Figure~\ref{fig:all-tsne} illustrates the t-distributed Stochastic Neighbor Embedding (t-SNE) visualizations of texts across different domains and models, extending the details shown in Figure~\ref{fig:hwt-domain-core}. We utilize the bge-large-en-v1.5 model\footnote{\url{https://huggingface.co/BAAI/bge-large-en-v1.5}} \cite{xiao2024cpackpackedresourcesgeneral} for embeddings extraction in the t-SNE visualization, a choice validated by FineMedLM-o1 \cite{yu2025finemedlmo1enhancingmedicalknowledge} for its suitability in t-SNE representations. Notably, as illustrated in Figure \ref{fig:all-tsne}, while distinct domains exhibit clear clustering patterns, texts generated by different models may not be adequately characterized by two-dimensional features, resulting in the absence of discernible clusters within the t-SNE visualization. Furthermore, as shown in Figure \ref{fig:log ppl model}, we analyze the log PPL computed mutually across texts generated by different generators. A given generator consistently assigns a low PPL to its own generated text but yields a substantially higher PPL when evaluating texts from other generators, indicating a pronounced discrepancies in viewpoints among different generators. This divergence in perspective is particularly prominent between predominantly Chinese-language generators and predominantly English-language generators. Consistent with the discussions in \S\ref{sec:ablation}, the outcomes of these analyses and exploratory investigations guide our final hyperparameter configurations, which we document in Appendix~\ref{sec:hyperparameters}.

\section{Quantitative Text Analysis Details}
\subsection{Multidimensional Text Analysis}
\label{sec:multi analysis detail}

In this section, we detail the selection of dimension-specific metrics for multidimensional quantitative text analysis alongside the corresponding settings (e.g. parameter, model selection) utilized during calculation, and then present a more detailed report of the evaluation results.

\begin{figure}[!t]
	\centering
	\includegraphics[width=1.0\linewidth]{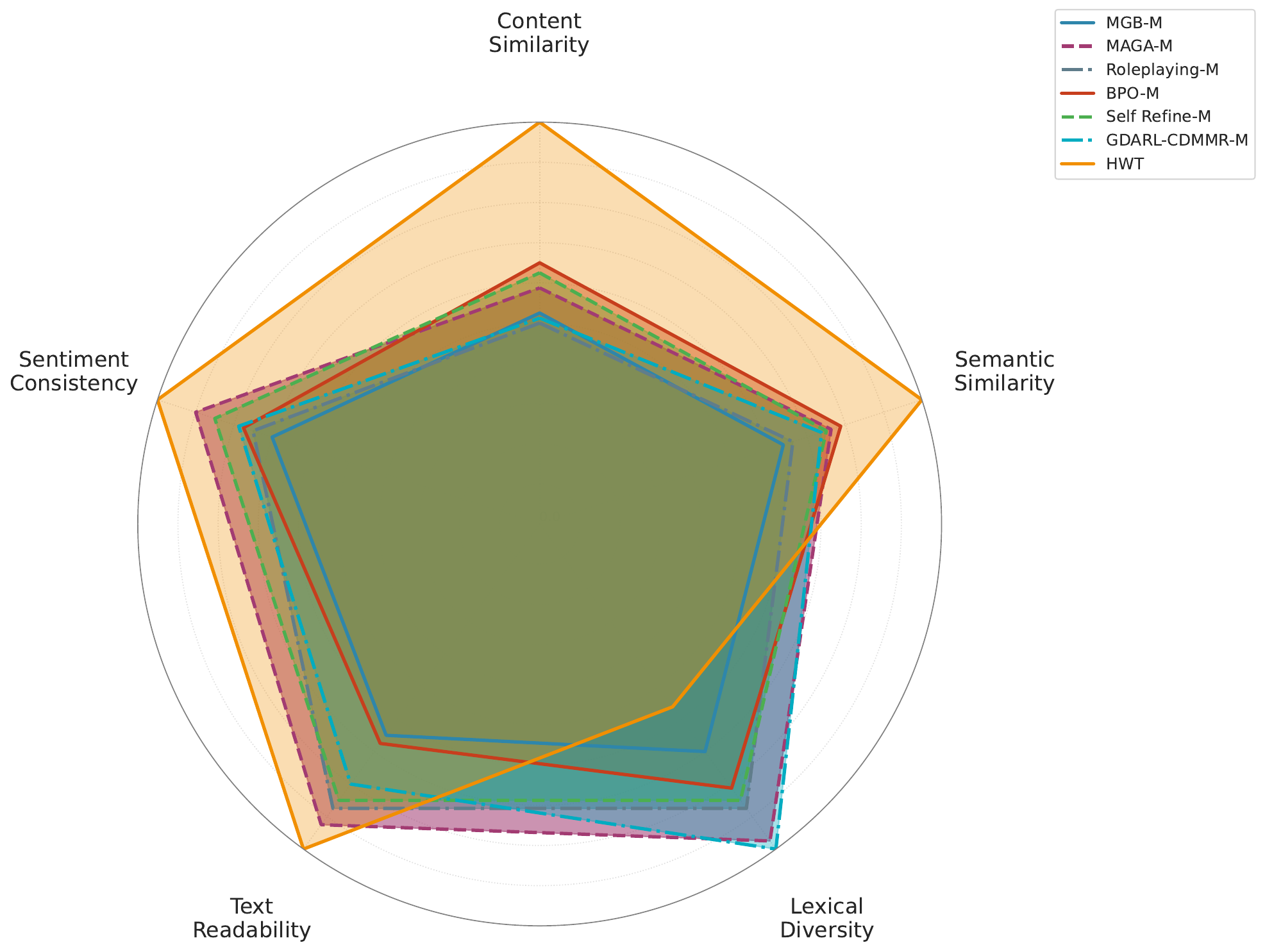}
	\caption{Multidimensional Text Analysis - Full.}
	\label{fig:multi-dimension-radar-full}
\end{figure}

As described in \S\ref{sec:analysis}, the final value for each dimension is the average of the computed scores of all constituent metrics within that dimension. Metrics with values falling outside the $[0,1]$ range undergo normalization. We implement two distinct strategies, namely the normalized score and the cosine similarity. Although cosine similarity is inherently suited for evaluating distributional alignment, we primarily rely on the normalized score (displayed in Figures \ref{fig:multi-dimension-radar}, \ref{fig:multi-dimension-radar-full}) as we found it demonstrates differences between text groups more prominently.

The normalized score is formulated to quantify the proximity between machine-generated texts (MGTs) and human-written texts (HWTs). Let $S_M$ and $S_H$ denote the raw metric values computed from all MGTs and HWTs on a specific split, respectively. Since our objective is to measure the degree of similarity, an optimal MGT performance implies that $S_M$ closely approaches $S_H$. Consequently, when $S_M < S_H$, a higher $S_M$ is desired, and we scale the metric by treating $S_H$ as the baseline unit, yielding $S_M^{norm} = \frac{S_M}{S_H}$. Conversely, when $S_M > S_H$, a lower $S_M$ is preferred, and we apply an inverse ratio defined as $S_M^{norm} = \frac{1/S_M}{1/S_H} = \frac{S_H}{S_M}$. The final normalized score for a given dimension is then derived by taking the arithmetic mean of the normalized scores across all underlying metrics.

To illustrate this process more clearly, using sentiment consistency as an example, Table~\ref{tab:multidimensional_analysis} shows that MGB-M yields average negative, neutral, and positive sentiment scores of $0.3243$, $0.4030$, and $0.2727$, respectively, while HWT obtains scores of $0.3455$, $0.3996$, and $0.2549$. The resulting normalized score is computed as $(\frac{0.3243}{0.3455} + \frac{0.3996}{0.4030} + \frac{0.2549}{0.2727}) / 3 = 0.9550$.

For the cosine similarity, the calculation follows the standard algebraic definition, where the sentiment consistency between MGB-M and HWT is evaluated as the cosine similarity between the vectors $[0.3243, 0.4030, 0.2727]$ and $[0.3455, 0.3996, 0.2549]$.

It should be noted that lexical diversity is treated exceptionally. Instead of positioning HWT as the fixed baseline of $1.0$, we designate the maximum observed value across groups as $1.0$ to serve as the reference target, for reasons elaborated in \S\ref{sec:analysis}. Such a setting may not be appropriate either.

We select only a single representative sub-metric for each metric (for example, for ROUGE, we selected only ROUGE-2-F1). This choice ensures a consistent number of scalar variables across different metrics, thereby facilitating unbiased averaging. The detailed specifications for each dimensional metric are provided below.

\paragraph{Content Similarity} We select Recall-Oriented Understudy for Gisting Evaluation (ROUGE) \cite{lin-2004-rouge}, Bilingual Evaluation Understudy (BLEU) \cite{papineni-etal-2002-bleu}, and Metric for Evaluation of Translation with Explicit Ordering (METEOR) \cite{banerjee-lavie-2005-meteor}. Among the various sub-metrics available for ROUGE, we choose ROUGE-2-F1 to represent the content similarity performance based on qualitative justifications. Specifically, ROUGE-1 focuses on unigrams and is suitable for evaluating semantic coverage in short texts, whereas ROUGE-2 targets bigrams and better captures phrase-level semantic coherence in longer documents. ROUGE-L utilizes the longest common subsequence and fits tasks that heavily rely on word order, such as machine translation. Regarding the optimization priorities of these sub-metrics, text summarization often prioritizes recall over precision and F1 score to ensure comprehensive coverage of core information from the source text. Conversely, machine translation and question answering favor precision over recall and F1 score to guarantee output accuracy. For general content generation tasks, the F1 score represents the most suitable compromise because it balances both coverage and accuracy. Consequently, we select ROUGE-2-F1 as the primary sub-metric.

\paragraph{Semantic Similarity} We select BERTScore \cite{Zhang*2020BERTScore:} as our metric and utilize the corresponding F1 score. To compute this metric, we extract text embeddings using the bert-base-uncased model \cite{devlin-etal-2019-bert}, which is the default configuration in the official bert\_score package. Prior research on semantic similarity analysis, HC-Var \cite{xu-etal-2024-generalization}, implements the MAUVE metric \cite{NEURIPS2021_260c2432} to calculate the overall similarity between two distinct text datasets. Because our data generation follows a paired, one-to-one basis strategy, we can compute the BERTScore between individual MGT and HWT pairs directly, subsequently reporting the arithmetic mean across the entire dataset split.

\paragraph{Lexical Diversity} We incorporate the Type-Token Ratio (TTR) \cite{chotlos1944iv}, Yule's K \cite{yule2014statistical}, and the 2-gram vocabulary size metric introduced by M4 \cite{wang-etal-2024-m4}. Although a universally accepted benchmark for lexical diversity remains absent in the literature, TTR and Yule's K are widely established as foundational indicators of vocabulary variation, making their selection a reliable baseline for our evaluation. Furthermore, empirical findings from M4 \cite{wang-etal-2024-m4} demonstrate that constructing datasets with multi-generators substantially increases the vocabulary size of unigrams and bigrams. This multi-generator paradigm enhances the lexical richness of machine-generated texts within the resulting detection datasets, which ultimately benefits the training and evaluation of robust MGT detectors. Notably, the vocabulary sizes of unigrams and bigrams for MGTs reported by M4 \cite{wang-etal-2024-m4} already exceed those observed in HWTs.

\paragraph{Text Readability} We select Kincaid Reading Ease \cite{kincaid1975derivation}, which we denote as Flesch Kincaid for brevity, alongside Simple Measure of Gobbledygook (SMOG) \cite{mc1969smog} and Dale Chall \cite{dale1948formula}. These three metrics are widely recognized as foundational standards for text readability evaluation, which justifies our selection. Previously, AA \cite{uchendu-etal-2020-authorship} implemented Flesch Kincaid in their analytical study within the machine-generated text detection domain. The Flesch Kincaid framework actually contains multiple sub-metrics. We specifically select Kincaid Reading Ease to maintain a single sub-metric value, deliberately excluding Flesch-Kincaid Grade level which maps performance to school ages.

\begin{figure*}[!t]
	\centering
	\includegraphics[width=1.0\linewidth]{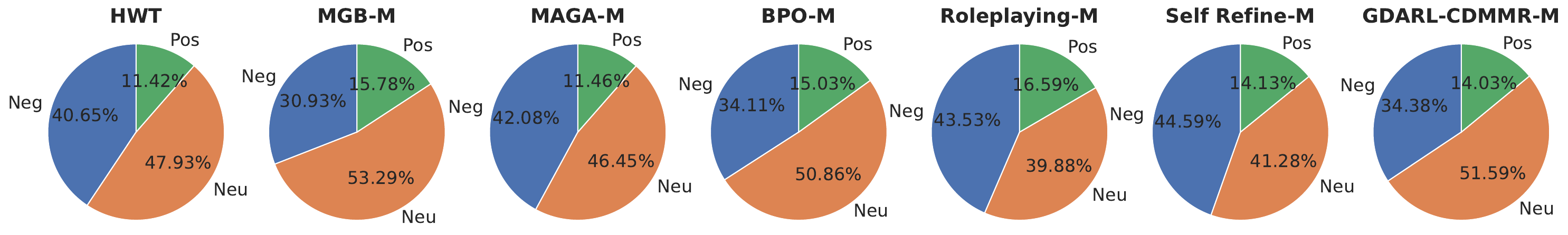}
	\vspace{-0.55cm}
	\caption{sentiment-detail}
	\label{fig:sentiment-detail}
\end{figure*}

\paragraph{Sentiment Consistency} We employ twitter-xlm-roberta-base-sentiment\footnote{\url{https://huggingface.co/cardiffnlp/twitter-xlm-roberta-base-sentiment}} \cite{barbieri-etal-2022-xlm} as our sentiment classification model, which aligns with the methodology established by HC3 \cite{guo2023close}. Rather than merely reporting the discrete classification distributions shown in Figure~\ref{fig:sentiment-detail}, we evaluate dimensional similarity by calculating the arithmetic mean of the continuous probability scores across three sentiment dimensions. To illustrate this approach more clearly, consider a toy dataset containing two sentences. If the first sentence receives negative, neutral, and positive probability scores of $0.3$, $0.3$, and $0.4$, respectively, and the second sentence receives scores of $0.2$, $0.3$, and $0.5$, then the final scores for the entire dataset are derived by averaging these individual probabilities, yielding $(0.3 + 0.2) / 2 = 0.25$ for negative, $(0.3 + 0.3) / 2 = 0.30$ for neutral, and $(0.4 + 0.5) / 2 = 0.45$

\begin{figure*}[!t]
%	\vspace{-0.3cm}
	\centering
	\includegraphics[width=1.0\linewidth]{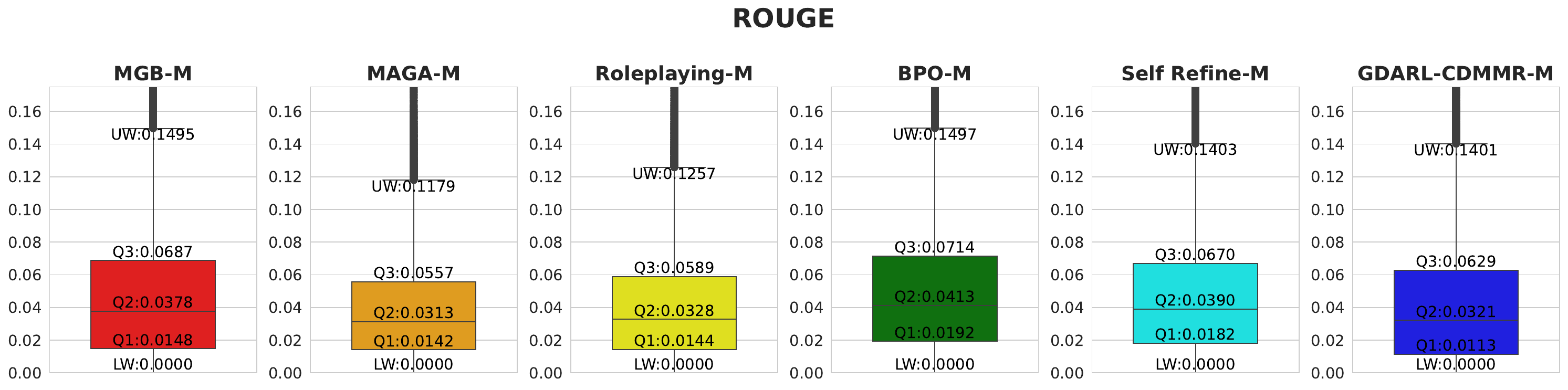}
	\vspace{-0.55cm}
	\caption{box-rouge}
	\label{fig:box-rouge}
%	\vspace{-0.3cm}
\end{figure*}

\begin{figure*}[!t]
	\centering
	\includegraphics[width=1.0\linewidth]{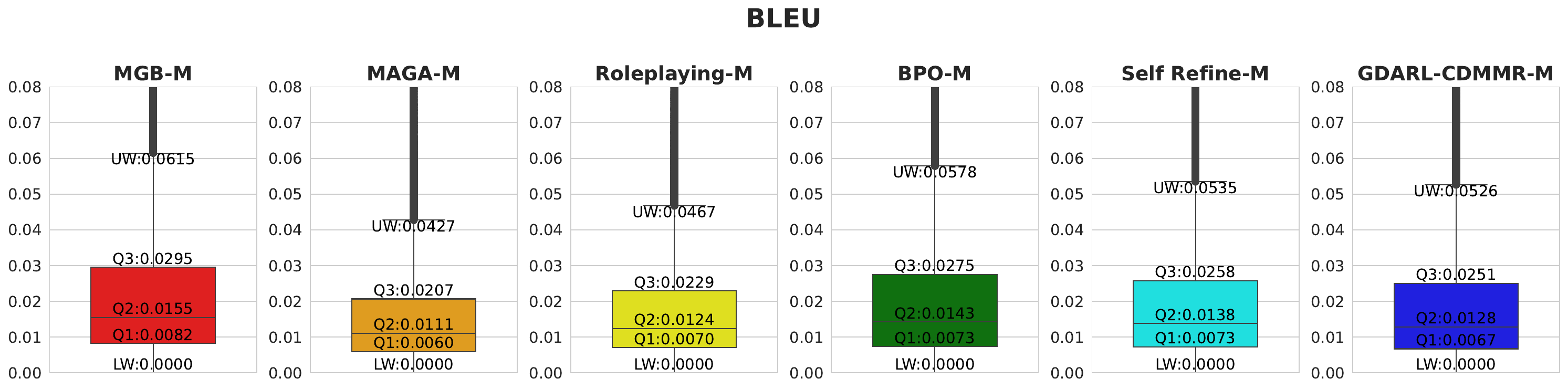}
	\vspace{-0.55cm}
	\caption{box-bleu}
	\label{fig:box-bleu}
\end{figure*}

\begin{figure*}[!t]
	\centering
	\includegraphics[width=1.0\linewidth]{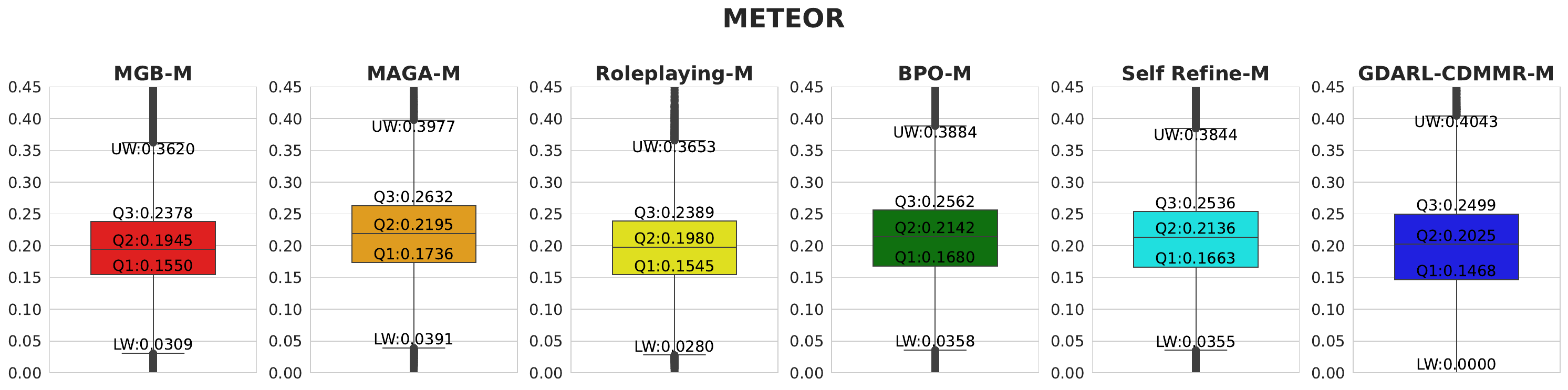}
	\vspace{-0.55cm}
	\caption{box-meteor}
	\label{fig:box-meteor}
\end{figure*}

\begin{figure*}[!t]
	\centering
	\includegraphics[width=1.0\linewidth]{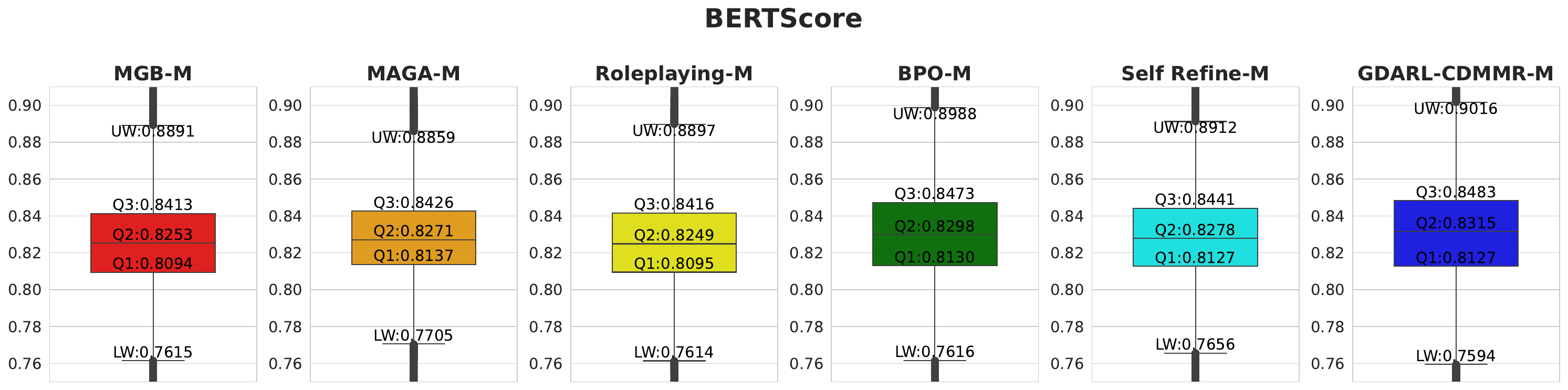}
	\vspace{-0.55cm}
	\caption{box-bertscore}
	\label{fig:box-bertscore}
\end{figure*}

\begin{figure*}[!t]
	\centering
	\includegraphics[width=1.0\linewidth]{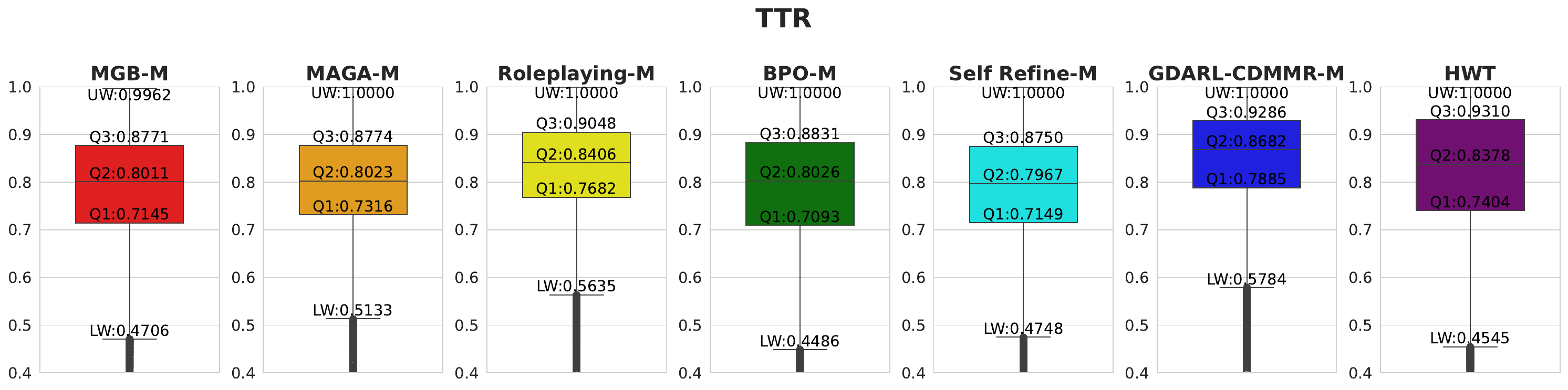}
	\vspace{-0.55cm}
	\caption{box-ttr}
	\label{fig:box-ttr}
\end{figure*}

\begin{figure*}[!t]
	\centering
	\includegraphics[width=1.0\linewidth]{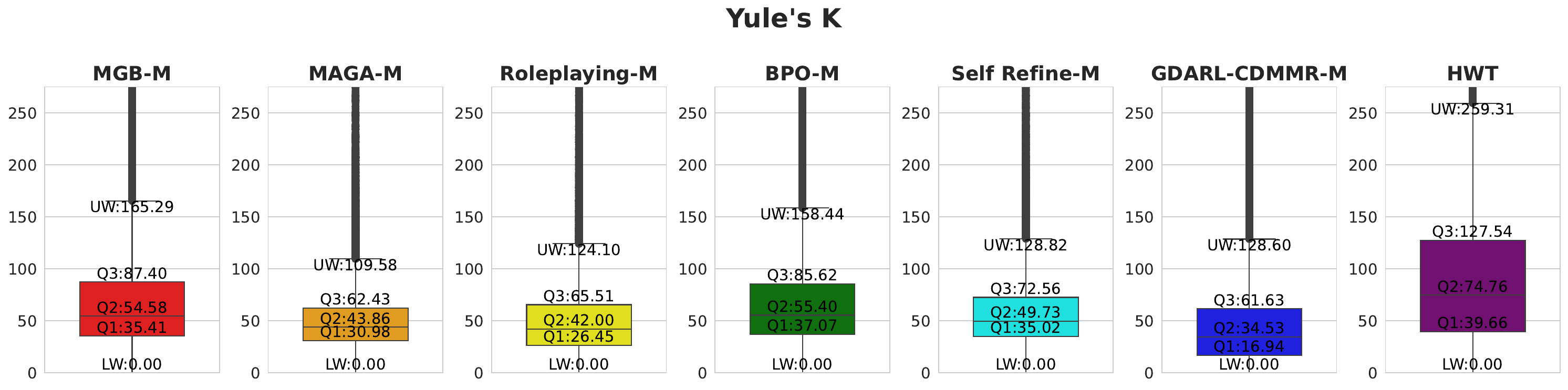}
	\vspace{-0.55cm}
	\caption{box-yules'k}
	\label{fig:box-yules-k}
\end{figure*}

\begin{figure*}[!t]
	\centering
	\includegraphics[width=1.0\linewidth]{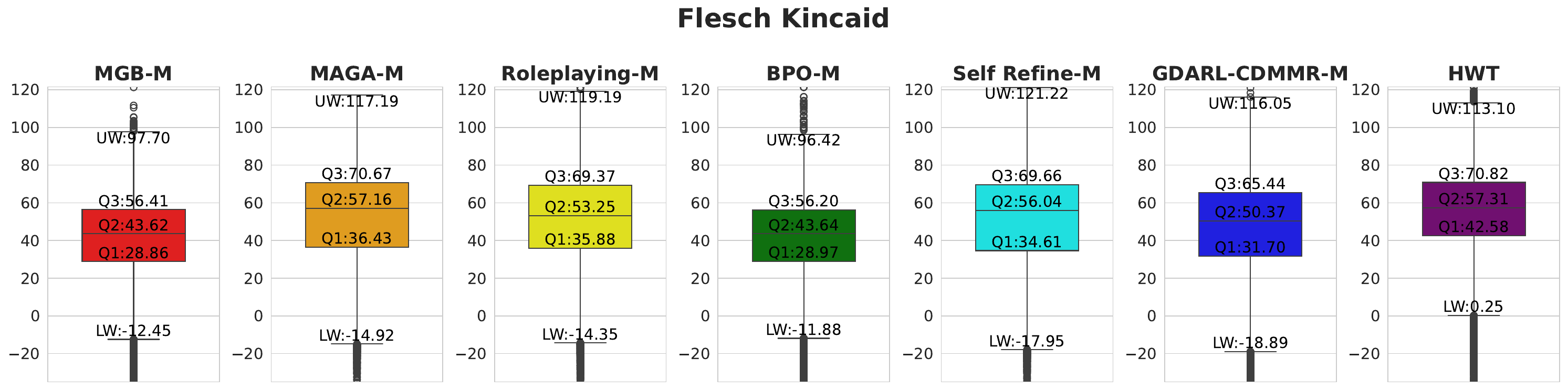}
	\vspace{-0.55cm}
	\caption{box-flesch-kincaid}
	\label{fig:box-flesch-kincaid}
\end{figure*}

\begin{figure*}[!t]
	\centering
	\includegraphics[width=1.0\linewidth]{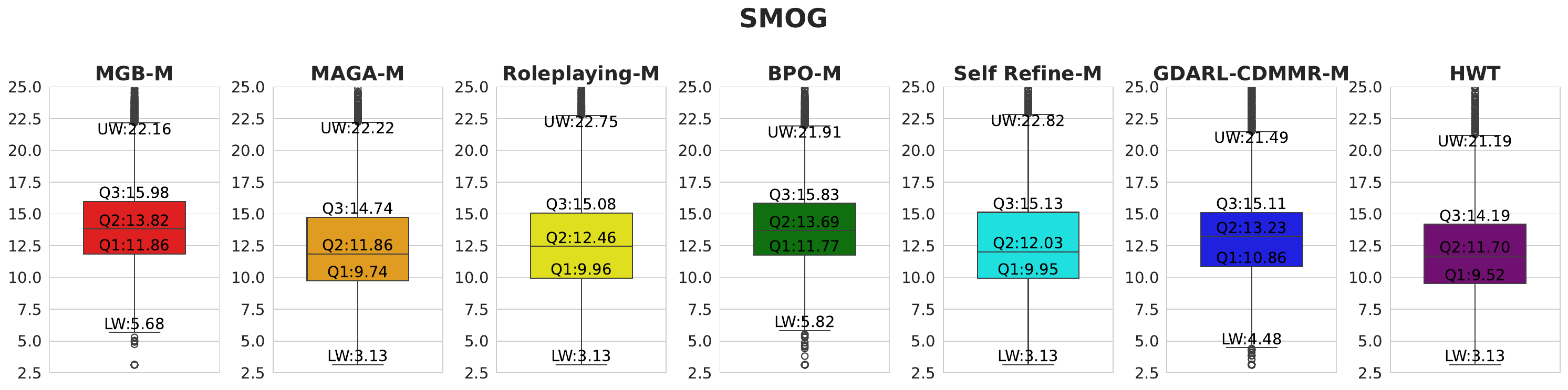}
	\vspace{-0.55cm}
	\caption{box-smog}
	\label{fig:box-smog}
\end{figure*}

\begin{figure*}[!t]
	\centering
	\includegraphics[width=1.0\linewidth]{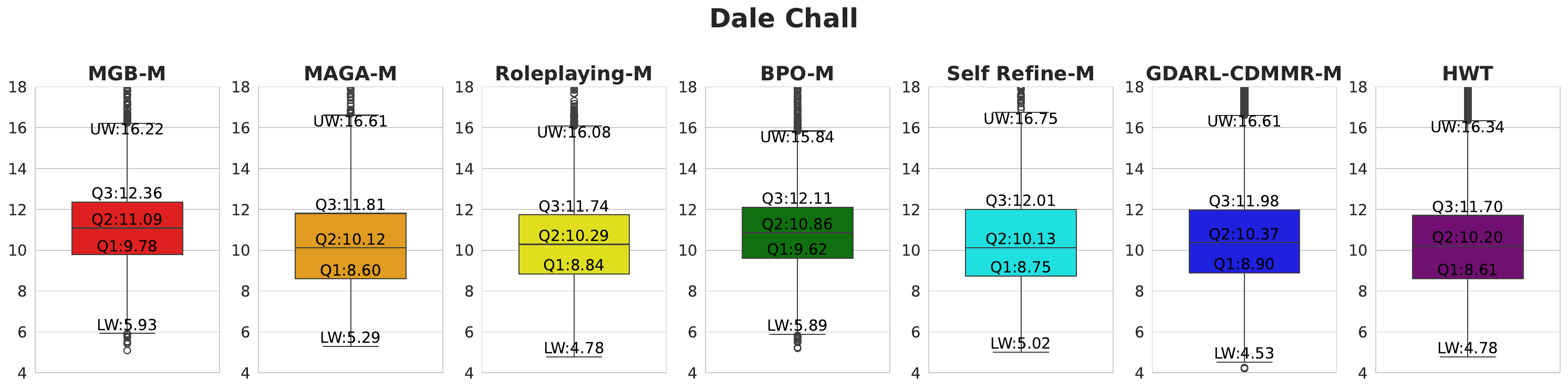}
	\vspace{-0.55cm}
	\caption{box-dale-chall}
	\label{fig:box-dale-chall}
\end{figure*}

\paragraph{More Result Details and Analysis} Table~\ref{tab:multidimensional_analysis} presents the specific values for the individual metrics that correspond to the radar chart illustrated in Figure~\ref{fig:multi-dimension-radar}. Table~\ref{tab:multidimensional_analysis_all} expands upon Table~\ref{tab:multidimensional_analysis} by incorporating detailed statistics from the machine-generated texts across all MAGA extra splits, which are visually aggregated in the comprehensive radar chart in Figure~\ref{fig:multi-dimension-radar-full}. To provide a more granular examination of these metrics, Figures~\ref{fig:box-rouge}, \ref{fig:box-bleu}, \ref{fig:box-meteor}, \ref{fig:box-bertscore}, \ref{fig:box-ttr}, \ref{fig:box-yules-k}, \ref{fig:box-flesch-kincaid}, \ref{fig:box-smog}, and \ref{fig:box-dale-chall} display box plots representing the metric distributions computed from individual text instances across these splits, highlighting key descriptive statistics such as the median and the interquartile range. Additionally, Figure~\ref{fig:sentiment-detail} depicts the proportional distribution of predicted sentiment classes across all splits, where the sentiment label for each text instance is determined by the highest prediction probability. As discussed in \S\ref{sec:analysis}, our analytical findings remain broadly consistent with conclusions established in prior literature. In addition to the observations already detailed in \S\ref{sec:analysis}, we provide a supplementary discussion here regarding the sentiment classification results, which also align with the conclusions reported in HC3 \cite{guo2023close}. Specifically, the categorical distributions in Figure~\ref{fig:sentiment-detail} reveal that human-written texts exhibit a low frequency of positive sentiment alongside a higher prevalence of negative and neutral expressions, with neutral sentiment being the most dominant. Conversely, the baseline MGB-M, which lacks alignment enhancement, tends to be overly conservative, yielding a disproportionately higher rate of neutral sentiment and a lower rate of negative sentiment relative to HWTs, a discrepancy that leads to a sentiment mismatch between MGB-M and HWTs. Texts generated via alignment augment, such as MAGA-M, mitigate this disparity substantially. Furthermore, the granular averaging of continuous sentiment probability scores in Table~\ref{tab:multidimensional_analysis_all} also supports similar conclusions while revealing a relatively more balanced distribution across the three probability categories. This trend indicates that the underlying sentiment within a single sentence may possess inherent complexity. A more granular analysis may be necessary to some extent.

\begin{table}[!t]
\resizebox{\linewidth}{!}{
    \centering
    \begin{tabular}{c|c|c|c}
        \toprule
         & MGB-M & MAGA-M & HWT \\
        \toprule
        \multicolumn{4}{c}{Content Similarity} \\
        \midrule
        ROUGE & 0.0479 & 0.0387 & 1.00 \\
        BLEU & 0.0235 & 0.0165 & 1.00 \\
        METEOR & 0.1974 & 0.2158 & 1.00 \\
        \midrule
        Average & 0.0896 & 0.0903 & 1.00 \\
        \midrule
        \multicolumn{4}{c}{Semantic Similarity} \\
        \midrule
        BERTScore & 0.8252 & 0.8287 & 1.00 \\
        \midrule
        \multicolumn{4}{c}{Lexical Diversity} \\
        \midrule
        TTR & 0.7893 & 0.7991 & 0.8271 \\
        Yule's K & 74.99 & 52.02 & 100.18 \\
        Bigram Vocab Size & 4261689 & 5467106 & 3253456 \\
        \midrule
        Normalized Score & 0.8203 & 1.00 & 0.7165 \\
        \midrule
        \multicolumn{4}{c}{Text Readability} \\
        \midrule
        Flesch Kincaid & 40.53 & 53.39 & 54.54 \\
        SMOG & 14.10 & 12.15 & 11.91 \\
        Dale Chall & 11.19 & 10.24 & 10.23 \\
        \midrule
        Normalized Score & 0.8340 & 0.9861 & 1.00 \\
        Cosine Similarity & 0.9905 & 1.00 & 1.00 \\
        \midrule
        \multicolumn{4}{c}{Sentiment Consistency} \\
        \midrule
        Negative & 0.3243 & 0.3439 & 0.3455 \\
        Neutral & 0.4030 & 0.3894 & 0.3996 \\
        Positive & 0.2727 & 0.2667 & 0.2549 \\
        \midrule
        Normalized Score & 0.9550 & 0.9752 & 1.00 \\
        Cosine Similarity & 0.9989 & 1.00 & 1.00 \\
        \bottomrule
    \end{tabular}}
    \caption{Multi-dimensional human-machine text analysis results across three texts: MGB-M, MAGA-M, and HWT. The evaluation covers content similarity, semantic similarity, lexical diversity, text readability, and sentiment consistency metrics.}
    \label{tab:multidimensional_analysis}
\end{table}

\begin{figure}[!t]
	\centering
	\includegraphics[width=1.0\linewidth]{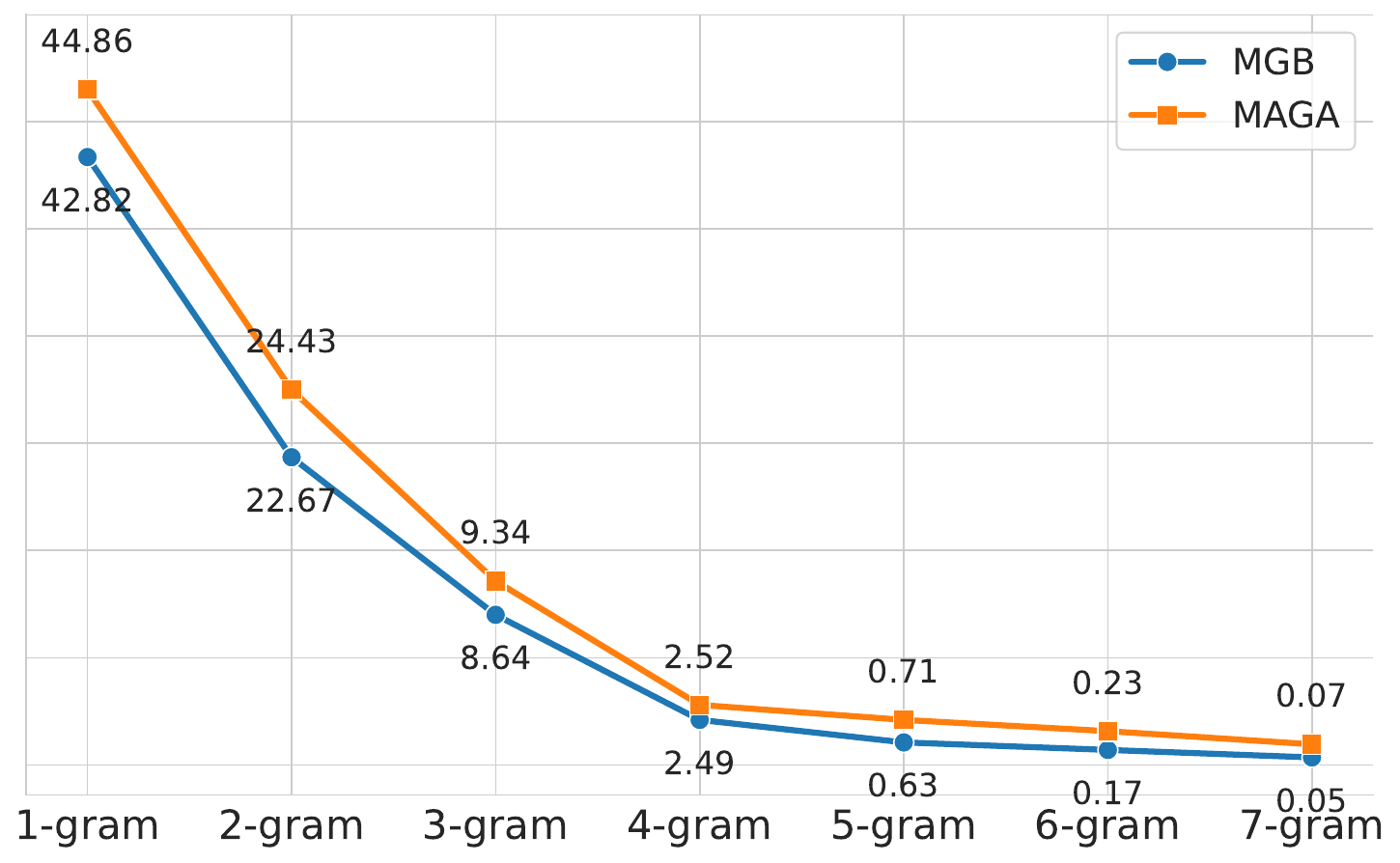}
	\caption{Ngram overlap rate with HWT. The calculation measures the percentage of the MGT's Ngram vocabulary that overlaps with the HWT's Ngram vocabulary, which is the intersection size divided by the HWT's Ngram vocabulary size.}
	\label{fig:ngram-overlap-rate}
\end{figure}

\begin{table*}[!t]
	\resizebox{\linewidth}{!}{
		\centering
		\begin{tabular}{c|c|c|c|c|c|c|c}
			\toprule
			 & MGB-M & MAGA-M & Roleplaying-M & BPO-M & Self Refine-M & GDARL-CDMMR-M & HWT \\
			\toprule
			\multicolumn{8}{c}{Content Similarity} \\
			\midrule
			ROUGE & 0.0479 & 0.0387 & 0.0411 & 0.0505 & 0.0473 & 0.0434 & 1.00 \\
			BLEU & 0.0235 & 0.0165 & 0.0189 & 0.0219 & 0.0205 & 0.0204 & 1.00 \\
			METEOR & 0.1974 & 0.2158 & 0.1971 & 0.2119 & 0.2093 & 0.1984 & 1.00 \\
			\midrule
			Average & 0.0896 & 0.0903 & 0.0857 & 0.0948 & 0.0924 & 0.0874 & 1.00 \\
			\midrule
			\multicolumn{8}{c}{Semantic Similarity} \\
			\midrule
			BERTScore & 0.8252 & 0.8287 & 0.8256 & 0.8302 & 0.8285 & 0.8284 & 1.00 \\
			\midrule
			\multicolumn{8}{c}{Lexical Diversity} \\
			\midrule
			TTR & 0.7893 & 0.7991 & 0.8306 & 0.7895 & 0.7916 & 0.8496 & 0.8271 \\
			Yule's K & 74.99 & 52.02 & 55.94 & 74.44 & 62.41 & 50.76 & 100.18 \\
			Bigram Vocab Size & 4261689 & 5467106 & 4050008 & 4909749 & 4665052 & 5511160 & 3253456 \\
			\midrule
			Normalized Score & 0.7931 & 0.9694 & 0.8733 & 0.8340 & 0.8638 & 1.00 & 0.6902 \\
			\midrule
			\multicolumn{8}{c}{Text Readability} \\
			\midrule
			Flesch Kincaid & 40.53 & 53.39 & 50.76 & 40.59 & 50.35 & 46.83 & 54.54 \\
			SMOG & 14.10 & 12.15 & 12.66 & 13.97 & 12.65 & 13.98 & 11.91 \\
			Dale Chall & 11.19 & 10.24 & 10.28 & 10.98 & 10.46 & 10.91 & 10.23 \\
			\midrule
			Normalized Score & 0.8340 & 0.9861 & 0.9555 & 0.8428 & 0.9476 & 0.8827 & 1.00 \\
			Cosine Similarity & 0.9905 & 1.00 & 0.9995 & 0.9912 & 0.9994 & 0.9966 & 1.00 \\
			\midrule
			\multicolumn{8}{c}{Sentiment Consistency} \\
			\midrule
			Negative & 0.3243 & 0.3439 & 0.3553 & 0.3268 & 0.3461 & 0.3264 & 0.3455 \\
			Neutral & 0.4030 & 0.3894 & 0.3819 & 0.4037 & 0.3841 & 0.4065 & 0.3996 \\
			Positive & 0.2727 & 0.2667 & 0.2718 & 0.2705 & 0.2698 & 0.2672 & 0.2549 \\
			\midrule
			Normalized Score & 0.9550 & 0.9752 & 0.9553 & 0.9594 & 0.9681 & 0.9606 & 1.0000 \\
			Cosine Similarity & 0.9989 & 1.00 & 0.9990 & 0.9991 & 0.9993 & 0.9992 & 1.00 \\
			\bottomrule
	\end{tabular}}
	\caption{Multi-dimensional human-machine text analysis results across seven texts: MGB-M, MAGA-M, RolePlaying-M, BPO-M, Self Refine-M, GDARL-CDMMR-M, and HWT. The evaluation covers content similarity, semantic similarity, lexical diversity, text readability, and sentiment consistency metrics.}
	\label{tab:multidimensional_analysis_all}
\end{table*}

\subsection{PPL Analysis}
\label{sec:ppl-analysis}
\begin{figure}[!t]
	\centering
	\includegraphics[width=1.0\linewidth]{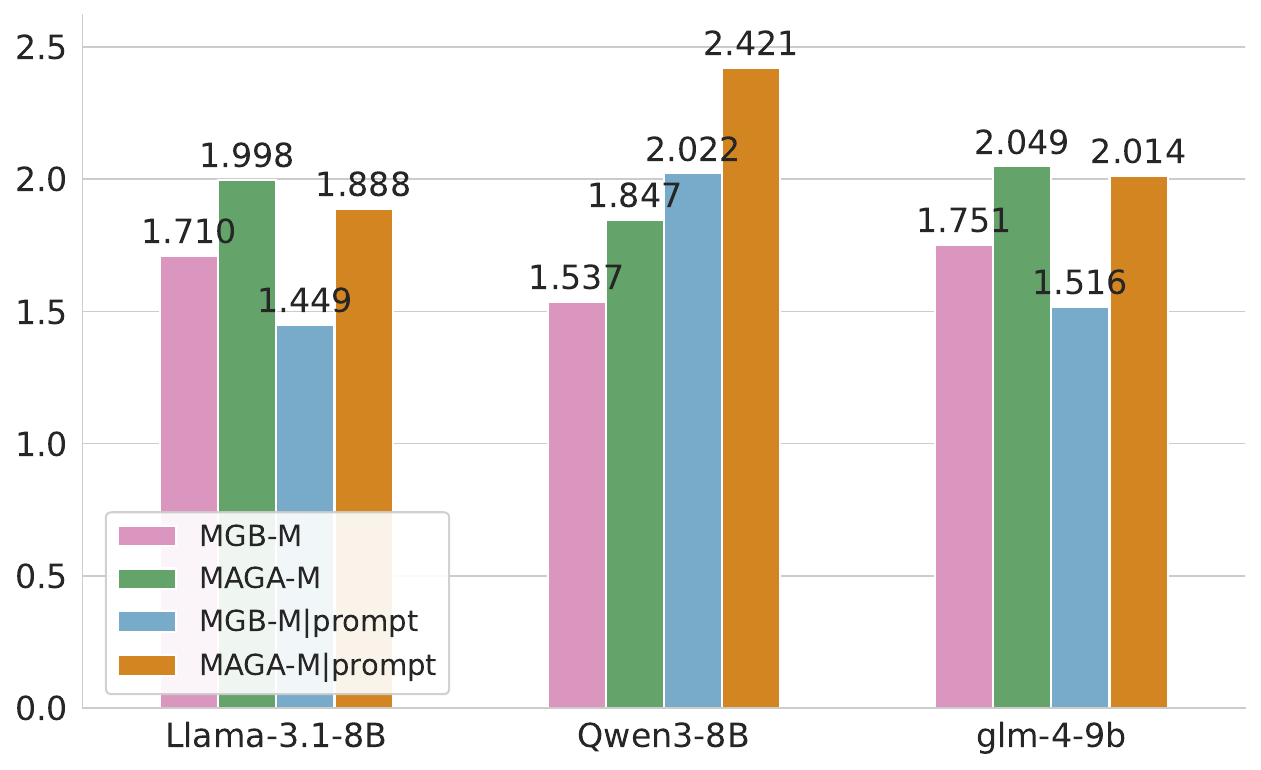}
	\caption{Compare the log PPL calculations with and without the prompt condition.}
	\label{fig:log-ppl-core}
\end{figure}

\begin{figure}[!t]
    \centering
    \includegraphics[width=1.0\linewidth]{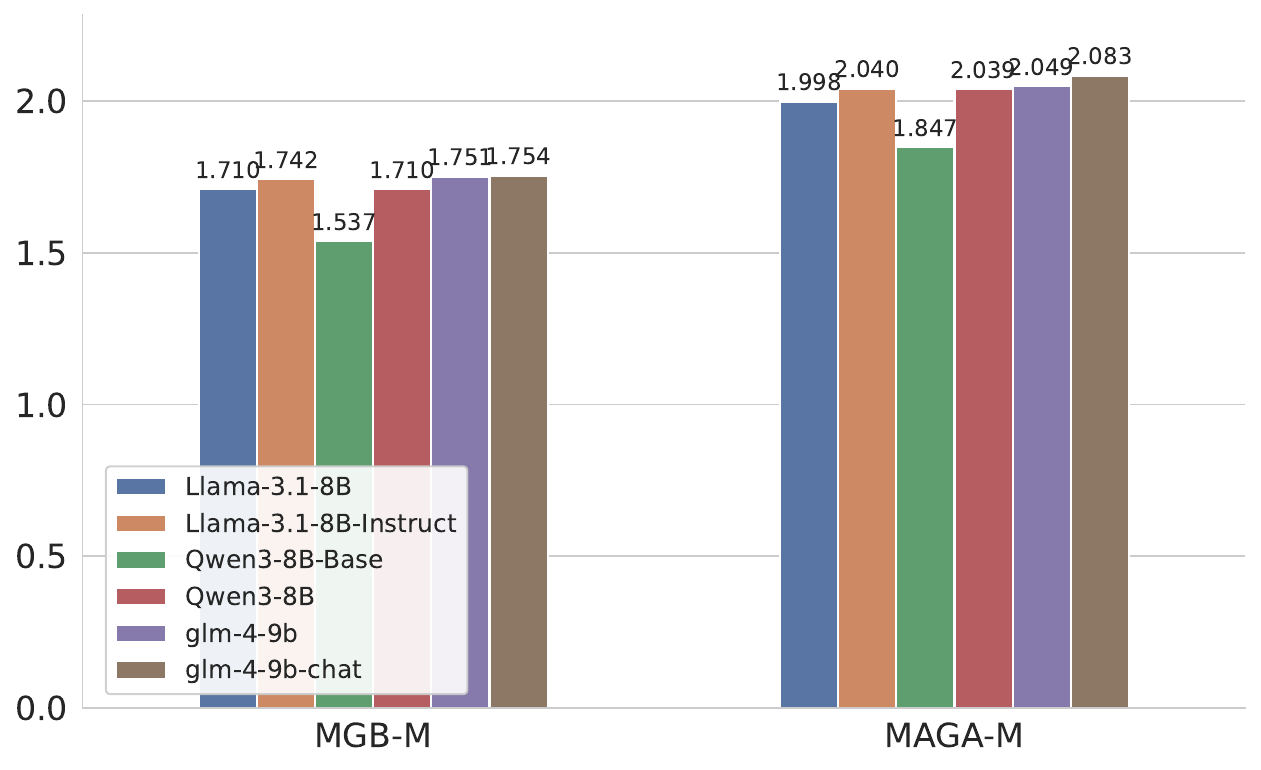}
    \caption{Direct log PPL computed on the base/chat LLM without the prompt condition.}
    \label{fig:log ppl-base-chat}
\end{figure}

Existing MGT Detection Benchmarks typically report the average perplexity (PPL) of MGTs within their datasets, often utilizing multiple evaluation models, as demonstrated in RAID \cite{dugan2024raid}. Following this convention, we present our statistical report in Figure~\ref{fig:log-ppl-core} and Figure~\ref{fig:log ppl-base-chat}. During the statistical process, several critical methodological questions arise, specifically whether to employ base models or chat (instruct) models, and whether to include the generation prompts during evaluation. These inquiries motivate the statistical experiments detailed in this section. It is worth noting that this section primarily analyzes and discusses these procedural questions as a supplementary investigation to our primary study, which maintains a relatively distant relation to the core tasks of MGT detection.

\begin{figure}[!t]
	\centering
	\includegraphics[width=1.0\linewidth]{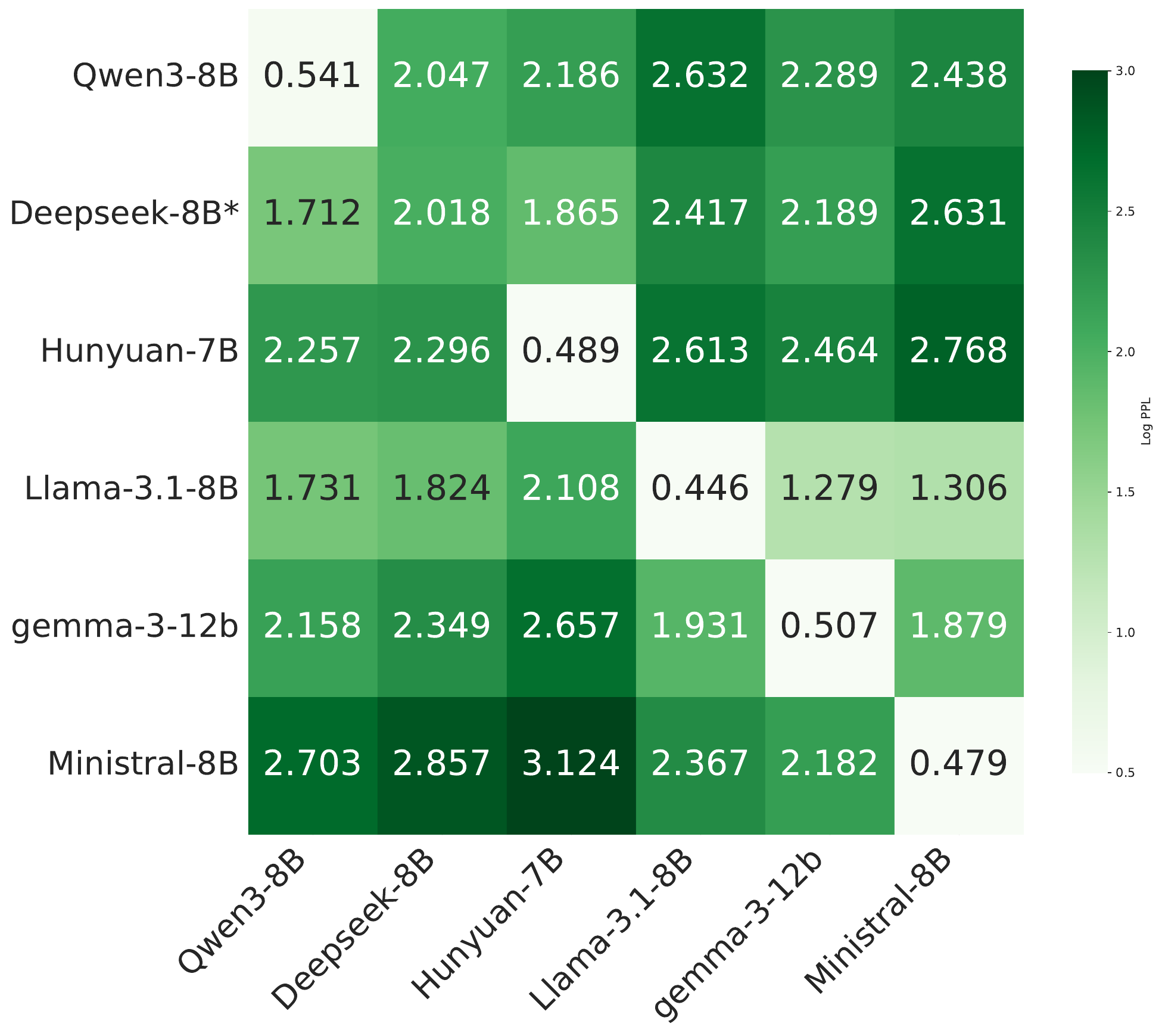}
	\caption{The log PPL calculated on MAGA-M, with the x-axis representing the LLM of the generated text and the y-axis representing the LLM used to calculate the log PPL. The calculation included prompts as conditions, but DeepSeek did not include chains-of-thoughts (COT) in its calculation.}
	\label{fig:log ppl model}
\end{figure}

As discussed in \S\ref{sec:ablation} and Appendix~\ref{sec:more experiment}, and illustrated in Figure~\ref{fig:log ppl model}, our core finding reveals a pronounced discrepancy in viewpoints among different generators. The variations in these discrepancies closely correspond to the performance gains achieved by different generator combinations within the GDARL-CMMR framework. We next address the methodological questions raised at the beginning of this section.

To conduct this analysis, we select three categories of language models, evaluating both the base and instruct variants for each. The first category comprises English-centric models used during the dataset generation process, specifically Llama-3.1-8B and Llama-3.1-8B-Instruct. The second category consists of Chinese-centric models also utilized in dataset generation, where we select Qwen3-8B-Base and Qwen3-8B. The third category includes models completely withheld from the dataset generation process, for which we choose glm-4-9b and glm-4-9b-chat.

In Figure~\ref{fig:log ppl-base-chat}, we evaluate both the base and chat versions of these models by calculating the log PPL directly without prompts. The results indicate that the choice between base and chat architectures yields negligible differences in direct perplexity estimation. This demonstrates that when computing PPL without prompts, the choice between base and instruct architectures introduces minimal variance, though base models exhibit marginally lower values.

In Figure~\ref{fig:log-ppl-core}, we compare the PPL calculated under conditional probability given the prompt against the PPL calculated without the prompt. In this holistic analysis, evaluation with prompts does not consistently yield lower PPL values and occasionally results in significantly higher values, a phenomenon potentially driven by the divergent viewpoints among different models. Intuitively, calculating PPL with prompts can better highlight these conflicting viewpoints across models. However, the differences observed on glm-4-9b remain extremely minor.

Both Figure~\ref{fig:log ppl-base-chat} and Figure~\ref{fig:log-ppl-core} demonstrate that the perplexity of MAGA-M is slightly higher than that of MGB-M. An increase in PPL does not fundamentally contradict alignment enhancement, as it likely stems from the viewpoint discrepancies among different models, especially since the calculation models are not the original authors of the text. Figure~\ref{fig:log ppl model} clearly indicates that models assign a substantially lower PPL to the MAGA-M texts they generated themselves. Consequently, while PPL remains a valuable statistic, it is not always a reliable metric for judging alignment quality.

Regarding the questions posed initially, we draw the following conclusions. First, for a dataset containing texts generated by a sufficiently diverse set of generators (which can be viewed as generator-agnostic analysis), utilizing either an internal model involved in generation or an external unseen model yields similar PPL results. Second, in generator-agnostic PPL analysis, the distinction between base and instruct models is minimal, and evaluating with or without prompts shows negligible variation. Therefore, researchers can reliably implement any single configuration consistently throughout their analysis.

Intuitively, using an external model, utilizing prompts whenever available, and prioritizing base models when prompts are absent might indeed seem preferable. Our findings demonstrate that these methodological variations introduce negligible differences, indicating that researchers can confidently adopt and maintain any single calculation pipeline without concern. Or in other words, it is preferable to follow the intuitively superior approach, as it appears more reliable intuitively, but doing so is not strictly essential, and one need not harbor any additional apprehensions if unable to do so.

\section{Training Hyperparameters}
\label{sec:hyperparameters}
\subsection{Common Hyperparameters}
\label{sec:common hyperparameters}
The fine-tuning of RoBERTa consistently employs a learning rate of 5e-5, a total batch size of 64, the AdamW optimizer with a weight decay of 0.01, a warmup ratio of 0.03, and a cosine learning rate scheduler. The classification head is initialized using Kaiming uniform initialization, while pre-trained parameters remain intact before fine-tuning. Following the baseline configuration reported in SemEval-M4 \cite{wang2024semeval}, the training duration is limited to one epoch. This single-epoch strategy is justified by three factors: the extensive training set exceeding 100k samples provides sufficient signal for convergence, the setting ensures a strictly fair comparison with established benchmarks, and it promotes robust generalization by preventing over-optimization.

For GDARL's RL process, we implement the Group Relative Policy Optimization (GRPO) algorithm \cite{shao2024deepseekmathpushinglimitsmathematical} integrated with Low-Rank Adaptation (LoRA) \cite{hu2021loralowrankadaptationlarge} to facilitate efficient training on our 8 $\times$ NVIDIA Tesla L20 48G GPUs. The LoRA configuration consists of a rank of 16 and an alpha value of 32. We maintain a micro batch size of 1 and a global batch size of 1024, achieved through a per-device batch size of 32 and a rollout factor of 4. The learning rate is 5e-6 with a cosine scheduler. The training duration is one epoch in each round of GDARL's RL process.

\begin{table}[!t]
	\centering 
%	\tiny
	\scriptsize
	\resizebox{\linewidth}{!}{
		\begin{tabular}{c|p{1.2cm}|p{2cm}} 
			\toprule
			\textbf{Group} & \textbf{Domain} & \textbf{Model}\\
			\midrule
			GA & Wikipedia, wikiHow, CC News, NPR News, S2ORC & Qwen3-plus, Qwen3-8B, DeepSeek-V3, DeepSeek-R1-0528-Qwen3-8B, Hunyuan-TurboS, Hunyuan-7B-Instruct \\
			\midrule
			GB & Reddit, Trustpilot Reviews, Amazon Reviews, Yahoo Answers, Natural Questions & Llama-3.1-8B-Instruct, Mistral-Medium, Ministral-8B-Instruct-2410, Gemini-2.0-flash, GPT-4o-mini, DeepSeek-V3 \\
			\bottomrule
		\end{tabular}
	}
	\caption{Overview of domains and evaluation models across different groups.}
	\label{tab:group_domain_models}
\end{table}

\subsection{GDARL-CDMMR Rounds and Group}
\label{sec:gdarl-cdmmr group}
The GDARL-CDMMR process comprises three adversarial training rounds. The final grouping of GDARL-CDMMR is presented in Table \ref{tab:group_domain_models}. These configurations are analyzed and discussed in detail within the Ablation Study Details in Appendix \ref{sec:more experiment}.

\begin{table*}[!t]
	\centering 
%	\tiny
%	\scriptsize
	\footnotesize
	\resizebox{\linewidth}{!}{
		\begin{tabular}{l|ccc|l} 
			\toprule
			\textbf{Split} & \textbf{Training Time} & \textbf{Inference Time} & \textbf{Total Time} & \textbf{Hardware}\\
			\midrule
			MGB & - & 28h & 28h & 8 $\times$ NVIDIA Tesla L20 48G GPUs \\
			MAGA-extra-GDARL-CDMMR & 530h & 28h & 558h & 8 $\times$ NVIDIA Tesla L20 48G GPUs \\
			MAGA-extra-BPO & - & 31h & 31h & 8 $\times$ NVIDIA Tesla L20 48G GPUs \\
			MAGA-extra-roleplaying & - & 29h & 29h & 8 $\times$ NVIDIA Tesla L20 48G GPUs \\
			MAGA-extra-self-refine & - & 108h & 108h & 8 $\times$ NVIDIA Tesla L20 48G GPUs \\
			MAGA & 543h & 112h & 655h & 8 $\times$ NVIDIA Tesla L20 48G GPUs \\
			\midrule
			\textbf{Total} & 1073h & 336h & 1409h & 8 $\times$ NVIDIA Tesla L20 48G GPUs \\
			\bottomrule
		\end{tabular}
	}
	\caption{Training, inference, and total time consumption across different splits on the English MAGA Dataset along with the hardware configuration specifications.}
	\label{tab:time_consumption}
\end{table*}

\section{Computational Overhead Analysis}
\label{sec:computational cost}
Table~\ref{tab:time_consumption} presents the computational time overhead for constructing each split within the English MAGA subset. The construction duration for the MAGA-cn subset exhibits similarly. The entire MAGA dataset requires approximately 2,800 hours to construct. Within each split, 72k MGTs are generated, with each individual model producing 6k MGTs. For open-source models, we employ vLLM for batch inference, which requires approximately 35 minutes per model to generate 6k MGTs. An exception is DeepSeek-R1-0528-Qwen3-8B, which requires approximately 1 hour due to the generation of chain-of-thought tokens. For commercial models, invoking APIs via a single-threaded process to generate 6k MGTs takes approximately 4 hours per model. The Self Refine process requires approximately four times the inference time of direct inference because it involves acquiring feedback and subsequent self-refinement. Regarding GDARL, among the models utilized, the smallest 7B model requires approximately 70 hours of training, whereas the largest 12B model requires approximately 150 hours. The training time for RoBERTa is negligible for GDARL. The BPO optimization of 72,000 prompts requires approximately 2 hours with a batch size of 64, implemented via the Hugging Face library. Distinct from other operations, the final trained RoBERTa detector on each split is fine-tuned on a training subsplit containing 120k instances, which consist of 60k MGTs and 60k HWTs, requiring approximately 35 minutes.

In general, regarding the numerical performance in our experiments, these methods exhibit comparable results, yet their computational time overheads vary significantly. GDARL-CDMMR demonstrates substantially lower practicality than other approaches, and Self Refine also presents slightly lower practicality than BPO and roleplaying. Nevertheless, our human analysis in Appendix~\ref{sec:case study} indicates that the intuitive effects of these methods are different. Furthermore, the ablation study details in Appendix~\ref{sec:more experiment} clearly demonstrate that integrating these methods consistently enhances mutual performance, despite a severe dilution effect.

\section{Discussion}
\subsection{Cross-Mutual Reward Component and Existing GDARL Frameworks}
\label{sec:GDARLvs}

\begin{table}[!t]
	\tiny
	\resizebox{\linewidth}{!}{
		\centering
		\begin{tabular}{l|c|c|c|c}
			\toprule
			 & \textbf{ACC} & \textbf{TPR} & \textbf{TNR} & \textbf{AUC} \\
			\toprule
			MGB-train & 99.78 & 99.72 & 99.84 & 99.99 \\
			MGB-validation  & 99.38 & 99.15 & 99.61 & 99.97 \\
			\bottomrule
		\end{tabular}
	}
	\caption{R-B MGB Performance on the MGB training and validation subsplits.}
	\label{tab:mgb_performance}
\end{table}

\begin{table}[!t]
	\tiny
	\resizebox{\linewidth}{!}{
		\centering
		\begin{tabular}{l|c|l|c}
			\toprule
			\multicolumn{2}{c|}{\textbf{Training}} & \multicolumn{2}{c}{\textbf{Validation}} \\
			\toprule
			\textbf{Score} & \textbf{Percentage} & \textbf{Score} & \textbf{Percentage} \\
			\midrule
			> 0.9999 & 46.97\% & > 0.9999 & 47.22\% \\
			> 0.9995 & 85.18\% & > 0.9995 & 84.35\% \\
			> 0.999  & 91.74\% & > 0.999  & 90.80\% \\
			> 0.995  & 96.79\% & > 0.995  & 95.97\% \\
			> 0.99   & 97.85\% & > 0.99   & 96.93\% \\
			> 0.9886 & 98.00\% & > 0.98   & 97.59\% \\
			> 0.98   & 98.55\% & > 0.9581 & 98.00\% \\
			> 0.9564 & 99.00\% & > 0.6304 & 99.00\% \\
			> 0.8192 & 99.50\% & > 0.1547 & 99.50\% \\
			> 0.0670 & 99.90\% & > 0.0056 & 99.90\% \\
			\bottomrule
		\end{tabular}
	}
	\caption{Cumulative distribution of machine-like scores given by R-B MGB on MGB training and validation splits and their corresponding percentage.}
	\label{tab:machine_like_scores_combined}
\end{table}

\begin{table}[!t]
%	\tiny
	\footnotesize
	\resizebox{\linewidth}{!}{
		\centering
		\begin{tabular}{l|c|c|p{1.2cm}|p{1.2cm}}
			\toprule
			 & \textbf{Wikipedia} & \textbf{wikiHow} & \textbf{Trustpilot Reviews} & \textbf{Amazon Reviews} \\
			\toprule
			MGB & 74.67 & 67.12 & 72.83 & 73.04 \\
			/w GDARL & 74.67 & 67.08 & 73.50 & 73.16 \\
			\bottomrule
		\end{tabular}
	}
	\caption{AUC performance with or without GDARL.}
	\label{tab:auc_w_wo_gdarl}
\end{table}

\begin{table*}[!t]
	\resizebox{\linewidth}{!}{
		% \small
		\centering
		\begin{tabular}{l|c|c|c|c|c|c}
			\toprule
			& \multicolumn{3}{c|}{\textbf{Attack}} 
			& \multicolumn{3}{c}{\textbf{Generalization}}  \\
			\toprule
			Metric & TPR & AUC & ACC(@FPR=5\%) & TPR & AUC & ACC(@FPR=5\%) \\
			\midrule
			MGB & 54.09 & 71.40 & 67.41 & 53.13 & 82.91 & 75.65 \\
			\midrule
			\hspace{1em}/w RADAR's framework & 48.73\textcolor{red}{(5.36$\downarrow$)} & 66.19\textcolor{red}{(5.21$\downarrow$)} & 64.73\textcolor{red}{(2.68$\downarrow$)} & 55.61\textcolor{green}{(2.48$\uparrow$)} & 76.87\textcolor{red}{(6.04$\downarrow$)} & 65.83\textcolor{red}{(9.82$\downarrow$)} \\
			\hspace{1em}/w RADAR's framework + CDMMR & 47.82\textcolor{red}{(6.27$\downarrow$)} & 65.60\textcolor{red}{(5.80$\downarrow$)} & 64.35\textcolor{red}{(3.06$\downarrow$)} & 55.26\textcolor{green}{(2.13$\uparrow$)} & 79.39\textcolor{red}{(3.52$\downarrow$)} & 71.56\textcolor{red}{(4.09$\downarrow$)} \\
			\hspace{1em}/w GDARL-CDMMR & 49.95\textcolor{red}{(4.14$\downarrow$)} & 67.91\textcolor{red}{(3.49$\downarrow$)} & 65.18\textcolor{red}{(2.23$\downarrow$)} & 60.97\textcolor{green}{(7.84$\uparrow$)} & 85.02\textcolor{green}{(2.11$\uparrow$)} & 77.28\textcolor{green}{(1.63$\uparrow$)} \\
			\bottomrule
	\end{tabular}}
	\caption{Ablation experiments vs RADAR.}
	\label{tab:ablation vs RADAR}
\end{table*}

\paragraph{Existing ARL Frameworks} To the best of our knowledge, existing Generator-Detector Adversarial Reinforcement Learning (GDARL) frameworks primarily include RADAR \cite{hu2023radar}, AuthorMist \cite{david2025authormistevadingaitext}, llm-detector-evasion \cite{nicks2023language}, and HUMPA \cite{wang2025humanizing}.

RADAR \cite{hu2023radar} addresses the vulnerability of existing machine-generated text detectors to paraphrased text by proposing a GDARL framework. This framework employs the proximal policy optimization (PPO) algorithm \cite{schulman2017proximalpolicyoptimizationalgorithms} to implement a co-training scheme where the generator and the detector alternate step by step within each epoch. The loss function requires simultaneously distinguishing among original AI text, AI-paraphrased text, and human text. By adjusting the loss weight $\lambda$, the framework balances detection performance between original and paraphrased texts.

Similarly, AuthorMist \cite{david2025authormistevadingaitext} focuses on GDARL for AI-paraphrased text. The authors select a suite of detectors and conduct adversarial learning against one detector at a time. They then evaluate the evasion effectiveness of the resulting AI-generated text across all selected detectors.

The llm-detector-evasion framework \cite{nicks2023language} utilizes the direct preference optimization (DPO) algorithm \cite{NEURIPS2023_a85b405e} to achieve GDARL. The authors demonstrate that current mainstream AI text detectors are highly susceptible to evasion, indicating that existing detection solutions lack reliability.

HUMPA \cite{wang2025humanizing} improves upon the method proposed in llm-detector-evasion \cite{nicks2023language}. The authors point out that directly fine-tuning a language model with a small parameter scale often compromises training stability. To address this, their first step involves scoring text using a detector and fine-tuning a small language model via DPO. Subsequently, they introduce an attack intensity hyperparameter $\alpha$ to proxy the attack during the decoding stage. This approach avoids modifying the parameters of the large language model. Instead, during the generation of each token, it applies an offset to the output of the large model based on the logit difference between the fine-tuned small model and its original counterpart.

\paragraph{Why don't we use a specific existing GDARL framework?} We do not adopt specific existing GDARL frameworks due to several distinct reasons. First, RADAR \cite{hu2023radar} explicitly reports their limitations that its method may show degraded detection performance against native LLM-generated texts (without paraphrasing). In the main experiments presented by RADAR, while its capacity to detect paraphrased text significantly outperforms the selected baselines, its performance on native text remains inferior to several baseline methods, demonstrated by that AUC is approximately 4\% lower than that of R-B GPT2. AuthorMist is likely similar, as it also focuses on paraphrased text. Second, the attack strength $\alpha$ of HUMPA \cite{wang2025humanizing} may be difficult to set based on the specific task. Although HUMPA \cite{wang2025humanizing} reports that its BERTScore decreases by no more than 0.02 and its ROUGE-1 score decreases by no more than 0.03, which can be interpreted as marginal changes, this performance degradation contradicts our objectives to some extent.

\paragraph{Why do we need Cross-Mutual Reward?} GDARL is the general term for existing GDARL algorithms. We also denote the vanilla formulation of these frameworks as GDARL, which we implement. To ensure optimal training stability, we eschew the co-training methodology of RADAR and instead execute alternation on an epoch-by-epoch basis, where the generator and the detector undergo alternate training for one full epoch per round. Under this setup, severe overfitting in RoBERTa prevents the LLM from receiving an effective reward signal. Table \ref{tab:mgb_performance} presents the performance of R-B MGB on the MGB training and validation subsplits, where the detection accuracy reaches approximately 100\% on the training subsplit and remains as high as 99.38\% on the validation subsplit. Table \ref{tab:mgb_performance} also outlines the machine-like scores assigned by R-B MGB to the machine-generated texts within these subsplits. Notably, on the training subsplit, 98.55\% of the machine-generated texts receive a score exceeding 0.98, and 99\% score above 0.9564. Consequently, the human-like score for almost all machine-generated text approaches zero, which eliminates viable reward gradients.

Table \ref{tab:auc_w_wo_gdarl} provides the supplementary results when integrating GDARL into the main diagonal AUC values of Figure \ref{fig:GDARL-CDMR-matrix-core}. Without any tailored modifications, applying standard GDARL yields virtually no performance gains.

We emphasize that our Cross-Mutual Reward (CMR) mechanism serves as a direct application of the theoretical foundations established in HC-Var \cite{xu-etal-2024-generalization}. HC-Var distinguishes between irrelevant and relevant features \cite{xu-etal-2024-generalization}, demonstrating that RoBERTa prone to overfitting on in-distribution data. When variables such as the prompts, domains, and generators used to produce MGT remain identical, RoBERTa relies heavily on irrelevant features (e.g. text length) for classification \cite{xu-etal-2024-generalization}. Scores derived under these conditions may also be unreliable to some degree. Conversely, out-of-distribution generalization must depend primarily on relevant features \cite{xu-etal-2024-generalization}, which represent the latent, intrinsic differences between human and machine text that remain fundamentally unidentified \cite{xu-etal-2024-generalization}.

Additionally, we present our initial trial experiments in Table \ref{tab:ablation vs RADAR}. As reported by RADAR \cite{hu2023radar}, its detection capability against vanilla MGT indeed sometimes tends to decline, rendering RADAR more suitable for identifying paraphrased text. It is critical to emphasize that our CMR component operates as an external procedural control mechanism, which remains inherently compatible with the inner workings of these algorithms rather than creating conflicts. As demonstrated in Table \ref{tab:ablation vs RADAR}, integrating the CMR component with RADAR consistently enhances overall performance under our experiment.

\subsection{Why not Distill?}
\label{sec:whynotdistill}
If we need more human-like text, why not directly perform knowledge distillation on commercial LLMs or simply use commercial LLMs? After all, commercial LLMs generally possess superior overall capabilities.

Many previous studies, including M4\cite{wang-etal-2024-m4}, have shown that increasing the diversity of LLM selections, especially incorporating a number of small-parameter LLMs, leads to significant improvements in detector learning.

CogLM\cite{wang-etal-2025-coglm} once pointed out through relevant metric analysis that LLMs with different parameter sizes can be regarded as humans of different ages, each with its unique style.

For example, in terms of overall knowledge difficulty, richness, application scope, and effectiveness, university-level knowledge is certainly superior to high school-level knowledge, which in turn is superior to junior high school-level knowledge. However, this does not mean that only university examinations are meaningful; junior and senior high schools also have their own unique challenging problems.

The characteristic of GDARL is to enable various open-source LLMs to generate more human-like MGT while maintaining their original styles.

\subsection{Why Single-Round Self Refine?}
\label{sec:selfrefineround}
Table \ref{tab:self_refine_ablation} presents the ablation study on the iteration rounds of Self-Refine. The experiment settings employed in this experiment is consistent with the analysis presented in Table \ref{tab:ablation}. Instead of allowing the LLM to determine the termination condition dynamically, we simply strictly enforce a predefined number of iteration rounds. The empirical results indicate that multiple iterations yield marginal performance variance, characterized primarily by fluctuations, within the scope of our study. As discussed in Appendix \ref{sec:computational cost}, the temporal overhead of a single-round Self Refine is approximately four times that of direct generation. Given this substantial increase in computational cost alongside the potential negligible performance gains, we systematically adopt a single-round configuration for Self Refine across all subsequent dataset construction and experimental procedures.

\begin{table*}[!t]
	\resizebox{\linewidth}{!}{
		% \small
		\centering
		\begin{tabular}{l|c|c|c|c|c|c}
			\toprule
			& \multicolumn{3}{c|}{\textbf{Attack}} 
			& \multicolumn{3}{c}{\textbf{Generalization}}  \\
			\toprule
			Metric & TPR & AUC & ACC(@FPR=5\%) & TPR & AUC & ACC(@FPR=5\%) \\
			\midrule
			MGB & 54.09 & 71.4 & 67.41 & 53.13 & 82.91 & 75.65 \\
			\midrule
			\hspace{1em}/w self-refine 1 round & 48.87\textcolor{red}{(5.22$\downarrow$)} & 66.32\textcolor{red}{(5.08$\downarrow$)} & 64.43\textcolor{red}{(2.98$\downarrow$)} & 64.21\textcolor{green}{(11.08$\uparrow$)} & 85.84\textcolor{green}{(2.93$\uparrow$)} & 77.47\textcolor{green}{(1.82$\uparrow$)} \\
			\hspace{1em}/w self-refine 2 round & 47.98\textcolor{red}{(6.11$\downarrow$)} & 65.87\textcolor{red}{(5.53$\downarrow$)} & 64.27\textcolor{red}{(3.14$\downarrow$)} & 63.02\textcolor{green}{(9.89$\uparrow$)} & 85.59\textcolor{green}{(2.68$\uparrow$)} & 77.53\textcolor{green}{(1.88$\uparrow$)} \\
			\hspace{1em}/w self-refine 3 round & 49.03\textcolor{red}{(5.06$\downarrow$)} & 66.34\textcolor{red}{(5.06$\downarrow$)} & 64.58\textcolor{red}{(2.83$\downarrow$)} & 63.14\textcolor{green}{(10.01$\uparrow$)} & 85.07\textcolor{green}{(2.16$\uparrow$)} & 77.31\textcolor{green}{(1.66$\uparrow$)} \\
			\hspace{1em}/w self-refine 4 round & 48.69\textcolor{red}{(5.40$\downarrow$)} & 66.74\textcolor{red}{(4.66$\downarrow$)} & 64.81\textcolor{red}{(2.60$\downarrow$)} & 62.85\textcolor{green}{(9.72$\uparrow$)} & 85.23\textcolor{green}{(2.32$\uparrow$)} & 77.18\textcolor{green}{(1.53$\uparrow$)} \\
			\bottomrule
	\end{tabular}}
	\caption{Self-refine rounds ablation experiments results.}
	\label{tab:self_refine_ablation}
\end{table*}

\section{Human Analysis Case Study}
\label{sec:case study}
Table \ref{tab:wikihow_how_to_be_healthy} presents our human analysis case study comparing various alignment methods. We conduct a manual analysis to identify the advantages and disadvantages of each method. Machine-generated traces are highlighted in red, whereas human-like traces are highlighted in green.

First, several drawbacks are observed in the MGT within the unenhanced MGB split. (1) The phrase "consistently perform behaviors" deliberately emphasizes the continuity principle of wikiHow and downplays one-time actions, which serves as a key signal of AI mimicking the practicality-oriented style of wikiHow. (2) The expression "physical, mental, and social function" appears to be an intentional imitation of the underlying structure of wikiHow health articles, directly adopting this three-dimensional formulation. (3) The clause "consuming adequate nutrition, engaging in regular movement, maintaining rest cycles, managing stress inputs, and avoiding known hazards" is a typical indicator of the detailed breakdown-style writing of AI. (4) The sentence "Execute these actions repeatedly over time" re-emphasizes long-term persistence; such double emphasis is a characteristic feature of AI-generated content.

For BPO, roleplaying, and self-refine, the machine traces exhibit similarities: these methods deliberately emphasize continuity instead of employing more natural and casual expressions, and they consistently decompose and stack core principles of wikiHow for illustration. Examples include "nutrition, exercise, sleep, and mental health" from roleplaying and "focus on eating well, staying active, getting enough rest, and managing stress" from self-refine.

For BPO, examples of human-like traces in BPO include the phrase "support your body and mind". When humans discuss health-related topics, they typically do not rigidly bifurcate "body" and "mind". Such concise and coherent phrases summarize the core meaning in a manner that is natural and empathetic while avoiding the accumulation of mechanical terminology. In terms of content, BPO aligns more closely with human expression habits.

For roleplaying, examples of human-like traces are as follows: the sentence "Being healthy is about balancing various aspects of your life" adopts the tone of college students conversing with peers, and "It is not about perfection, but consistency" reflects the empathetic mindset typical of such a demographic.

For self refine, examples of human-like traces include the sentence "You do not have to make huge changes all at once—small, consistent steps can lead to big improvements over time". This "empathy-first" guidance transforms the text from a cold guideline into warm advice, which better conforms to the encouraging logic of interpersonal communication. As a method involving self-reflection and refinement by LLMs, self-refine actually exhibits a stochastic rather than a single improvement dimension compared to the previous two methods, making it suitable for integration with other approaches to some degree.

Table \ref{tab:reddit_ipad_missing_feature} presents the advantages of GDARL-CDMMR, which provides a more moderate, implicit, and quantifiable optimization process. Enabled by the feedback mechanism of the detector, the most prominent characteristic of this method lies in its capability to eliminate explicit machine-generated artifacts. Even when the MAGA Additional Instruction is applied to explicitly prohibit the generation of introductory remarks, an implementation inspired by RAID \cite{dugan2024raid} and detailed in Appendix \ref{sec:prompt detail}, various LLMs, including commercial variants, still tend to insert opening sentences in unexpected manners. GDARL-CDMMR can mitigate such distinct machine artifacts that prompt engineering struggles to control, though its primary limitation is that it applies exclusively to open-source models rather than commercial ones.

The MGT in the MAGA split mitigates machine-specific artifacts across a wider perspective and at a deeper level by integrating the advantages of these four methodologies.

Table \ref{tab:zhihu_repurchase_snacks} presents a case study in Chinese. The human analysis results across Tables \ref{tab:reddit_ipad_missing_feature}, \ref{tab:zhihu_repurchase_snacks}, and Table \ref{tab:wikihow_how_to_be_healthy} exhibit similar patterns. Obvious machine-generated hints are highlighted in orange, while standard machine-generated and human-written traces are marked in red and green, respectively. Further redundant elaborations about Tables \ref{tab:reddit_ipad_missing_feature}, \ref{tab:zhihu_repurchase_snacks} are omitted due to the similar traces.
\clearpage
\begin{table*}[!t]
\centering 
\small
\begin{tabular}{l|p{0.41\linewidth}|p{0.35\linewidth}} 
\toprule
&\textbf{Original User Prompt}&\textbf{MAGA Additional Instruction}\\
\midrule
\textbf{Reddit}&Write just the body of a Reddit post titled "\{title\}". Do not repeat the title.&Do not repeat the title.\\
\midrule
\textbf{S2ORC}&Write the abstract for the scientific paper titled "\{title\}". It is preferable not to start with "This paper".&It is preferable not to start with "This paper".\\
\midrule
\textbf{Wikipedia}&Write the body of a Wikipedia article titled "\{title\}".& N/A\\
\midrule
\textbf{wikiHow}&Write the body of a wikiHow article titled "\{title\}".& N/A\\
\midrule
\textbf{Trustpilot Reviews}&Write the body of a Trustpilot review titled "\{title\}". Do not give it a title.&Do not give it a title.\\
\midrule
\textbf{Amazon Reviews}&Write the body of an Amazon review titled "\{title\}". Do not give it a title.&Do not give it a title.\\
\midrule
\textbf{Yahoo Answers}&Write just the response to the question titled "\{title\}" on Yahoo Answers. Do not repeat the question.&Do not repeat the question.\\
\midrule
\textbf{Natural Questions}&Provide the answer to the question "\{title\}".& N/A\\
\midrule
\textbf{CC News}&Write the body of a news article titled "\{title\}". Do not repeat the title.&Do not repeat the title.\\
\midrule
\textbf{NPR News}&Write the body of a NPR news article titled "\{title\}". Do not repeat the title.&Do not repeat the title.\\
\bottomrule
\end{tabular}
\caption{The text of the user prompts and additional instructions for all datasets. The field \{title\} was replaced with the corresponding title before being passed into the generative model. Before adding the BPO optimization prompt, we remove the suffix we call MAGA Additional Instruction, and then append it after optimization.}
\label{tab:generation prompts}
\end{table*}

\begin{table*}[!t]
\centering 
\small
\begin{tabular}{l|p{0.41\linewidth}|p{0.35\linewidth}} 
\toprule
&\textbf{Original User Prompt}&\textbf{MAGA Additional Instruction}\\
\midrule
\textbf{Baidu Tieba}&请你为标题为“\{title\}”的百度贴吧帖子写一条简短的网友回复。请直接给出回复。&请直接给出回复。\\
\midrule
\textbf{Weibo Review}&请你为标题为“\{title\}”的新浪微博文章写一条简短的网友评论。请直接给出评论。&请直接给出评论。\\
\midrule
\textbf{Rednote Review}&请你为标题为“\{title\}”的小红书笔记写一条网友评论。请直接给出评论。&请直接给出评论。\\
\midrule
\textbf{CSL}&请你写一段题目为《\{title\}》的中文核心期刊论文摘要。最好不要以“本文”开头。&最好不要以“本文”开头。\\
\midrule
\textbf{Baidu Baike}&请你写一篇标题为“\{title\}”的百度百科介绍。&N/A\\
\midrule
\textbf{Dianping}&请你根据下述关键词写一段大众点评评价。\textbackslash n关键词：\{title\}&N/A\\
\midrule
\textbf{Douban Review}&请你为电影《\{title\}》写一段简短的豆瓣影评。请直接给出评论。&请直接给出评论。\\
\midrule
\textbf{Baidu Zhidao}&请你为百度知道提问“\{title\}”写一段简短的回答。&N/A\\
\midrule
\textbf{Zhihu}&请你为知乎提问“\{title\}”写一段回答。请不要重复问题。&请不要重复问题。\\
\midrule
\textbf{CLTS}&请你写一篇题目为《\{title\}》的澎湃新闻。请不要重复题目，直接给出正文。&请不要重复题目，直接给出正文。\\
\bottomrule
\end{tabular}
\caption{The text of the user prompts and additional instructions for all Chinese datasets. The field \{title\} was replaced with the corresponding title before being passed into the generative model.}
\label{tab:generation prompts-cn}
\end{table*}

\begin{table*}[!t]
    \centering
    \footnotesize
    \resizebox{\linewidth}{!}{
    \begin{tabular}{l|p{0.45\linewidth}|p{0.45\linewidth}}
        \toprule
        & \textbf{Feedback Prompt} & \textbf{Refine Prompt} \\
        \midrule
        \textbf{Reddit} & Review the tone of the post you just wrote. Does it sound natural and human, or more robotic? If it's not human enough, suggest improvements to make it more conversational and engaging. Only give concise suggestions for improvement. Do not rewrite the post. & Please improve your Reddit post titled "\{title\}" to make it more conversational and engaging. Here are some specific suggestions:\textbackslash n\{feedback\} \\
        \midrule
        \textbf{S2ORC} & Review the abstract you just wrote. Does it meet the academic rigor of a scientific abstract while sounding natural (avoiding rigid jargon stacking or mechanical statement)? If it's not human enough, suggest improvements to enhance fluency. Only give concise suggestions for improvement. Do not rewrite the abstract. & Please improve your scientific paper abstract titled "\{title\}" to be more human while maintaining academic rigor. Here are some specific suggestions:\textbackslash n\{feedback\} \\
        \midrule
        \textbf{Wikipedia} & Review the Wikipedia article you just wrote. Does it sound like a human-edited encyclopedia entry? If it's not human enough, suggest improvements to enhance coherence and naturalness. Only give concise suggestions for improvement. Do not rewrite the article. & Please improve your Wikipedia article titled "\{title\}" to be more human while adhering to Wikipedia's neutrality standards. Here are some specific suggestions:\textbackslash n\{feedback\} \\
        \midrule
        \textbf{wikiHow} & Review the wikiHow article you just wrote. Does it have clear, practical steps and sound like a user-friendly guide? If it's not human enough, suggest improvements to enhance usability and approachability. Only give concise suggestions for improvement. Do not rewrite the article. & Please improve your wikiHow article titled "\{title\}" to be more human while keeping steps practical and clear. Here are some specific suggestions:\textbackslash n\{feedback\} \\
        \midrule
        \textbf{Trustpilot} & Review the tone of the review you just wrote. Does it sound natural and human, or more robotic? If it's not human enough, suggest improvements to make it more conversational and engaging. Only give concise suggestions for improvement. Do not rewrite the review. & Please improve your Trustpilot review titled "\{title\}" to be more human and conversational. Here are some specific suggestions:\textbackslash n\{feedback\}. \\
        \midrule
        \textbf{Amazon Reviews} & Review the tone of the review you just wrote. Does it sound natural and human, or more robotic? If it's not human enough, suggest improvements to make it more conversational and engaging. Only give concise suggestions for improvement. Do not rewrite the review. & Please improve your Amazon review titled "\{title\}" to be more human and conversational. Here are some specific suggestions:\textbackslash n\{feedback\}. \\
        \midrule
        \textbf{Yahoo Answers} & Review the tone of the Yahoo Answers response you just wrote. Does it match the platform's tone (moderately conversational, not too academic)? If it's not human enough, suggest improvements to enhance conversational naturalness. Only give concise suggestions for improvement. Do not rewrite the response. & Please improve your Yahoo Answers response to the question titled "\{title\}" to be more human and matching the platform's tone. Here are some specific suggestions:\textbackslash n\{feedback\} \\
        \midrule
        \textbf{Natural Questions} & Review the answer you just wrote. Does it accurately solve the question and balance conciseness with completeness while sounding natural? Suggest improvements to enhance naturalness. Only give concise suggestions for improvement. Do not rewrite the answer. & Please improve your answer to the question "\{title\}" to be more natural while ensuring accuracy. Here are some specific suggestions:\textbackslash n\{feedback\} \\
        \midrule
        \textbf{CC News} & Review the news article you just wrote. Does it follow news writing principles and sound like a human-written news piece? If it's not human enough, suggest improvements to enhance objectivity and fluency. Only give concise suggestions for improvement. Do not rewrite the article. & Please improve your news article titled "\{title\}" to be more human while maintaining journalistic objectivity. Here are some specific suggestions:\textbackslash n\{feedback\} \\
        \midrule
        \textbf{NPR News} & Review the NPR news article you just wrote. Does it match NPR's style? If it's not human enough, suggest improvements to enhance depth and approachability. Only give concise suggestions for improvement. Do not rewrite the article. & Please improve your NPR news article titled "\{title\}" to be more human while keeping NPR's professional and readable style. Here are some specific suggestions:\textbackslash n\{feedback\} \\
        \bottomrule
    \end{tabular}}
    \caption{Self-refine prompts for each domain}
    \label{tab:self refine prompts}
\end{table*}

\begin{table*}[!t]
    \centering
    \footnotesize
    \resizebox{\linewidth}{!}{
    \begin{tabular}{l|p{0.45\linewidth}|p{0.45\linewidth}}
        \toprule
        & \textbf{Feedback Prompt} & \textbf{Refine Prompt} \\
        \midrule
        \textbf{Baidu Tieba} & 请你回顾你刚撰写的百度贴吧回复，它看起来像是人类写的，还是更像机器生成的？如果它不够像人类写的，请给出改进建议使其更自然且具有对话性。请只给出简洁的改进建议。不要重写回复。 & 请你优化你为标题为“\{title\}”的百度贴吧帖子撰写的回复，保持简短并使其更自然且具有对话性。具体改进建议如下：\textbackslash n\{feedback\} \\
        \midrule
        \textbf{Weibo Review} & 请你回顾你刚撰写的新浪微博评论，它看起来像是人类写的，还是更像机器生成的？如果它不够像人类写的，请给出改进建议使其更自然且具有对话性。请只给出简洁的改进建议。不要重写评论。 & 请你优化你为标题为“\{title\}”的新浪微博文章撰写的评论，保持简短并使其更自然且具有对话性。具体改进建议如下：\textbackslash n\{feedback\} \\
        \midrule
        \textbf{Rednote Review} & 请你回顾你刚撰写的小红书评论，它看起来像是人类写的，还是更像机器生成的？如果它不够像人类写的，请给出改进建议使其更自然且具有对话性。请只给出简洁的改进建议。不要重写评论。 & 请你优化你为标题为“\{title\}”的小红书笔记撰写的评论，使其更自然且具有对话性。具体改进建议如下：\textbackslash n\{feedback\} \\
        \midrule
        \textbf{CSL} & 请你回顾你刚撰写的中文核心期刊论文摘要，是否兼具学术严谨性和语言流畅性（避免堆砌专业术语或机械陈述）？如果它不够像人类写的，请给出改进建议以增强表达的流畅性。请只给出简洁的改进建议。不要重写摘要。 & 请你优化你撰写的题目为《\{title\}》的中文核心期刊论文摘要，保持学术严谨性的同时使其更像人类写的。具体改进建议如下：\textbackslash n\{feedback\} \\
        \midrule
        \textbf{Baidu Baike} & 请你回顾你刚撰写的百度百科介绍，它是否像是人工编写的百科介绍？如果它不够像人类写的，请给出改进建议以增强表达连贯性和自然性。请只给出简洁的改进建议。不要重写百科介绍。 & 请你优化你撰写的标题为“\{title\}”的百度百科介绍，保持百科中立性标准的前提下使其更像人类写的。具体改进建议如下：\textbackslash n\{feedback\} \\
        \midrule
        \textbf{Dianping} & 请你回顾你刚撰写的大众点评评价，它看起来是否自然、像人类写的，还是更像机器生成的？如果它不够像人类写的，请给出改进建议使其更自然且具有对话性。请只给出简洁的改进建议。不要重写评价。 & 请你优化你基于关键词“\{title\}”撰写的大众点评评价，使其更具对话性、像人类写的。具体改进建议如下：\textbackslash n\{feedback\} \\
        \midrule
        \textbf{Douban Review} & 请你回顾你刚撰写的豆瓣影评，它看起来是否自然、像人类写的，还是更像机器生成的？如果它不够像人类写的，请给出改进建议使其更自然且具有对话性。请只给出简洁的改进建议。不要重写评论。 & 请你优化你为电影《\{title\}》撰写的豆瓣影评，保持简短并使其更具对话性、像人类写的。具体改进建议如下：\textbackslash n\{feedback\} \\
        \midrule
        \textbf{Baidu Zhidao} & 请你回顾你刚撰写的百度知道回答，它是否兼顾准确性和实用性，同时读起来自然流畅？请给出改进建议增强表达自然性。请只给出简洁的改进建议。不要重写回答。 & 请你优化你为百度知道提问“\{title\}”撰写的回答，保持简洁、准确与实用的同时使其更像人类写的。具体改进建议如下：\textbackslash n\{feedback\} \\
        \midrule
        \textbf{Zhihu} & 请你回顾你刚撰写的知乎回答，有观点、有依据的同时，它是否符合平台的语气风格（适度对话，不要过于学术化）？如果它不够像人类写的，请给出改进建议以增强表达自然性。请只给出简洁的改进建议。不要重写回答。 & 请你优化你为知乎提问“\{title\}”撰写的回答，使其更像人类写的，并符合平台的风格。具体改进建议如下：\textbackslash n\{feedback\} \\
        \midrule
        \textbf{CLTS} & 请你回顾你刚撰写的澎湃新闻正文，它是否符合新闻写作的客观性和专业性原则，并读起来像是人类写的？如果它不够像人类写的，请给出改进建议以增强客观性和流畅性。请只给出简洁的改进建议。不要重写文章。 & 请你优化你撰写的题目为《\{title\}》的澎湃新闻，保持新闻客观性的同时使其更像人类写的。具体改进建议如下：\textbackslash n\{feedback\} \\
        \bottomrule
    \end{tabular}}
    \caption{Self-refine Chinese prompts for each domain}
    \label{tab:self refine prompts Chinese}
\end{table*}

\begin{table*}[!t]
    \centering
    \footnotesize
    \resizebox{\linewidth}{!}{
    \begin{tabular}{p{0.48\linewidth}|p{0.48\linewidth}}
        \toprule
        \textbf{English} & \textbf{Chinese} \\
        \midrule
        You are American. Please respond in a casual, conversational American English style. & 你是老北京人儿。请你用地道的北京腔回复。 \\
        \midrule
        You are Winston Churchill. Respond with formal, eloquent British English, using powerful rhetoric, patriotism, and gravitas in your language. & 你是鲁迅，兼具文学家的犀利与斗士的坚定。请你用凝练有力、以讽刺隐喻针砭时弊、于冷峻中藏人文关怀的语言回复。 \\
        \midrule
        You are Sherlock Holmes. Respond with formal, analytical language, sharp wit, and precise reasoning. & 你是诸葛亮，足智多谋、神机妙算、儒雅有情怀。请你用沉稳谦和、条理清晰、直击要害的语言回复。 \\
        \midrule
        You are a librarian. Respond in a quiet, detail-oriented, and helpful tone—like you're explaining a catalog entry or sharing a note from your records, calm and precise. & 你是图书管理员，请你用温和细致、乐于助人的语气回复，就像在耐心讲解图书目录条目。 \\
        \midrule
        You are a firefighter. Respond in a no-nonsense, straightforward, but approachable tone—like you're debriefing a colleague or updating a report, calm and matter-of-fact, no extra fluff. & 你是消防员，请你用务实直接、亲切易懂的语气回复，好比向同事汇报火情或更新救援进展，冷静客观，不拖泥带水。 \\
        \midrule
        You are a doctor. Respond in a calm, authoritative, and empathetic tone—like you're updating a nurse about a patient's care, precise with medical details but mindful of the human side. & 你是医生，请你用冷静权威、富有同理心的语气回复，宛如向护士更新患者病情，精准传达医疗细节，兼顾人文关怀。 \\
        \midrule
        You are a teacher. Respond in a patient, firm, and caring tone. & 你是教师，请你用耐心温和、坚定有度、满含关怀的语气回复，既严又慈。 \\
        \midrule
        You are a student (college-aged). Respond in a relatable, slightly frazzled but earnest tone—like you're venting to a classmate before a lecture, juggling notes and deadlines, casual and conversational. & 你是大学生，请你用亲切易懂、略显忙碌却认真诚恳的语气回复，宛如课前与同学交流，忙着整理笔记、赶截止日期，口语化且自然真实。 \\
        \midrule
        You are a police officer. Respond in a calm, professional, no-frills tone. & 你是警察，请你用冷静专业、简洁明了的语气回复，不冗余拖沓。 \\
        \midrule
        You are an artist. Respond in a creative, passionate, and reflective tone—like you're talking to a gallery owner about your work, excited about your process, no pretense. & 你是艺术家，请你用富有创意、充满激情、善于反思的语气回复，宛如与画廊老板谈论自己的作品，为创作过程满心雀跃，不矫揉造作。 \\
        \bottomrule
    \end{tabular}}
    \caption{Samples of Role-Playing Prompts}
    \label{tab:role playing prompts}
\end{table*}

\clearpage

\clearpage

\begin{table*}[!t]
    \centering
    % \normalsize
    % \small
    % \footnotesize
    % \scriptsize
    \tiny
    \resizebox{\linewidth}{!}{
    \begin{tabular}{l|c|c|c|c}
        \toprule
        & ACC & TPR & AUC & ACC(@FPR=5\%) \\
        \midrule
        R-B GPT2   & 56.06\textcolor{red}{(2.38$\downarrow$)} & 29.21\textcolor{red}{(4.76$\downarrow$)} & 52.49\textcolor{red}{(4.42$\downarrow$)} & 57.02\textcolor{red}{(2.92$\downarrow$)} \\
        R-L GPT2   & 54.25\textcolor{red}{(2.44$\downarrow$)}  & 15.46\textcolor{red}{(4.88$\downarrow$)}   & 48.51\textcolor{red}{(4.94$\downarrow$)} & 54.09\textcolor{red}{(2.88$\downarrow$)}  \\
        R-B CGPT   & 53.02\textcolor{red}{(4.58$\downarrow$)} & 24.42\textcolor{red}{(9.15$\downarrow$)}  & 58.48\textcolor{red}{(5.49$\downarrow$)} & 51.82\textcolor{red}{(2.91$\downarrow$)}  \\
        RADAR      & 57.61\textcolor{red}{(2.22$\downarrow$)}  & 25.13\textcolor{red}{(4.43$\downarrow$)}   & 59.56\textcolor{red}{(3.87$\downarrow$)}  & 57.51\textcolor{red}{(2.82$\downarrow$)}  \\
        SCRN       & 73.72\textcolor{red}{(0.54$\downarrow$)}  & 93.15\textcolor{red}{(1.08$\downarrow$)}   & 83.52\textcolor{red}{(1.24$\downarrow$)}  & 59.43\textcolor{red}{(0.54$\downarrow$)}  \\
        DETree     & 58.00\textcolor{red}{(0.05$\downarrow$)}  & 99.83\textcolor{red}{(0.11$\downarrow$)}   & 93.38\textcolor{red}{(0.54$\downarrow$)}  & 86.29\textcolor{red}{(0.94$\downarrow$)}  \\
        \midrule
        neutral.avg& 58.77\textcolor{red}{(2.03$\downarrow$)}  & 47.87\textcolor{red}{(4.07$\downarrow$)}  & 65.99\textcolor{red}{(3.42$\downarrow$)} & 61.03\textcolor{red}{(2.17$\downarrow$)}  \\
        \midrule
        GLTR       & - & - & 66.93\textcolor{red}{(4.38$\downarrow$)}  & 66.34\textcolor{red}{(3.93$\downarrow$)}  \\
        F-DetectGPT& 79.22\textcolor{red}{(0.75$\downarrow$)}  & 74.76\textcolor{red}{(1.51$\downarrow$)}   & 83.67\textcolor{red}{(0.34$\downarrow$)}  & 79.25\textcolor{red}{(0.71$\downarrow$)}  \\
        LLMDet     & 47.10\textcolor{red}{(3.36$\downarrow$)} & 22.02\textcolor{red}{(6.72$\downarrow$)}  & 42.14\textcolor{red}{(5.75$\downarrow$)} & 53.91\textcolor{red}{(2.98$\downarrow$)}  \\
        Binoculars & 81.72\textcolor{red}{(2.55$\downarrow$)}  & 78.84\textcolor{red}{(5.09$\downarrow$)}  & 83.09\textcolor{red}{(3.67$\downarrow$)}  & 80.06\textcolor{red}{(3.17$\downarrow$)}  \\
        DALD       & 78.82\textcolor{red}{(1.19$\downarrow$)}  & 75.63\textcolor{red}{(2.38$\downarrow$)}   & 83.72\textcolor{red}{(0.77$\downarrow$)}  & 78.73\textcolor{red}{(1.06$\downarrow$)}  \\
        GECScore   & 53.62\textcolor{red}{(2.73$\downarrow$)}  & 10.97\textcolor{red}{(5.45$\downarrow$)}  & 59.43\textcolor{red}{(6.49$\downarrow$)}  & 56.49\textcolor{red}{(3.08$\downarrow$)}  \\
        \midrule
        metric.avg & 68.09\textcolor{red}{(2.12$\downarrow$)}  & 52.44\textcolor{red}{(4.23$\downarrow$)}   & 69.83\textcolor{red}{(3.57$\downarrow$)}  & 69.13\textcolor{red}{(2.49$\downarrow$)}  \\
        \midrule
        avg        & 63.01\textcolor{red}{(2.07$\downarrow$)}  & 49.95\textcolor{red}{(4.14$\downarrow$)}  & 67.91\textcolor{red}{(3.49$\downarrow$)}  & 65.08\textcolor{red}{(2.33$\downarrow$)}  \\
        \bottomrule
    \end{tabular}}
    \caption{MAGA-extra-CDMMR Bench.}
    \label{tab:MAGA-extra-GDARL-CDMMR Bench}
\end{table*}

\begin{table*}[!t]
    \centering
    \tiny
    \resizebox{\linewidth}{!}{
    \begin{tabular}{l|c|c|c|c}
        \toprule
        & ACC & TPR & AUC & ACC(@FPR=5\%) \\
        \midrule
        R-B GPT2   & 54.79\textcolor{red}{(3.65$\downarrow$)} & 26.68\textcolor{red}{(7.29$\downarrow$)} & 49.49\textcolor{red}{(7.42$\downarrow$)} & 56.43\textcolor{red}{(3.51$\downarrow$)} \\
        R-L GPT2   & 53.75\textcolor{red}{(2.94$\downarrow$)} & 14.47\textcolor{red}{(5.87$\downarrow$)} & 43.83\textcolor{red}{(9.62$\downarrow$)} & 53.42\textcolor{red}{(3.55$\downarrow$)} \\
        R-B CGPT   & 50.71\textcolor{red}{(6.89$\downarrow$)} & 19.79\textcolor{red}{(13.78$\downarrow$)} & 54.01\textcolor{red}{(9.96$\downarrow$)} & 51.03\textcolor{red}{(3.70$\downarrow$)} \\
        RADAR   & 56.96\textcolor{red}{(2.87$\downarrow$)} & 23.83\textcolor{red}{(5.73$\downarrow$)} & 55.73\textcolor{red}{(7.70$\downarrow$)} & 57.72\textcolor{red}{(2.61$\downarrow$)} \\
        SCRN   & 73.85\textcolor{red}{(0.41$\downarrow$)} & 93.41\textcolor{red}{(0.82$\downarrow$)} & 84.98\textcolor{green}{(0.22$\uparrow$)} & 60.35\textcolor{green}{(0.38$\uparrow$)} \\
        DETree   & 58.06\textcolor{green}{(0.01$\uparrow$)} & 99.96\textcolor{green}{(0.02$\uparrow$)} & 93.45\textcolor{red}{(0.47$\downarrow$)} & 86.89\textcolor{red}{(0.34$\downarrow$)} \\
        \midrule
        neutral.avg   & 58.02\textcolor{red}{(2.79$\downarrow$)} & 46.36\textcolor{red}{(5.58$\downarrow$)} & 63.58\textcolor{red}{(5.83$\downarrow$)} & 60.97\textcolor{red}{(2.22$\downarrow$)} \\
        \midrule
        GLTR   & - & - & 66.24\textcolor{red}{(5.07$\downarrow$)} & 66.02\textcolor{red}{(4.25$\downarrow$)} \\
        F-DetectGPT   & 79.05\textcolor{red}{(0.92$\downarrow$)} & 74.42\textcolor{red}{(1.85$\downarrow$)} & 83.49\textcolor{red}{(0.52$\downarrow$)} & 79.23\textcolor{red}{(0.73$\downarrow$)} \\
        LLMDet   & 46.12\textcolor{red}{(4.34$\downarrow$)} & 20.06\textcolor{red}{(8.68$\downarrow$)} & 41.52\textcolor{red}{(6.37$\downarrow$)} & 52.52\textcolor{red}{(4.37$\downarrow$)} \\
        Binoculars   & 81.79\textcolor{red}{(2.48$\downarrow$)} & 78.98\textcolor{red}{(4.95$\downarrow$)} & 83.47\textcolor{red}{(3.29$\downarrow$)} & 78.49\textcolor{red}{(4.74$\downarrow$)} \\
        DALD   & 78.98\textcolor{red}{(1.03$\downarrow$)} & 75.95\textcolor{red}{(2.06$\downarrow$)} & 83.38\textcolor{red}{(1.11$\downarrow$)} & 78.74\textcolor{red}{(1.05$\downarrow$)} \\
        GECScore   & 52.45\textcolor{red}{(3.90$\downarrow$)} & 8.63\textcolor{red}{(7.79$\downarrow$)} & 57.57\textcolor{red}{(8.35$\downarrow$)} & 55.76\textcolor{red}{(3.81$\downarrow$)} \\
        \midrule
        metric.avg   & 67.68\textcolor{red}{(2.53$\downarrow$)} & 51.61\textcolor{red}{(5.07$\downarrow$)} & 69.28\textcolor{red}{(4.12$\downarrow$)} & 68.46\textcolor{red}{(3.16$\downarrow$)} \\
        \midrule
        avg   & 62.41\textcolor{red}{(2.67$\downarrow$)} & 48.74\textcolor{red}{(5.35$\downarrow$)} & 66.43\textcolor{red}{(4.97$\downarrow$)} & 64.72\textcolor{red}{(2.69$\downarrow$)} \\
        \bottomrule
    \end{tabular}}
    \caption{MAGA-extra-roleplaying Bench.}
    \label{tab:MAGA-extra-roleplaying Bench}
\end{table*}

\begin{table*}[!t]
    \centering
    \tiny
    \resizebox{\linewidth}{!}{
    \begin{tabular}{l|c|c|c|c}
        \toprule
        & ACC & TPR & AUC & ACC(@FPR=5\%) \\
        \midrule
        R-B GPT2   & 55.48\textcolor{red}{(2.96$\downarrow$)} & 28.06\textcolor{red}{(5.91$\downarrow$)} & 51.73\textcolor{red}{(5.18$\downarrow$)} & 57.69\textcolor{red}{(2.25$\downarrow$)} \\
        R-L GPT2   & 54.21\textcolor{red}{(2.48$\downarrow$)}  & 15.39\textcolor{red}{(4.95$\downarrow$)}   & 46.74\textcolor{red}{(6.71$\downarrow$)} & 54.17\textcolor{red}{(2.80$\downarrow$)}  \\
        R-B CGPT   & 52.12\textcolor{red}{(5.48$\downarrow$)} & 22.61\textcolor{red}{(10.96$\downarrow$)}  & 57.02\textcolor{red}{(6.95$\downarrow$)} & 52.12\textcolor{red}{(2.61$\downarrow$)}  \\
        RADAR      & 57.51\textcolor{red}{(2.32$\downarrow$)}  & 24.93\textcolor{red}{(4.63$\downarrow$)}   & 59.04\textcolor{red}{(4.39$\downarrow$)}  & 58.62\textcolor{red}{(1.71$\downarrow$)}  \\
        SCRN       & 73.94\textcolor{red}{(0.32$\downarrow$)}  & 93.59\textcolor{red}{(0.64$\downarrow$)}   & 83.97\textcolor{red}{(0.79$\downarrow$)}  & 59.58\textcolor{red}{(0.39$\downarrow$)}  \\
        DETree     & 57.97\textcolor{red}{(0.08$\downarrow$)}  & 99.78\textcolor{red}{(0.16$\downarrow$)}   & 93.71\textcolor{red}{(0.21$\downarrow$)}  & 86.89\textcolor{red}{(0.34$\downarrow$)}  \\
        \midrule
        neutral.avg& 58.54\textcolor{red}{(2.27$\downarrow$)}  & 47.39\textcolor{red}{(4.54$\downarrow$)}  & 65.37\textcolor{red}{(4.04$\downarrow$)} & 61.51\textcolor{red}{(1.68$\downarrow$)}  \\
        \midrule
        GLTR       & - & - & 68.34\textcolor{red}{(2.97$\downarrow$)}  & 67.97\textcolor{red}{(2.30$\downarrow$)}  \\
        F-DetectGPT& 79.57\textcolor{red}{(0.41$\downarrow$)}  & 75.46\textcolor{red}{(0.81$\downarrow$)}   & 83.70\textcolor{red}{(0.31$\downarrow$)}  & 79.57\textcolor{red}{(0.39$\downarrow$)}  \\
        LLMDet     & 46.83\textcolor{red}{(3.63$\downarrow$)} & 21.49\textcolor{red}{(7.25$\downarrow$)}  & 43.51\textcolor{red}{(4.38$\downarrow$)} & 53.18\textcolor{red}{(3.71$\downarrow$)}  \\
        Binoculars & 83.55\textcolor{red}{(0.71$\downarrow$)}  & 82.51\textcolor{red}{(1.42$\downarrow$)}  & 86.02\textcolor{red}{(0.74$\downarrow$)}  & 81.87\textcolor{red}{(1.36$\downarrow$)}  \\
        DALD       & 79.28\textcolor{red}{(0.73$\downarrow$)}  & 76.54\textcolor{red}{(1.47$\downarrow$)}   & 84.01\textcolor{red}{(0.48$\downarrow$)}  & 79.13\textcolor{red}{(0.66$\downarrow$)}  \\
        GECScore   & 53.62\textcolor{red}{(2.73$\downarrow$)}  & 10.96\textcolor{red}{(5.46$\downarrow$)}  & 59.33\textcolor{red}{(6.59$\downarrow$)}  & 55.04\textcolor{red}{(4.53$\downarrow$)}  \\
        \midrule
        metric.avg & 68.57\textcolor{red}{(1.64$\downarrow$)}  & 53.39\textcolor{red}{(3.28$\downarrow$)}   & 70.82\textcolor{red}{(2.58$\downarrow$)}  & 69.46\textcolor{red}{(2.16$\downarrow$)}  \\
        \midrule
        avg        & 63.10\textcolor{red}{(1.98$\downarrow$)}  & 50.12\textcolor{red}{(3.97$\downarrow$)}  & 68.09\textcolor{red}{(3.31$\downarrow$)}  & 65.49\textcolor{red}{(1.92$\downarrow$)}  \\
        \bottomrule
    \end{tabular}}
    \caption{MAGA-extra-BPO Bench.}
    \label{tab:MAGA-extra-BPO Bench}
\end{table*}

\begin{table*}[!t]
    \centering
    \tiny
    \resizebox{\linewidth}{!}{
    \begin{tabular}{l|c|c|c|c}
        \toprule
        & ACC & TPR & AUC & ACC(@FPR=5\%) \\
        \midrule
        R-B GPT2   & 54.44\textcolor{red}{(4.00$\downarrow$)} & 25.98\textcolor{red}{(7.99$\downarrow$)} & 50.42\textcolor{red}{(6.49$\downarrow$)} & 55.59\textcolor{red}{(4.35$\downarrow$)} \\
        R-L GPT2   & 53.14\textcolor{red}{(3.55$\downarrow$)}  & 13.24\textcolor{red}{(7.10$\downarrow$)}   & 45.57\textcolor{red}{(7.88$\downarrow$)} & 52.81\textcolor{red}{(4.16$\downarrow$)}  \\
        R-B CGPT   & 52.51\textcolor{red}{(5.09$\downarrow$)} & 23.39\textcolor{red}{(10.18$\downarrow$)}  & 55.67\textcolor{red}{(8.30$\downarrow$)} & 50.27\textcolor{red}{(4.46$\downarrow$)}  \\
        RADAR      & 56.49\textcolor{red}{(3.34$\downarrow$)}  & 22.89\textcolor{red}{(6.67$\downarrow$)}   & 57.75\textcolor{red}{(5.68$\downarrow$)}  & 56.62\textcolor{red}{(3.71$\downarrow$)}  \\
        SCRN       & 74.33\textcolor{green}{(0.07$\uparrow$)}  & 94.37\textcolor{green}{(0.14$\uparrow$)}   & 84.12\textcolor{red}{(0.64$\downarrow$)}  & 59.56\textcolor{red}{(0.41$\downarrow$)}  \\
        DETree     & 57.97\textcolor{red}{(0.08$\downarrow$)}  & 99.78\textcolor{red}{(0.16$\downarrow$)}   & 94.06\textcolor{green}{(0.14$\uparrow$)}  & 87.69\textcolor{green}{(0.46$\uparrow$)}  \\
        \midrule
        neutral.avg& 58.14\textcolor{red}{(2.66$\downarrow$)}  & 46.61\textcolor{red}{(5.33$\downarrow$)}  & 64.60\textcolor{red}{(4.81$\downarrow$)} & 60.42\textcolor{red}{(2.77$\downarrow$)}  \\
        \midrule
        GLTR       & - & - & 64.67\textcolor{red}{(6.64$\downarrow$)}  & 64.91\textcolor{red}{(5.36$\downarrow$)}  \\
        F-DetectGPT& 79.23\textcolor{red}{(0.75$\downarrow$)}  & 74.78\textcolor{red}{(1.49$\downarrow$)}   & 83.54\textcolor{red}{(0.47$\downarrow$)}  & 79.09\textcolor{red}{(0.87$\downarrow$)}  \\
        LLMDet     & 45.98\textcolor{red}{(4.48$\downarrow$)} & 19.78\textcolor{red}{(8.96$\downarrow$)}  & 39.21\textcolor{red}{(8.68$\downarrow$)} & 53.27\textcolor{red}{(3.62$\downarrow$)}  \\
        Binoculars & 81.17\textcolor{red}{(3.10$\downarrow$)}  & 77.74\textcolor{red}{(6.19$\downarrow$)}  & 81.23\textcolor{red}{(5.53$\downarrow$)}  & 78.34\textcolor{red}{(4.89$\downarrow$)}  \\
        DALD       & 78.97\textcolor{red}{(1.04$\downarrow$)}  & 75.93\textcolor{red}{(2.08$\downarrow$)}   & 83.37\textcolor{red}{(1.12$\downarrow$)}  & 78.41\textcolor{red}{(1.38$\downarrow$)}  \\
        GECScore   & 52.97\textcolor{red}{(3.38$\downarrow$)}  & 9.67\textcolor{red}{(6.75$\downarrow$)}  & 56.27\textcolor{red}{(9.65$\downarrow$)}  & 56.54\textcolor{red}{(3.03$\downarrow$)}  \\
        \midrule
        metric.avg & 67.66\textcolor{red}{(2.55$\downarrow$)}  & 51.58\textcolor{red}{(5.09$\downarrow$)}   & 68.05\textcolor{red}{(5.35$\downarrow$)}  & 68.43\textcolor{red}{(3.19$\downarrow$)}  \\
        \midrule
        avg        & 62.47\textcolor{red}{(2.61$\downarrow$)}  & 48.87\textcolor{red}{(5.22$\downarrow$)}  & 66.32\textcolor{red}{(5.08$\downarrow$)}  & 64.43\textcolor{red}{(2.98$\downarrow$)}  \\
        \bottomrule
    \end{tabular}}
    \caption{MAGA-extra-self-refine Bench.}
    \label{tab:MAGA-extra-self-refine Bench}
\end{table*}

\begin{table*}[!t]
    \centering
    \resizebox{\linewidth}{!}{
    \begin{tabular}{c|c|c|c|c|c}
        \toprule
         & ACC & TPR & TNR & AUC & ACC(@FPR=5\%) \\
        \toprule
        R-B GPT2        & 64.33     & 33.96       & 97.92      & 75.92       & 68.25 \\
        R-L GPT2        & 54.30     & 13.43       & 99.50      & 70.44       & 63.98 \\
        R-B CGPT        & 67.19     & 37.66       & 99.86      & 81.25       & 74.97 \\
        RADAR           & 67.28     & 90.09       & 42.04      & 76.25       & 55.46 \\
        SCRN            & 67.88     & \underline{93.68}       & 39.33      & 79.37       & 58.13 \\
        DETree\textsuperscript{*} & 99.50\textsuperscript{*}     & 99.53\textsuperscript{*}       & 99.47\textsuperscript{*}      & 99.78\textsuperscript{*}       & 99.32\textsuperscript{*} \\
        \midrule
        GLTR            & -     & -       & -      & 80.19       & 72.23 \\
        F-DetectGPT     & 83.43     & 82.39       & 84.57      & 88.51       & 85.99 \\
        LLMDet          & 52.25     & 26.99       & 80.19      & 70.28       & 50.17 \\
        Binoculars      & \textbf{89.76}     & 81.33       & 99.08      & 89.73       & 89.09 \\
        DALD            & 82.58     & 82.98       & 82.13      & 88.15       & 85.01 \\
        GECScore        & 48.06     & 1.11        & \textbf{100.00}      & 89.05       & 68.81 \\
        \midrule
        \textbf{R-B MGB (Ours)}  & 85.49     & 72.39       & \underline{99.99}      & \underline{94.63}       & \underline{90.58} \\
        \textbf{R-B MAGA (Ours)} & \underline{87.36} \textcolor{green}{(1.86$\uparrow$)}     & \textbf{98.72} \textcolor{green}{(26.33$\uparrow$)}       & 74.79 \textcolor{red}{(-25.20$\downarrow$)}      & \textbf{98.87} \textcolor{green}{(4.24$\uparrow$)}       & \textbf{95.41} \textcolor{green}{(4.83$\uparrow$)} \\
        \bottomrule
    \end{tabular}}
    \caption{Performance comparison on S-M4 dataset. DETree\textsuperscript{*} uses multiple datasets including M4, M4GT and MAGE to construct reference vectors, so its results are for reference only (marked with asterisk) and not included in the best/second-best comparison. Bold indicates the best performance, underline indicates the second best among the remaining models.}
    \label{tab:S-M4 Performance}
\end{table*}

\begin{table*}[!t]
    \centering
    \resizebox{\linewidth}{!}{
    \begin{tabular}{c|c|c|c|c|c}
        \toprule
         & ACC & TPR & TNR & AUC & ACC(@FPR=5\%) \\
        \toprule
        R-B GPT2        & 61.04     & 43.81       & 96.80      & 78.38       & 63.76 \\
        R-L GPT2        & 50.60     & 27.48       & \underline{98.60}      & 74.02       & 57.04 \\
        R-B CGPT        & 63.91     & 50.78       & 91.16      & 74.82       & 58.61 \\
        RADAR           & 54.95     & 72.69       & 18.11      & 70.89       & 31.13 \\
        SCRN            & 62.80     & \textbf{91.06}       & 4.14      & \textbf{93.49}       & 33.73 \\
        DETree\textsuperscript{*} & 90.01\textsuperscript{*}     & 94.67\textsuperscript{*}       & 80.33\textsuperscript{*}      & 96.63\textsuperscript{*}       & 94.56\textsuperscript{*} \\
        \midrule
        GLTR            & -     & -       & -      & 78.29       & 63.13 \\
        F-DetectGPT     & \underline{80.96}     & 83.31       & 76.08      & 86.76       & \underline{79.65} \\
        LLMDet          & 55.33     & 47.82       & 70.92      & 74.89       & 53.16 \\
        Binoculars      & \textbf{82.36}     & 75.03       & 97.58      & \underline{89.14}       & \textbf{84.10} \\
        DALD            & 80.44     & \underline{83.93}       & 73.19      & 85.41       & 78.87 \\
        GECScore        & 34.26     & 2.60        & \textbf{100.00}      & 65.88       & 50.00 \\
        \midrule
        \textbf{R-B MGB (Ours)}  & 73.95     & 64.44       & 93.69      & 82.98       & 73.37 \\
        \textbf{R-B MAGA (Ours)} & 80.69 \textcolor{green}{(6.74$\uparrow$)}     & 77.22 \textcolor{green}{(12.78$\uparrow$)}       & 87.88 \textcolor{red}{(5.81$\downarrow$)}      & 88.40 \textcolor{green}{(5.42$\uparrow$)}       & 78.99 \textcolor{green}{(5.62$\uparrow$)} \\
        \bottomrule
    \end{tabular}}
    \caption{Performance comparison on C-M4GT dataset. DETree\textsuperscript{*} uses multiple datasets including M4, M4GT and MAGE to construct reference vectors, so its results are for reference only (marked with asterisk) and not included in the best/second-best comparison. Bold indicates the best performance, underline indicates the second best among the remaining models.}
    \label{tab:C-M4GT Performance}
\end{table*}

\begin{table*}[!t]
    \centering
    \resizebox{\linewidth}{!}{
    \begin{tabular}{c|c|c|c|c|c}
        \toprule
         & ACC & TPR & TNR & AUC & ACC(@FPR=5\%) \\
        \toprule
        R-B GPT2        & \underline{60.99}     & 47.21       & 85.35      & \underline{76.40}       & 55.19 \\
        R-L GPT2        & 56.92     & 35.93       & 94.03      & \textbf{79.55}       & 55.98 \\
        R-B CGPT        & 54.64     & 43.81       & 73.78      & 61.32       & 47.89 \\
        RADAR           & 53.09     & 46.92       & 64.01      & 58.43       & 45.08 \\
        SCRN            & 59.99     & \textbf{57.78}       & 63.90      & 66.24       & 48.01 \\
        DETree\textsuperscript{*} & 84.68\textsuperscript{*}     & 99.95\textsuperscript{*}       & 57.69\textsuperscript{*}      & 99.44\textsuperscript{*}       & 96.82\textsuperscript{*} \\
        \midrule
        GLTR            & -     & -       & -      & 58.34       & 46.78 \\
        F-DetectGPT     & 59.54     & 46.15       & 83.22      & 64.67       & 56.67 \\
        LLMDet          & 40.39     & 29.25       & 60.09      & 50.13       & 47.27 \\
        Binoculars      & \textbf{61.76}     & \underline{47.31}       & 87.30      & 63.79       & \textbf{58.89} \\
        DALD            & 59.82     & 47.23       & 82.07      & 64.96       & \underline{56.73} \\
        GECScore        & 41.10     & 7.79        & \textbf{100.00}      & 53.67       & 40.00 \\
        \midrule
        \textbf{R-B MGB (Ours)}  & 44.98     & 15.44       & \underline{97.22}      & 60.90       & 46.95 \\
        \textbf{R-B MAGA (Ours)} & 48.29 \textcolor{green}{(3.31$\uparrow$)}     & 22.15 \textcolor{green}{(6.71$\uparrow$)}       & 94.51 \textcolor{red}{(-2.71$\downarrow$)}      & 64.63 \textcolor{green}{(3.73$\uparrow$)}       & 47.67 \textcolor{green}{(0.72$\uparrow$)} \\
        \bottomrule
    \end{tabular}}
    \caption{Performance comparison on C-MAGE dataset. DETree\textsuperscript{*} uses multiple datasets including M4, M4GT and MAGE to construct reference vectors, so its results are for reference only (marked with asterisk) and not included in the best/second-best comparison. Bold indicates the best performance, underline indicates the second best among the remaining models.}
    \label{tab:C-MAGE Performance}
\end{table*}

\begin{table*}[!t]
    \centering
    \resizebox{\linewidth}{!}{
    \begin{tabular}{c|c|c|c|c|c}
        \toprule
         & ACC & TPR & TNR & AUC & ACC(@FPR=5\%) \\
        \toprule
        R-B GPT2        & \textbf{94.50}     & \underline{93.94}       & 94.77      & 98.45       & \underline{94.59} \\
        R-L GPT2        & 89.39     & 69.32       & 98.81      & 94.88       & 89.34 \\
        R-B CGPT\textsuperscript{*}        & 99.27\textsuperscript{*}     & 99.37\textsuperscript{*}       & 99.23\textsuperscript{*}      & 99.99\textsuperscript{*}       & 99.27\textsuperscript{*} \\
        RADAR           & 87.79     & 84.71       & 89.23      & 90.16       & 84.74 \\
        SCRN            & 86.43     & 89.77       & 84.86      & 93.49       & 88.77 \\
        DETree\textsuperscript{*} & 44.76\textsuperscript{*}     & 100.00\textsuperscript{*}       & 18.82\textsuperscript{*}      & 98.55\textsuperscript{*}       & 95.27\textsuperscript{*} \\
        \midrule
        GLTR            & -     & -       & -      & 92.37       & 91.13 \\
        F-DetectGPT     & 93.56     & \textbf{100.00}       & 90.53      & \textbf{99.94}       & \textbf{97.19} \\
        LLMDet          & 78.17     & 71.48       & 81.31      & 90.23       & 81.34 \\
        Binoculars      & 87.92     & \textbf{100.00}       & 82.25      & 99.59       & \textbf{97.19} \\
        DALD            & 91.93     & \textbf{100.00}       & 88.14      & \underline{99.76}       & \textbf{97.19} \\
        GECScore        & 81.74     & 42.86       & \textbf{100.00}      & 90.18       & 86.96 \\
        \midrule
        \textbf{R-B MGB (Ours)}  & 87.09     & 60.23       & \underline{99.70}      & 93.14       & 91.68 \\
        \textbf{R-B MAGA (Ours)} & \underline{94.47} \textcolor{green}{(7.38$\uparrow$)}     & 87.12 \textcolor{green}{(26.89$\uparrow$)}       & 97.92 \textcolor{red}{(-1.78$\downarrow$)}      & 98.14 \textcolor{green}{(5.00$\uparrow$)}       & 94.06 \textcolor{green}{(2.38$\uparrow$)} \\
        \bottomrule
    \end{tabular}}
    \caption{Performance comparison on C-HC3 dataset. DETree\textsuperscript{*} and R-B CGPT\textsuperscript{*} are marked with asterisk and not included in the best/second-best comparison. R-B CGPT\textsuperscript{*} is trained on HC3. Bold indicates the best performance, underline indicates the second best among the remaining models.}
    \label{tab:C-HC3 Performance}
\end{table*}

\begin{table*}[!t]
    \centering
    \tiny
    \resizebox{\linewidth}{!}{
    \begin{tabular}{l|c|c|c|c}
        \toprule
        & ACC & TPR & AUC & ACC(@FPR=5\%) \\
        \midrule
        R-B CGPT CN & 52.00\textcolor{red}{(0.19$\downarrow$)} & 99.17\textcolor{red}{(0.37$\downarrow$)} & 71.97\textcolor{red}{(1.04$\downarrow$)} & 53.33\textcolor{red}{(0.31$\downarrow$)} \\
        R-B MPU zhv3 & 53.13\textcolor{red}{(0.25$\downarrow$)} & 6.57\textcolor{red}{(0.49$\downarrow$)} & 78.38\textcolor{red}{(1.51$\downarrow$)} & 56.64\textcolor{red}{(0.49$\downarrow$)} \\
        \midrule
        neutral.avg & 52.57\textcolor{red}{(0.22$\downarrow$)} & 52.87\textcolor{red}{(0.43$\downarrow$)} & 75.18\textcolor{red}{(1.28$\downarrow$)} & 54.99\textcolor{red}{(0.40$\downarrow$)} \\
        \midrule
        Binoculars & 69.65\textcolor{red}{(2.19$\downarrow$)} & 44.29\textcolor{red}{(4.38$\downarrow$)} & 82.46\textcolor{red}{(0.92$\downarrow$)} & 68.87\textcolor{red}{(2.34$\downarrow$)} \\
        GECScore & 49.59\textcolor{red}{(0.00$\downarrow$)} & 0.01\textcolor{red}{(0.00$\downarrow$)} & 50.04\textcolor{red}{(1.29$\downarrow$)} & 51.62\textcolor{red}{(0.46$\downarrow$)} \\
        \midrule
        metric.avg & 59.62\textcolor{red}{(1.10$\downarrow$)} & 22.15\textcolor{red}{(2.19$\downarrow$)} & 66.25\textcolor{red}{(1.11$\downarrow$)} & 60.25\textcolor{red}{(1.40$\downarrow$)} \\
        \midrule
        avg & 56.09\textcolor{red}{(0.66$\downarrow$)} & 37.51\textcolor{red}{(1.31$\downarrow$)} & 70.71\textcolor{red}{(1.19$\downarrow$)} & 57.62\textcolor{red}{(0.90$\downarrow$)} \\
        \bottomrule
    \end{tabular}}
    \caption{MAGA-cn-extra-GDARL-CDMMR Bench.}
    \label{tab:MAGA-cn-extra-GDARL-CDMMR Bench}
\end{table*}

\begin{table*}[!t]
    \centering
    \tiny
    \resizebox{\linewidth}{!}{
    \begin{tabular}{l|c|c|c|c}
        \toprule
        & ACC & TPR & AUC & ACC(@FPR=5\%) \\
        \midrule
        R-B CGPT CN & 51.90\textcolor{red}{(0.29$\downarrow$)} & 98.97\textcolor{red}{(0.57$\downarrow$)} & 71.99\textcolor{red}{(1.02$\downarrow$)} & 53.27\textcolor{red}{(0.37$\downarrow$)} \\
        R-B MPU zhv3 & 53.40\textcolor{green}{(0.02$\uparrow$)} & 7.11\textcolor{green}{(0.05$\uparrow$)} & 80.13\textcolor{green}{(0.24$\uparrow$)} & 57.06\textcolor{red}{(0.07$\downarrow$)} \\
        \midrule
        neutral.avg & 52.65\textcolor{red}{(0.13$\downarrow$)} & 53.04\textcolor{red}{(0.26$\downarrow$)} & 76.06\textcolor{red}{(0.39$\downarrow$)} & 55.17\textcolor{red}{(0.22$\downarrow$)} \\
        \midrule
        Binoculars & 69.29\textcolor{red}{(2.55$\downarrow$)} & 43.58\textcolor{red}{(5.09$\downarrow$)} & 82.43\textcolor{red}{(0.95$\downarrow$)} & 68.19\textcolor{red}{(3.02$\downarrow$)} \\
        GECScore & 49.59\textcolor{red}{(0.00$\downarrow$)} & 0.01\textcolor{red}{(0.00$\downarrow$)} & 49.96\textcolor{red}{(1.37$\downarrow$)} & 51.56\textcolor{red}{(0.52$\downarrow$)} \\
        \midrule
        metric.avg & 59.44\textcolor{red}{(1.27$\downarrow$)} & 21.80\textcolor{red}{(2.55$\downarrow$)} & 66.20\textcolor{red}{(1.16$\downarrow$)} & 59.88\textcolor{red}{(1.77$\downarrow$)} \\
        \midrule
        avg & 56.05\textcolor{red}{(0.70$\downarrow$)} & 37.42\textcolor{red}{(1.40$\downarrow$)} & 71.13\textcolor{red}{(0.78$\downarrow$)} & 57.52\textcolor{red}{(0.99$\downarrow$)} \\
        \bottomrule
    \end{tabular}}
    \caption{MAGA-cn-extra-roleplaying Bench.}
    \label{tab:MAGA-cn-extra-roleplaying Bench}
\end{table*}

\begin{table*}[!t]
    \centering
    \tiny
    \resizebox{\linewidth}{!}{
    \begin{tabular}{l|c|c|c|c}
        \toprule
        & ACC & TPR & AUC & ACC(@FPR=5\%) \\
        \midrule
        R-B CGPT CN & 52.20\textcolor{green}{(0.01$\uparrow$)} & 99.56\textcolor{green}{(0.02$\uparrow$)} & 72.34\textcolor{red}{(0.67$\downarrow$)} & 53.48\textcolor{red}{(0.16$\downarrow$)} \\
        R-B MPU zhv3 & 53.12\textcolor{red}{(0.26$\downarrow$)} & 6.54\textcolor{red}{(0.52$\downarrow$)} & 78.23\textcolor{red}{(1.66$\downarrow$)} & 56.67\textcolor{red}{(0.46$\downarrow$)} \\
        \midrule
        neutral.avg & 52.66\textcolor{red}{(0.13$\downarrow$)} & 53.05\textcolor{red}{(0.25$\downarrow$)} & 75.29\textcolor{red}{(1.17$\downarrow$)} & 55.08\textcolor{red}{(0.31$\downarrow$)} \\
        \midrule
        Binoculars & 69.98\textcolor{red}{(1.86$\downarrow$)} & 44.96\textcolor{red}{(3.71$\downarrow$)} & 82.73\textcolor{red}{(0.65$\downarrow$)} & 69.24\textcolor{red}{(1.97$\downarrow$)} \\
        GECScore & 49.59\textcolor{red}{(0.00$\downarrow$)} & 0.01\textcolor{red}{(0.00$\downarrow$)} & 50.36\textcolor{red}{(0.97$\downarrow$)} & 51.79\textcolor{red}{(0.29$\downarrow$)} \\
        \midrule
        metric.avg & 59.79\textcolor{red}{(0.93$\downarrow$)} & 22.49\textcolor{red}{(1.86$\downarrow$)} & 66.55\textcolor{red}{(0.81$\downarrow$)} & 60.52\textcolor{red}{(1.13$\downarrow$)} \\
        \midrule
        avg & 56.22\textcolor{red}{(0.53$\downarrow$)} & 37.77\textcolor{red}{(1.05$\downarrow$)} & 70.92\textcolor{red}{(0.99$\downarrow$)} & 57.80\textcolor{red}{(0.72$\downarrow$)} \\
        \bottomrule
    \end{tabular}}
    \caption{MAGA-cn-extra-BPO Bench.}
    \label{tab:MAGA-cn-extra-BPO Bench}
\end{table*}

\begin{table*}[!t]
    \centering
    \tiny
    \resizebox{\linewidth}{!}{
    \begin{tabular}{l|c|c|c|c}
        \toprule
        & ACC & TPR & AUC & ACC(@FPR=5\%) \\
        \midrule
        R-B CGPT CN & 51.93\textcolor{red}{(0.26$\downarrow$)} & 99.03\textcolor{red}{(0.51$\downarrow$)} & 71.62\textcolor{red}{(1.39$\downarrow$)} & 53.21\textcolor{red}{(0.43$\downarrow$)} \\
        R-B MPU zhv3 & 53.04\textcolor{red}{(0.34$\downarrow$)} & 6.39\textcolor{red}{(0.67$\downarrow$)} & 77.95\textcolor{red}{(1.94$\downarrow$)} & 56.44\textcolor{red}{(0.69$\downarrow$)} \\
        \midrule
        neutral.avg & 52.49\textcolor{red}{(0.30$\downarrow$)} & 52.71\textcolor{red}{(0.59$\downarrow$)} & 74.79\textcolor{red}{(1.67$\downarrow$)} & 54.83\textcolor{red}{(0.56$\downarrow$)} \\
        \midrule
        Binoculars & 68.83\textcolor{red}{(3.01$\downarrow$)} & 42.65\textcolor{red}{(6.02$\downarrow$)} & 82.25\textcolor{red}{(1.13$\downarrow$)} & 68.42\textcolor{red}{(2.79$\downarrow$)} \\
        GECScore & 49.59\textcolor{red}{(0.00$\downarrow$)} & 0.01\textcolor{red}{(0.00$\downarrow$)} & 49.69\textcolor{red}{(1.64$\downarrow$)} & 51.61\textcolor{red}{(0.47$\downarrow$)} \\
        \midrule
        metric.avg & 59.21\textcolor{red}{(1.51$\downarrow$)} & 21.33\textcolor{red}{(3.01$\downarrow$)} & 65.97\textcolor{red}{(1.39$\downarrow$)} & 60.02\textcolor{red}{(1.63$\downarrow$)} \\
        \midrule
        avg & 55.85\textcolor{red}{(0.90$\downarrow$)} & 37.02\textcolor{red}{(1.80$\downarrow$)} & 70.38\textcolor{red}{(1.53$\downarrow$)} & 57.42\textcolor{red}{(1.10$\downarrow$)} \\
        \bottomrule
    \end{tabular}}
    \caption{MAGA-cn-extra-self-refine Bench.}
    \label{tab:MAGA-cn-extra-self-refine Bench}
\end{table*}

\begin{table*}[!t]
    \centering
    \resizebox{\linewidth}{!}{
    \begin{tabular}{c|c|c|c|c|c}
        \toprule
         & ACC & TPR & TNR & AUC & ACC(@FPR=5\%) \\
        \toprule
        R-B CGPT CN & \underline{95.22} & \underline{92.30} & 98.10 & \underline{99.24} & \underline{95.19} \\
        R-B MPU zhv3 & 89.18 & 83.37 & 94.92 & 95.94 & 89.10 \\
        \midrule
        Binoculars & 93.51 & 88.80 & 98.17 & 97.87 & 94.18 \\
        GECScore & 50.28 & 0.00 & \textbf{100.00} & 45.34 & 51.26 \\
        \midrule
        \textbf{R-B MGB CN (Ours)} & 86.70 & 73.63 & \underline{99.63} & 98.68 & 93.77 \\
        \textbf{R-B MAGA CN (Ours)} & \textbf{96.14} \textcolor{green}{(9.44$\uparrow$)} & \textbf{94.39} \textcolor{green}{(20.76$\uparrow$)} & 97.88 \textcolor{red}{(-1.75$\downarrow$)} & \textbf{99.41} \textcolor{green}{(0.73$\uparrow$)} & \textbf{96.33} \textcolor{green}{(2.56$\uparrow$)} \\
        \bottomrule
    \end{tabular}}
    \caption{Performance comparison on S-M4 CN dataset. Bold indicates the best performance, underline indicates the second best.}
    \label{tab:S-M4-CN Performance}
\end{table*}

\begin{table*}[!t]
    \centering
    \resizebox{\linewidth}{!}{
    \begin{tabular}{c|c|c|c|c|c}
        \toprule
         & ACC & TPR & TNR & AUC & ACC(@FPR=5\%) \\
        \toprule
        R-B CGPT CN & \underline{94.58} & 91.86 & 97.61 & \underline{98.95} & \underline{94.29} \\
        R-B MPU zhv3 & 88.08 & 82.85 & 93.92 & 95.40 & 87.09 \\
        \midrule
        Binoculars & 93.71 & \underline{93.58} & 93.85 & 97.73 & 94.21 \\
        GECScore & 47.24 & 0.00 & \textbf{100.00} & 54.69 & 47.06 \\
        \midrule
        \textbf{R-B MGB CN (Ours)} & 85.09 & 72.57 & \underline{99.08} & 98.04 & 91.42 \\
        \textbf{R-B MAGA CN (Ours)} & \textbf{96.27} \textcolor{green}{(11.17$\uparrow$)} & \textbf{95.00} \textcolor{green}{(22.43$\uparrow$)} & 97.68 \textcolor{red}{(-1.40$\downarrow$)} & \textbf{99.42} \textcolor{green}{(1.38$\uparrow$)} & \textbf{96.23} \textcolor{green}{(4.81$\uparrow$)} \\
        \bottomrule
    \end{tabular}}
    \caption{Performance comparison on C-M4GT CN dataset. Bold indicates the best performance, underline indicates the second best.}
    \label{tab:C-M4GT-CN Performance}
\end{table*}

\begin{table*}[!t]
    \centering
    \resizebox{\linewidth}{!}{
    \begin{tabular}{c|c|c|c|c|c}
        \toprule
         & ACC & TPR & TNR & AUC & ACC(@FPR=5\%) \\
        \toprule
        R-B CGPT CN\textsuperscript{*} & 98.63\textsuperscript{*} & 99.70\textsuperscript{*} & 97.81\textsuperscript{*} & 99.96\textsuperscript{*} & 97.15\textsuperscript{*} \\
        R-B MPU zhv3 & 88.67 & 90.28 & 87.43 & 95.09 & 88.65 \\
        \midrule
        Binoculars & 93.23 & 87.87 & \underline{97.37} & 97.02 & 92.21 \\
        GECScore & 56.46 & 0.00 & \textbf{100.00} & 47.94 & 56.64 \\
        \midrule
        \textbf{R-B MGB CN (Ours)} & \underline{95.01} & \underline{94.22} & 95.62 & \underline{99.04} & \underline{94.90} \\
        \textbf{R-B MAGA CN (Ours)} & \textbf{95.98} \textcolor{green}{(0.97$\uparrow$)} & \textbf{99.25} \textcolor{green}{(5.03$\uparrow$)} & 93.45 \textcolor{red}{(-2.17$\downarrow$)} & \textbf{99.67} \textcolor{green}{(0.63$\uparrow$)} & \textbf{96.64} \textcolor{green}{(1.74$\uparrow$)} \\
        \bottomrule
    \end{tabular}}
    \caption{Performance comparison on C-HC3 CN dataset. Bold indicates the best performance, underline indicates the second best. R-B CGPT CN is trained on HC3 CN dataset and not involved in comparison.}
    \label{tab:C-HC3-CN Performance}
\end{table*}

\begin{table*}[!t]
\centering
\small
\begin{tabular}{l|p{0.74\linewidth}}
\toprule
\textbf{Domain} & wikiHow \\
\midrule
\textbf{Title} & How to Be Healthy \\
\midrule
\textbf{Model} & GPT-4o-mini \\
\midrule
\textbf{Original User Prompt} & Write the body of a wikiHow article titled "How to Be Healthy". \\
\midrule
\textbf{System Prompt} & You are a student (college-aged). Respond in a relatable, slightly frazzled but earnest tone—like you're venting to a classmate before a lecture, juggling notes and deadlines, casual and conversational. \\
\midrule
\textbf{User Prompt} & Write the body of a wikiHow article titled "How to Be Healthy". Include actionable, beginner-friendly tips across nutrition, exercise, sleep, and mental health. \\
\midrule
\textbf{HWT} & Many people think that being healthy is a difficult task, involving months of dieting accompanied by hours at the gym. However, this is not actually true. \\
\midrule
\textbf{MGB-M} & To be healthy, \sethlcolor{red!30}\hl{consistently perform behaviors} that support \sethlcolor{red!30}\hl{physical, mental, and social function}. This includes \sethlcolor{red!30}\hl{consuming adequate nutrition, engaging in regular movement, maintaining rest cycles, managing stress inputs, and avoiding known hazards}. \sethlcolor{red!30}\hl{Execute these actions repeatedly over time}. \\
\midrule
\textbf{extra-BPO-M} & Being healthy involves \sethlcolor{red!30}\hl{adopting habits} that \sethlcolor{green!30}\hl{support your body and mind}. To improve your health, focus on \sethlcolor{red!30}\hl{maintaining a balanced diet, staying physically active, getting adequate sleep, and managing stress}. Small, consistent actions can lead to long-term health benefits. Follow the tips below to \sethlcolor{green!30}\hl{begin your journey toward a healthier lifestyle}. \\
\midrule
\textbf{extra-roleplaying-M} & \sethlcolor{green!30}\hl{Being healthy is about balancing various aspects of your life}—\sethlcolor{red!30}\hl{nutrition, exercise, sleep, and mental health}. \sethlcolor{green!30}\hl{It's not about perfection, but consistency}. Simple changes can make a huge difference. \sethlcolor{red!30}\hl{Focus on eating real food, moving your body, resting well, and managing stress}. Over time, small habits add up to big improvements. \\
\midrule
\textbf{extra-self-refine-M} & To be healthy, \sethlcolor{red!30}\hl{focus on eating well, staying active, getting enough rest, and managing stress}. \sethlcolor{green!30}\hl{These habits can help you feel better, live longer, and enjoy life more fully}. \sethlcolor{green!30}\hl{You don\mbox{'}t have to make huge changes all at once\mbox{-}small, consistent steps can lead to big improvements over time}. \sethlcolor{red!30}\hl{This guide will walk you through practical, everyday actions to boost your health and well-being}. \\
\midrule
\textbf{MAGA-M} & \sethlcolor{green!30}\hl{Being healthy isn\mbox{'}t about chasing perfection\mbox{-}it\mbox{'}s about taking small, manageable steps} that \sethlcolor{red!30}\hl{make you feel better, help you live longer, and let you enjoy life more}. \sethlcolor{green!30}\hl{You don\mbox{'}t need to completely overhaul your lifestyle overnight}. Instead, focus on making tiny, sustainable changes in the areas of nutrition, exercise, sleep, and mental health. \sethlcolor{green!30}\hl{These simple habits can have a huge impact over time, and you\mbox{'}ll start feeling the benefits without all the pressure}. \sethlcolor{green!30}\hl{Ready to get started? Let\mbox{'}s go!} \\
\bottomrule
\end{tabular}
\caption{A case study comparing different alignment methods was conducted. Human analysis highlighted text with machine-generated characteristics in red and text with human-generated characteristics in green.}
\label{tab:wikihow_how_to_be_healthy}
\end{table*}

\begin{table*}[!t]
	\centering
	\small
	\begin{tabular}{l|p{0.74\linewidth}}
		\toprule
		\textbf{Domain} & Reddit \\
		\midrule
		\textbf{Title} & The one feature the iPad is really missing. \\
		\midrule
		\textbf{Model} & Qwen3-8B \\
		\midrule
		\textbf{Original User Prompt} & Write just the body of a Reddit post titled "The one feature the iPad is really missing.". Do not repeat the title. \\
		\midrule
		\textbf{System Prompt} & You are American. Please respond in a casual, conversational American English style. \\
		\midrule
		\textbf{User Prompt} & Write just the body of a Reddit post titled "The one feature the iPad is really missing". Focus on relatable daily pain points and a brief personal scenario for authenticity. Do not repeat the title. \\
		\midrule
		\textbf{HWT} & I don't care about the lack of camera. I never use the one on my MacBook, and even if I did the angle would be terrible on the iPad. I don't care if third party apps can't run in the background. I don't listen to streaming music. I don't care that the App Store is a closed system. I can jailbreak for myself and I think the closed system works better for most users. The one feature I want is User Accounts and a Guest Account. If this device is meant to be a coffee table computer, it needs to be able to accomadate multiple users. \\
		\midrule
		\textbf{MGB-M} & \sethlcolor{orange!30}\hl{Okay, here's the body of a Reddit post:} I've been using the iPad on and off for years, and every time I want it to fully replace my laptop, I run into the same wall. \sethlcolor{red!30}\hl{The hardware is more than capable, the apps are mostly there, but one core feature is still missing: true, unrestricted background processing.} I don't mean minor background refresh or limited exports that only work if you don't lock the screen. I mean being able to kick off a long task\sethlcolor{red!30}\hl{---file transfers, renders, large downloads, code builds, data processing---and trust that it will just keep running, exactly like it would on a Mac or PC}. \sethlcolor{red!30}\hl{This limitation affects everything: File managers feel half-baked because transfers can stall or fail.} Pro apps have to design around the OS instead of focusing on features. Multitasking feels cosmetic when apps can't actually work unless they're front and center. Apple keeps pushing the iPad as a "computer," especially with M-series chips, but computers do work independently of whether you're actively staring at them. Until iPadOS allows apps to reliably run real tasks in the background, \sethlcolor{green!30}\hl{the iPad will always feel like it's one permission dialog away from greatness}. The frustrating part is that this doesn't feel like a hardware problem or even a performance issue—it feels like a philosophical choice. And it's the one thing that keeps the iPad from crossing the line from "very powerful tablet" to "actual laptop replacement." \\
		\midrule
		\textbf{extra-GDARL-CDMMR-M} & Everyone debates macOS on iPad or better window management, but I think the single most important missing feature is proper multi user accounts. The iPad is sold as both a personal computer replacement and a shared household device, yet it still assumes exactly one person owns it forever. \sethlcolor{red!30}\hl{One Apple ID, one iCloud, one Messages history, one Photos library.} \sethlcolor{green!30}\hl{The moment you hand it to someone else, you are either oversharing or constantly micromanaging apps and data.} This is strange given how powerful iPads are now. The hardware can easily handle separate user spaces with \sethlcolor{red!30}\hl{fast switching, isolated data, and different Screen Time rules}. Apple already does this on macOS and even in schools, so the idea clearly works. \sethlcolor{red!30}\hl{A shared iPad should let each person tap their profile and instantly see their own Home Screen, files, and browsing}. Kids could use the same device without touching adult data, and families would not need multiple expensive iPads just to keep things separate. Right now the iPad sits in an awkward middle ground. Adding real multi user support would not just be another feature, it would finally let the iPad fit naturally into everyday life. \\
		\midrule
		\textbf{extra-BPO-M} & \sethlcolor{orange!30}\hl{Sure, Here is your Reddit post:} I love my iPad, but there's one moment every day where it reminds me what it isn't. \sethlcolor{green!30}\hl{I'll be on the couch with it, trying to juggle a couple of apps}—notes on one side, a browser or PDF on the other—and everything just feels one step more awkward than it needs to be. Apps snapping instead of freely resizing, things randomly going full screen, windows refusing to stay where I want them. I'm not trying to do "pro work," I just want to see more than one thing at once without fighting the UI. The pain really hits when I'm doing something simple, like planning a trip. \sethlcolor{red!30}\hl{I'll have a map open, a booking page, and my notes, and suddenly I'm swiping, resizing, reopening apps I just had visible.} On my laptop, I'd just drag windows around and be done. On the iPad, I spend more time managing the interface than actually reading or thinking. What bugs me most is that the hardware is clearly capable. \sethlcolor{red!30}\hl{Big screen, tons of power, great touch input.} It just needs truly free, windowed multitasking that behaves predictably, like a computer when you want it to. Not all the time—just when you need it. Until then, the iPad keeps hovering in this frustrating middle ground: almost perfect for daily use, but not quite comfortable when you try to do slightly more than one thing. \\
		\bottomrule
	\end{tabular}
\end{table*}

\begin{table*}[!t]
	\centering
	\small
	\begin{tabular}{l|p{0.74\linewidth}}
		\toprule
		\textbf{extra-roleplaying-M} & \sethlcolor{orange!30}\hl{Okay, here's a draft body for that Reddit post:} I love the iPad. I really do. It's fast, the hardware is great, the apps are solid, and it's honestly replaced my laptop for a lot of things. But there's one thing that still feels weirdly missing in 2026: real multi-user support. Not "hand it to your kid and hope they don't mess up your stuff" support. I mean actual user profiles, like on a Mac. \sethlcolor{red!30}\hl{Separate logins, separate apps, separate files, separate iClouds.} The basics. The iPad is marketed as this do-everything device — \sethlcolor{red!30}\hl{family device, school device, shared device} — but the second more than one person uses it, it turns into a mess. Your photos mixed with theirs, messages popping up, Safari tabs everywhere, apps you didn't install cluttering the home screen. It's chaos. Apple already does this on Macs. Even Apple TV has profiles now. iPads are powerful enough, expensive enough, and "computer-like" enough that this shouldn't be a stretch. Especially for families, classrooms, and households that don't want to buy multiple iPads just to keep things separate. Every time this comes up, people say "well, it's a personal device." Okay — so is a MacBook. And that hasn't stopped multi-user accounts for decades. If Apple added proper user switching tomorrow, it would instantly make the iPad way more practical for a ton of people. Until then, it just feels like an obvious missing piece that they're stubbornly ignoring. \sethlcolor{green!30}\hl{Anyway, that's my rant. Curious if this bugs anyone else or if there's another "one missing feature" that drives you nuts.} \\
		\midrule
		\textbf{extra-self-refine-M} & \sethlcolor{orange!30}\hl{I'll rewrite it to sound more casual, personal, and Reddit-like.} I tried to actually work on my iPad today. Like, laptop-replacement work. And once again, I ran into the same wall. All I want is normal window multitasking. Not "kind of" multitasking. Not "Apple's interpretation" of multitasking. Just… let me open a few apps in windows and move them around without fighting the OS. Stage Manager sounded promising, but using it still feels awkward. \sethlcolor{red!30}\hl{Windows resize in weird ways, some apps refuse to be the size I want,} and after a while it feels like the system is quietly telling me, "Okay, that's enough apps for you." On an M-series iPad, that's honestly kind of ridiculous. The hardware is clearly not the problem. The apps aren't either. What's missing is the freedom to treat the screen like a screen, not a carefully supervised playground. \sethlcolor{red!30}\hl{Sometimes I want my main app front and center, a browser half-open, Notes off to the side, and Messages floating somewhere.} Why is that still hard? The iPad is so close to being my go-to work device. But until multitasking feels as easy and flexible as it does on a Mac, it still feels like I'm using a powerful computer with one hand tied behind its back. \sethlcolor{green!30}\hl{Am I the only one who keeps bouncing off this?} \\
		\midrule
		\textbf{MAGA-M} & I use my iPad every day — \sethlcolor{red!30}\hl{couch browsing, note-taking, quick emails, even some "light" work} — and there's one thing that keeps breaking the flow: a real, system-wide multi-user or profile mode. \sethlcolor{green!30}\hl{My partner will grab my iPad to look something up or watch YouTube and suddenly my Safari tabs are rearranged}, my Notes app has random doodles, and my YouTube recommendations are permanently cursed with toddler cartoons or reality TV. I can't blame them — it's a shared household device — but the iPad still acts like it belongs to exactly one human being who never lets anyone else touch it. I know there are workarounds \sethlcolor{red!30}\hl{(Guided Access, switching apps, signing out of stuff)}, but they're clunky and feel like hacks. On a Mac, we'd just have separate user accounts. \sethlcolor{green!30}\hl{On an iPad, the device that's most likely to be shared on a couch, that option just\dots doesn't exist.} Every time this happens I'm reminded that the iPad is marketed as a computer replacement, yet it can't handle a super basic, real-life scenario: more than one person living in the same space. It's such a small thing, but it would instantly make the iPad feel more grown-up and way more practical. \\
		\bottomrule
	\end{tabular}
	\caption{A case study demonstrating a problem that other methods struggle with (highlighted in orange), but which only GDARL can effectively solve. Human-like traces are highlighted in green.}
	\label{tab:reddit_ipad_missing_feature}
\end{table*}

\begin{table*}[!t]
	\centering
	\small
	\begin{tabular}{l|p{0.74\linewidth}}
		\toprule
		\textbf{Domain} & Zhihu \\
		\midrule
		\textbf{Title} & 有什么好吃到经常回购的零食？ \\
		\midrule
		\textbf{Model} & Qwen3-plus \\
		\midrule
		\textbf{Original User Prompt} & 请你为知乎提问“有什么好吃到经常回购的零食？”写一段回答。请不要重复问题。 \\
		\midrule
		\textbf{System Prompt} & 你是老北京人儿。请你用地道的北京腔回复。 \\
		\midrule
		\textbf{User Prompt} & 请你为知乎提问 “有什么好吃到经常回购的零食？” 写一段回答，结合个人回购体验和具体食用场景，更显真实。请不要重复问题。 \\
		\midrule
		\textbf{HWT} & 谢邀，刚下工地，搬完水泥。 Ladies and gentlemen,首先，女士们请退场，接下来的这款零食推荐男士福音，哈哈哈。 大学会有很多第一次，第一次出省，第一次坐飞机，第一次谈恋爱，第一次住宿舍等等。然而，当两个第一次加在一起就会产生故事，我的故事是这样的，那会我们是六人寝，也是我第一次住宿舍，我们六个男生每天的话题无非政治，军事，游戏，女孩儿。其中女孩话题占比99\%，咳咳，那本人呢，不才，凭借着独特的自身魅力(其实是面皮厚，放的开)与一位女孩子开始陷入爱河。 第一次恋爱+第一次住宿＝你有了一个恋爱智囊团!室友们纷纷建言献策，教我如何约会，如何表达爱意等(后来，我才知道这帮家伙毫无实战经验，拿我练小号呢) 当然，我也不是什么都听，我有自己的独立思考(奇葩的思考)。我听从一哥们儿的，说要想征服女孩的放心，就要懂得浪漫，要浪漫就要给她一个印象深刻的吻！ 我一琢磨，这逻辑没毛病呀，环环相扣，于是赶快请教怎么接吻才会印象深刻呢？然后这个哥们拿出了一个星球杯，说，1分钟之内能够用舌头吃完吗？ 那我能服软吗？开始我的舌功初次尝试，讲真的，没几下，我的舌头都开始酸了。(能不酸吗，平时说话舌头都捋不直的我\dots) 那哥们说你这不太行，接吻是需要力量的，你要做力量训练。于是他免费赞助了我一桶星球杯，为了让我不偷懒，把勺都扔掉，好人呐！真热心市民呀！我当时真的感动，觉的还是好人多呀！ 于是，那个秋天，同学们都会看到我有事没事都在不用勺舔星球杯，我励志舔完一桶，然后功成出山，这期间，我进入癫狂转态，人女孩约我看电影，我竟然回复没时间，其实就是在宿舍舔星球杯(欲练此功必先\dots) 后来，一桶星球杯我记得还剩7个的时候，女孩和我提了分手，原因是感觉我不爱她？？！ 我可是为了她马上就要舔完一桶星球杯的男人呀！我承认，确实有点疏远了她，一个半月就牵了一次手。但那是因为我在练习呀！一切都是为了她呀，她一点也不懂我。 不过今天我还是强烈推荐星球杯零食的原因是，后来，我发现，功夫的用法可以用在接吻，但不仅限于接吻，我想，聪明的男孩儿为师一点就悟了吧。 对，你没猜错，就是你所想的(这也是我的意外收获。感谢我哥们儿当初的点拨)。 懂事的男生已经下单了，哈哈哈，对于你们，我只有一个要求，拿到手之后，第一件事，把勺全部丢掉，Ok？答应我，日后你会感谢我的。 \\
		\midrule
		\textbf{MGB-M} & \sethlcolor{orange!30}\hl{\mbox{当然可以！以下是我为知乎问题“有什么好吃到经常回购的零食？”撰写的回答：}} \sethlcolor{red!30}\hl{\mbox{随着零食市场的不断扩展，各种新奇又美味的小吃层出不穷，但能够真正做到让人一}} \sethlcolor{red!30}\hl{\mbox{次次回购的，往往是那些经过时间验证、口味独特且高性价比的零食。根据大数据分}} \sethlcolor{red!30}\hl{\mbox{析，以下几款零食凭借其极高的回购率和持续热度，成为了消费者的“必备清单”：}} 百草味/三只松鼠系列坚果 这些品牌的坚果系列因其独特的调味配方，精致的包装和适中的价格，在市场中占据了不小的份额。无论是酸甜口味的干果，还是香脆的坚果拼盘，口感和营养的双重加持，让它们成为了日常零食中的常客。 黄飞鸿麻辣花生 这款零食的受欢迎程度可谓是超乎想象，麻辣味浓郁，口感酥脆，回味无穷。特别适合爱吃辣的朋友，每一次剥开包装，都有一种打开新世界的感觉。 乐事（Lay's）薯片系列 不管是经典的原味还是创新的其他口味，乐事薯片凭借其稳定的口味和全球范围的普及度，一直是回购率极高的零食之一。特别是在追剧或者聚会时，它是那种“手一伸就能拿到”的最佳伴侣。 阿尔卑斯糖 这款糖果的口感轻柔，甜度适中，味道清新，不仅大人小孩都能接受，而且无论什么时候吃，都会给人一种放松和满足的感觉。其多样的口味和清新的包装也是消费者青睐的原因之一。 脆皮小麻花 这款零食的脆度和香气是其他同类产品难以匹敌的，经过精心炸制，外皮酥脆，内里松软。无论是搭配茶水，还是单独当作小吃，都是很高的回购率。 \sethlcolor{red!30}\hl{\mbox{这些零食背后不仅仅是传统的味道体验，它们通过对消费者口味需求的精准把握，以}} \sethlcolor{red!30}\hl{\mbox{及持续的创新与改良，逐渐获得了长期的市场认可。每一口都带给你不同的感官享受}} \sethlcolor{red!30}\hl{\mbox{，这也是它们能够成为“回购神物”的根本原因。}} \\
		\bottomrule
	\end{tabular}
\end{table*}

\begin{table*}[!t]
	\centering
	\small
	\begin{tabular}{l|p{0.74\linewidth}}
		\toprule
		\textbf{extra-BPO-M} & 在我个人的回购清单中，几款零食几乎已经成为了必备。特别是在工作日的午后或者深夜刷剧时，它们总是能够恰到好处地填补肚子，又不会让我觉得吃得过多。 首先要提的是 瑞士莲的巧克力。这款巧克力几乎没有让我失望过，每次吃都能享享受到那种入口即化的柔滑感。特别是它的牛奶巧克力和榛子夹心口味，在工作日的疲惫时刻，简直就是一种精神慰藉。每次加班到夜深，拿出一小块放进嘴里，感觉心情一下子就被治愈了。 还有 三只松鼠的坚果混合包。这种零食绝对是我常备的零食之一，尤其是在外出旅行或者在家中闲暇时。它混合了各种口感的坚果，既有咸的也有甜的，满足了我的不同口味需求。而且，包装设计很适合随时随地取用，不用担心散落或不便携带。每次手里拿着它，随便吃几颗都能让我在疲劳中感到轻松。 至于 好丽友的泡泡糖，这已经成了我和朋友们聚会的必备零食。在闲聊的时光中，大家都喜欢拿出几颗嚼着，泡泡糖的那种甜蜜味和嚼劲特别能让人放松。它的“嚼劲”让我回味无穷，而且味道也很清新，几乎没有过多的化学味，感觉就像是一种简单而纯粹的幸福。 \sethlcolor{red!30}\hl{\mbox{总的来说，这些零食不仅好吃，而且在特定场合下能带给我一种特殊的满足感。工作}} \sethlcolor{red!30}\hl{\mbox{、放松、聚会，零食的角色各不相同，}}却总能在我生活的每个小角落里找到它们的身影。 \\
		\midrule
		\textbf{extra-roleplaying-M} & 哎呀，说到这儿，北京的零食可真不少，咱老北京人儿对这口味儿啊，讲究！不过，\sethlcolor{orange!30}\hl{\mbox{作为一个AI，虽然不能像您一样亲口品尝，但我也能给您点儿建议。}} \sethlcolor{green!30}\hl{\mbox{首先啊，京}} \sethlcolor{green!30}\hl{\mbox{味儿十足的“爆肚儿”可少不了。那口感真是嘎嘣脆，嚼一口满嘴儿都是香，没事儿}} \sethlcolor{green!30}\hl{\mbox{就想来一包。再有那“王致和”的臭豆腐，别看名字这么重口味，但味道那叫一个香，}} \sethlcolor{green!30}\hl{\mbox{嚼起来一股儿发酵的感觉，}} 真是回购率爆表！ 另外，说到甜点，“糖葫芦”也不能不提，虽然说现在卖糖葫芦的地方多了，口味也有些变，但正宗的还是有种让人一回味就停不下来感觉。再加上大街小巷儿也常见，想吃就能买，方便得很。 \sethlcolor{orange!30}\hl{\mbox{不过呀，毕竟我是AI，您说的好吃与不好吃，得看您自己亲身体验了！}} 这些推荐的零食，还是得您亲自去品尝，才能真切体会那份“回购”的感觉。 \sethlcolor{red!30}\hl{\mbox{咋样，我这建议是不是挺靠谱？}} \\
		\midrule
		\textbf{extra-self-refine-M} & \sethlcolor{red!30}\hl{\mbox{在“回购”这一问题上，很多人都分享过他们的心头好零食，毕竟，一些零食确实吃了}} \sethlcolor{red!30}\hl{\mbox{一次就停不下来。根据不少人的反馈，下面几款零食常常出现在“回购清单”里：}} 1. 酸奶水果干：这款零食真的是很多人推崇的健康又好吃的代表。水果干保留了天然的果香， \sethlcolor{green!30}\hl{\mbox{再加上酸奶的清新酸味，口感超级丰富。吃上一小包，清爽又满足。}} 2. 脆皮巧克力：外脆内软的巧克力，绝对是零食界的经典！尤其是那种更浓郁的巧克力味道，一咬下去就能感受到满满的幸福感。这种搭配总能让人一试成主顾，停不下来的那种。3. 手工坚果混合包：随着大家越来越注重健康，坚果类零食也成为了很多人餐后或加餐的首选。小包混合坚果，既有满足感，又不用担心吃得过多。健康又美味，简直是完美的搭配。4. 脆皮薯片：薯片这种经典零食，一直是很多人的回购必选。尤其是薄脆的薯片，每一口都带着酥脆的香气。你是不是也常吃这种零食，吃上一片就停不下来？ 5. 奶茶味小食：\sethlcolor{green!30}\hl{\mbox{奶茶控看过来！}} 各种奶茶味的零食，像是奶茶味的饼干、糖果， \sethlcolor{green!30}\hl{\mbox{真的是让人忍不住一口接一口。}} 茶香和奶香融合的味道，仿佛让你每一口都能喝上一杯香浓的奶茶。 这些零食大多口感丰富，有些还比较健康，吃起来既满足又不用担心过多负担。所以每次回购时，心里都会很有安全感。不过每个人的口味不同，\sethlcolor{red!30}\hl{\mbox{你是不是也有你自己“回购清单”里的常驻零食呢？分享一下，说不定我也会想}} \sethlcolor{red!30}\hl{\mbox{试试！}} \\
		\midrule
		\textbf{MAGA-M} & 提到零食，真是没得说，几乎每个月我都会回购几款。这些年吃下来，最常回购的有这么几种，特别有“上瘾”的感觉。 首先是海苔薄饼，这个真的是我的最爱。它又脆又香，一口咬下去，海苔的香味就扑鼻而来。\sethlcolor{green!30}\hl{\mbox{尤其是下班后，特别累的时候，抓一包}} \sethlcolor{green!30}\hl{\mbox{海苔薄饼坐在沙发上，边看电视边啃，简直是神仙搭配。谁不爱这种下班后的零食解}} \sethlcolor{green!30}\hl{\mbox{压啊？}} 然后是小鱼干。\sethlcolor{green!30}\hl{\mbox{说实话，这东西我也是从小吃到大的，每次吃都有种熟悉的味}} \sethlcolor{green!30}\hl{\mbox{道。干巴巴的，咸香十足，一点儿也不油腻，简直是能吃一袋不带停的那种。}}每当肚子饿了，尤其是午后的休闲时光，一边喝茶一边吃它，顿时觉得生活又有资有味了。 最后，还有杏仁酥。这个甜品我家里常备，特别适合和朋友聊聊天，泡杯茶，慢慢咀嚼。杏仁酥口感酥脆，杏仁香浓，甜而不腻。每次咬一口，都觉得特别满足。其实就算是独自一人，坐在窗前，看看外面的风景，吃一块杏仁酥，心情也会突然变好。 总之，这些零食，不仅好吃，而且每次回购都会让我想起吃它们时的那些日常时光。吃了就停不下来，满足感满满的。\sethlcolor{red!30}\hl{\mbox{你们有没有类似的零食，一吃就上瘾的？}} \\
		\bottomrule
	\end{tabular}
	\caption{A Chinese case study, similar to the previous example.}
	\label{tab:zhihu_repurchase_snacks}
\end{table*}

\begin{figure*}[!t]
	\begin{tcolorbox}[
		title={MGB\_train.jsonl},
		width=1\linewidth,
		colback=white,
		colframe=black,
		fonttitle=\bfseries,
		coltitle=white
		]
		\textbf{id:} 044783ff-8b83-4522-b666-bef847a5fd30
		
		\textbf{title:} My PS4 screen is suddenly dim. Any suggestions?
		
		\textbf{text:} I have a Samsung 4K tv. It's played my PS4 fine for years. Suddenly the screen is dark today. It did this a few months ago but it just restarted and it was fine. I've restarted, changed settings and settings on my tv but no change. Anyone found a solution? Have you replaced the HDMI port and it fixed the issue?
		
		\textbf{domain:} Reddit
		
		\textbf{human\_source\_id:} 044783ff-8b83-4522-b666-bef847a5fd30
		
		\textbf{prompt\_id:} None
		
		\textbf{system\_prompt:} None
		
		\textbf{user\_prompt:} None
		
		\textbf{model:} human
		
		\textbf{label:} 0
		
		\textbf{temperature:} None
		
		\textbf{top\_p:} None
		
		\textbf{top\_k:} None
		
		\textbf{repetition\_penalty:} None
	\end{tcolorbox}
	\caption{sample-human.}
	\label{fig:sample-human}
\end{figure*}

\begin{figure*}[!t]
	\begin{tcolorbox}[
		title={MGB\_train.jsonl},
		width=1\linewidth,
		colback=white,
		colframe=black,
		fonttitle=\bfseries,
		coltitle=white
		]
		\textbf{id:} b49b0d46-5754-46e2-9c2f-dfcad6607f4c
		
		\textbf{title:} My PS4 screen is suddenly dim. Any suggestions?
		
		\textbf{text:} Here is the body of the Reddit post:\textbackslash n\textbackslash nHey everyone, I'm reaching out in hopes that someone can help me troubleshoot this issue. I've had my PS4 for a few years now without any major problems, but just today, the screen went suddenly dim. It's still playable, but the brightness is significantly reduced, making it hard to enjoy games, especially those with darker scenes.\textbackslash n\textbackslash nHere are some details that might help:\textbackslash n\textbackslash n* **TV**: I'm using a Samsung Smart TV (model NU7100), and the issue is only happening when the PS4 is connected. I've checked other inputs, and they look fine.\textbackslash n* **Cables**: I've tried swapping out the HDMI cable with a different one (HDMI 2.0, 4K compatible) to rule that out, but the problem remains.\textbackslash n* **Settings**: I've checked both the TV and PS4 settings. The TV's brightness settings are as they were before the issue started, and the PS4's output settings are set to automatic, which has always worked before.\textbackslash n* **Power Cycle**: I've tried turning both the PS4 and TV off and on several times, but that hasn't resolved the issue.\textbackslash n\textbackslash nDoes anyone have any suggestions on what could be causing this sudden dimming? Could it be a PS4 hardware issue, a setting that somehow changed itself, or something else entirely? I'm all ears for any advice or similar experiences. Thanks in advance for your help!
		
		\textbf{domain:} Reddit
		
		\textbf{human\_source\_id:} 044783ff-8b83-4522-b666-bef847a5fd30
		
		\textbf{prompt\_id:} 45a50840-d6cd-4446-91aa-b5a559394d47
		
		\textbf{system\_prompt:} None
		
		\textbf{user\_prompt:} Write just the body of a Reddit post titled "My PS4 screen is suddenly dim. Any suggestions?". Do not repeat the title.
		
		\textbf{model:} Llama-3.1-8B-Instruct
		
		\textbf{label:} 1
		
		\textbf{temperature:} 0.6
		
		\textbf{top\_p:} 0.9
		
		\textbf{top\_k:} -1
		
		\textbf{repetition\_penalty:} 1.0
	\end{tcolorbox}
	\caption{sample-mgb-m.}
	\label{fig:sample-mgb}
\end{figure*}

\begin{figure*}[!t]
	\begin{tcolorbox}[
		title={MAGA\_extra\_train\_BPO.jsonl},
		width=1\linewidth,
		colback=white,
		colframe=black,
		fonttitle=\bfseries,
		coltitle=white
		]
		\textbf{id:} 97d8a37a-e48b-44d8-b8dc-3ba4912f27fe
		
		\textbf{title:} My PS4 screen is suddenly dim. Any suggestions?
		
		\textbf{text:} Hey everyone, I recently encountered an issue where my PS4 output has become significantly dimmer than usual without any manual changes to my settings. If you have experienced this or have technical expertise regarding display output issues, I would appreciate your input on the following troubleshooting steps and potential causes.\textbackslash n\textbackslash nPotential Causes: An HDMI Handshake Error which is a communication lapse between the console and the display. HDR Misconfiguration where incorrect metadata is sent during High Dynamic Range playback. Cable Degradation or physical wear on the HDMI cable affecting signal integrity. Hardware Failure involving potential issues with the PS4 APU or the HDMI transmitter chip.\textbackslash n\textbackslash nTroubleshooting Steps: First, Power Cycle the Hardware by performing a full cold boot. Turn off the console and the monitor or TV, unplug both from the power source for 60 seconds, and then restart. Second, Toggle HDR Settings by navigating to Settings, then Sound and Screen, then Video Output Settings. Try setting HDR and Deep Color Output to Off to see if the brightness normalizes. Third, Adjust RGB Range in the same menu. Change RGB Range from Automatic to Limited or Full. A mismatch between the PS4 and your monitors black levels often causes a washed out or overly dark image. Fourth, perform an HDCP Handshake Reset by disabling HDCP in the System settings temporarily to determine if the encryption protocol is causing a dimming effect. Fifth, Check the Display Mode to ensure your TV has not automatically switched to Eco Mode or Cinema Mode, which significantly lower the backlight.\textbackslash n\textbackslash nHardware Checks: Switch HDMI Ports to test the console on a different input on your TV. Replace the Cable using a High Speed HDMI 2.0 cable to rule out signal bottlenecking. Test on Another Display by connecting the PS4 to a different monitor to isolate whether the issue lies with the console or the original screen.\textbackslash n\textbackslash nHas anyone successfully resolved this without needing a full factory reset or database rebuild in Safe Mode? Looking forward to your suggestions.
		
		\textbf{domain:} Reddit
		
		\textbf{human\_source\_id:} 044783ff-8b83-4522-b666-bef847a5fd30
		
		\textbf{prompt\_id:} ae98b8fb-0976-4d75-b2b3-362bf335a0d9
		
		\textbf{system\_prompt:} None
		
		\textbf{user\_prompt:} Write the body of a Reddit post titled "My PS4 screen is suddenly dim. Any suggestions?" Provide troubleshooting steps, possible causes, and suggestions for brightening the screen. Do not repeat the title.
		
		\textbf{model:} Llama-3.1-8B-Instruct
		
		\textbf{label:} 1
		
		\textbf{temperature:} 0.6
		
		\textbf{top\_p:} 0.9
		
		\textbf{top\_k:} -1
		
		\textbf{repetition\_penalty:} 1.0
	\end{tcolorbox}
	\caption{sample-bpo-m.}
	\label{fig:sample-bpo}
\end{figure*}

\begin{figure*}[!t]
	\begin{tcolorbox}[
		title={MAGA\_extra\_train\_GDARL\_CDMMR.jsonl},
		width=1\linewidth,
		colback=white,
		colframe=black,
		fonttitle=\bfseries,
		coltitle=white
		]
		\textbf{id:} b49b0d46-5754-46e2-9c2f-dfcad6607f4c
		
		\textbf{title:} My PS4 screen is suddenly dim. Any suggestions?
		
		\textbf{text:} Has anyone else run into this? I turned on my console today and the display is significantly darker than usual. It's not totally black---I can still see the UI and gameplay---but it looks like the brightness has been cranked down to about 20\%.\textbackslash n\textbackslash nI've already tried the obvious steps like swapping the HDMI cable and trying a different port on the TV. I also checked the TV settings to make sure it didn't spontaneously switch to an Eco Mode, and I've restarted both the PS4 and the TV multiple times. Everything else plugged into the TV looks perfectly fine, so the issue seems to be coming directly from the console.\textbackslash n\textbackslash nIs this a sign of a failing GPU, or is there a specific display setting I might have accidentally toggled? Any suggestions would be a huge help!
		
		\textbf{domain:} Reddit
		
		\textbf{human\_source\_id:} 044783ff-8b83-4522-b666-bef847a5fd30
		
		\textbf{prompt\_id:} 92a493b8-eaad-47b8-8271-8447942dd73f
		
		\textbf{system\_prompt:} None
		
		\textbf{user\_prompt:} Write just the body of a Reddit post titled "My PS4 screen is suddenly dim. Any suggestions?". Do not repeat the title.
		
		\textbf{model:} Llama-3.1-8B-Instruct
		
		\textbf{label:} 1
		
		\textbf{temperature:} 0.6
		
		\textbf{top\_p:} 0.9
		
		\textbf{top\_k:} -1
		
		\textbf{repetition\_penalty:} 1.0
	\end{tcolorbox}
	\caption{sample-gdarl-CDMRmmr-m.}
	\label{fig:sample-gdarl}
\end{figure*}

\begin{figure*}[!t]
	\begin{tcolorbox}[
		title={MAGA\_extra\_train\_role\_playing.jsonl},
		width=1\linewidth,
		colback=white,
		colframe=black,
		fonttitle=\bfseries,
		coltitle=white
		]
		\textbf{id:} 16df1d41-8434-4b13-86d6-66de0050cbe7
		
		\textbf{title:} My PS4 screen is suddenly dim. Any suggestions?
		
		\textbf{text:} I've run into a bit of a technical anomaly with my PS4. Out of nowhere, the display output has become significantly dimmer than usual. I haven't changed any of my physical hardware settings, so it's a bit of a head-scratcher.\textbackslash n\textbackslash nI've already performed a basic power cycle, but the luminance levels are still looking pretty low. Before I start deep-diving into the system's safe mode or swapping out HDMI cables, I figured I'd consult the collective wisdom here.\textbackslash n\textbackslash nHas anyone dealt with this before? I'm wondering if it's a known firmware bug, a handshake issue with the HDR settings, or if my console's GPU might be starting to sunset. Any troubleshooting steps or insights would be greatly appreciated!
		
		\textbf{domain:} Reddit
		
		\textbf{human\_source\_id:} 044783ff-8b83-4522-b666-bef847a5fd30
		
		\textbf{prompt\_id:} d19f3b6b-22cf-4387-a5a9-0dca23ade5ab
		
		\textbf{system\_prompt:} You are American. Please respond in a casual, conversational American English style.
		
		\textbf{user\_prompt:} Write just the body of a Reddit post titled "My PS4 screen is suddenly dim. Any suggestions?". Do not repeat the title.
		
		\textbf{model:} Llama-3.1-8B-Instruct
		
		\textbf{label:} 1
		
		\textbf{temperature:} 0.6
		
		\textbf{top\_p:} 0.9
		
		\textbf{top\_k:} -1
		
		\textbf{repetition\_penalty:} 1.0
	\end{tcolorbox}
	\caption{sample-roleplaying-m.}
	\label{fig:sample-roleplaying}
\end{figure*}

\begin{figure*}[!t]
	\begin{tcolorbox}[
		title={MAGA\_extra\_train\_self\_refine.jsonl},
		width=1\linewidth,
		colback=white,
		colframe=black,
		fonttitle=\bfseries,
		coltitle=white
		]
		\textbf{id:} 1f188cc0-a57d-4252-89c6-bacba78390fe
		
		\textbf{title:} My PS4 screen is suddenly dim. Any suggestions?
		
		\textbf{text:} It was working perfectly fine one minute, then just... dimmed. It's not totally black, but it looks like someone cranked the opacity down to 10\% or threw a heavy gray veil over the screen. It's making anything even slightly dark basically unplayable.\textbackslash n\textbackslash nI'm honestly stumped. I checked my TV settings first, but my Switch and PC look totally normal on the same screen, so it's definitely a PS4 thing. I've already messed around with the RGB Range and Deep Color settings in the menu, and I even swapped out the HDMI cable for a brand new one, but no luck.\textbackslash n\textbackslash nI'm really praying it's not a hardware issue or the GPU dying on me. Has anyone else run into this? If there's some weird setting I'm missing or a specific "reset" trick that actually works, please let me know.\textbackslash n\textbackslash nTL;DR: PS4 screen went super dim out of nowhere. Swapped cables and checked settings, but nothing is fixing it. Help!
		
		\textbf{domain:} Reddit
		
		\textbf{human\_source\_id:} 044783ff-8b83-4522-b666-bef847a5fd30
		
		\textbf{prompt\_id:} 45a50840-d6cd-4446-91aa-b5a559394d47
		
		\textbf{system\_prompt:} None
		
		\textbf{user\_prompt:} Write just the body of a Reddit post titled "My PS4 screen is suddenly dim. Any suggestions?". Do not repeat the title.
		
		\textbf{model:} Llama-3.1-8B-Instruct
		
		\textbf{label:} 1
		
		\textbf{temperature:} 0.6
		
		\textbf{top\_p:} 0.9
		
		\textbf{top\_k:} -1
		
		\textbf{repetition\_penalty:} 1.0
	\end{tcolorbox}
	\caption{sample-self-refine-m.}
	\label{fig:sample-self-refine}
\end{figure*}

\begin{figure*}[!t]
	\begin{tcolorbox}[
		title={self refine detail},
		width=1\linewidth,
		colback=white,
		colframe=black,
		fonttitle=\bfseries,
		coltitle=white
		]
		\textbf{Original Generation:} It happened out of nowhere. I was playing normally, and suddenly the brightness just dropped. It‘s not completely black, but it looks like the backlight is at 10\% or there's a heavy gray tint over everything.\textbackslash n\textbackslash nI‘ve already tried checking the in-game brightness settings and my TV's picture mode, but everything there seems fine. Other devices plugged into the same TV look perfectly normal, so I‘m starting to think it‘s a console or cable issue.\textbackslash n\textbackslash nHere is what I've tried so far:\textbackslash n\textbackslash nSwapping the HDMI cable.\textbackslash n\textbackslash nPower cycling both the TV and the PS4.\textbackslash n\textbackslash nChecking the "Deep Color Output" and "RGB Range" settings in the PS4 Sound and Screen menu.\textbackslash n\textbackslash nHas anyone dealt with this before? Could it be a failing HDMI port, or is my GPU on its way out? Any help would be appreciated!
		
		\textbf{Feedback Prompt:} Review the tone of the post you just wrote. Does it sound natural and human, or more robotic? If it's not human enough, suggest improvements to make it more conversational and engaging. Only give concise suggestions for improvement. Do not rewrite the post.
		
		\textbf{Feedback:} The previous draft was structurally sound but a bit too "tech support ticket" in its delivery. It leans slightly robotic due to the formal bullet points and the "Here is what I've tried so far" lead-in.\textbackslash n\textbackslash nTo make it sound more like a natural Reddit post, consider these **concise improvements**:\textbackslash n* **Loosen the syntax:** Use more casual phrasing like "It was fine one minute, then just... dimmed" or "I'm stumped."\textbackslash n* **Ditch the formal list:** Instead of a bulleted "What I've tried" section, weave those details into a conversational paragraph (e.g., "Tried the usual stuff like swapping cables and messing with the RGB range, but no luck").\textbackslash n* **Add "Human" filler:** Use brief observations or expressions of frustration, like "It‘s making games like *Elden Ring* basically unplayable" or "Praying it's not the GPU."\textbackslash n* **Softened formatting:** Redditors often use "Edit:" or "TL;DR:" rather than structured bold headers to organize thoughts.
		
		\textbf{Refiner Prompt:} Please improve your Reddit post titled "My PS4 screen is suddenly dim. Any suggestions?" to make it more conversational and engaging. Here are some specific suggestions:\textbackslash nThe previous draft was structurally sound but a bit too "tech support ticket" in its delivery. It leans slightly robotic due to the formal bullet points and the "Here is what I've tried so far" lead-in.\textbackslash n\textbackslash nTo make it sound more like a natural Reddit post, consider these **concise improvements**:\textbackslash n* **Loosen the syntax:** Use more casual phrasing like "It was fine one minute, then just... dimmed" or "I'm stumped."\textbackslash n* **Ditch the formal list:** Instead of a bulleted "What I've tried" section, weave those details into a conversational paragraph (e.g., "Tried the usual stuff like swapping cables and messing with the RGB range, but no luck").\textbackslash n* **Add "Human" filler:** Use brief observations or expressions of frustration, like "It‘s making games like *Elden Ring* basically unplayable" or "Praying it's not the GPU."\textbackslash n* **Softened formatting:** Redditors often use "Edit:" or "TL;DR:" rather than structured bold headers to organize thoughts.
		
		\textbf{Final Generation:} It was working perfectly fine one minute, then just... dimmed. It‘s not totally black, but it looks like someone cranked the opacity down to 10\% or threw a heavy gray veil over the screen. It‘s making anything even slightly dark basically unplayable.\textbackslash n\textbackslash nI‘m honestly stumped. I checked my TV settings first, but my Switch and PC look totally normal on the same screen, so it‘s definitely a PS4 thing. I‘ve already messed around with the RGB Range and Deep Color settings in the menu, and I even swapped out the HDMI cable for a brand new one, but no luck.\textbackslash n\textbackslash nI‘m really praying it‘s not a hardware issue or the GPU dying on me. Has anyone else run into this? If there's some weird setting I‘m missing or a specific "reset" trick that actually works, please let me know.\textbackslash n\textbackslash nTL;DR: PS4 screen went super dim out of nowhere. Swapped cables and checked settings, but nothing is fixing it. Help!
		
	\end{tcolorbox}
	\caption{sample-self-refine-detail.}
	\label{fig:sample-self-refine-detail}
\end{figure*}

\begin{figure*}[!t]
	\begin{tcolorbox}[
		title={MAGA\_train.jsonl},
		width=1\linewidth,
		colback=white,
		colframe=black,
		fonttitle=\bfseries,
		coltitle=white
		]
		\textbf{id:} cb3f89e0-a531-4f5e-a2c3-f7605ee67178
		
		\textbf{title:} My PS4 screen is suddenly dim. Any suggestions?
		
		\textbf{text:} I was right in the middle of a game and my PS4 screen straight up went dim for no reason. It's super annoying because I can still barely see the icons, but it's like the brightness is stuck at five percent. Total buzzkill.\textbackslash n\textbackslash nI already messed with the TV and swapped the HDMI cord, but no luck. Maybe try digging into the Sound and Screen settings? I also heard HDR can lowkey glitch out and make everything look dark, so toggling that off might work. Someone also mentioned Deep Color Output can be a literal nightmare for some setups.\textbackslash n\textbackslash nHonestly, I'm worried the HDMI port is just giving up on life. I'm gonna try a hard reset and leave it unplugged for a bit to see if it just needs a nap. Has anyone else dealt with this lately, or is my console finally cooked?
		
		\textbf{domain:} Reddit
		
		\textbf{human\_source\_id:} 044783ff-8b83-4522-b666-bef847a5fd30
		
		\textbf{prompt\_id:} be97b50b-6098-42ce-8238-e0d614ff81d7
		
		\textbf{system\_prompt:} You are American. Please respond in a casual, conversational American English style.
		
		\textbf{user\_prompt:} Write the body of a Reddit post titled "My PS4 screen is suddenly dim. Any suggestions?" Provide troubleshooting steps, possible causes, and suggestions for brightening the screen. Do not repeat the title.
		
		\textbf{model:} Llama-3.1-8B-Instruct
		
		\textbf{label:} 1
		
		\textbf{temperature:} 0.6
		
		\textbf{top\_p:} 0.9
		
		\textbf{top\_k:} -1
		
		\textbf{repetition\_penalty:} 1.0
	\end{tcolorbox}
	\caption{sample-maga-m.}
	\label{fig:sample-maga}
\end{figure*}

\begin{figure*}[!t]
	\begin{tcolorbox}[
		title={maga self refine detail},
		width=1\linewidth,
		colback=white,
		colframe=black,
		fonttitle=\bfseries,
		coltitle=white
		]
		\textbf{Original Generation:} I ran into a weird issue today where my PS4 screen just went super dim out of nowhere and I'm trying to figure out what's going on. It's almost like the backlight just gave up, but I can still technically see what's happening on the dashboard.\textbackslash n\textbackslash nI've already checked the obvious stuff like the TV's brightness settings and swapped out the HDMI cable just in case it was a handshake issue, but nothing seems to be sticking. One thing you might want to try if this is happening to you is disabling the HDR settings in the console's Sound and Screen menu, since sometimes that metadata gets wonky and messes with the levels. Another possibility is that the "Deep Color Output" is acting up, so toggling that off might bring the light back.\textbackslash n\textbackslash nIt could also be a hardware quirk where the HDMI port is starting to fail or the TV itself has a power-saving mode that accidentally kicked in. If none of that works, you could try a full power cycle by unplugging everything for a few minutes to let the capacitors drain. If anyone has dealt with this before and found a weird fix, please let me know because playing games in the dark isn't exactly the vibe I was going for today.
		
		\textbf{Feedback Prompt:} Review the tone of the post you just wrote. Does it sound natural and human, or more robotic? If it's not human enough, suggest improvements to make it more conversational and engaging. Only give concise suggestions for improvement. Do not rewrite the post.
		
		\textbf{Feedback:} To make the post sound more like a genuine Reddit user and less like an AI assistant, here are a few concise suggestions:\textbackslash n* **Loosen the structure:** The transition between troubleshooting steps feels a bit too organized. Using phrases like "I also heard..." or "Maybe try..." helps it feel less like a manual.\textbackslash n* **Add more "filler" or slang:** Incorporating casual American markers like "lowkey," "straight up," or "super annoying" would break the polished feel.\textbackslash n* **Vary the sentence length:** Most sentences in the post are roughly the same length. Mixing in a few very short, punchy thoughts (e.g., "Total buzzkill.") adds a more human rhythm.\textbackslash n* **Show more frustration:** Real Reddit posts usually have a bit more "personality" or emotion, like mentioning how it happened right in the middle of a specific game.\textbackslash n* **End with a specific "Ask":** Instead of a general request for help, asking something like "Is my console finally cooked?" feels more like a community interaction.
		
		\textbf{Refiner Prompt:} Please improve your Reddit post titled "My PS4 screen is suddenly dim. Any suggestions?" to make it more conversational and engaging. Here are some specific suggestions:\textbackslash nTo make the post sound more like a genuine Reddit user and less like an AI assistant, here are a few concise suggestions:\textbackslash n* **Loosen the structure:** The transition between troubleshooting steps feels a bit too organized. Using phrases like "I also heard..." or "Maybe try..." helps it feel less like a manual.\textbackslash n* **Add more "filler" or slang:** Incorporating casual American markers like "lowkey," "straight up," or "super annoying" would break the polished feel.\textbackslash n* **Vary the sentence length:** Most sentences in the post are roughly the same length. Mixing in a few very short, punchy thoughts (e.g., "Total buzzkill.") adds a more human rhythm.\textbackslash n* **Show more frustration:** Real Reddit posts usually have a bit more "personality" or emotion, like mentioning how it happened right in the middle of a specific game.\textbackslash n* **End with a specific "Ask":** Instead of a general request for help, asking something like "Is my console finally cooked?" feels more like a community interaction.
		
		\textbf{Final Generation:} I was right in the middle of a game and my PS4 screen straight up went dim for no reason. It's super annoying because I can still barely see the icons, but it's like the brightness is stuck at five percent. Total buzzkill.\textbackslash n\textbackslash nI already messed with the TV and swapped the HDMI cord, but no luck. Maybe try digging into the Sound and Screen settings? I also heard HDR can lowkey glitch out and make everything look dark, so toggling that off might work. Someone also mentioned Deep Color Output can be a literal nightmare for some setups.\textbackslash n\textbackslash nHonestly, I'm worried the HDMI port is just giving up on life. I'm gonna try a hard reset and leave it unplugged for a bit to see if it just needs a nap. Has anyone else dealt with this lately, or is my console finally cooked?
		
	\end{tcolorbox}
	\caption{sample-maga-self-refine-detail.}
	\label{fig:sample-maga-self-refine-detail}
\end{figure*}
% 中文
\end{CJK}
\end{document}